\documentclass{article}

\PassOptionsToPackage{dvipsnames,xcdraw,table}{xcolor}

\providecommand{\ispreprint}{1}
\ifnum\ispreprint=1
  \usepackage[preprint]{tmlr}
\else
  \usepackage{tmlr}
\fi

\usepackage{xspace}
\usepackage{graphicx}
\usepackage{amsmath}
\usepackage{amssymb}
\usepackage{booktabs}
\usepackage[format=plain,labelformat=simple,font=small,compatibility=false]{caption}
\usepackage[font=footnotesize,skip=3pt,subrefformat=parens]{subcaption}
\usepackage[hyphens]{url}
\usepackage{enumitem}
\setlist[itemize]{noitemsep,leftmargin=*,topsep=0em}
\setlist[enumerate]{noitemsep,leftmargin=*,topsep=0em}

\usepackage[dvipsnames,xcdraw,table]{xcolor}
\usepackage{tabularx}
\usepackage{array}
\usepackage{float}
\usepackage{pgfplots}
\pgfplotsset{compat=1.17}
\usepgfplotslibrary{groupplots}
\usepackage[most]{tcolorbox}
\usepackage[accsupp]{axessibility}

\definecolor{cOurs}{HTML}{2A78D6}
\definecolor{cInst}{HTML}{EB6834}
\definecolor{cDiff}{HTML}{1BAF7A}
\definecolor{cAtiss}{HTML}{EDA100}
\definecolor{cOursS}{HTML}{86B6EF}

\definecolor{tmlrblue}{rgb}{0.21,0.49,0.74}
\usepackage[breaklinks,colorlinks,citecolor=tmlrblue,linkcolor=tmlrblue]{hyperref}
\usepackage[capitalize]{cleveref}

\ifnum\ispreprint=1
\fi

\title{SceneNAT: Masked Generative Modeling for Language-Guided Indoor Scene Synthesis}

\ifnum\ispreprint=1
  \author{\name Jeongjun Choi$^*$ \email lojol2327@snu.ac.kr \\
        \addr Seoul National University
        \AND
        \name Yeonsoo Park$^*$ \email ys\_park@snu.ac.kr \\
        \addr Seoul National University
        \AND
        \name H. Jin Kim \email hjinkim@snu.ac.kr \\
        \addr Seoul National University}
\else
  \author{\name Anonymous Author \email anonymous@example.com \\
        \addr Anonymous Institution}
\fi

\AtBeginDocument{\renewcommand{\textcolor}[2]{#2}}

\begin{document}

\maketitle

\ifnum\ispreprint=1
  \let\thefootnote\relax\footnotetext{$^*$Equal contribution.}
\fi

\begin{abstract}
We present SceneNAT, a masked non-autoregressive Transformer for 3D indoor scene synthesis from natural language instructions.
It generates complete scenes in a few parallel decoding passes, improving both quality and efficiency over prior methods.
SceneNAT is trained via masked modeling over fully discretized representations of both semantic and spatial attributes.
By applying a masking strategy at both the attribute level and the instance level, the model can better capture intra-object and inter-object structure.
To boost relational reasoning, SceneNAT employs a relational reasoning module (RRM) that captures implicit spatial constraints.
By formulating relation modeling as a set prediction task, it extracts structure-aware features to guide the layout generation without explicit sequential parsing.
Extensive experiments on 3D-FRONT show that SceneNAT outperforms state-of-the-art autoregressive and diffusion baselines in both semantic compliance and spatial arrangement accuracy while using substantially lower computational cost, enabling high-throughput generation of diverse scenes at scale.
\end{abstract}

\begin{center}
  \includegraphics[width=0.95\textwidth]{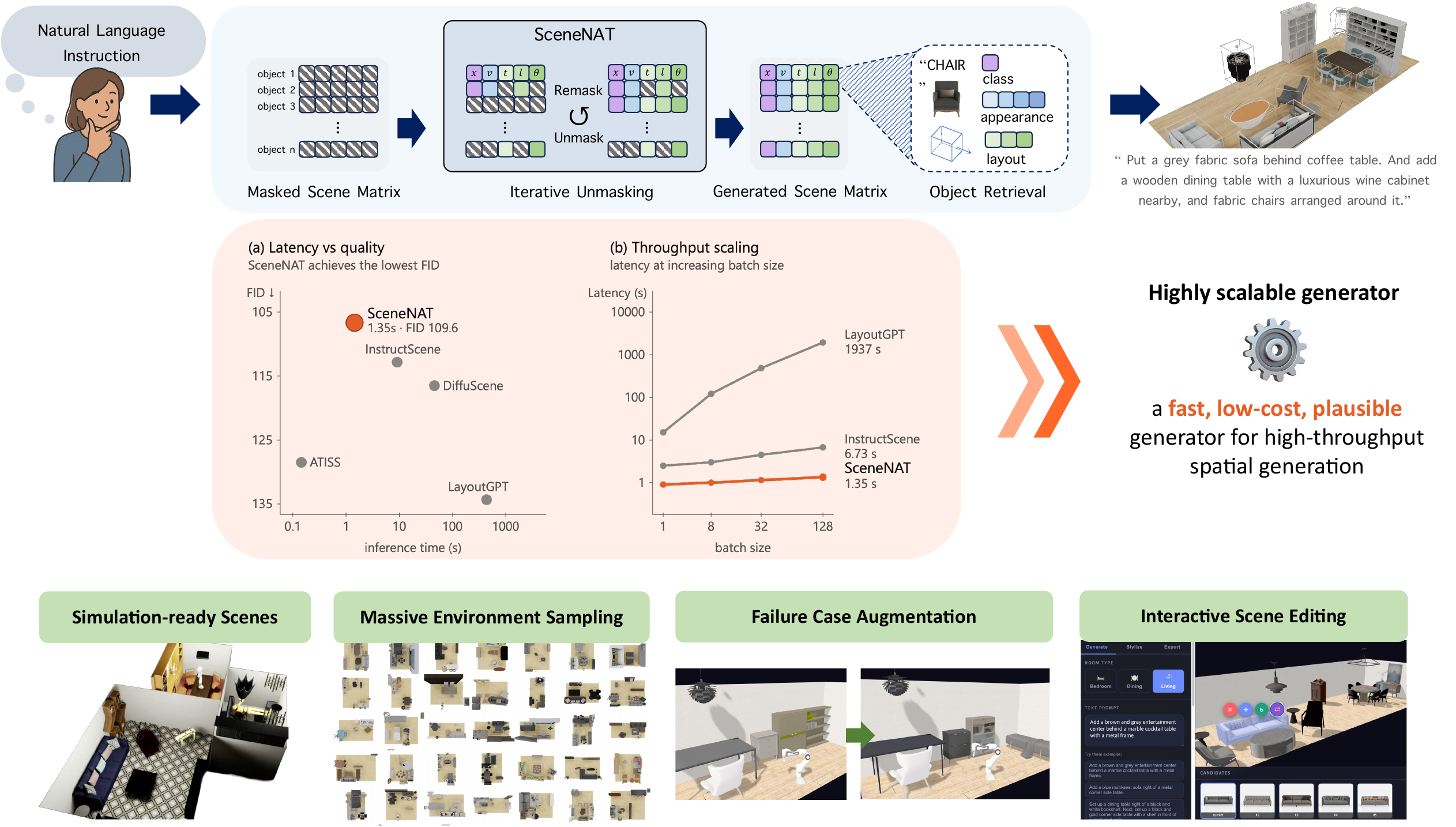}
  \vspace{-8pt}
  \captionsetup{type=figure,hypcap=false}
  \caption{%
  {{An illustration of the proposed method for text-driven 3D indoor scene generation.
Given a text instruction, our model iteratively predicts the attributes (class,
appearance, layout) for a set of initially masked objects in a non-autoregressive
manner (top). The resulting set of object attributes guides the retrieval and
placement of 3D assets to construct the scene. The middle row relates cost to
quality: (a) plots FID against inference time per scene, where SceneNAT attains
the lowest FID at 1.35 s, and (b) plots latency against batch size, where SceneNAT
stays nearly flat. The bottom row shows downstream uses
this efficiency enables: simulation-ready scenes, massive environment sampling,
failure-case augmentation, and interactive scene editing.}
}
  }
  \label{fig1}
\end{center}
\vspace{8pt}

\section{Introduction}
\sloppy

\vspace{-5pt}
\label{sec:intro}
Generating realistic 3D indoor scenes that faithfully adhere to text instructions is essential for applications, ranging from creative design tools like virtual staging \citep{ji2023virtual, tukur2023spider, tukur2024virtual} and AR/VR content creation \citep{dionisio20133d, Coelho, li2020automatic} to large-scale synthetic environments for robot learning and embodied AI \citep{fernandez2022robot, deitke2022, huang2023embodied, fu2024scene, yang2024physcene, khanna2024habitat, kim2026molmospaces}.
For these applications to be practically useful—especially in interactive design or simulation settings—users must be able to flexibly and reliably steer the output to match their intent.
This requires fine-grained control over both object content and spatial layout through natural language.
Such instructions would encode complex object attributes, spatial relationships, and implicit commonsense constraints, making it particularly difficult to ground language accurately and generate structured layouts in a controllable manner.

As a result, many prior works have proposed and advanced approaches based on either autoregressive or diffusion models.
Representative autoregressive approaches \citep{wang2021sceneformer, paschalidou2021atiss, leimer2022layoutenhancer, liu2023clip,para2023cofs,  zhao2024roomdesigner} formulate scene synthesis as set generation conditioned on room type and layout, enabling diverse and structured layout generation.
However, their sequential decoding suffers from inefficient generation and error accumulation, and it also prevents bidirectional reasoning due to reliance on pre-defined decoding orders across scene elements (e.g., category $\to$ attributes $\to$ layout).
More recently, diffusion models \citep{ho2020denoising, nichol2021improved} have been widely adopted in this domain \citep{zhai2023commonscenes, wei2023lego, tang2024diffuscene, lin2024instructscene, maillard2024debara, hu2024mixed, zhai2025echoscene, bai2025freescene}, harnessing their strengths in capturing fine-grained semantics and spatial detail.
However, their reliance on hundreds of sampling steps leads to high computational cost and latency, limiting their applicability in real-time settings.
Beyond these data-driven paradigms, LLM- and VLM-based agentic frameworks \citep{yang2024holodeck, ccelen2024design, yang2025sceneweaver, berdoz2025text, gao2025disco, xia2026sage, pfaff2026scenesmith} have emerged as another line of approach, offering high flexibility and generality through iterative reasoning. However, their sequential API querying and multi-step physical optimization make them ill-suited for large-scale scene production, where throughput becomes critical. We view these directions as serving different roles: agent-based methods excel at open-ended, high-flexibility scenarios, whereas large-scale conditional generation calls for a complementary approach optimized for throughput.

Orthogonal to the choice of generative paradigm,
another important line of research has studied controllable 3D scene generation, aiming to better align outputs with detailed user input.
Prior works \citep{paschalidou2021atiss, wang2021sceneformer, hu2024mixed, zhai2023commonscenes, zhai2025echoscene, feng2023layoutgpt, tang2024diffuscene, lin2024instructscene, sun2024layoutvlm, bai2025freescene} have sought to enhance controllability using various conditioning inputs, including floorplans, scene graphs, and natural language instructions.
Although notable progress has been made, reflecting detailed user intent is further complicated when instructions contain \textit{multiple} spatial relations, nested object dependencies, or detailed numerical constraints \citep{ye2024relscene, tam2025sceneeval}.
Consequently, generated layouts frequently deviate from the desired configuration, especially as instruction complexity increases.

In this paper, we introduce a novel Masked Indoor Scene Modeling (MISM) approach for instruction-conditioned 3D layout generation using a non-autoregressive transformer designed for parallel decoding.
This modeling strategy progressively reconstructs complete 3D scenes from fully masked tokens through iterative refinement, enabling efficient inference and bidirectional context modeling.
To address the challenges of complex instructions, we also introduce a dedicated relational reasoning module (RRM). This module uses a transformer decoder to produce relation-aware features with auxiliary supervision from sparse symbolic constraints formatted as (subject, predicate, object) triplets. At inference time, SceneNAT does not decode symbolic triplets; only the learned relational features are used to guide layout generation. This design provides robust and interpretable guidance, leading to more structured and accurate scene layouts.

Our contributions are as follows:

\begin{itemize}[itemsep=0pt,parsep=0pt,topsep=0pt]
    \item We present SceneNAT, the first framework to leverage a non-autoregressive transformer with carefully designed masked modeling for language-guided 3D indoor scene synthesis. Our model achieves superior results while significantly reducing computational costs compared to diffusion-based approaches.
    \item We propose a novel approach that disentangles relational reasoning from scene generation using a dedicated relational reasoning module (RRM). By explicitly modeling inter-object dependencies as a direct set prediction problem with bipartite matching under auxiliary triplet supervision, our approach overcomes a fundamental limitation of prior methods, enabling our model to achieve superior controllability and fidelity with complex language instructions.
    \item Extensive experiments on high-quality 3D scene benchmark demonstrate that SceneNAT achieves state-of-the-art performance, outperforming representative baselines in spatial accuracy, semantic alignment, and overall scene quality, while offering substantially improved scalability for large-scale scene generation.
\end{itemize}

\section{Related Works}
\label{sec:relwork}

\noindent \textbf{3D Scene Synthesis}
The synthesis of realistic 3D layouts has been a long-standing challenge. 
Early approaches utilized convolutional networks \citep{wang2018deep, wang2019planit, ritchie2019fast} and graph encoders \citep{li2019grains, zhou2019scenegraphnet}, while recent works incorporate hierarchical structures \citep{pun2025hsm} or autoregressive Transformers \citep{wang2021sceneformer, paschalidou2021atiss} to sequentially generate scene elements. 
More recently, diffusion models \citep{huang2023diffusion, maillard2024debara, lin2024instructscene, zhai2023commonscenes, bai2025freescene} have been introduced. These approaches, including those leveraging geometric scene graphs \citep{ruiz2025geoscenegraph, yang2025mmgdreamer} or semantic layouts \citep{sun2025semlayoutdiff}, excel at generating diverse and physically plausible layouts but are computationally expensive due to their iterative sampling process.
In parallel, recent works have explored leveraging large language models (LLMs) \citep{feng2023layoutgpt, yang2024llplace, sun2024layoutvlm, ran2025direct, bucher2025respace, yang2025sceneweaver} to directly map text into structured 3D scene representations.

\noindent \textbf{Controllable 3D Scene Generation}
A critical aspect of scene generation is controllability, enabling users to specify objects, their arrangement, and style. Various control signals have been explored, including floor plans \citep{paschalidou2021atiss, wang2021sceneformer}, scene graphs \citep{zhou2019scenegraphnet, zhai2023commonscenes, gao2026incremental3d}, etc. Among these, natural language offers the most intuitive way to specify detailed configurations. However, its free-form nature makes accurate interpretation highly challenging, particularly for fine-grained object relations and spatial constraints. Most text-conditioned frameworks \citep{wang2021sceneformer, lin2024instructscene} rely on global text representations, which often fail to capture these details. 
Recent LLM-based planners and agentic frameworks \citep{yang2024holodeck, ccelen2024design, yang2025sceneweaver, berdoz2025text, gao2025disco, xia2026sage, pfaff2026scenesmith} can interpret high-level language and align with user preferences through iterative reasoning, but their sequential querying and multi-step physical optimization incur high latency compared to one-pass generative models. Since these API calls cannot be batched or parallelized across scenes, the cost scales linearly with the number of scenes generated, making them ill-suited for large-scale production.

\noindent \textbf{Non-Autoregressive Transformers (NATs)}
Originally developed for machine translation \citep{ghazvininejad2019mask, gu2020fully}, NATs have since been adopted across various generation tasks such as image \citep{chang2022maskgit, lezama2022improved, chang2023muse, li2023mage, ni2024revisiting, bai2024meissonic}, video editing \citep{ma2024maskint}, motion synthesis \citep{guo2024momask}, and object-centric visual generation \citep{nakano2025efficient}, where fast and scalable inference is essential.
As an alternative to sequential and diffusion-based models, they enable fast and scalable inference by decoding all elements in parallel, while still supporting bidirectional context modeling with fewer steps.
Despite their potential, NATs remain largely unexplored for 3D scene layout generation, where aligning object semantics with spatial structure is inherently difficult, and further complicated by the need to satisfy complex language instructions.
\begin{figure}[t]
    \centering
    \includegraphics[width=\textwidth]{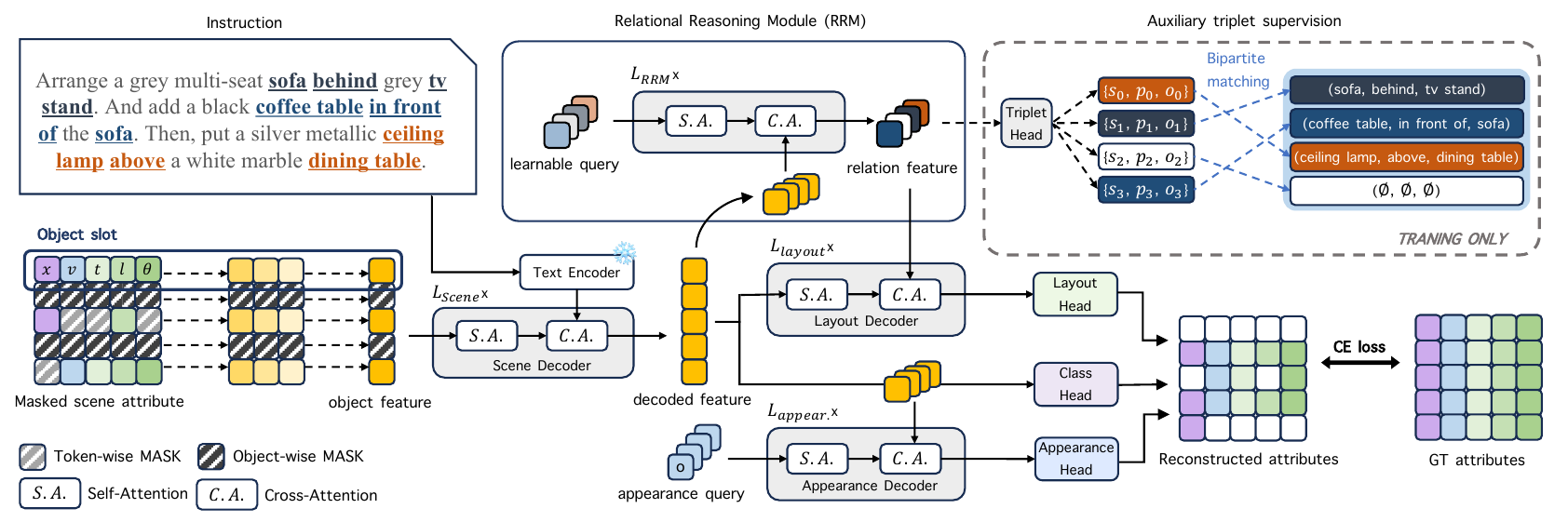}
    \caption{
    An overview of the SceneNAT framework. To achieve both visual quality and controllability, our model employs two specialized components. A scene decoder uses masked modeling to generate a globally coherent scene structure. Concurrently, relational reasoning module (RRM) transforms a set of learnable queries into relation-aware features, guided by constraints parsed from the text instruction. These intermediate relational features are fused into the layout decoder via cross-attention layers to generate the final indoor scene attributes.}
    \label{fig2}
    \vspace{-8pt}
\end{figure}

\section{Method}

\subsection{Problem Definition and Scene Representation}
\label{problem definition}
We formulate 3D scene synthesis from natural language as a masked modeling task over structured object-level attributes.
Each scene $r_i$ is represented as a set of $n_i$ objects, where a scene consists of a maximum of $N$ objects.
Each object is described by its category $x \in \mathcal{C}$, vector-quantized appearance tokens $v \in \mathbb{Z}^4$, and discretized layout parameters: 3D position $t \in \mathbb{Z}^3$, scale $l \in \mathbb{Z}^3$, and yaw rotation $\theta \in \mathbb{Z}$.
We obtain the discrete appearance token $v$ from features extracted by a pretrained OpenShape encoder~\citep{liu2023openshape} using the vector-quantized tokenizer from InstructScene~\citep{lin2024instructscene}.
The layout variables are also uniformly discretized into fixed bins.
Collectively, these attributes define a structured scene representation $r_i=\{x_j, v_j, t_j, l_j, \theta_j\}_{j=1}^{n_i}$, including both semantic and geometric information. Given a text instruction $I$, the objective is to model the conditional distribution $P(R|I)$ and generate a plausible scene consistent with the described configuration.
Following common practice, the instruction is encoded using a pretrained CLIP text encoder \citep{radford2021learning}.

\subsection{SceneNAT Framework}
Our goal is to establish a MISM framework that enables controllable 3D layout generation conditioned on text instruction.
To this end, we formulate the task as masked modeling. 
Given a partially masked scene representation $R_{\sim M}$, the model learns to infer the masked tokens $R_M$ by modeling the conditional distribution $P(R \mid R_{\sim M}, I),$
where $M$ denotes the set of masked positions.
An overview of the pipeline is provided in \cref{fig2}, and we detail its components in the following subsections.

\noindent \textbf{Object Attribute Embedding}
Each object is represented by discrete tokens for $x$, $v$, $t$, $l$, and $\theta$ (\cref{problem definition}).
Each attribute type has its own embedding table.
These embedding features are aggregated across attributes to form a representative feature vector for each object.
We reuse the embedding tables in the final attribute prediction heads via weight tying \citep{press2017using,inan2017tying}, reducing parameters while keeping input and output token spaces consistent.

\noindent \textbf{Decoder Architecture}
The decoder stage of SceneNAT operates on the object embeddings to produce the final scene layout.
The scene decoder reconstructs masked object slots by leveraging contextual information from visible slots and guidance from the text instruction to capture the global structure. However, attribute decoding alone is often insufficient for grounding complex instructions that involve inter-object relations.
To support such relation-aware modeling, our architecture incorporates RRM.
This module separates relational reasoning from the primary attribute reconstruction task.
Instead of treating relations as simple text tokens, this module is trained to encode relational semantics into a set of learnable queries.
These queries are supervised using ground-truth subject-predicate-object triplets parsed from training instructions, effectively inducing the model to capture symbolic constraints within the latent features.
In these triplets, the predicates from a predefined set $\mathcal{P}$ specify how objects are arranged.
We provide further details on this module in \cref{triplet}.

\noindent \textbf{Attribute Prediction Heads}
The decoded features are passed to attribute-specific heads to generate the complete scene.
The class head determines the object's category, while the appearance head recovers fine-grained style.
The layout head predicts translation, scale, and rotation by integrating both the contextual features from the scene decoder and the \textit{relational embeddings} from RRM.
This ensures the final object placement complies with the symbolic constraints expressed in the instruction.
Together, these predictions complete the scene matrix, yielding a structured 3D layout that coherently reflects both semantics and geometry.

\subsection{Relational Reasoning via Direct Set Prediction}
\label{triplet}

To explicitly model the inter-object relationships dictated by the language instruction, we frame relational reasoning as a direct set prediction problem \citep{carion2020end}.
This formulation allows us to capture the permutation-invariant nature of scene relationships without imposing an artificial order.
Our RRM, illustrated in \cref{fig2}, uses a transformer decoder to transform a fixed set of $N_{\text{q}}$ learnable \textit{triplet queries} into relation-aware features $F_{\text{rel}}$.
While we employ auxiliary heads (FFNs) to map these features to symbolic triplets (subject, predicate, object) for supervision, the primary role of this module during inference is to inject the learned $F_{\text{rel}}$ into the scene decoder to guide layout generation.
Each head's output includes an additional class to represent a ``no-relation'' ($\emptyset$) class, allowing the model to handle scenes with fewer than $N_{\text{q}}$ constraints.

Since the relational constraints form a set where the order is arbitrary, we employ a permutation-invariant set-based loss that finds an optimal bipartite matching between the ground-truth and predicted triplets. Let the ground-truth set be $Y = \{y_j\}_{j=1}^{N_t}$, where each $y_j = (s_j, p_j, o_j)$. Let the predicted set be $\hat{Y} = \{\hat{y}_k\}_{k=1}^{N_q}$, where each prediction $\hat{y}_k = (\hat{s}_k, \hat{p}_k, \hat{o}_k)$ is a bundle containing the logits for a subject, predicate, and object.
We use the Hungarian algorithm to find the optimal matching $\hat{\sigma}$ that minimizes the overall matching cost:
\begin{equation}
\hat{\sigma} = \underset{\sigma \in \mathfrak{S}_{N_{\text{q}}}}{\arg\min} \sum_{j=1}^{N_\text{t}} \mathcal{C}_{\text{match}}(y_j, \hat{y}_{\sigma(j)})
\end{equation}
The pair-wise matching cost $\mathcal{C}_{\text{match}}$ is defined as a linear combination of the negative log-probabilities for the ground-truth subject, predicate, and object classes.
For instance, the predicate cost is $\mathcal{C}_p(p_j, \hat{p}_k) = -\mathbb{P}_{\hat{p}_k}(p_j)$, where $\mathbb{P}_{\hat{p}_k}$ is the probability distribution from the softmax of the predicate logits $\hat{p}_k$.
Once the optimal matching $\hat{\sigma}$ is established, we define the target for each of the $N_q$ prediction queries to compute the triplet loss.
The target for each prediction $\hat{y}_k$, denoted $y'_k$, is the matched ground-truth triplet $y_j$ if a match exists (i.e., $k = \hat{\sigma}(j)$), and the null triplet $(\emptyset, \emptyset, \emptyset)$ otherwise.
The final triplet loss is a weighted sum of cross-entropy losses over all $N_q$ predictions against their corresponding targets, $y'_k = (s'_k, p'_k, o'_k)$. The loss is formulated as:
\begin{equation}
\begin{aligned}
\mathcal{L}_{\text{triplet}}
= \sum_{k=1}^{N_{\text{q}}} \big[
&\lambda_s \mathcal{L}_{\text{CE}}(s'_k, \hat{s}_{k})
+ \lambda_p \mathcal{L}_{\text{CE}}(p'_k, \hat{p}_{k}) \\
&+ \lambda_o \mathcal{L}_{\text{CE}}(o'_k, \hat{o}_{k})
\big].
\end{aligned}
\end{equation}
To balance the training, we apply a lower weight to the loss for the $\emptyset$ class.
This entire mechanism enables end-to-end learning of relational structures, producing robust feature guidance for the layout generation task.

\subsection{Training and Inference}
\label{train}
We optimize SceneNAT with a token-level reconstruction loss over masked scene attributes. For each masked token $i \in M$, the model is trained to maximize the likelihood of the ground-truth token $r_i$ given the unmasked context $R_{\sim M}$ and the instruction $I$. This reconstruction objective is a form of cross-entropy loss, applied separately to each attribute type.
Formally, the reconstruction loss is a weighted sum of cross-entropy losses for each attribute, defined as:
\begin{equation}
\begin{aligned}
\mathcal{L}_{\text{recon}}
=\;& \lambda_x \mathcal{L}_{\text{CE}}(x, \hat{x})
+ \lambda_v \mathcal{L}_{\text{CE}}(v, \hat{v})
+ \lambda_t \mathcal{L}_{\text{CE}}(t, \hat{t}) \\
&+ \lambda_l \mathcal{L}_{\text{CE}}(l, \hat{l})
+ \lambda_\theta \mathcal{L}_{\text{CE}}(\theta, \hat{\theta}).
\end{aligned}
\end{equation}
where $x, v, t, l, \theta$ are the ground-truth attributes and $\hat{x}, \hat{v}, \hat{t}, \hat{l}, \hat{\theta}$ are the model's predictions.
The $\lambda$ terms are hyperparameters used for attribute-specific loss weighting.
This reconstruction loss is defined to be equivalent to minimizing the negative log-likelihood of the ground-truth tokens, and it can be expressed as:
\begin{equation}
\mathcal{L}_{\text{recon}} = \sum_{i \in M} -\log P(r_i \mid R_{\sim M}, I)
\end{equation}
where $r_i$ is the ground-truth token at position $i$.
The overall training objective combines this reconstruction loss with the relational reasoning loss from \cref{triplet} and is given by
$
\mathcal{L}_{\text{total}} = \mathcal{L}_{\text{recon}} + \lambda_{\text{triplet}} \mathcal{L}_{\text{triplet}},
$
where $\lambda_{\text{triplet}}$ is a hyperparameter to balance two terms.

\begin{figure}[t]
  \centering
  \begin{minipage}{\textwidth}   
    \centering
    \input{figure/fig_sampling}
  \end{minipage}
  \caption{Visualization of the iterative sampling process.
\textbf{Left:} the initial mask state at step~0, where each row of the scene
matrix corresponds to an object and each column to an attribute, with all
elements masked and appearing dark.
\textbf{Right:} the 50-step sampling trajectories for a bedroom and a dining room scene, shown at every 10th step. For each scene, the upper row displays the top-down renders and the lower row
shows the corresponding mask state, with fixed (unmasked) elements highlighted
in color.}
  \label{fig:sampling}
\end{figure}

\noindent \textbf{Masking Strategy}
Following prior works \citep{chang2022maskgit, chang2023muse, guo2024momask}, we use a cosine-based dynamic masking schedule, where the masking ratio is sampled from $\gamma(\tau) = \cos\left(\frac{\pi \tau}{2}\right)$ with $\tau \sim \mathcal{U}(0, 1)$. To better capture the multi-modal nature of 3D scenes and disentangle local and global context, we stochastically split the total masking ratio between object-level and token-level masking. Specifically, we sample a proportion $p_{\text{obj}} \sim \mathcal{U}(0, 1)$ and assign $p_{\text{obj}} \cdot \gamma(\tau)$ of the masking to the object level, with the remainder assigned to the token level. This stochastic masking schedule encourages the model to handle various levels of corruption during training, from fully observed to nearly fully masked inputs, enabling joint learning of representation and generation \citep{li2023mage}. Additionally, we apply a BERT-style \citep{devlin2019bert} replace-and-remask policy, simulating noise-injected refinement during inference.

\noindent \textbf{Sampling Strategy}
For inference, our model starts from a fully masked scene representation and performs parallel decoding conditioned on the instruction $I$, progressively refining its predictions to reconstruct a complete scene layout.
During this phase, the auxiliary subject/predicate/object classification heads are not used; only the relational features $F_{\text{rel}}$ from RRM are injected into layout decoding.
We adopt an iterative masked token prediction strategy inspired by MaskGIT~\citep{chang2022maskgit}, where uncertain tokens are progressively refined through parallel decoding.  
At each iteration, the model remasks and re-predicts tokens with low confidence, measured by the maximum softmax probability.  
This process is repeated for a fixed number of steps, with the remasking ratio at each step determined by a cosine schedule, consistent with the masking strategy used during training.
This iterative refinement gradually improves the quality and consistency of the generated scene.  
\cref{fig:sampling} visualizes this process, showing how the fully masked
scene matrix is progressively filled over 50 sampling steps.
\section{Experiment}

\subsection{Experimental Setup}
\noindent \textbf{Dataset} We conduct all experiments on the extended 3D-FRONT dataset introduced in InstructScene~\citep{instructscene_dataset}, which augments 3D-FRONT~\citep{fu20213d_front} and 3D-FUTURE~\citep{fu20213d_future} with instruction annotations.
To improve instruction quality and coverage, we enhance the original templated instructions with more diverse sentence structures and increase the number of relations per instruction.
We support up to four constraints to accommodate more complex guidance, whereas InstructScene~\citep{lin2024instructscene} typically uses only one or two per instruction.
We set this limit on the number of relations based on the maximum token length of the CLIP~\citep{radford2021learning} text encoder.
Instructions with more than four relations typically exceed this length, leading to a loss of information through truncation.
Further details of the instruction refinement process are provided in \cref{prompting}.

\noindent \textbf{Baselines}
We compare against three baseline models, including state-of-the-art methods that cover diverse architectural paradigms in 3D indoor scene generation.
ATISS~\citep{paschalidou2021atiss} is an autoregressive transformer model that sequentially generates object attributes conditioned on previously generated outputs.
DiffuScene~\citep{tang2024diffuscene} represents scene components as a fixed-size 2D matrix and generates the matrix through a continuous diffusion process.
InstructScene~\citep{lin2024instructscene} adopts a semantic scene graph with object relation edges as an intermediate representation and employs a \textit{two-stage} cascaded diffusion model.
To ensure fair comparison, we reimplemented all baselines following InstructScene and trained them from scratch using our instruction set.
As a more recent method, FreeScene~\citep{bai2025freescene} introduces a mixed-graph diffusion framework for 3D scene synthesis from free-form prompts. Since its official training code is not publicly available, we report its metrics from the original paper and provide a separate comparison.
Implementation details for both baselines and our method are provided in \cref{implementation}.

\begin{figure}[t]
\centering
\includegraphics[width=0.96\textwidth]{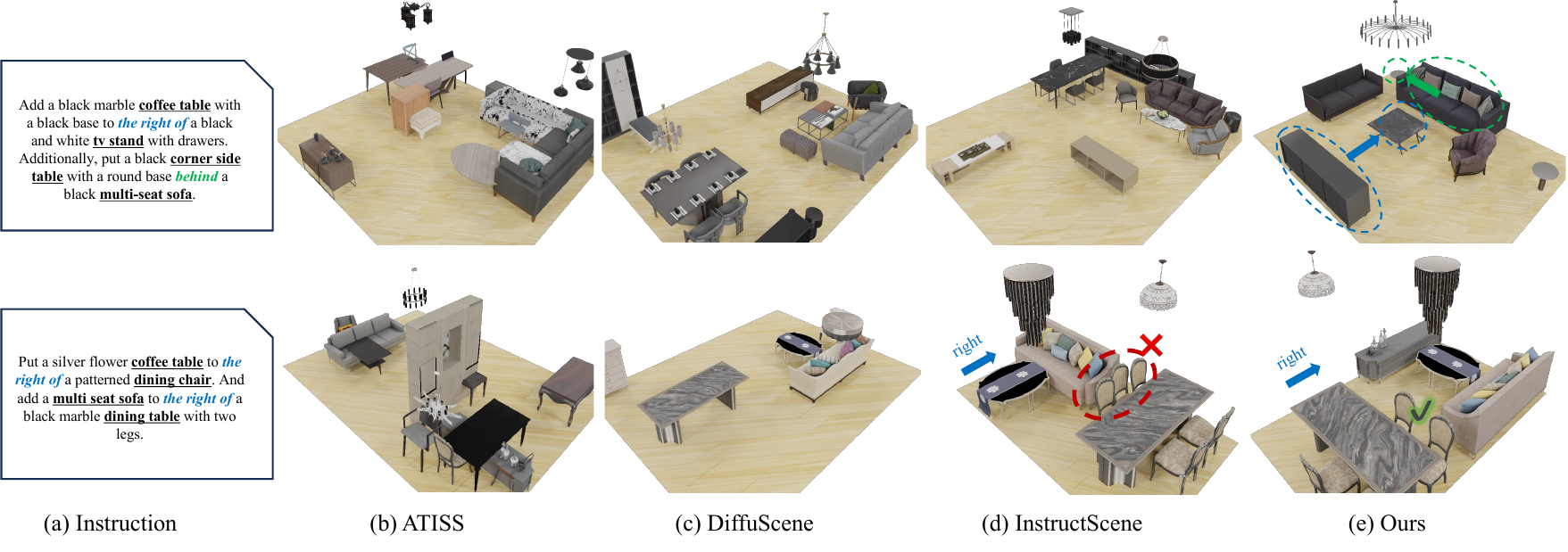}
\caption{Comparison of qualitative results on language-guided scene generation. SceneNAT demonstrates superior performance in generating realistic 3D scenes that faithfully adhere to complex instructions. Additional qualitative results can be found in \cref{qualitative_appendix}.}
\label{fig3}
\vspace{-6pt}
\end{figure}

\begin{figure}[t!]
    \centering
    \pgfplotsset{
        ATISS/.style={color=cAtiss, mark=*, mark size=1.1pt, thick},
        DiffuScene/.style={color=cDiff, mark=*, mark size=1.1pt, thick},
        InstructScene/.style={color=cInst, mark=*, mark size=1.1pt, thick},
        Ours/.style={color=cOurs, mark=*, mark size=1.4pt, very thick},
    }
    \begin{tikzpicture}
        \begin{groupplot}[
            group style={group size=3 by 1, horizontal sep=1.15cm},
            width=0.34\textwidth,
            height=3.6cm,
            xlabel={$\#$ of relations},
            xmin=0.8, xmax=6.2,
            xtick={1, 2, 3, 4, 5, 6},
            tick label style={font=\tiny},
            label style={font=\tiny},
            title style={font=\small\bfseries},
            legend style={
                at={(0.5, -0.58)},
                anchor=north,
                legend columns=-1,
                font=\tiny,
                draw=none,
            },
            grid=major,
            grid style={dashed, gray!25},
            axis on top,
        ]

        \nextgroupplot[title=Bedroom, ylabel={iRecall (\%)}, ymin=15, ymax=90]
            \fill[black!7] (axis cs:4.5,15) rectangle (axis cs:6.2,90);
            \addplot [ATISS] coordinates {(1,36.81) (2,33.82) (3,29.22) (4,29.80) (5,31.71) (6,31.78)};
            \addplot [DiffuScene] coordinates {(1,47.45) (2,47.62) (3,47.38) (4,43.17) (5,43.13) (6,48.22)};
            \addplot [InstructScene] coordinates {(1,64.58) (2,68.89) (3,68.68) (4,64.19) (5,64.69) (6,62.43)};
            \addplot [Ours] coordinates {(1,81.71) (2,74.21) (3,68.20) (4,68.54) (5,68.08) (6,69.16)};

        \nextgroupplot[title=Living room, ymin=10, ymax=65]
            \fill[black!7] (axis cs:4.5,10) rectangle (axis cs:6.2,65);
            \addplot [ATISS] coordinates {(1,21.05) (2,21.21) (3,21.36) (4,19.18) (5,20.73) (6,20.12)};
            \addplot [DiffuScene] coordinates {(1,34.62) (2,26.23) (3,26.42) (4,26.77) (5,27.53) (6,28.72)};
            \addplot [InstructScene] coordinates {(1,58.70) (2,48.98) (3,48.73) (4,43.85) (5,43.21) (6,41.69)};
            \addplot [Ours] coordinates {(1,60.93) (2,52.97) (3,47.16) (4,47.67) (5,47.56) (6,50.15)};
            \legend{ATISS, DiffuScene, InstructScene, Ours}

        \nextgroupplot[title=Dining room, ymin=20, ymax=70]
            \fill[black!7] (axis cs:4.5,20) rectangle (axis cs:6.2,70);
            \addplot [ATISS] coordinates {(1,32.89) (2,30.38) (3,34.45) (4,26.92) (5,27.66) (6,31.33)};
            \addplot [DiffuScene] coordinates {(1,41.67) (2,39.47) (3,36.17) (4,34.13) (5,37.01) (6,33.70)};
            \addplot [InstructScene] coordinates {(1,51.10) (2,47.01) (3,48.10) (4,43.69) (5,46.17) (6,44.94)};
            \addplot [Ours] coordinates {(1,63.60) (2,57.65) (3,56.52) (4,53.31) (5,53.08) (6,53.64)};
        \end{groupplot}
    \end{tikzpicture}
    \caption{iRecall against the number of relational constraints. The shaded band marks
    5 and 6 constraints, which exceed the maximum of 4 seen during training.}
    \label{fig:rel_vs_irecall}
\end{figure}

\subsection{Metrics}
We evaluate generated scenes based on instruction adherence, visual quality, and physical plausibility.
Instruction adherence is measured using iRecall~\citep{lin2024instructscene}, which reports the recall of spatial relation triplets specified in the instruction that are realized in the generated scene.
Following prior works~\citep{paschalidou2021atiss, liu2023clip, tang2024diffuscene, lin2024instructscene}, we assess the visual plausibility of rendered scenes using three metrics.
FID~\citep{heusel2017gans} measures feature distribution differences, $\text{FID}^{\text{CLIP}}$~\citep{kynkaanniemi2022role} assesses perceptual alignment using CLIP embeddings, and KID~\citep{gruber2023bias} provides an unbiased kernel-based similarity estimate.
For all metrics, lower values indicate a closer alignment to the real scene distribution.
Finally, to assess physical plausibility, we report SceneEval-style layout metrics: non-accessibility violation (NAV), object collision (COL), accessibility (ACC), and floor penetration (Floor Pen.). Additional collision metrics and a more comprehensive evaluation are provided in \cref{sec:collision_analysis}.

\begin{table}[t!]
    \centering
    \scriptsize
    \caption{Quantitative comparison on text-conditioned scene synthesis on three room types. We report the mean and standard deviation over 10 independent trials for generative quality and instruction adherence metrics. For all metrics, lower is better except for iRecall, NAV, and ACC. The best results are highlighted in bold, and the second-best are underlined.}
    \label{table:main}
    \setlength{\tabcolsep}{2.7pt}
    \resizebox{\textwidth}{!}{
    \fontsize{8}{8}\selectfont
    \begin{tabular}{@{}llcccccccc@{}}
    \toprule
                &                     & iRecall(\%) ($\uparrow$) & FID ($\downarrow$)     & FID-CLIP ($\downarrow$) & KID$_{\times1e3}$ ($\downarrow$) & NAV(\%) ($\uparrow$) & COL(\%) ($\downarrow$) & ACC(\%) ($\uparrow$) & Floor Pen. ($\downarrow$) \\ \midrule
    Bedroom     & LayoutGPT           & 45.32                     & 133.31                    & 8.77                    & 3.49                     & 73.45 & 51.64 & 41.65 & 0.0274 \\ \cmidrule(lr){2-10}
                & ATISS               & 31.30 (1.76)              & 128.50 (1.28)              & 7.50 (0.14)               & 3.59 (0.71)               & 78.7 & 52.6 & 53.3 & 0.056 \\
                & DiffuScene          & 45.98 (2.30)              & 119.37 (1.53)              & 6.71 (0.14)               & 1.04 (0.82)               & \underline{83.6} & 34.8 & 54.5 & \textbf{0.014} \\
                & InstructScene       & \underline{66.72} (2.00)  & \underline{115.76} (1.71)  & \underline{6.50} (0.19)   & \underline{-0.33} (0.49)  & 79.9 & \underline{29.3} & 53.0 & \underline{0.016} \\
    \rowcolor[HTML]{ECF4FF}\cellcolor{white}
    & \cellcolor{white}\textbf{Ours}  & \textbf{70.45} (1.92)     & \textbf{109.55} (1.36)     & \textbf{6.19} (0.12)      & \textbf{-1.18} (0.16)     & \textbf{84.9} & \textbf{23.9} & \textbf{58.0} & 0.019 \\ \midrule
    Living room & LayoutGPT           & 40.65                     & 142.34                    & 9.94                    & 22.93                    & 83.78 & 44.18 & 61.98 & 0.0348 \\ \cmidrule(lr){2-10}
                & ATISS               & 20.46 (1.41)              & 134.71 (2.75)              & 8.46 (0.22)               & 52.26 (3.93)              & \textbf{96.3} & 54.9 & \textbf{78.7} & 0.029 \\
                & DiffuScene          & 27.39 (1.57)              & 115.09 (1.62)              & 5.64 (0.13)               & 14.03 (2.11)              & \underline{93.4} & \underline{34.6} & \underline{74.5} & \textbf{0.014} \\
                & InstructScene       & \underline{47.97} (2.71)  & \underline{111.58} (1.07)  & \textbf{5.31} (0.06)      & \underline{9.30} (0.95)   & 93.1 & 36.0 & 71.0 & \underline{0.018} \\
    \rowcolor[HTML]{ECF4FF}\cellcolor{white}
    & \cellcolor{white}\textbf{Ours}  & \textbf{50.01} (2.25)     & \textbf{110.28} (1.18)     & \underline{5.49} (0.09)   & \textbf{6.18} (1.11)      & 94.3 & \textbf{26.5} & 71.8 & \underline{0.018} \\ \midrule
    Dining room & LayoutGPT           & 38.44                     & 155.04                    & 10.68                   & 20.43                    & 76.68 & 46.09 & 59.49 & 0.0301 \\ \cmidrule(lr){2-10}
                & ATISS               & 30.52 (3.01)              & 157.60 (2.07)              & 10.65 (0.31)              & 61.31 (4.38)              & \textbf{91.3} & 55.3 & \textbf{79.7} & 0.041 \\
                & DiffuScene          & 36.68 (2.14)              & 132.97 (0.70)              & 7.93 (0.23)               & 16.61 (1.53)              & \underline{87.9} & \underline{30.5} & \underline{73.8} & \textbf{0.011} \\
                & InstructScene       & \underline{46.54} (3.22)  & \underline{132.91} (1.37)  & \underline{7.64} (0.20)   & \underline{14.81} (1.87)  & 86.4 & 34.4 & 68.9 & 0.019 \\
    \rowcolor[HTML]{ECF4FF}\cellcolor{white}
    & \cellcolor{white}\textbf{Ours}  & \textbf{56.29} (2.47)     & \textbf{129.65} (1.68)     & \textbf{7.51} (0.17)      & \textbf{12.26} (0.99)     & 87.2 & \textbf{23.8} & 71.9 & \underline{0.017} \\ \bottomrule
    \end{tabular}
    }
    \vspace{-10pt}
\end{table}

\begin{table}[t]
    \centering
    \scriptsize
    \renewcommand{\arraystretch}{0.8}
    \setlength{\tabcolsep}{3pt}
    \caption{Additional comparison with recent state-of-the-art (FreeScene vs. Ours). FreeScene values are directly reported from the original paper \citep{bai2025freescene}. For direct comparability with prior settings, iRecall of Ours is additionally evaluated with up to two relational constraints ($N_{rel}\le2$).}
    \label{tab:concurrent_main}
    {\fontsize{8}{8.0}\selectfont
    \resizebox{0.6\linewidth}{!}{
    \begin{tabular}{@{}cccccc@{}}
        \toprule
         &  & iRecall ($\uparrow$) & FID ($\downarrow$) & FID$_\text{CLIP}$ ($\downarrow$) & KID$_{\times1e3}$ ($\downarrow$) \\
        \midrule
        Bedroom & FreeScene & 73.69 & 111.21 & 6.43 & 0.35 \\
    \rowcolor[HTML]{ECF4FF}\cellcolor{white}
         & \cellcolor{white}Ours & 77.96 & 109.55 & 6.19 & -1.18 \\
        \midrule
        Living room & FreeScene & 58.16 & 110.55 & 5.83 & 7.95 \\
    \rowcolor[HTML]{ECF4FF}\cellcolor{white}
         & \cellcolor{white}Ours & 59.56 & 110.28 & 5.49 & 6.18 \\
        \midrule
        Dining room & FreeScene & 63.39 & 127.28 & 8.01 & 14.83 \\
    \rowcolor[HTML]{ECF4FF}\cellcolor{white}
         & \cellcolor{white}Ours & 61.89 & 129.65 & 7.51 & 12.26 \\
        \bottomrule
    \end{tabular}
    }
    }
\end{table}

\subsection{3D Indoor Scene Synthesis from Natural Language}

\cref{table:main} reports quantitative results across three room categories. LayoutGPT provides an LLM-based reference point, while ATISS, DiffuScene, and InstructScene are data-driven baselines.
SceneNAT achieves the best iRecall and visual quality in all room types, while maintaining competitive physical plausibility under NAV, COL, ACC, and Floor Pen. metrics.
\cref{tab:concurrent_main} additionally compares with FreeScene using the metrics reported in the original paper, where SceneNAT remains competitive under the $N_{rel}\le2$ setting.

Compared with LayoutGPT, SceneNAT is substantially stronger on instruction adherence and rendered-scene quality, indicating that in-domain training remains important for text-conditioned indoor layout generation.
Among data-driven baselines, ATISS is physically plausible on some accessibility metrics but lacks language controllability, while DiffuScene and InstructScene improve generation quality at the cost of either slow inference or weaker relation handling.
SceneNAT offers a better balance by improving iRecall and FID/FID-CLIP while also reducing object collisions across all room types.

\cref{fig:rel_vs_irecall} further shows that SceneNAT maintains strong iRecall as relational complexity increases.
This robustness comes from our masked modeling strategy, which captures realistic scene distributions, and RRM, which injects relation-aware features learned through auxiliary triplet supervision.
The qualitative comparisons are presented in \cref{fig3}.

\noindent \textbf{Robustness to Unseen Relational Complexity}
To evaluate highly complex scenarios, we test instructions with 5 and 6 relational constraints, exceeding the maximum of 4 constraints used during training. We prune non-essential descriptive words to avoid text-encoder truncation while preserving core relational information.

As shown in \cref{fig:rel_vs_irecall}, SceneNAT maintains high iRecall on these unseen complexity levels, whereas InstructScene degrades more clearly as constraints increase.
This suggests that the auxiliary relational supervision in RRM helps the decoder generalize beyond the relation counts observed during training.

\begin{table}[t!]
    \centering
    \setlength{\tabcolsep}{5pt} 
    \scriptsize
    \caption{Comparison of computational costs. The reported TFLOPs are calculated by multiplying the TFLOPs per step by the number of sampling timesteps required for each model.}
    \label{table:cost}
    \begin{tabular}{lccccc} 
        \toprule
        \textbf{Model} & Inference time (s) & Params (M) & TFLOPs & FID & iRecall (\%) \\
        \midrule
        ATISS & 0.14 & 33.5 & 0.2 & 128.50 & 31.30 \\
        DiffuScene & 33.28 & 63.4 & 63.5 & 119.37 & 45.98\\
        InstructScene & 6.73 & 87.7 & 44.3 & 115.76 & 66.72 \\
        \midrule
        SceneNAT-S ($L_\text{scene}=4$) & 1.02 & 53.1 & 7.9 & \underline{110.91} & \underline{67.06} \\
        SceneNAT-B ($L_\text{scene}=8$) & 1.35 & 69.9 & 10.3 & \textbf{109.55} & \textbf{70.45}\\
        \bottomrule
    \end{tabular}
\end{table}

\begin{figure}[t]
\centering
\pgfplotsset{
    DiffuScene/.style={color=cDiff, mark=*, mark size=1.1pt, thick},
    InstructSceneA/.style={color=cInst, mark=*, mark size=1.1pt, thick},
    InstructSceneB/.style={color=cInst, mark=*, mark size=1.3pt, thick},
    Ours/.style={color=cOurs, mark=*, mark size=1.4pt, very thick},
    SceneNATB/.style={color=cOurs, mark=*, mark size=1.4pt, very thick},
    SceneNATS/.style={color=cOursS, mark=*, mark size=1.3pt, very thick, dashed},
}

\begin{minipage}[t]{0.48\textwidth}
\centering
\resizebox{\linewidth}{!}{%
\begin{tikzpicture}
    \begin{groupplot}[
        group style={
            group size=1 by 2,
            vertical sep=0.72cm,
        },
        width=6cm,
        height=3.1cm,
        xmode=log,
        log basis x=10,
        xmin=9, xmax=1100,
        xtick={10, 20, 30, 50, 100, 500, 1000},
        tick label style={font=\tiny},
        log ticks with fixed point,
        label style={font=\tiny},
        grid=major,
        grid style={dashed, gray!25},
        legend style={
            at={(0.5, 1.03)},
            anchor=south,
            legend columns=-1,
            font=\tiny,
            draw=none,
            fill=none,
        }
    ]

    \nextgroupplot[ylabel={FID}, ymin=90, ymax=200]
        \addplot [DiffuScene] coordinates {(10,185.95) (20,175.52) (30,167.47) (50,160.70) (100,148.23) (500,120.72) (1000,120.02)};
        \addplot [InstructSceneA] coordinates {(10,123.23) (20,114.98) (30,116.78) (50,116.45) (100,113.76)};
        \addplot [Ours] coordinates {(10,111.51) (20,108.94) (30,110.81) (50,108.21) (100,108.54)};
        \legend{DiffuScene, InstructScene, Ours}

    \nextgroupplot[ylabel={iRecall (\%)}, xlabel={Inference Steps}, ymin=10, ymax=80]
        \addplot [DiffuScene] coordinates {(10,11.25) (20,17.85) (30,20.78) (50,20.05) (100,26.16) (500,41.56) (1000,45.48)};
        \addplot [InstructSceneA] coordinates {(10,58.44) (20,63.57) (30,64.30) (50,64.55) (100,62.10)};
        \addplot [Ours] coordinates {(10,68.46) (20,69.93) (30,70.17) (50,71.12) (100,70.42)};
    \end{groupplot}
\end{tikzpicture}%
}
\subcaption{FID and iRecall vs.\ inference steps.}
\label{fig:inference_step}
\end{minipage}%
\hfill
\begin{minipage}[t]{0.5\textwidth}
\centering
\resizebox{\linewidth}{!}{%
\begin{tikzpicture}
    \begin{axis}[
        xlabel={Batch Size},
        ylabel={Latency (s)},
        xmode=log,
        log basis x=2,
        xtick={1, 8, 32, 64, 128},
        xmin=0.8, xmax=130,
        tick label style={font=\tiny},
        log ticks with fixed point,
        label style={font=\tiny},
        grid=major,
        grid style={dashed, gray!25},
        ymin=0.0, ymax=7.5,
        legend style={
            at={(0.5, 1.03)},
            anchor=south,
            legend columns=-1,
            font=\tiny,
            draw=none,
            fill=none,
        },
        width=6.5cm,
        height=5.65cm
    ]

    \addplot [InstructSceneB] coordinates {
        (1, 2.46) (8, 2.52) (32, 2.66) (64, 3.89) (128, 6.73)
    };
    \addplot [SceneNATB] coordinates {
        (1, 0.64) (8, 0.72) (32, 0.70) (64, 0.79) (128, 1.35)
    };
    \addplot [SceneNATS] coordinates {
        (1, 0.49) (8, 0.51) (32, 0.52) (64, 0.62) (128, 1.02)
    };
    \legend{InstructScene, SceneNAT-B, SceneNAT-S}

    \end{axis}
\end{tikzpicture}%
}
\subcaption{Inference latency vs.\ batch size.}
\label{fig:latency_scaling}
\end{minipage}

\caption{Scaling behavior of the proposed method. (a) Generative quality and
instruction adherence as a function of the number of inference steps.
(b) Inference latency (in seconds) as a function of batch size.}
\label{fig:scaling}
\end{figure}

\subsection{Computational Efficiency}

\cref{table:cost} summarizes the computational efficiency of SceneNAT against key baselines on the bedroom dataset.
For clarity, SceneNAT-B is our full model, and SceneNAT-S is its smaller variant that uses only half of the scene decoder layers.
For fair comparison, inference time and TFLOPs are measured at batch size 128 over 50 iterations after warmup.
While ATISS is fast, its generative quality and instruction adherence are significantly weaker.
In contrast, the superiority of SceneNAT is clearly evidenced in comparison with the powerful diffusion-based models.
As illustrated in \cref{fig:inference_step}, we achieve superior scene quality and instruction adherence in just 30 steps, whereas diffusion models such as DiffuScene and InstructScene require 1,000 and 110 (100+10) steps, respectively.
This efficiency leads to a dramatic reduction in inference time, making our model approximately 5x faster than InstructScene and over 24.7x faster than DiffuScene.
Furthermore, this efficiency is achieved with over 20\% fewer parameters than InstructScene, requiring 4.3x lower TFLOPs than InstructScene and 6x lower than DiffuScene.
Notably, SceneNAT-S also surpasses the performance of InstructScene with 6.6x faster inference speed and 5.6x lower TFLOPs.
This advantage holds across throughput regimes, as shown in \cref{fig:latency_scaling}.
Even at batch size 1, SceneNAT-S (0.49s) is roughly $5\times$ faster than InstructScene (2.46s), and the gap widens as scenes are processed in parallel: at batch size 128, InstructScene's latency surges to 6.73s while SceneNAT-B reaches only 1.35s.
Together, these findings highlight the effectiveness of our non-autoregressive parallel decoding for instruction-guided scene synthesis, and its scalability for real-world, high-demand deployment.

\vspace{-5pt}
\subsection{Generalization Capability}
\begin{figure}[t!]
\centering
\includegraphics[page=2,width=0.80\textwidth,trim=76.75bp 320.83bp 72.86bp 35.86bp,clip]{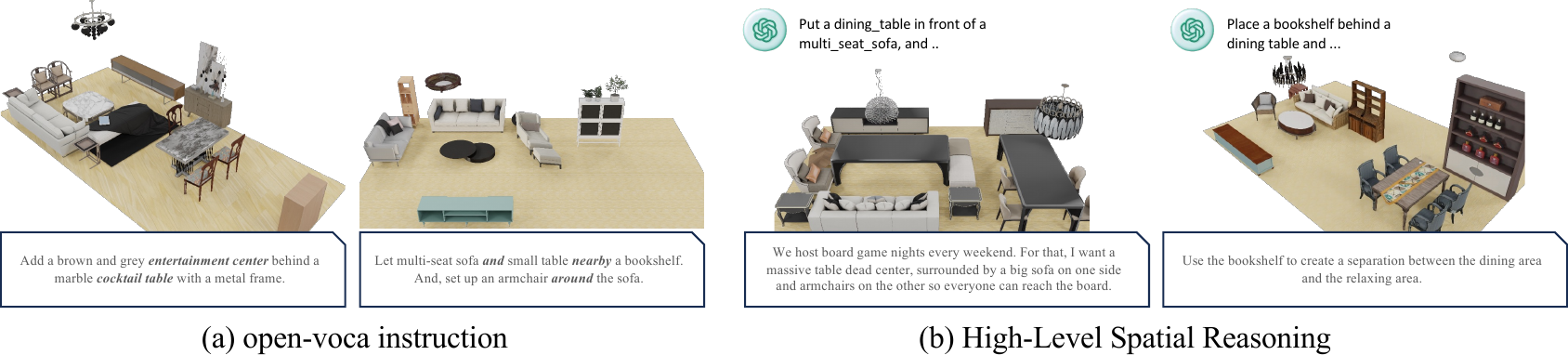}
\caption{Generalization results on unseen instructions. SceneNAT handles unseen object/relation expressions and LLM-assisted rewritten prompts.}
\label{fig6}
\vspace{-14pt}
\end{figure}

\begin{figure}[t]
\centering
\includegraphics[page=3,width=0.80\columnwidth,trim=131.75bp 166.61bp 280.50bp 179.14bp,clip]{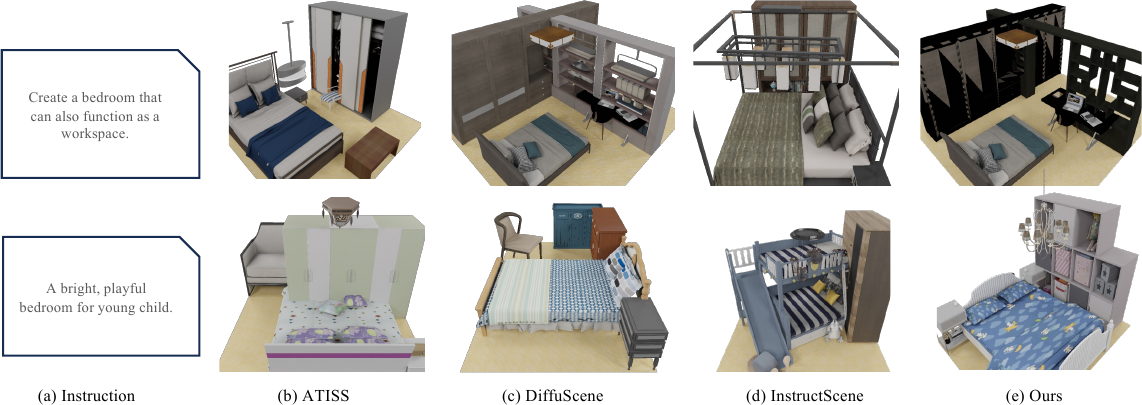}
\caption{Additional qualitative comparisons on unseen ``in-the-wild'' prompts.}
\label{fig6_additional}
\vspace{-12pt}
\end{figure}

Robustness to free-form and ``in-the-wild'' instructions is essential for practical language-conditioned scene generation.
Although SceneNAT is trained on templated instructions, it generalizes well to unseen lexical and semantic patterns.
In \cref{fig6}(a), the model correctly handles unseen object/relation expressions, and \cref{fig6_additional} shows similarly stable object selection and spatial arrangement across diverse open-vocabulary prompts.
We also evaluate an LLM-assisted inference mode in which a general-purpose LLM rewrites raw prompts into structured relational instructions; as shown in \cref{fig6}(b), this lightweight preprocessing improves robustness on long, functional, and abstract inputs while preserving layout consistency.
Together, these results indicate two complementary generalization paths, direct CLIP-based semantic transfer and lightweight LLM-based instruction rewriting for unconstrained prompts.

\subsection{Ablation Study}
\label{sec:ablation}
To validate the effectiveness of our model's key components, we conduct a series of ablation studies.
We systematically remove or alter parts of our architecture and training strategy to observe the impact on performance. The results are summarized in \cref{tab:ablation}.

\noindent \textbf{Relational Reasoning Module}
RRM is fundamental to our model's ability to reason about relational structures. Removing this module causes a significant decrease in iRecall.
This highlights its critical role in translating high-level semantic constraints into a coherent and accurate spatial arrangement of objects.
Further validation of our design choice is detailed in \cref{abstractor}.

\noindent \textbf{Full Relation Supervision}
We also evaluate full relation supervision, a method similar to the entire scene graph approach used in InstructScene.
Our findings suggest that it is more beneficial to supervise RRM primarily on the relations explicitly provided in the input, rather than on an entire scene graph.
Our approach allows the scene decoder to more effectively manage the natural layout and implicit relations, leading to higher iRecall and more globally consistent scenes.

\noindent \textbf{Masking Strategy}
The masking strategy is another cornerstone of our approach.
The replace-and-remasking is essential for stable non-autoregressive training, and removing it severely degrades generation quality.
Our approach further decomposes this strategy into token-level and object-level masking.
While both strategies contribute to a better FID, object-level masking is more critical for improving iRecall.
The full model utilizing both strategies achieves the best performance across both FID and iRecall.

\noindent \textbf{Discretization Granularity}
We vary the number of bins for layout attributes ($t, l, \theta$) from 32 to 256 and observe that SceneNAT remains robust across granularities (\cref{tab:bin_ablation}).
Coarser bins slightly reduce iRecall, while denser settings bring only marginal gains; we therefore use 64 bins as a practical balance between generation quality and search-space complexity.

\begin{table}[t]
    \centering
    \begin{minipage}[t]{0.475\textwidth}
        \centering
        \scriptsize
        \setlength{\tabcolsep}{3pt}
        \captionof{table}{Ablation studies on key components of the SceneNAT architecture.}
        \label{tab:ablation}
        \resizebox{\linewidth}{!}{
        \begin{tabular}{p{0.54\linewidth}cc}
            \toprule
            \textbf{Configuration} & FID & iRecall (\%) \\
            \midrule
            Full relation supervision & 110.76 & 65.16 \\
            w/o replace/remasking & 191.69 & - \\
            w/o RRM & 110.91 & 62.77 \\
            w/o object masking & 110.06 & 66.11 \\
            w/o token masking & 111.20 & 68.26 \\
            \textbf{Ours (Full)} & \textbf{109.55} & \textbf{70.45} \\
            \bottomrule
        \end{tabular}
        }
    \end{minipage}
    \hfill
    \begin{minipage}[t]{0.475\textwidth}
        \centering
        \scriptsize
        \setlength{\tabcolsep}{3pt}
        \captionof{table}{Ablation study on spatial discretization granularity.}
        \label{tab:bin_ablation}
        \resizebox{\linewidth}{!}{
        \begin{tabular}{lccccc}
            \toprule
            & \# of Bins & 32 & 64 & 128 & 256 \\
            \midrule 
            Bedroom     & iRecall ($\%$) & 70.17 & \textbf{70.45} & 69.93 & 68.02 \\
                        & FID    & 110.28 & 109.55 & 109.69 & \textbf{108.56} \\
            \midrule
            Living room & iRecall ($\%$) & 49.19 & 50.01 & 50.80 & \textbf{52.60} \\
                        & FID   & 109.30 & 110.28 & \textbf{108.98} & 110.41 \\
            \midrule
            Dining room & iRecall ($\%$) & 53.36 & 56.29 & 56.02 & \textbf{59.52} \\
                        & FID   & 130.70 & 129.65 & \textbf{127.72} & 129.92 \\
            \bottomrule
        \end{tabular}
        }
    \end{minipage}
\end{table}

\section{Conclusion}
We introduced SceneNAT, a masked non-autoregressive transformer for controllable 3D indoor scene generation. Its key design combines masked indoor scene modeling (MISM) with a dedicated RRM for complex relational instructions, trained with auxiliary triplet supervision and used at inference through relation-aware features only. This design consistently outperforms strong diffusion-based baselines while reducing computational cost. These results support masked generative modeling as a practical direction for scalable 3D scene synthesis, with future extensions to multimodal inputs and larger, relation-rich environments.

\section*{Broader Impact Statement}
We acknowledge that our framework, by utilizing pretrained models (e.g., CLIP~\citep{radford2021learning}, OpenShape~\citep{liu2023openshape}), may inherit societal biases from broad and unfiltered web data.
However, these potential risks are inherently limited. This is primarily due to our focus on the low-risk domain of indoor scenes, the use of public and non-sensitive training data, and a model architecture that prioritizes structural plausibility over unconstrained generation.
In line with our commitment to research integrity, we will release our work under a restrictive license to prevent misuse.

\section*{Reproducibility Statement}
We provide comprehensive details in the Appendix to ensure the reproducibility of our work. We offer a full breakdown of our model's architecture, scene representation, masking strategy, and the complete training procedure, including all hyperparameters in Appendix~\ref{implementation}. Additionally, we describe our methodology for generating an enhanced instruction dataset with more complex relations in Appendix~\ref{prompting} and detail our re-implementation of baselines to ensure a fair comparison in Appendix~\ref{baseline}.
\ifnum\ispreprint=1
  The source code is publicly available at \url{https://github.com/lojol2327/SceneNAT-official}.
\else
  The source code is included in the supplementary material.
\fi

\bibliographystyle{tmlr}
\bibliography{main}

\appendix
\providecommand{\trainref}{\cref{train}}
\section{More Details} \label{implementation}

\subsection{Implementation Details}

SceneNAT is based on the standard transformer architecture, utilizing the SiLU activation function~\citep{ramachandran2017searching} in its feed-forward networks.
\textcolor{blue}{For the main experiments, we set the common hyperparameters to an attention dimension of 512, 8 attention heads, and a dropout rate of 0.1. The decoding components include 2 RRM layers, 2 appearance decoder layers, and 1 layout decoder layer. The two model variants are distinguished by the scene decoder depth: SceneNAT-B utilizes 8 scene decoder layers, compared to 4 layers for SceneNAT-S.
We set the number of learnable triplet queries, $N_q$, to 4.}

\paragraph{Dataset}
We use the dataset filtered by InstructScene \citep{lin2024instructscene}, including 4,041 bedrooms, 813 living rooms, and 900 dining rooms.
For bedrooms, the number of objects per scene ranges from 3 to 12, across 21 distinct object categories.
In contrast, for living and dining rooms, the object count per scene  varies from 3 to 21, with a larger set of 24 object categories.
More detailed information can be found in InstructScene paper.

\paragraph{Scene Representation}
Object shape features are extracted using a pre-trained VQ-VAE~\citep{van2017neural} from the 3D-FRONT~\citep{fu20213d_front} dataset, following \citet{lin2024instructscene}. To prepare the data for our transformer-based model, we discretize continuous spatial attributes: object positions and sizes are quantized into 64 bins, while orientations are discretized into intervals of 10 degrees.

\paragraph{VQ-VAE Training}
For VQ-VAE used in the scene representation above, frozen OpenShape encoder (PointBERT-ViT-G14)~\citep{liu2023openshape} is adopted to extract 1280-dimensional semantic features from 3D point clouds. These features are processed by a Transformer-based encoder-decoder architecture with a vector-quantization bottleneck. A codebook of size 64 with 512-dimensional vectors encodes each object into a sequence of 4 ordered discrete tokens, following InstructScene.
Training this VQ-VAE module requires approximately 10 hours on a single NVIDIA RTX A5000 GPU using the 4,000+ objects from the filtered 3D-FRONT dataset.

\paragraph{Object Retrieval}
After iterative sampling, we perform a nearest-neighbor retrieval to select the best-matching 3D asset for each object to construct a scene.
First, we filter the asset database (3D-FUTURE~\citep{fu20213d_future}) to select candidate objects that match the class label predicted by our Class Head. 
Then, the sequence of 4 discrete appearance tokens generated by the Appearance Head is mapped back to their corresponding codebook vectors and decoded through the pre-trained VQ-VAE decoder to reconstruct a continuous 1280-dimensional semantic feature. 
Finally, we compute the cosine similarity between this reconstructed feature and the pre-computed OpenShape features of all candidate assets, and the asset with the highest similarity is retrieved and placed into the scene.

\paragraph{Masking Strategy}
Our masking strategy is inspired by the methodologies of MaskGIT~\citep{chang2022maskgit} and BERT~\citep{devlin2019bert}. Diverging from BERT's fixed 15\% masking ratio, we employ a dynamic masking ratio determined by a cosine schedule, which varies the proportion of masked tokens at each training step. The masking is applied at both an object-level and a token-level. Initially, a portion of the total masking budget is used to mask entire objects. The remaining budget is then met by randomly masking individual tokens from the objects that were not previously masked.
For each token selected for masking, the following procedure is applied: 10\% are substituted with a random token from the vocabulary. Of the remaining 90\%, 88\% are replaced with a special \texttt{[MASK]} token, and the rest are left unchanged with their original values.

\subsection{Training}

\paragraph{Setup}
We implement our model using the PyTorch~\citep{paszke2019pytorch} framework.
All experiments were conducted on a single NVIDIA RTX A5000 GPU.

\paragraph{Optimization}
We use the AdamW~\citep{loshchilov2017decoupled} optimizer with a learning rate of 1e-4 and a weight decay of 0.02.
We employ a cosine annealing scheduler with a linear warmup for the initial 50 epochs, after which the learning rate is decayed to a minimum of 1e-5.
All models are trained for 2000 epochs with a batch size of 256.
To ensure stability, we apply gradient clipping with a maximum norm of 1.0.
The coefficients for the triplet loss components ($\lambda_s$, $\lambda_p$, and $\lambda_o$), as well as the reconstruction loss components ($\lambda_x$, $\lambda_v$, $\lambda_t$, $\lambda_l$, and $\lambda_\theta$), are all set to 1.0.
We found that a triplet loss weight of $\lambda_{\text{triplet}}=1.0$ yielded the best performance on our validation set after exploring values of 0.2, 0.5, 1.0, and 2.0.
Also, we set the weight for null triplet to 0.1.
After training, we select the model checkpoint with the best performance on our validation set for all evaluations.

\paragraph{Classifier-Free Guidance (CFG)}
To improve instruction fidelity without sacrificing diversity, we incorporate classifier-free guidance~\citep{ho2022classifier, wang2024cfgschedulers, bradley2025cfgpc} into the training phase.
During training, we randomly remove the conditioning input by nulling out CLIP \citep{radford2021learning} embedding of instruction \( I \) with probability \( p = 0.1 \), enabling the model to learn both conditional and unconditional generation.
At inference, guidance is applied by linearly combining conditional and unconditional logits \( \omega_c, \omega_u \) before softmax:
\begin{equation}
\omega = (1 + \rho) \cdot \omega_c - \rho \cdot \omega_u,
\end{equation}
where \( \rho \) denotes the guidance scale.

\subsection{Baseline Implementation} \label{baseline}
To ensure a fair comparison under our redefined problem setup, we reimplement and retrain all baselines from scratch using our refined dataset annotations (see \cref{prompting}).
For our work, we adapt three key baselines, ATISS~\citep{paschalidou2021atiss}, DiffuScene~\citep{tang2024diffuscene}, and InstructScene~\citep{lin2024instructscene}.
We modify both ATISS and DiffuScene to predict discrete VQ-based appearance features and condition on CLIP~\citep{radford2021learning} text embeddings.
For ATISS, we replace the original global context token (a room mask feature) with a CLIP instruction embedding, while DiffuScene utilizes the text-based cross-attention from its original implementation.
We also retrain InstructScene on our refined dataset using its original configuration, as its architecture does not require modifications.
For training, we follow the optimization procedures, training steps, and hyperparameter configurations from the respective papers. We evaluate all baselines under the same experimental settings to ensure a fair comparison with our method.

\subsection{Spatial Relation Definitions} \label{rel_def}

We adopt the set of 11 geometric relationship definitions from InstructScene~\citep{lin2024instructscene} to describe the spatial arrangement of objects. These definitions are based on the properties of object bounding boxes, assuming the XY-plane represents the ground and Z is the vertical axis. Let $s$ denote a ``subject'' and $o$ an ``object'' in a relationship. The definitions rely on three core metrics:
\begin{itemize}
    \item Relative Orientation ($\theta_{so}$): The angle between the subject and object on the ground plane, computed as $\theta_{so} = \text{atan2}(Y_s - Y_o, X_s - X_o)$.
    \item Ground Distance ($d(s, o)$): The Euclidean distance between the centers of $s$ and $o$ projected onto the XY-plane.
    \item Overlap Predicate ($\text{inside}(s, o)$): A boolean function that is true if the subject's center lies within the 2D projection of the object.
\end{itemize}

Based on these metrics, the relationships are determined by the following rules:

\paragraph{Horizontal Relationships}
Defined by ground distance and relative orientation, categorized into two ranges.
For objects at a medium distance ($1 < d(s, o) \le 3$), the four primary directional relationships are defined as:
\begin{itemize}
    \item Right of: $-\frac{\pi}{4} \le \theta_{so} < \frac{\pi}{4}$.
    \item In front of: $\frac{\pi}{4} \le \theta_{so} < \frac{3\pi}{4}$.
    \item Left of: $\theta_{so} \ge \frac{3\pi}{4}$ or $\theta_{so} < -\frac{3\pi}{4}$.
    \item Behind: $-\frac{3\pi}{4} \le \theta_{so} < -\frac{\pi}{4}$.
\end{itemize}
For objects in close proximity ($d(s, o) \le 1$), four corresponding `closely' variants (e.g., closely right of) apply. These use the exact same angular constraints as their primary counterparts.

\paragraph{Vertical Relationships}
\begin{itemize}
    \item Above: A subject $s$ is above an object $o$ if
    \[
    \begin{aligned}
    (Center_{Z_s} - Center_{Z_o}) &> \frac{Height_s + Height_o}{2}, \\
    \text{and}\quad \text{Inside}(s, o) &\lor \text{Inside}(o, s).
    \end{aligned}
    \]
    \item Below: A subject $s$ is below an object $o$ if the inverse condition is met:
    \[
    \begin{aligned}
    (Center_{Z_o} - Center_{Z_s}) &> \frac{Height_s + Height_o}{2}, \\
    \text{and}\quad \text{Inside}(s, o) &\lor \text{Inside}(o, s).
    \end{aligned}
    \]
\end{itemize}

\paragraph{Null Relationship}
\begin{itemize}
    \item None: No direct spatial relationship exists if the ground distance is greater than 3 units: $d(s, o) > 3$.
\end{itemize}

\begin{table}[t]
\centering
\caption{\textcolor{blue}{Ablation studies on the architectural design of our model, comparing the number of transformer layers in RRM and the use of relational attention without bipartite matching. Performance is evaluated using Fréchet Inception Distance (FID) and instruction Recall (iRecall).}}
\label{tab:design_choice}
\small
\begin{tabular}{ccc}
\toprule
\# of layers & FID $\downarrow$ & iRecall $\uparrow$ \\
\midrule
w/ relational attention & 110.05 & 65.04 \\
\midrule
\textcolor{blue}{w/o RRM} & 110.91 & 62.77 \\
1 & 110.06 & 65.87 \\
2 & \textbf{109.55} & \textbf{70.45} \\
3 & 110.48 & 68.50 \\
4 & 109.80 & 67.54 \\
\bottomrule
\end{tabular}
\end{table}

\subsection{Design Choice for Relation Modeling} \label{abstractor}
To further validate our design choices for relation modeling, we conduct two analyses.

\paragraph{\textcolor{blue}{Ablation on RRM Layers}}
\textcolor{blue}{We report an ablation study on the performance impact of the number of layers in RRM. As shown in \cref{tab:design_choice}, all configurations with this module outperform the baseline without it, demonstrating its overall effectiveness. The best performance is achieved when using two transformer layers in terms of both FID and iRecall.}

\paragraph{Comparison with Relational Attention (RA)}
We present a comparative experiment to demonstrate that learning separate subject, predicate, and object embeddings via triplet-based bipartite matching is more effective than using a combined triplet embedding.
\textcolor{blue}{To provide a fair comparison, we replace RRM, which is built on basic transformer blocks, with a specialized transformer architecture that uses relational attention (RA), as proposed by \citet{altabaa2023abstractors, altabaa2024disentangling}.}
The RA is a module specifically designed to learn relational features disentangled from sensory information.

The RA's operation is based on two key components, slot attention and relational attention. First, slot attention maps each object's feature vector $z$ obtained from the scene decoder to a set of learned slots that represent its semantic role. This is defined by the following equation:
\begin{equation}
\begin{aligned}
\operatorname{SlotAttention}(z) &= \operatorname{concat}\!\left(s^{(1)}, \ldots, s^{(n_h)}\right), \\
\text{where} \quad
s^{i} &= \operatorname{Softmax}\!\left((z W_q^{i})(F_{\text{lib}}^{i})^\top\right) S_{\text{lib}}^{i}.
\end{aligned}
\end{equation}
Here, $F_{\text{lib}}$ and $S_{\text{lib}}$ each denotes a set of learnable feature templates and slot vectors.
The resulting slot features $s$ are used in a relational attention head to compute interactions between objects, passing messages along fixed relational channels. Formally, this process is defined by the following equation:
\begin{equation}
\begin{aligned}
\operatorname{RelAttn}(z)
&= \sum_{i=1}^{n} A_i(z)\left(r_i W_r + s_i W_s\right), \\
A_i(z)
&= \operatorname{Softmax}\!\left((z W_q^{i})(z W_k^{i})^\top\right).
\end{aligned}
\end{equation}
This dual-attention approach explicitly encodes relational information between objects, effectively disentangling it from object-specific priors.

\textcolor{blue}{Unlike RRM, which is trained with auxiliary supervision to decode explicit symbolic constraints and learn separate embeddings for the subject, predicate, and object via triplet-based bipartite matching, the transformer with RA is trained to predict a single, combined triplet embedding.}
The output of the trained transformer then serves as a conditioning signal for the layout decoder.
\textcolor{blue}{As shown in \cref{tab:design_choice}, RRM achieves superior performance on iRecall, validating our design choice.}

\subsection{Ablation Study on Classifier-Free Guidance Scales} \label{cfg}

To analyze the impact of the classifier-free guidance (CFG) scale on the trade-off between instruction fidelity and generation quality, we give an ablation study by varying the scale $\rho$ from $0.0$ to $10.0$. The results are summarized in \cref{tab:cfg_ablation}.

We observe a distinct trade-off across the guidance scales.
While unconditional generation ($\rho=0$) expectedly yields poor instruction adherence, increasing the scale up to $2.0$ reveals a clear trade-off where iRecall improves at the cost of a slight degradation in visual quality. However, increasing the scale beyond $2.0$ leads to performance drops in both metrics.
Based on these results, we set $\rho=1.0$ as our default setting to maintain the balance between high-quality generation and faithful instruction following.

\begin{table}[t]
    \centering
    \caption{Ablation study on classifier-free guidance scales ($\rho$). The best results for each metric are highlighted in \textbf{bold}.}
    \label{tab:cfg_ablation}
    \resizebox{\linewidth}{!}{
    \begin{tabular}{lccccccccc}
        \toprule
        Guidance Scale ($\rho$) & 0.0 & 0.5 & 1.0 & 1.5 & 2.0 & 3.0 & 4.0 & 5.0 & 10.0 \\
        \midrule
        iRecall ($\uparrow$) & 11.98 & 66.26 & 70.45 & 71.39 & \textbf{72.62} & 68.22 & 63.81 & 66.99 & 43.52 \\
        FID ($\downarrow$) & 128.17 & 111.61 & \textbf{109.55} & 110.10 & 111.49 & 113.25 & 116.99 & 120.15 & 133.45 \\
        \bottomrule
    \end{tabular}
    }
\end{table}


\subsection{Quantitative Analysis on Physical Plausibility}
\label{sec:collision_analysis}
To evaluate the physical plausibility of the generated scenes, we conduct an additional quantitative evaluation focusing on object collisions. To ensure precise detection, intersections were computed using oriented bounding boxes. We employed three specific metrics to capture different aspects of physical plausibility.
\paragraph{Total Intersection Volume ($\mathcal{V}_{\cap}^{\text{sum}}$)} This represents the aggregate sum of intersection volumes for all object pairs in the scene. It quantifies the absolute magnitude of physical interference, serving as a global measure of layout validity.
\paragraph{Average Intersection Volume ($\mathcal{V}_{\cap}^{\text{avg}}$)} This metric calculates the mean intersection volume across all colliding object pairs. It serves as an indicator of the average collision density within a generated scene.
\paragraph{Intersection-over-Minimum ($\text{IoMin}$)} Defined as the average ratio of the intersection volume to the volume of the smaller object in a pair ($\frac{V_{\cap}}{\min(V_A, V_B)}$), this metric captures the \textit{degree} of collisions. Unlike standard IoU, $\text{IoMin}$ effectively handles size imbalance, such as detecting when a small object is completely submerged within a larger one.

\begin{table}[t]
    \centering
    \caption{Quantitative comparison of physical plausibility. We report the average intersection volume ($\mathcal{V}_{\cap}^{\text{avg}}$), total intersection volume ($\mathcal{V}_{\cap}^{\text{sum}}$), and IoMin Score. \textbf{Bold} indicates the best result, and \underline{underline} indicates the second best. Lower values are better ($\downarrow$) for all metrics.}
    \label{tab:collision_metrics}
    \resizebox{0.6\linewidth}{!}{
    \begin{tabular}{lcccc}
        \toprule
        \textbf{Room Type} & \textbf{Method} & $\mathcal{V}_{\cap}^{\text{sum}}$ ($\downarrow$) & $\mathcal{V}_{\cap}^{\text{avg}}$ ($\downarrow$) & IoMin ($\downarrow$) \\
        \midrule
        Bedroom     & ATISS         & 241.25            & 0.041              & 0.384            \\
                    & DiffuScene    & 93.26             & \textbf{0.012}     & \textbf{0.227}    \\
                    & InstructScene & \underline{79.62} & 0.018              & \underline{0.242} \\
                    & \textbf{Ours} & \textbf{69.58}    & \underline{0.017}  & 0.257 \\
        \midrule
        Living Room & ATISS         & 472.68            & 0.008             & 0.193 \\
                    & DiffuScene    & 249.61            & \underline{0.007}             & 0.158 \\
                    & InstructScene & \underline{204.91}& \textbf{0.006} & \textbf{0.152} \\
                    & \textbf{Ours} & \textbf{151.24}   & \underline{0.007}    & \underline{0.153} \\
        \midrule
        Dining Room & ATISS         & 367.71             & 0.011             & 0.259 \\
                    & DiffuScene    & 240.46             & \underline{0.010} & \underline{0.218} \\
                    & InstructScene & \underline{187.95} & \underline{0.010} & 0.222 \\
                    & \textbf{Ours} & \textbf{169.31}    & \textbf{0.008}    & \textbf{0.207} \\
        \bottomrule
    \end{tabular}
    }
\end{table}

The quantitative comparisons presented in \cref{tab:collision_metrics} indicate that while baselines such as DiffuScene and InstructScene achieve competitive performance in certain metrics, SceneNAT consistently demonstrates a strong capability in minimizing overall physical violations. Most notably, our model achieves the lowest $\mathcal{V}_{\cap}^{\text{sum}}$ across all three room types. This suggests that the global context modeling of SceneNAT is particularly effective at reducing the aggregate magnitude of inter-object collisions within a scene.
It is noteworthy that SceneNAT achieves these results without explicit collision supervision, whereas DiffuScene incorporates an IoU loss during training. This demonstrates that our masked generative approach effectively learns to adhere to physical constraints directly from the data distribution.

\section{Instruction Prompting} \label{prompting}
We adapt the dataset introduced by InstructScene, which augments 3D-FRONT \citep{fu20213d_front} and 3D-FUTURE \citep{fu20213d_future} with relation annotations, object captions, and filtered natural descriptions. While we adopt the same preparation pipeline, we refine the instruction generation stage to support richer multi-relation prompts. Instruction generation converts annotated subject-predicate-object relations into natural-language prompts by randomly sampling a subset of relations and filling them into textual templates. InstructScene limits each prompt to at most two relations, whereas in our setup we extend this to a maximum of four to reflect more complex guidance. To further refine the generated instructions, we incorporate (i) filtering of redundant symmetric pairs, (ii) discourse-aware ordering so that successive relations preferentially reuse previously mentioned objects, and (iii) regulation of referring expressions (e.g., ``the'' vs. ``another''). These changes yield instructions that are longer, more coherent, and better aligned with the multi-relation setting of our experiments, while preserving compatibility with the relational annotations provided by InstructScene.

\section{Limitations and Future Work}

While our framework demonstrates strong performance, we identify several limitations for future research.
\textcolor{blue}{First, the scalability of RRM to a larger number of simultaneous relational constraints has yet to be explored.}
Second, our model relies on a uniform discretization for layout variables ($t, l, \theta$), and the optimality of this fixed quantization for representing continuous 3D space is not strictly guaranteed.
Third, the framework involves a number of hyperparameters governing both the training and inference process (e.g., remasking ratios, guidance scale), which require careful tuning.
Finally, SceneNAT is trained and evaluated exclusively on synthetic assets and relies on a closed retrieval library. We note that this reliance on synthetic data is a common constraint shared by most recent state-of-the-art baselines in this domain, limiting direct applicability to real-world environments.

These limitations also highlight promising directions.
To address hyperparameter sensitivity, one could explore alternating optimization techniques, proposed in AutoNAT~\citep{ni2024revisiting}, for automatically tuning key hyperparameters such as the remasking ratio and guidance scale.
Furthermore, a significant extension would be to move beyond the current retrieval-based approach. By integrating a generative mesh decoder, our framework could be enhanced to synthesize novel object geometries from scratch conditioned on text instruction, rather than relying solely on a predefined library of shapes.
In addition, incorporating visual or structural cues from real environments such as partial scans, images, or coarse floorplans as additional conditioning signals could open up a path toward more grounded and real-world-aware scene generation.

\section{User Study}
\label{sec:user_study}

We conducted a user study involving 42 participants to assess the perceptual quality and instruction adherence of the generated scenes.
We presented 18 sets of qualitative comparisons where each set displayed the text instruction alongside scenes generated by SceneNAT and three baselines including ATISS, DiffuScene, and InstructScene.

Participants were asked to rank the 1st and 2nd results based on a holistic evaluation of two criteria: (1) \textit{Instruction Alignment} which measures how faithfully the scene adheres to the spatial and semantic constraints described in the text, and (2) \textit{Overall Quality} which assesses the visual plausibility and aesthetic coherence of the layout.
The results are summarized in \cref{fig:user_study}. SceneNAT was selected as the most preferred method in 54.8\% of the cases which outperforms the strongest baseline, InstructScene (38.1\%).

\begin{figure}[t]
    \centering
    \begin{tikzpicture}
        \begin{axis}[
            ybar stacked,   
            bar width=15pt,
            width=0.6\linewidth,
            height=4.5cm,
            ymin=0, ymax=100,
            ylabel={User Preference (\%)},
            symbolic x coords={SceneNAT, InstructScene, DiffuScene, ATISS},
            xtick=data,
            xticklabel style={text height=1.5ex, font=\scriptsize, rotate=20, anchor=east},
            legend style={
                at={(0.5,1.25)}, 
                anchor=north, 
                legend columns=-1, 
                draw=none, 
                /tikz/every even column/.append style={column sep=0.5cm}
            },
            grid=major,
            grid style={dashed, gray!20},
            axis on top
        ]

        \addplot[fill=NavyBlue, draw=none, text=white] coordinates {
            (SceneNAT, 54.8)
            (InstructScene, 38.1)
            (DiffuScene, 7.1)
            (ATISS, 0.0)
        };

        \addplot[fill=CornflowerBlue!60, draw=none, text=black] coordinates {
            (SceneNAT, 35.7)
            (InstructScene, 50.0)
            (DiffuScene, 9.5)
            (ATISS, 4.8)
        };

        \legend{\textbf{1st}, \textbf{2nd}}
        
        \end{axis}
    \end{tikzpicture}
    \caption{The user study across different methods. The stacked bars indicate the percentage of times each model was selected as the 1st and 2nd choice by 42 human evaluators with 18 comparison sets. SceneNAT shows a distinct advantage in capturing user intent and generation quality.}
    \label{fig:user_study}
\end{figure}

\begin{figure}[t]
    \centering
    \footnotesize
    \begin{minipage}{0.92\linewidth}
    \begin{tcolorbox}[
        colback=blue!3!white,
        colframe=blue!60!black,
        coltitle=white,
        title=\textbf{LLM Prompt for Spatial Reasoning},
        arc=1mm,
        boxrule=0.5pt,
        fonttitle=\bfseries
    ]
    
    \textbf{System:} You are an expert spatial reasoning assistant that converts implicit natural language descriptions of indoor scenes into explicit spatial relational constraints.
    
    \vspace{2mm}
    
    STRICT GUIDELINES:
    \begin{enumerate}
        \item You MUST ONLY use the following predefined categories. Do NOT invent any other objects or relations.
        \begin{itemize}
            \item \textit{Object types:} armchair, bookshelf, cabinet, ceiling lamp, chaise longue sofa, chinese chair, coffee table, console table, corner side table, desk, dining chair, dining table, l shaped sofa, lazy sofa, lounge chair, loveseat sofa, multi seat sofa, pendant lamp, round end table, shelf, stool, tv stand, wardrobe, wine cabinet
            \item \textit{Spatial relations:} above, left of, in front of, closely left of, closely in front of, below, right of, behind, closely right of, closely behind
        \end{itemize}
        
        \item Do NOT write explanatory text, conversational filler, or complex grammar (e.g., do not use words like "between", "to create", or "because"). The output MUST strictly be a sequence of simple pairs in the exact format separated by commas or "and": \\
        \texttt{"Place a [Object A] [Relation] a [Object B]"}
    \end{enumerate}
    \vspace{2mm}
    Example Input: I want to watch TV comfortably from the couch, but make sure the coffee table isn't blocking the pathway.
    Example Output: Place a multi seat sofa in front of a tv stand, and place a coffee table closely right of the multi seat sofa.
    \vspace{2mm}
    Process the following instructions and return exactly one rewritten instruction per line.
    \\
    \\
    \textbf{User:} Use the bookshelf to create a separation between the dining area and the relaxing area.
     \\
    \\
    \textbf{Assistant:} Place a bookshelf behind a dining table and place a bookshelf in front of a multi seat sofa.
    \end{tcolorbox}
    \end{minipage}
    \caption{Example of the LLM-assisted spatial reasoning prompt. The LLM is strictly constrained to use predefined object types and spatial relations to decompose complex instructions for integration with the SceneNAT generation pipeline.}
    \label{fig:llm_prompt}
\end{figure}

\section{Zero-shot Downstream Tasks}
\label{sec:supp_downstream}

Following InstructScene, we evaluate zero-shot downstream tasks including stylization, re-arrangement, completion, and unconditional generation, and further introduce \textit{layout-to-object} to test bidirectional context modeling.
In layout-to-object, the model predicts the most suitable object attributes ($x,v$) for a given 3D box ($t,l,\theta$). We report InstructScene's stylization score $\Delta$ and summarize all results in \cref{downstream}.

SceneNAT is consistently competitive with or better than InstructScene, and clearly outperforms ATISS and DiffuScene on most settings.
For layout-to-object, ATISS and InstructScene are not applicable due to their directional/cascaded generation pipelines; against the only applicable baseline (DiffuScene), SceneNAT shows substantially better FID in all room types.
\cref{downstream} reports the complete quantitative comparison for all downstream tasks and room types.

\begin{table}[t]
    \centering
    \caption{Zero-shot generalization across downstream scene synthesis tasks. Higher is better for all metrics except FID, where lower is better. The best results are highlighted in bold, and the second-best are underlined.}
    \label{downstream}
    \renewcommand{\arraystretch}{0.8}
    {\fontsize{6.8}{8.6}\selectfont
    \resizebox{0.86\textwidth}{!}{
    \begin{tabular}{@{}ccccccccc@{}}
    \toprule
    & & Stylization & \multicolumn{2}{c}{Re-arrangement} & \multicolumn{2}{c}{Completion} & Uncond. & Layout-to-Object \\
    \cmidrule(l){3-9}
    & & $\Delta_{\times1e-3}$ & iRecall (\%) & FID & iRecall (\%) & FID & FID & FID \\
    \midrule
    Bedroom     & ATISS         & 2.97 & 43.77 & 113.19 & 22.25 & 120.54 & \textbf{124.81} & -- \\
            & DiffuScene    & 3.53 & 45.48 & 118.54 & 55.50 & 96.30 & 131.72 & \underline{81.41} \\
                & InstructScene & \textbf{7.03} & \underline{75.55} & \underline{100.95} & \underline{64.06} & \underline{94.36} & \underline{125.47} & -- \\
    \rowcolor[HTML]{ECF4FF}
        & \textbf{Ours}      &   \underline{5.58}   &\textbf{77.26} &\textbf{100.07}&\textbf{67.97 }&\textbf{92.58}&   128.17       & \textbf{62.68} \\
    \midrule
    Living room & ATISS         & 0.70 & 26.73 & 115.65 & 36.73 & 127.37 & 139.40 & -- \\
           & DiffuScene    & 0.38 & 31.63 & 129.55 & 34.49 & \textbf{95.91} & 128.60 & \underline{92.86} \\
           & InstructScene & \underline{0.71} & \textbf{58.16} & \underline{110.56} & \textbf{44.49} & \underline{96.36} & \textbf{117.59} & -- \\
    \rowcolor[HTML]{ECF4FF}
        & \textbf{Ours}      &\textbf{1.42}& \underline{54.89} &\textbf{108.03}&  \underline{40.61}     &   99.85     & \underline{120.67} & \textbf{86.07} \\
    \midrule
    Dining room & ATISS         & -0.31 & 34.89 & 141.59 & \underline{47.33} & 147.80 & 166.23 & -- \\
           & DiffuScene    & \underline{-0.15} & 41.56 & 142.52 & 45.78 & \underline{120.88} & 147.73 & \underline{107.35} \\
           & InstructScene & \textbf{5.02} & \underline{56.22} & \underline{133.71} & \textbf{53.56} & \textbf{118.10} & \textbf{138.32} & -- \\
    \rowcolor[HTML]{ECF4FF}
        & \textbf{Ours}      &  \textbf{5.02}    &\textbf{62.00}&\textbf{124.34}&   46.00     &  121.91      &    \underline{139.95}    & \textbf{98.64} \\
    \bottomrule
    \end{tabular}
    }
    }
\end{table}

\section{Additional Results} \label{qualitative_appendix}
In this section, we present various qualitative results. \cref{text-to-scene} reports various visualizations of text-conditioned scene generation.
From \cref{tri1} to \cref{tri4}, we report visualization results for instructions with one to four relational constraints, respectively. 
Additionally, from \cref{Stylization} to \cref{fig:uncond_rooms}, we report qualitative results for downstream tasks including stylization, re-arrangement, completion, layout-to-object and unconditional generation.
\cref{arbitration} visualizes the system's behavior when encountering physically implausible or contradictory instructions, demonstrating the prioritization of learned physical priors.
\cref{openvoca} further demonstrates the model's generalization capacity with open vocabulary instructions, including unseen spatial predicates and descriptive object attributes.

Finally, \cref{in-the-wild} presents qualitative results of the model's performance on diverse ``in-the-wild'' instructions. We observe strong generalization in handling functional roles and aesthetic descriptions by mapping abstract concepts to concrete objects via the semantic embedding space.
However, the model exhibits variable capability with negative constraints and high-level spatial reasoning.
We attribute the inconsistent adherence to negative constraints to the dominance of positive samples in the training data which causes the model to overlook negation tokens, while the challenge with high-level spatial reasoning stems from the absence of explicit relational triggers required to ground abstract spatial logic into geometric constraints.

\section{Failure Cases}
We present a qualitative analysis of typical failure modes observed in our framework. \cref{fail_rel} illustrates instances of relation violations, where the generated layout fails to strictly satisfy the spatial constraints specified in the instruction. \cref{fail_desc} demonstrates cases of incomplete adherence to visual descriptions, where the model overlooks specific object attributes such as color or style. Finally, \cref{fail_layout} showcases examples of compromised layout plausibility, exhibiting physical inconsistencies such as object collisions or geometric misalignments.

\newpage
\begin{figure}[p]
  \vspace{-10pt}
  \centering
\makebox[.16\linewidth][c]{t=10}\hfill
\makebox[.16\linewidth][c]{t=20}\hfill
\makebox[.16\linewidth][c]{t=30}\hfill
\makebox[.16\linewidth][c]{t=40}\hfill
\makebox[.16\linewidth][c]{t=50}

  \begin{subfigure}[b]{\textwidth}
    \centering
    \includegraphics[width=.16\linewidth]{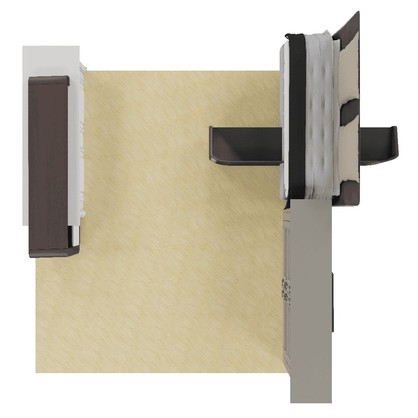}\hfill
    \includegraphics[width=.16\linewidth]{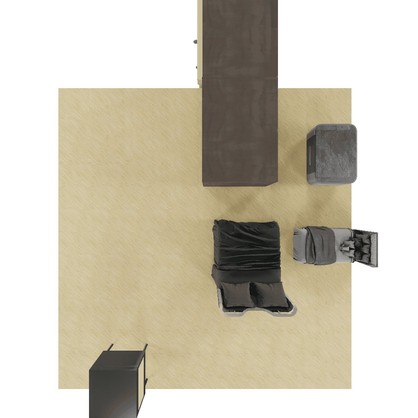}\hfill
    \includegraphics[width=.16\linewidth]{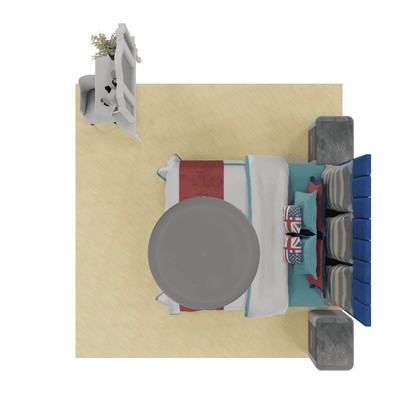}\hfill
    \includegraphics[width=.16\linewidth]{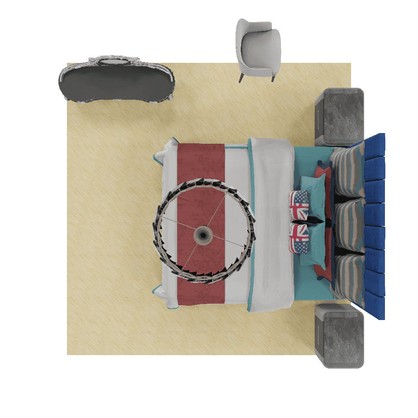}\hfill
    \includegraphics[width=.16\linewidth]{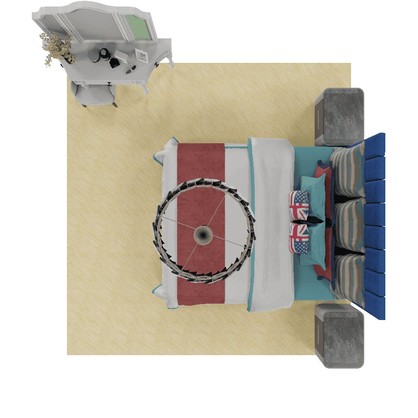}
  \end{subfigure}
  
  \vspace{-10pt}

  \begin{subfigure}[b]{\textwidth}
    \centering
    \includegraphics[width=.16\linewidth]{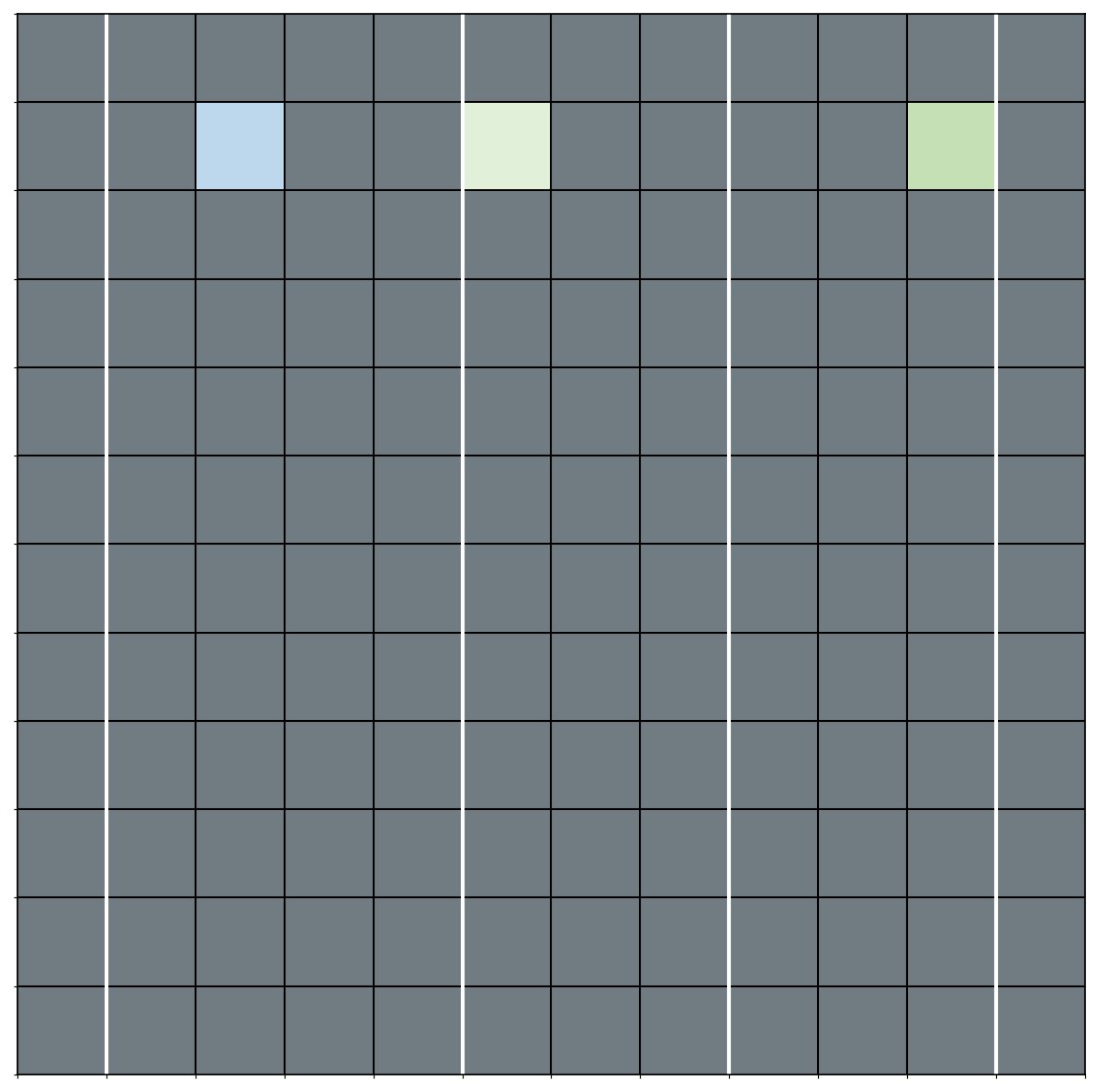}\hfill
    \includegraphics[width=.16\linewidth]{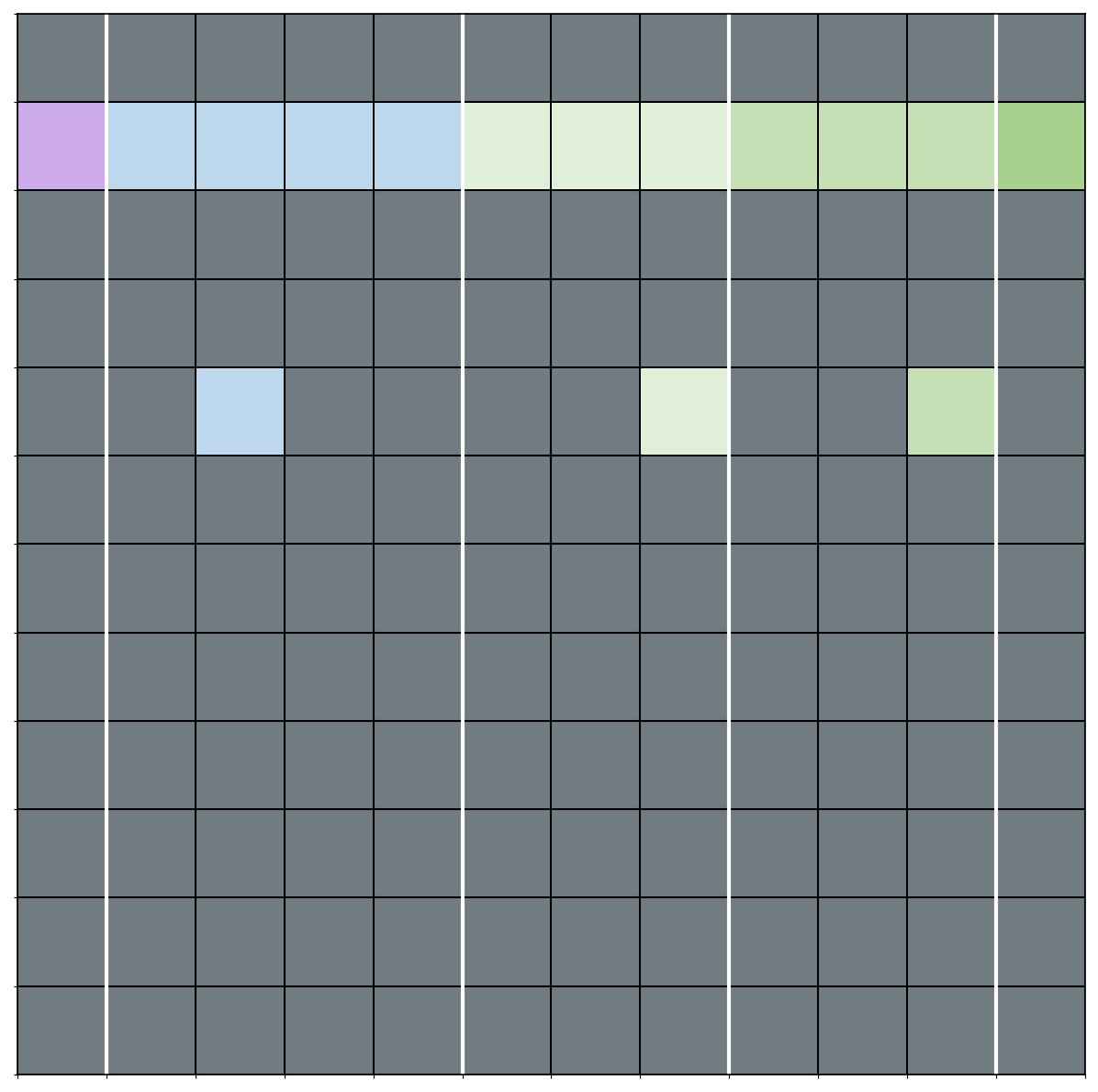}\hfill
    \includegraphics[width=.16\linewidth]{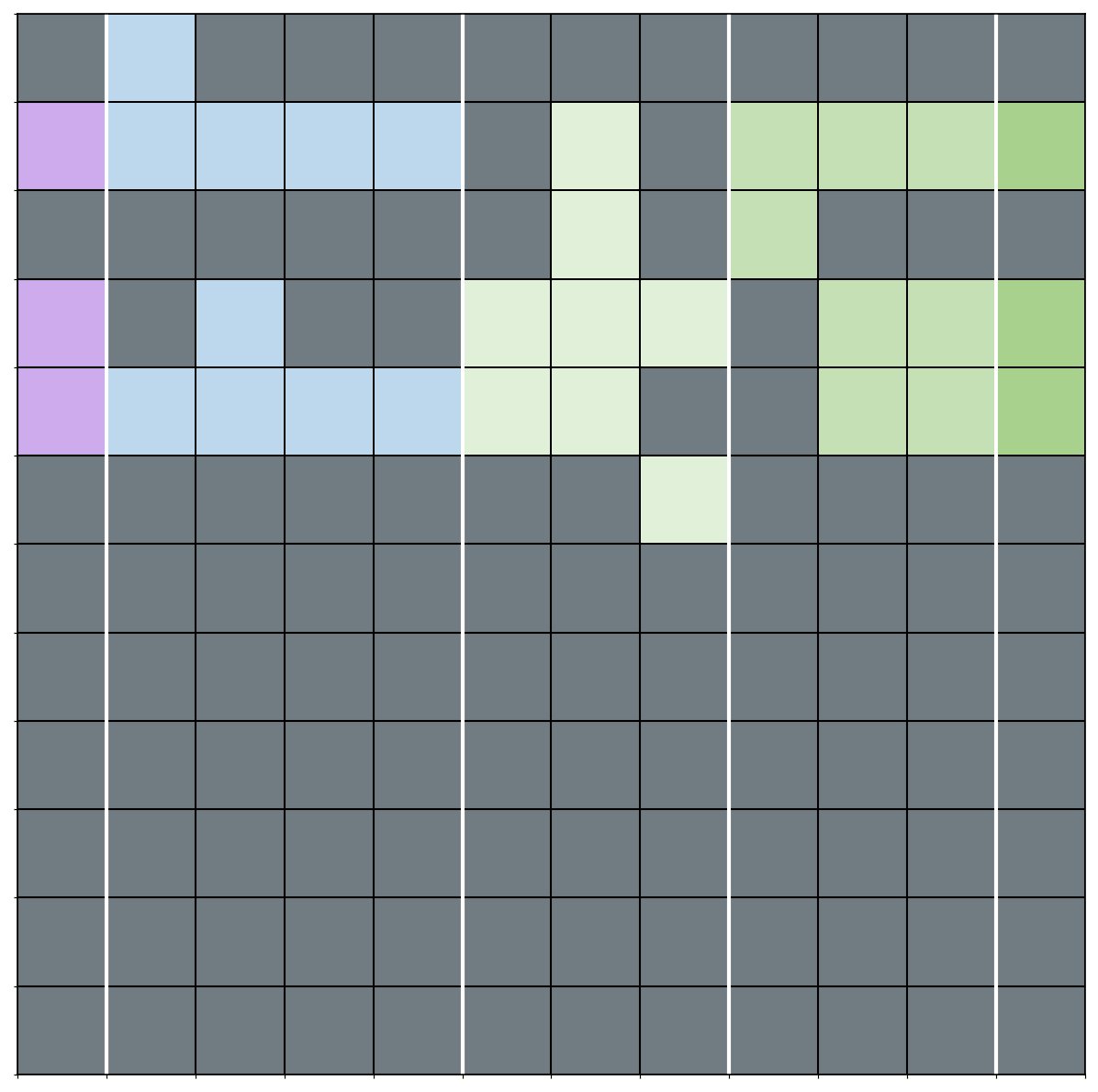}\hfill
    \includegraphics[width=.16\linewidth]{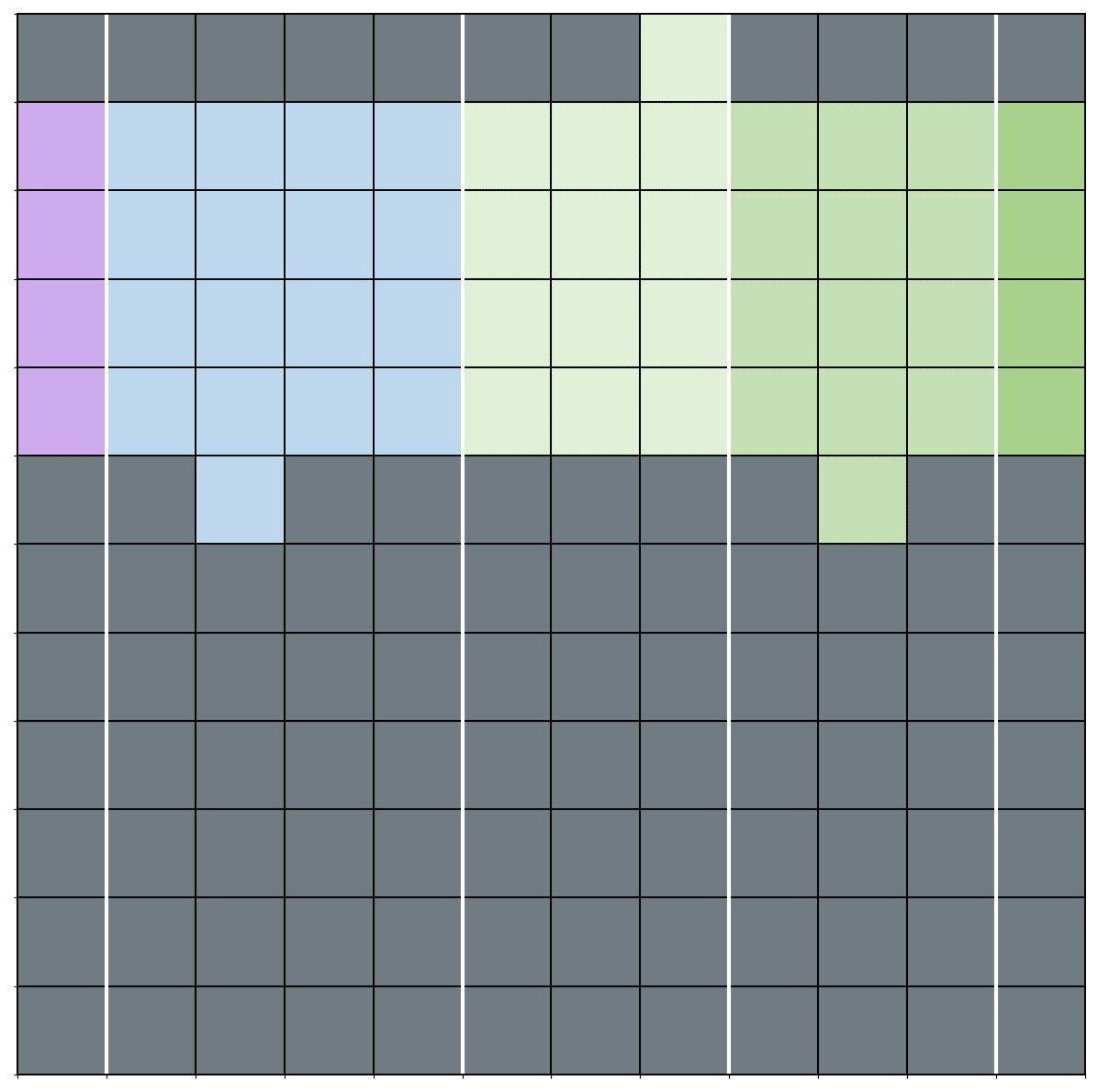}\hfill
    \includegraphics[width=.16\linewidth]{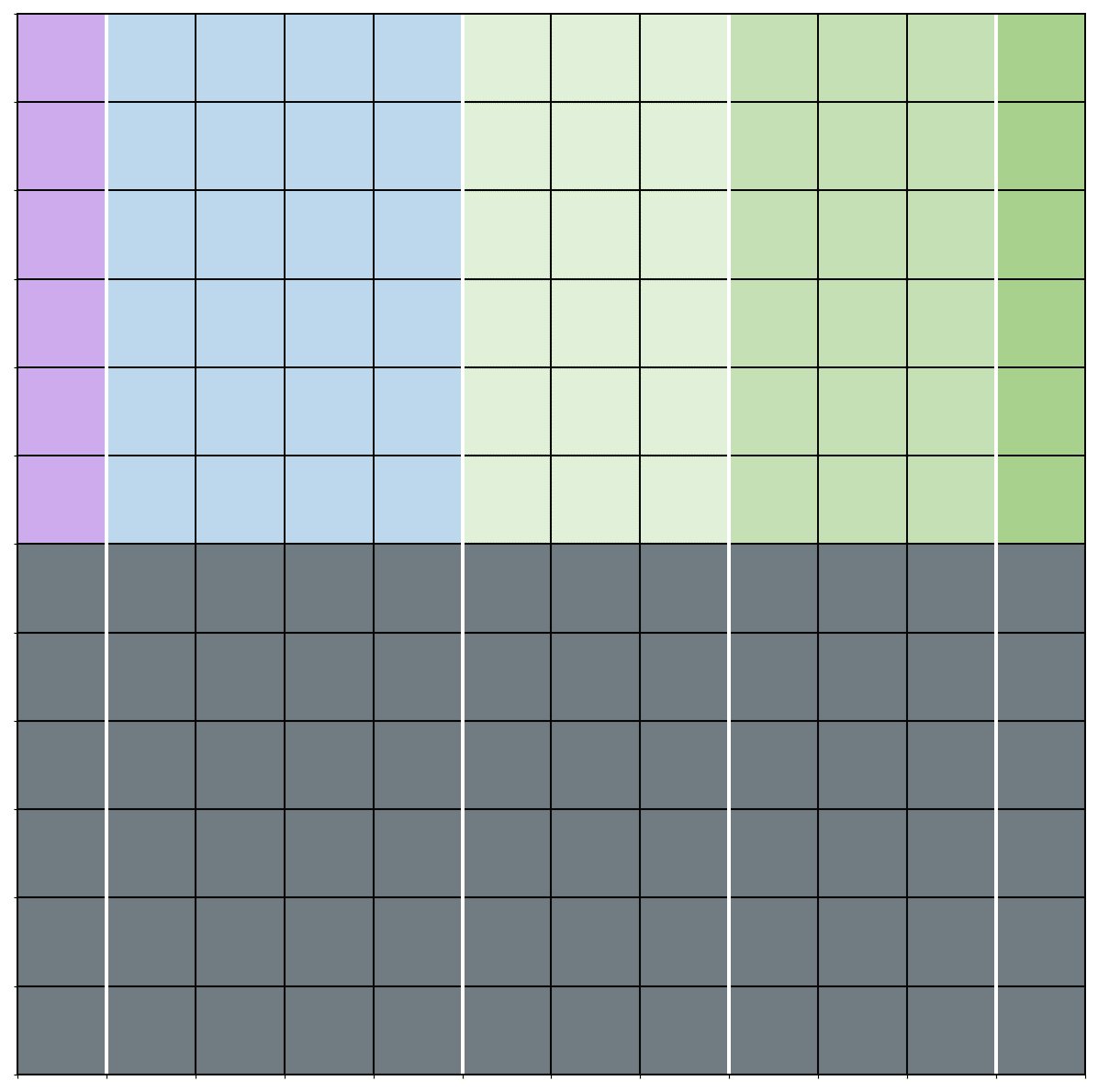}
  \end{subfigure}
 \begin{subfigure}[b]{\textwidth}
  \vspace{-10pt}
    \centering
    \includegraphics[width=.16\linewidth]{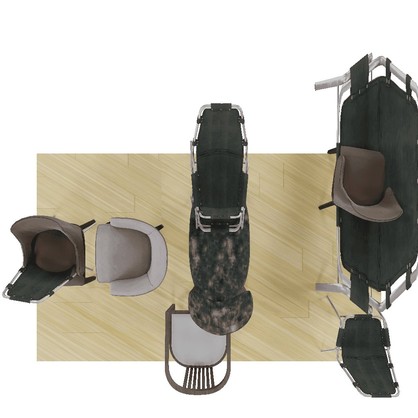}\hfill
    \includegraphics[width=.16\linewidth]{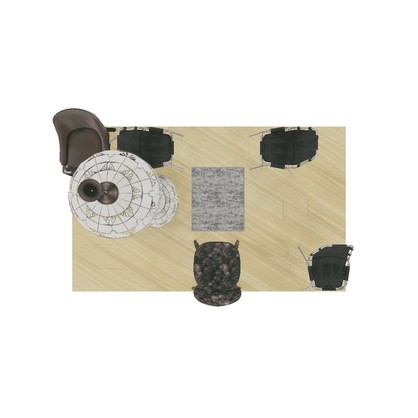}\hfill
    \includegraphics[width=.16\linewidth]{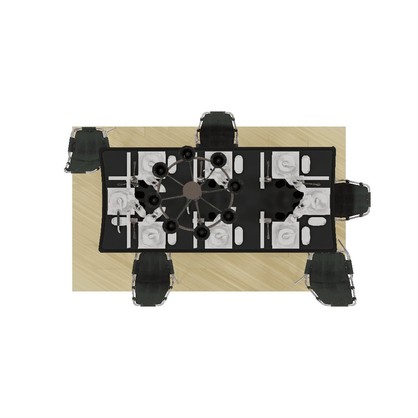}\hfill
    \includegraphics[width=.16\linewidth]{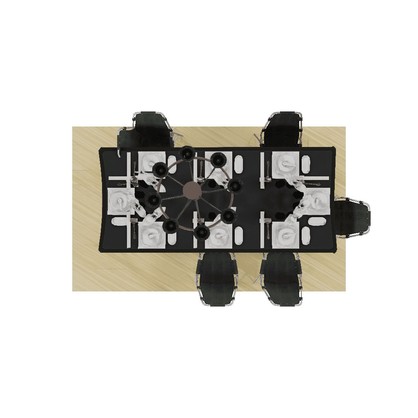}\hfill
    \includegraphics[width=.16\linewidth]{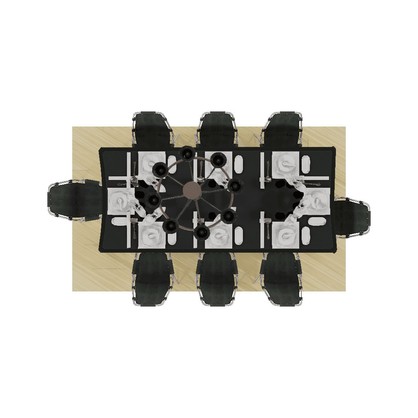}
  \end{subfigure}
  
  \vspace{-10pt}

  \begin{subfigure}[b]{\textwidth}
    \centering
    \includegraphics[width=.16\linewidth]{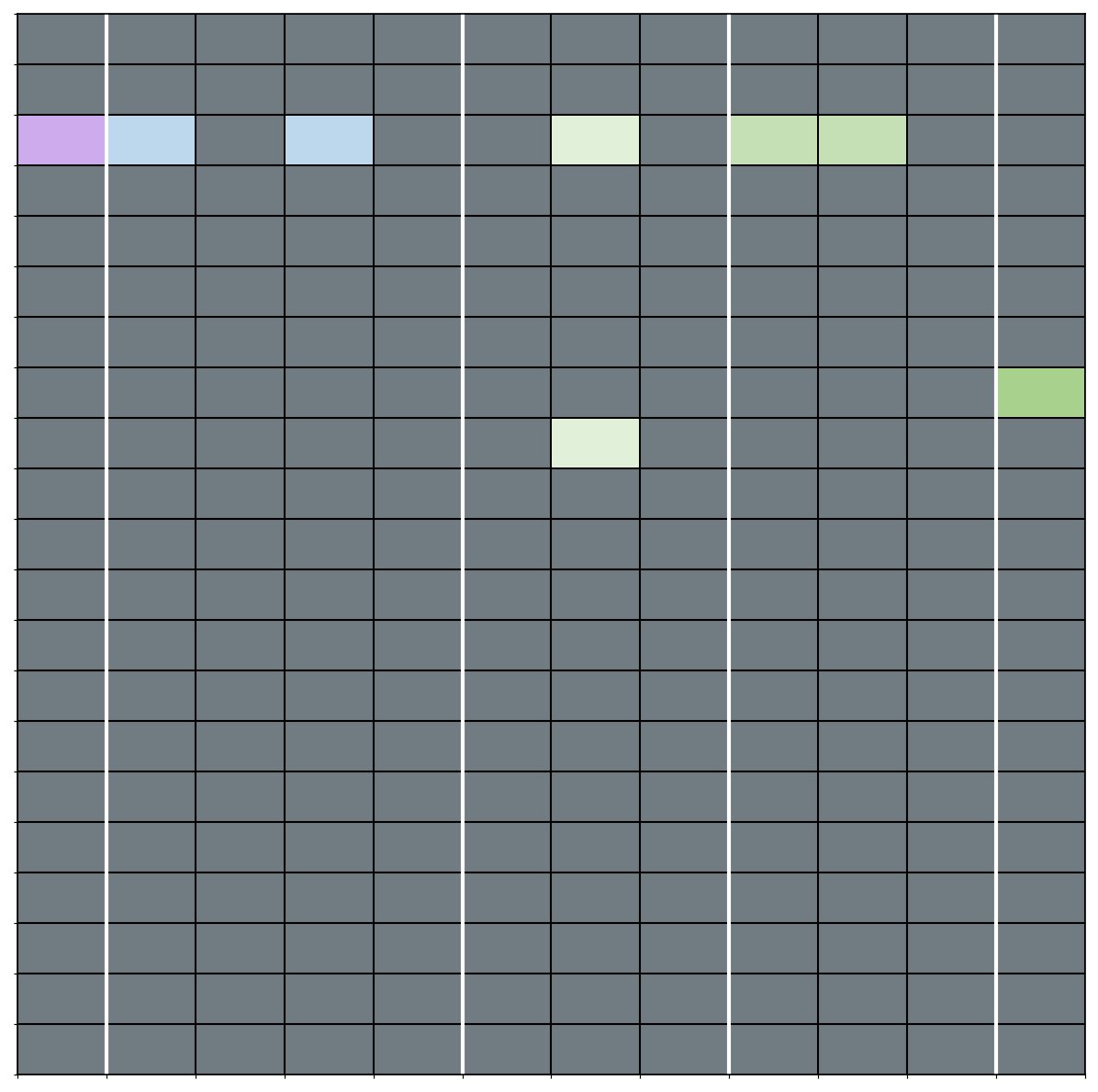}\hfill
    \includegraphics[width=.16\linewidth]{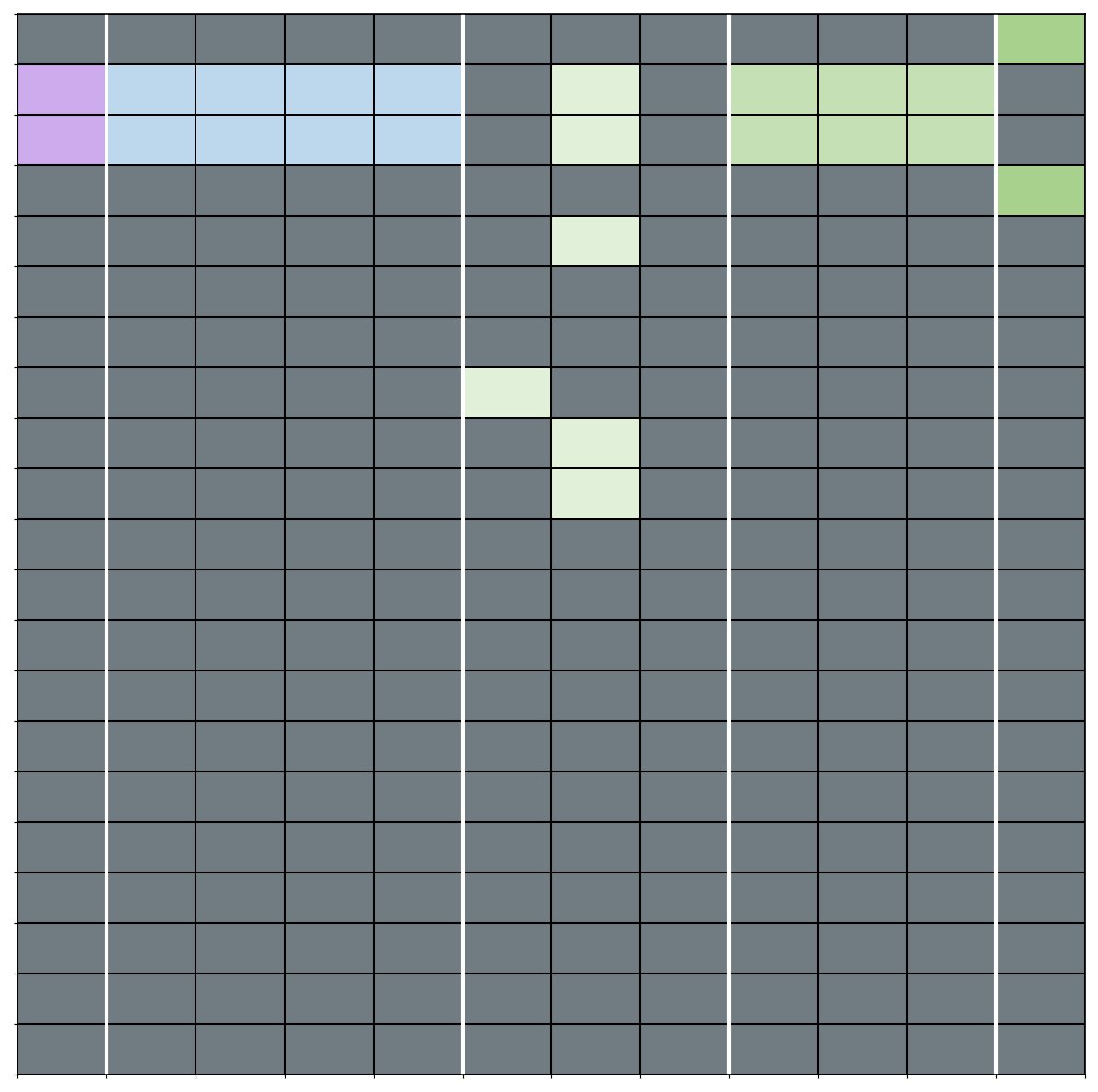}\hfill
    \includegraphics[width=.16\linewidth]{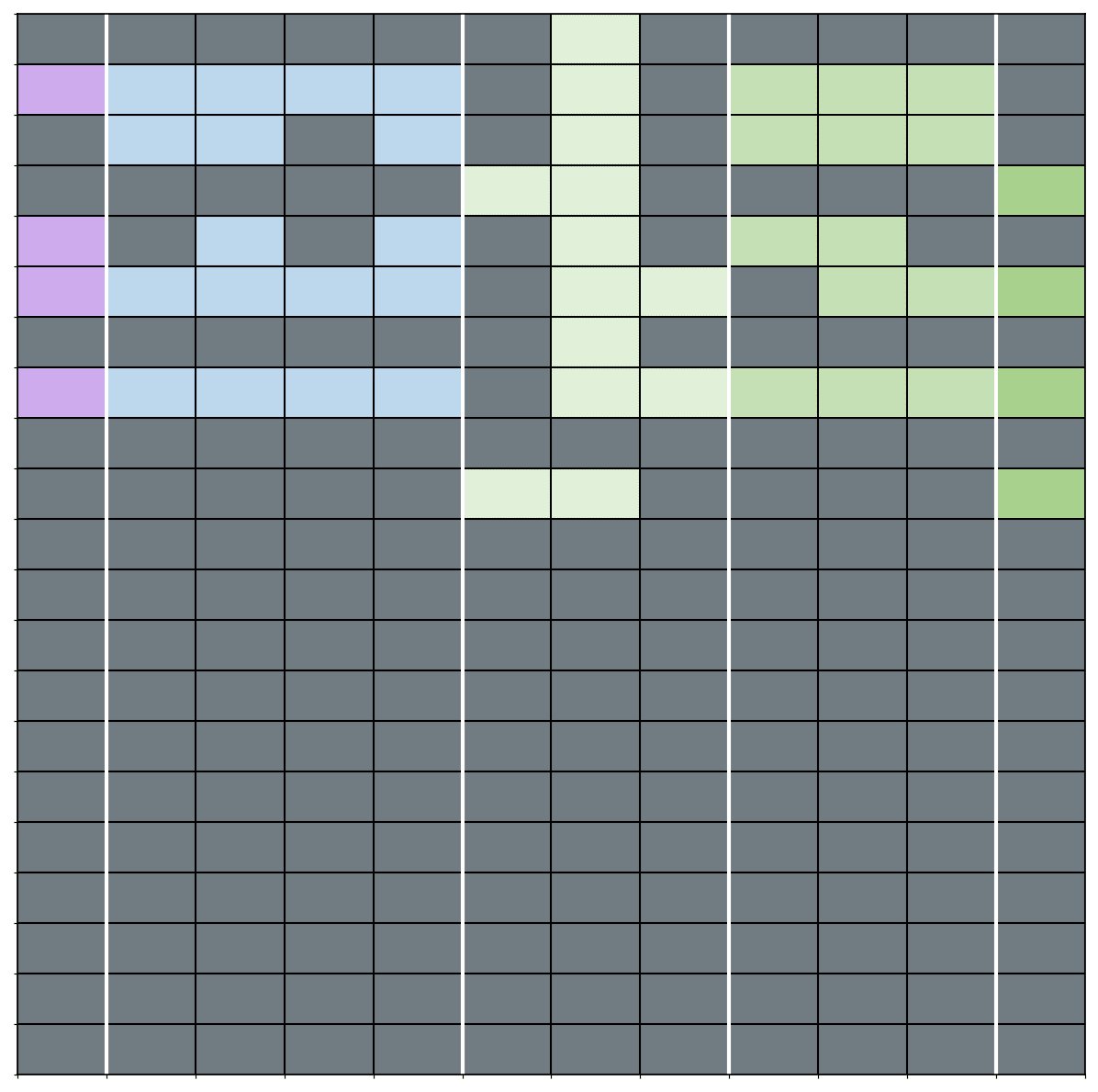}\hfill
    \includegraphics[width=.16\linewidth]{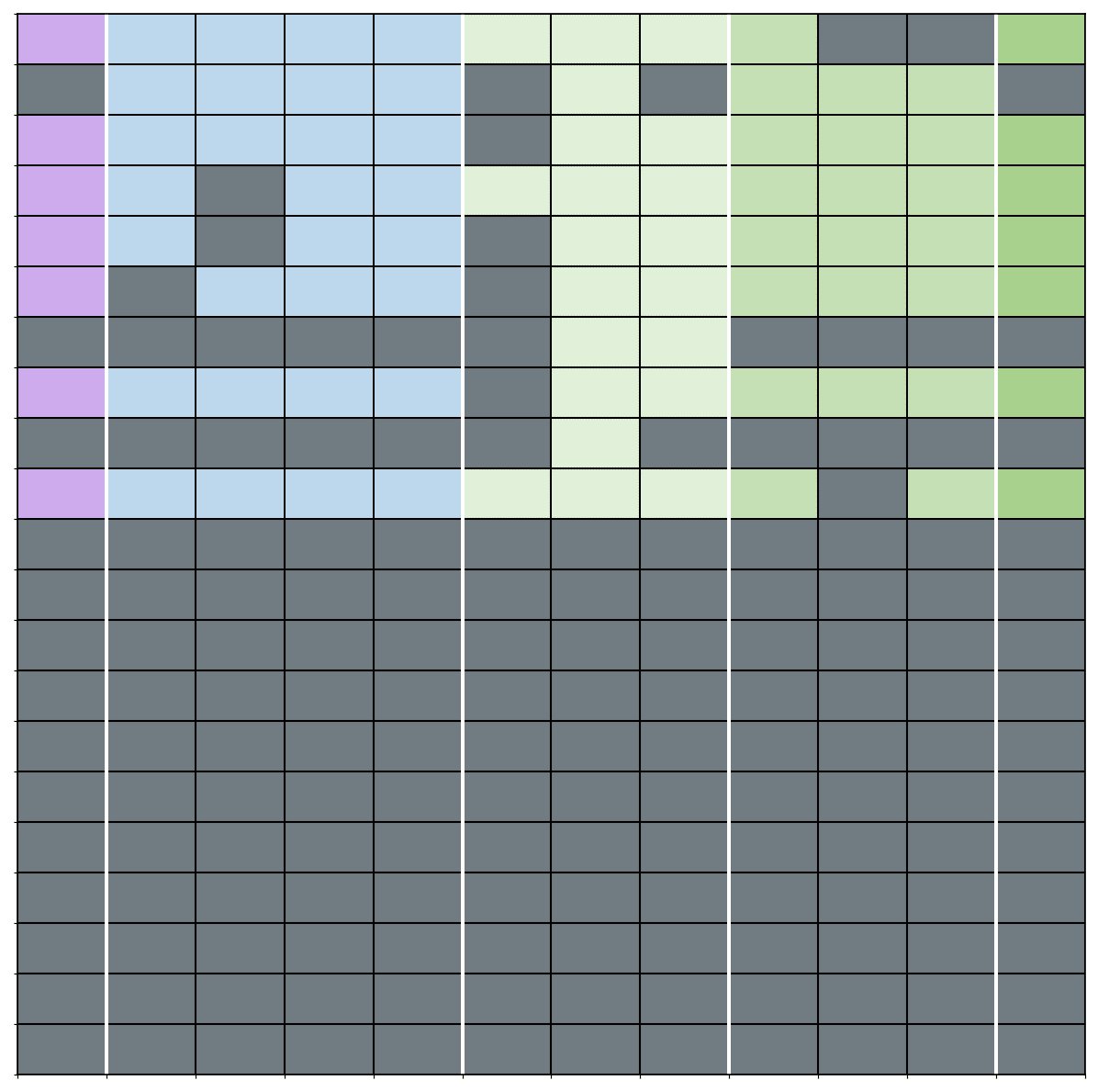}\hfill
    \includegraphics[width=.16\linewidth]{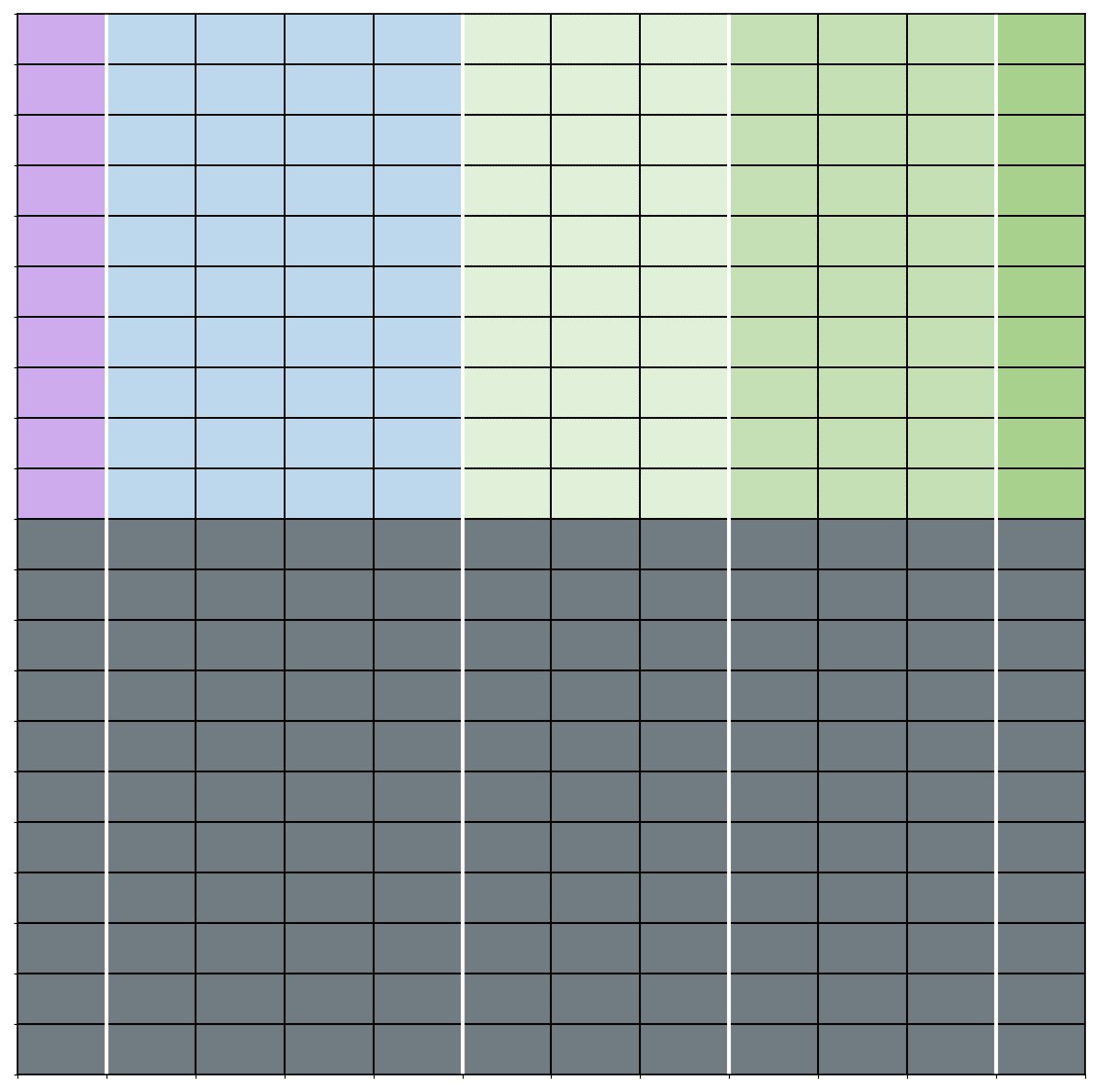}
  \end{subfigure}
 \begin{subfigure}[b]{\textwidth}
    \centering
    \includegraphics[width=.16\linewidth]{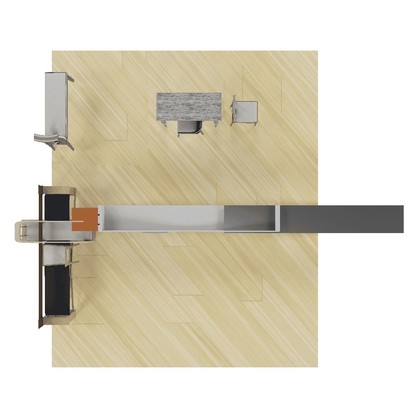}\hfill
    \includegraphics[width=.16\linewidth]{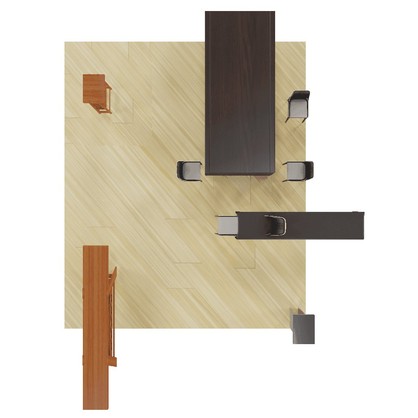}\hfill
    \includegraphics[width=.16\linewidth]{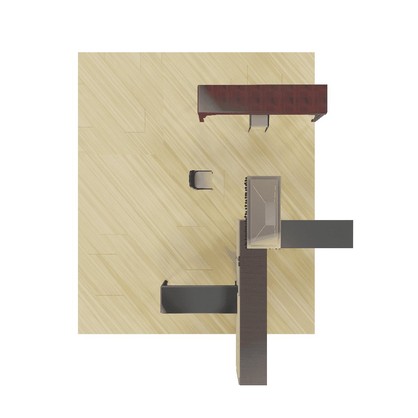}\hfill
    \includegraphics[width=.16\linewidth]{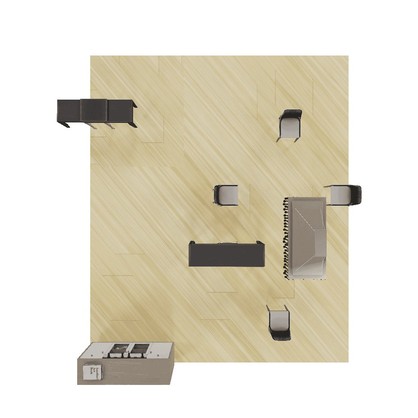}\hfill
    \includegraphics[width=.16\linewidth]{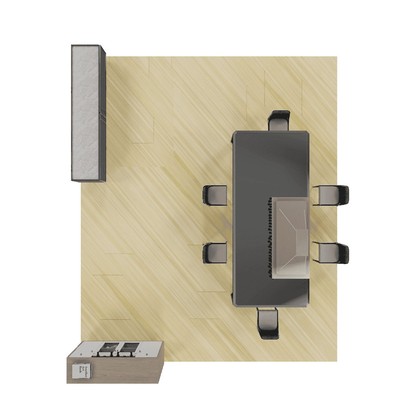}
  \end{subfigure}
  
  \vspace{-10pt}

  \begin{subfigure}[b]{\textwidth}
    \centering
    \includegraphics[width=.16\linewidth]{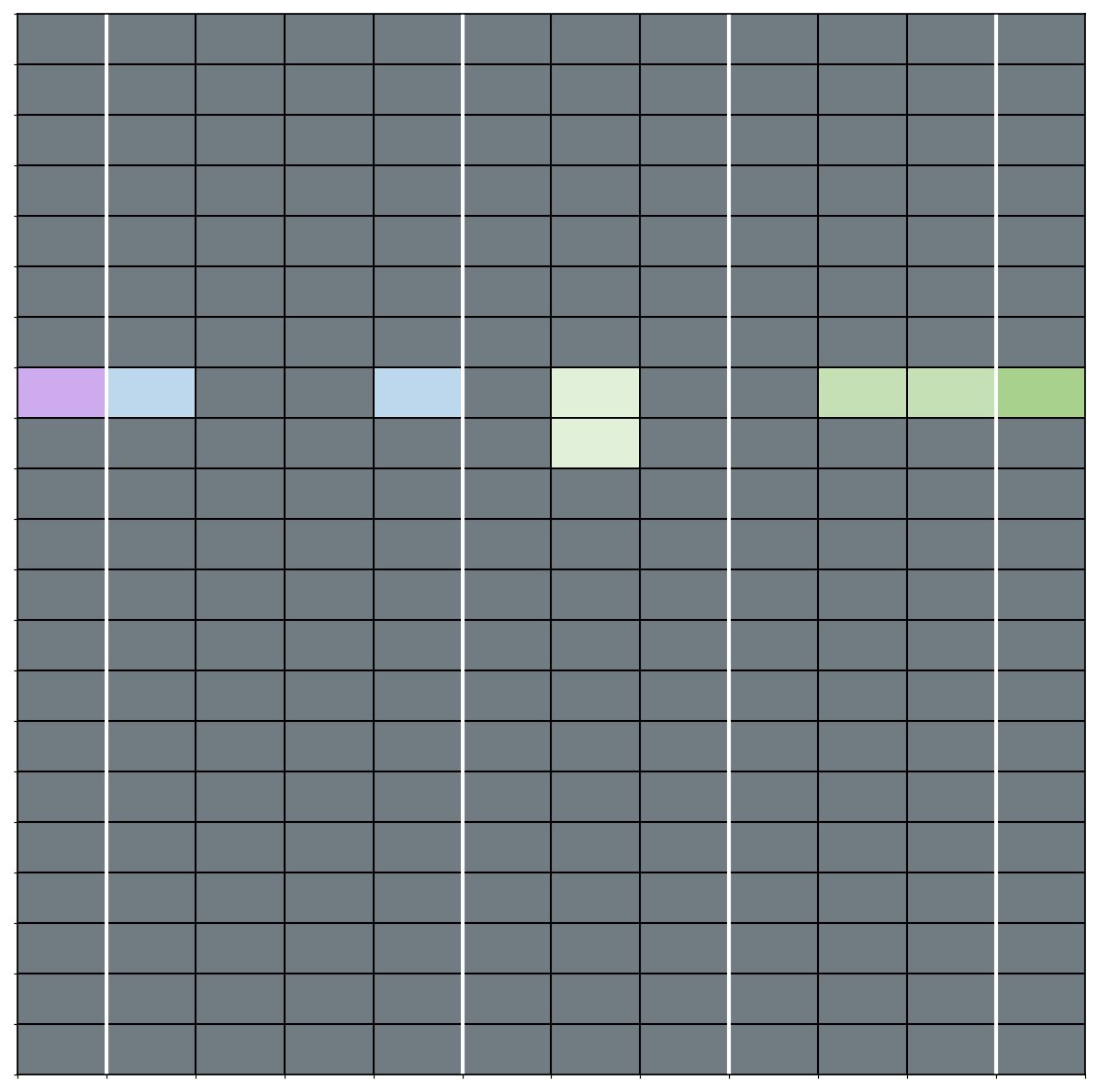}\hfill
    \includegraphics[width=.16\linewidth]{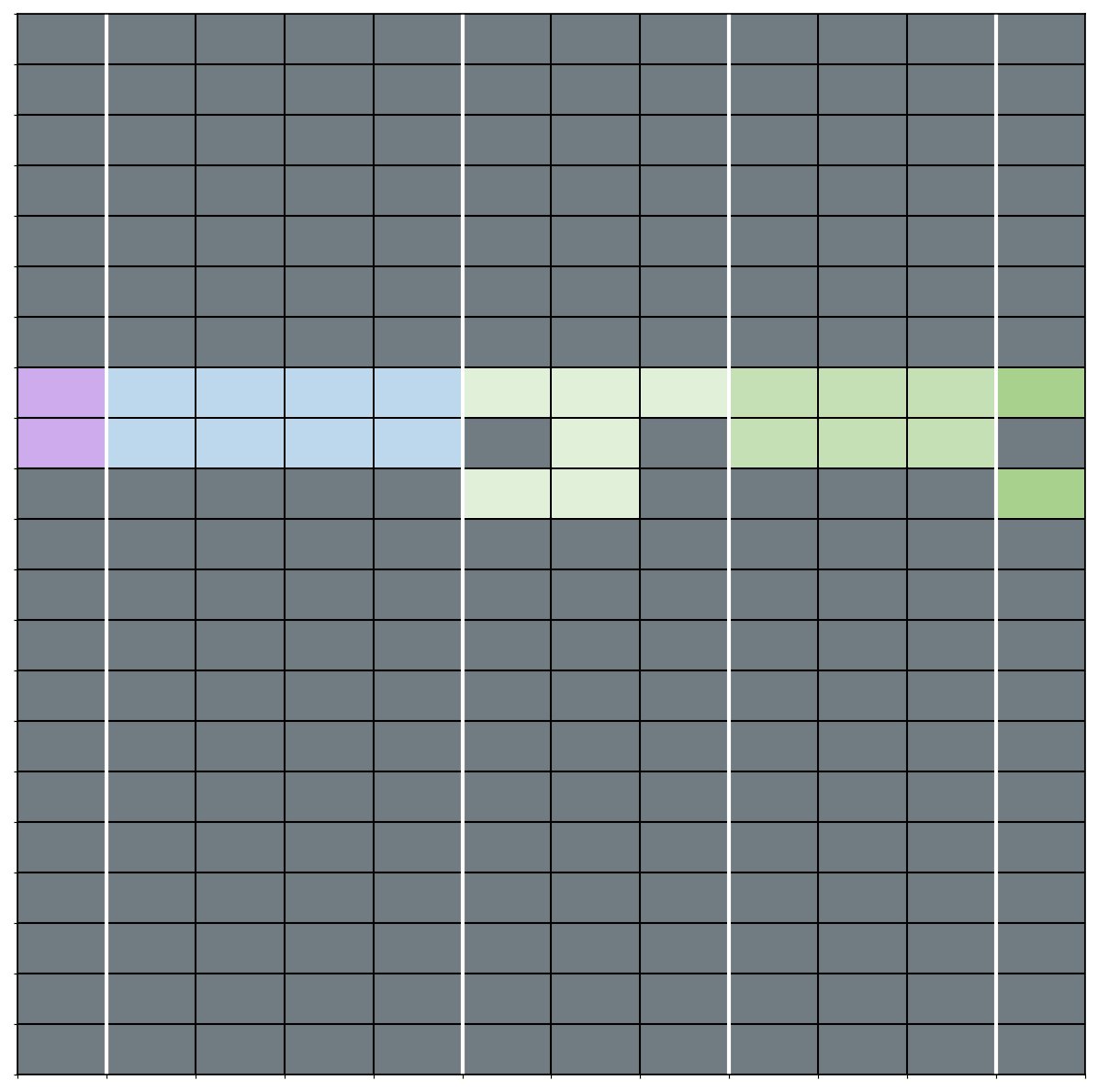}\hfill
    \includegraphics[width=.16\linewidth]{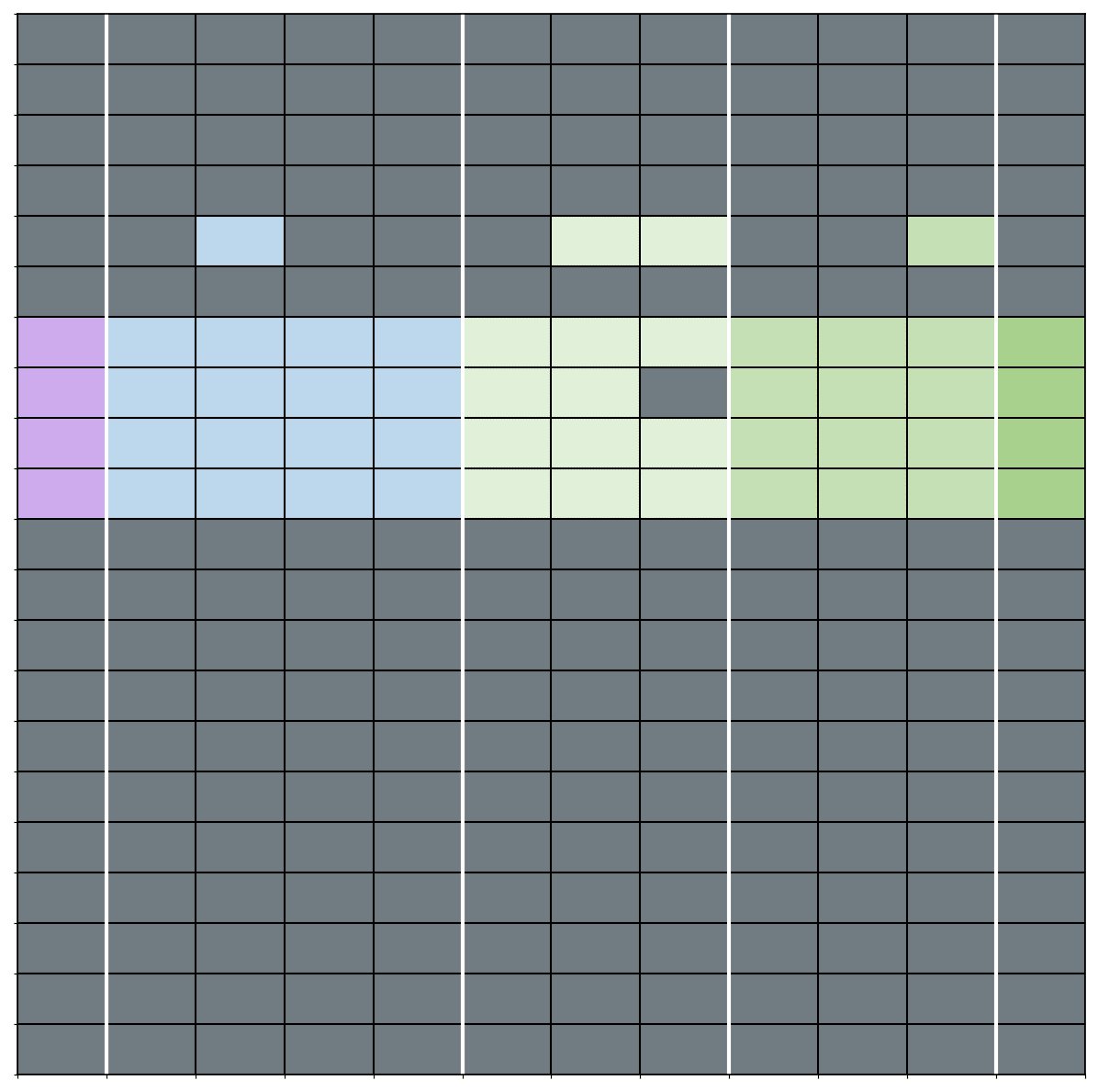}\hfill
    \includegraphics[width=.16\linewidth]{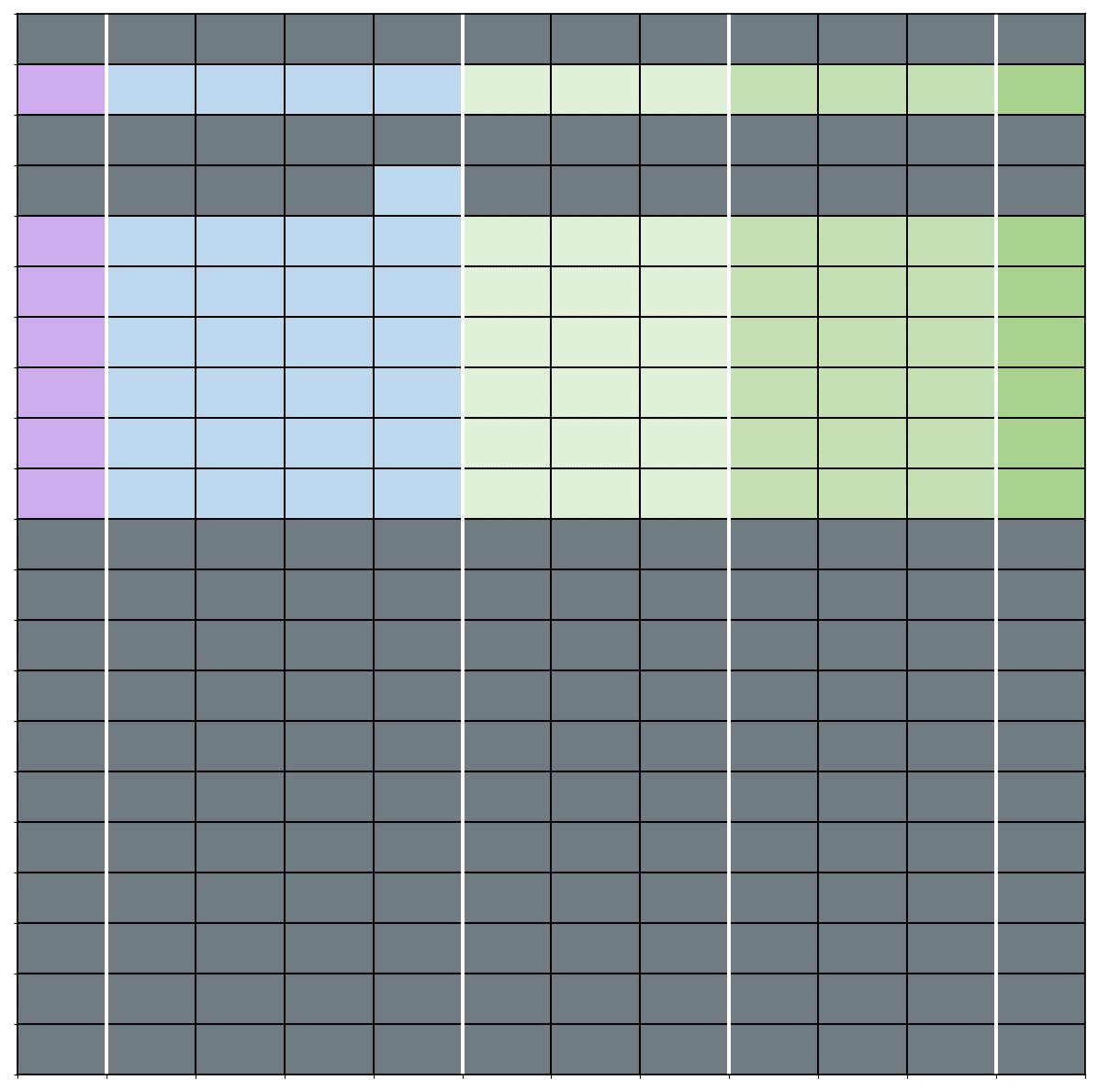}\hfill
    \includegraphics[width=.16\linewidth]{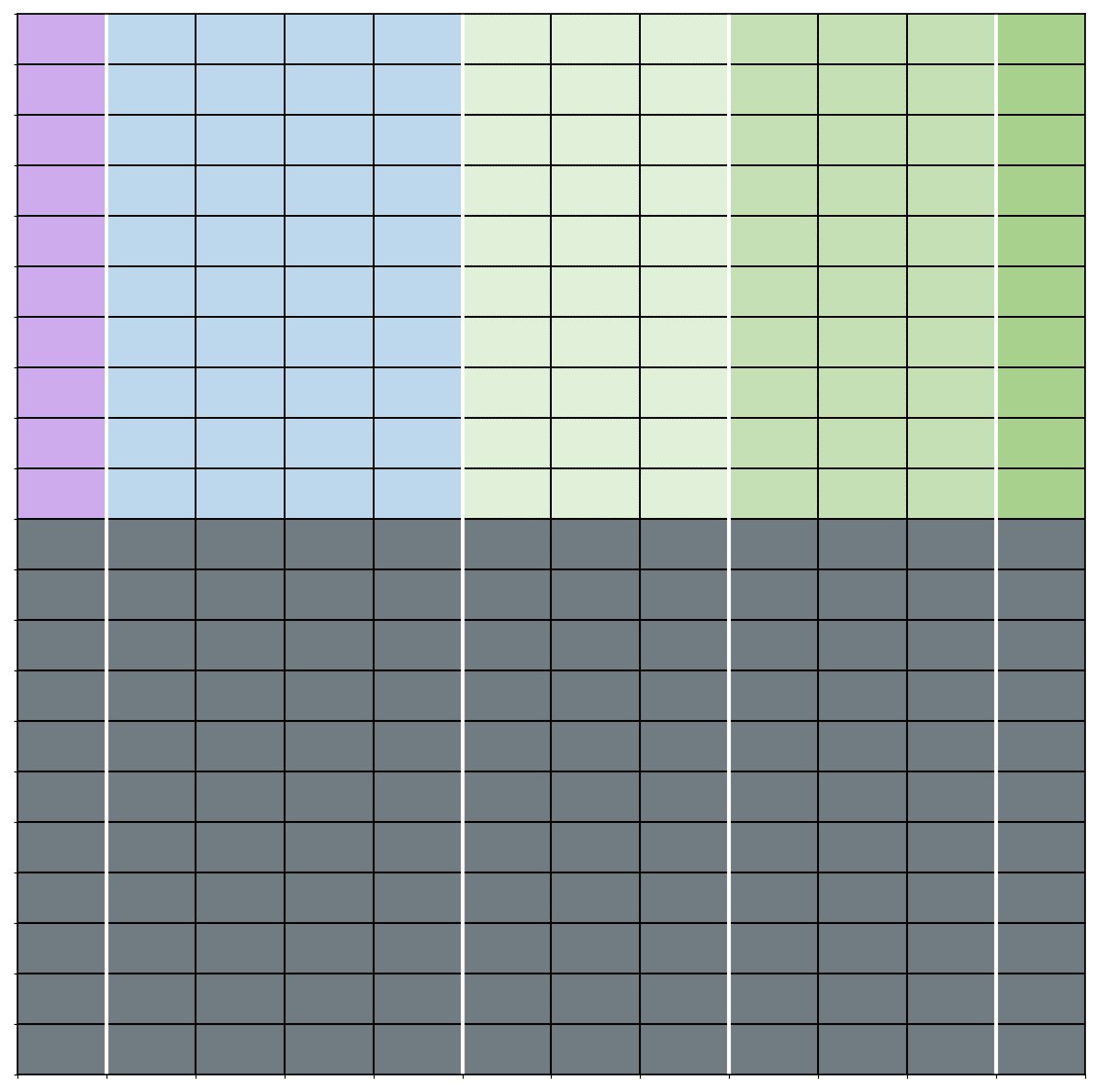}
  \vspace{-10pt}
  \end{subfigure}
 \begin{subfigure}[b]{\textwidth}
    \centering
    \includegraphics[width=.16\linewidth]{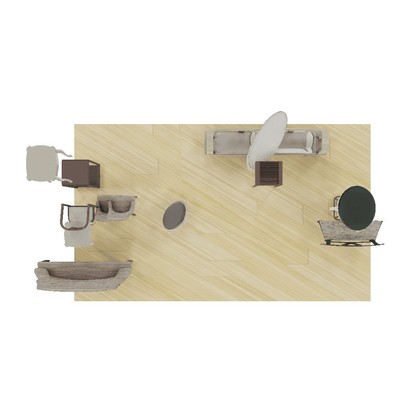}\hfill
    \includegraphics[width=.16\linewidth]{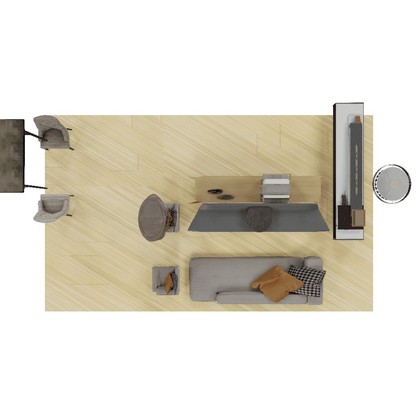}\hfill
    \includegraphics[width=.16\linewidth]{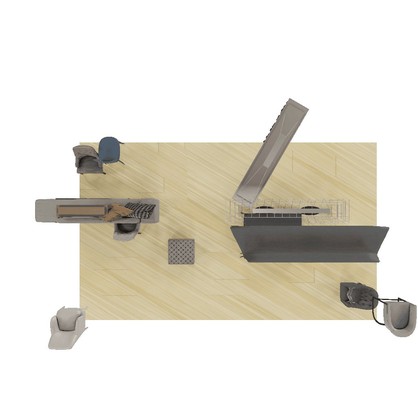}\hfill
    \includegraphics[width=.16\linewidth]{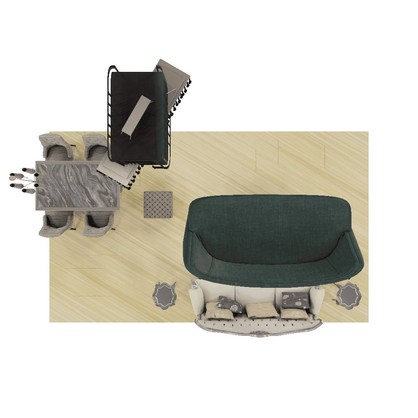}\hfill
    \includegraphics[width=.16\linewidth]{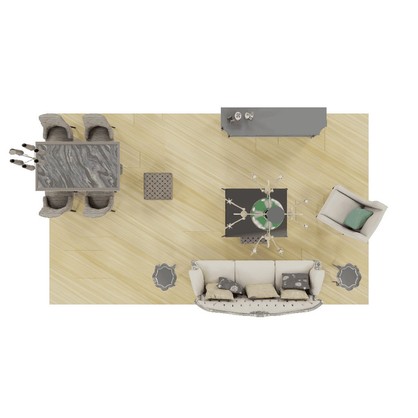}
  \end{subfigure}
  \vspace{-20pt}

  \begin{subfigure}[b]{\textwidth}
    \centering
    \includegraphics[width=.16\linewidth]{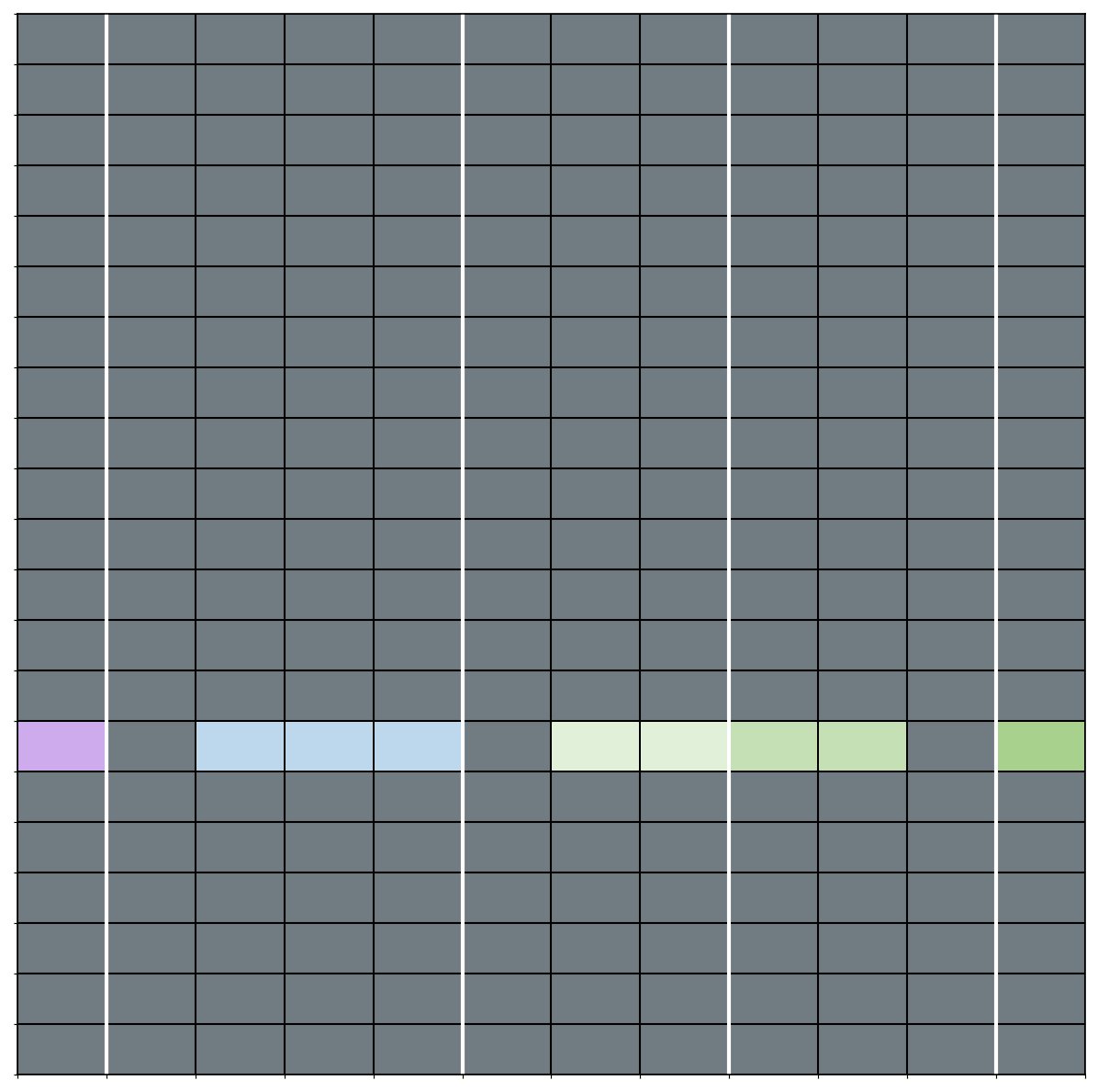}\hfill
    \includegraphics[width=.16\linewidth]{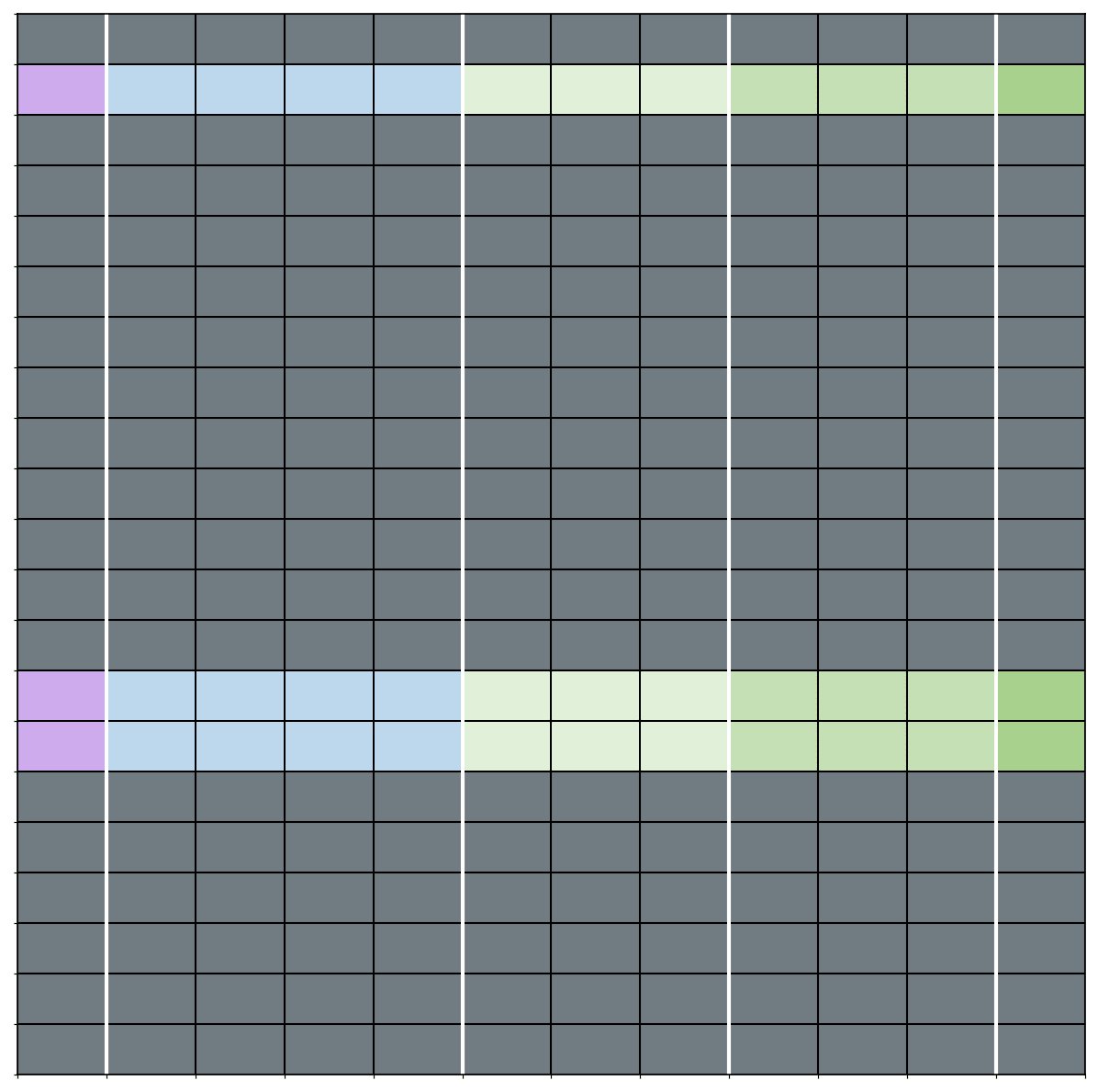}\hfill
    \includegraphics[width=.16\linewidth]{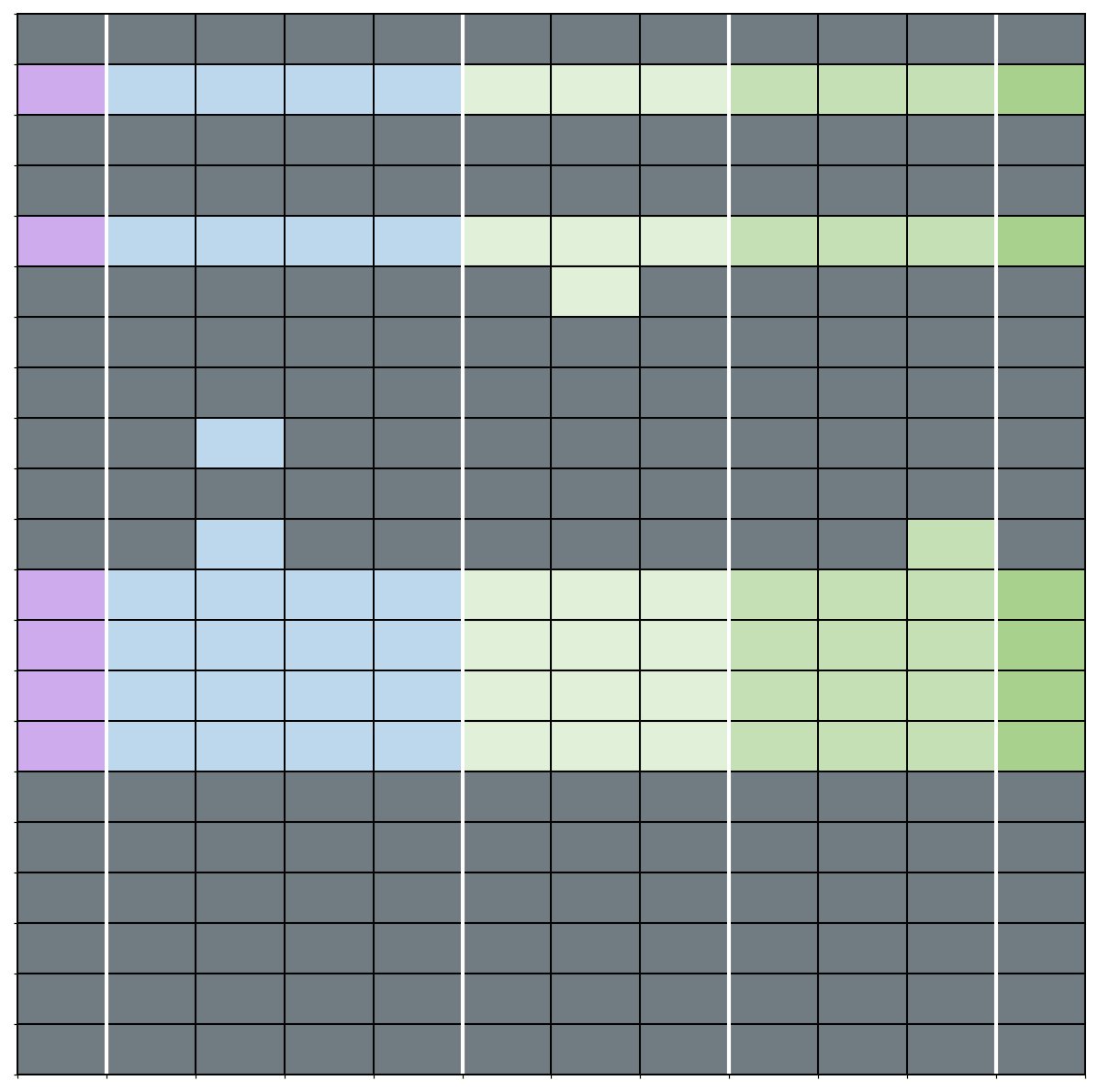}\hfill
    \includegraphics[width=.16\linewidth]{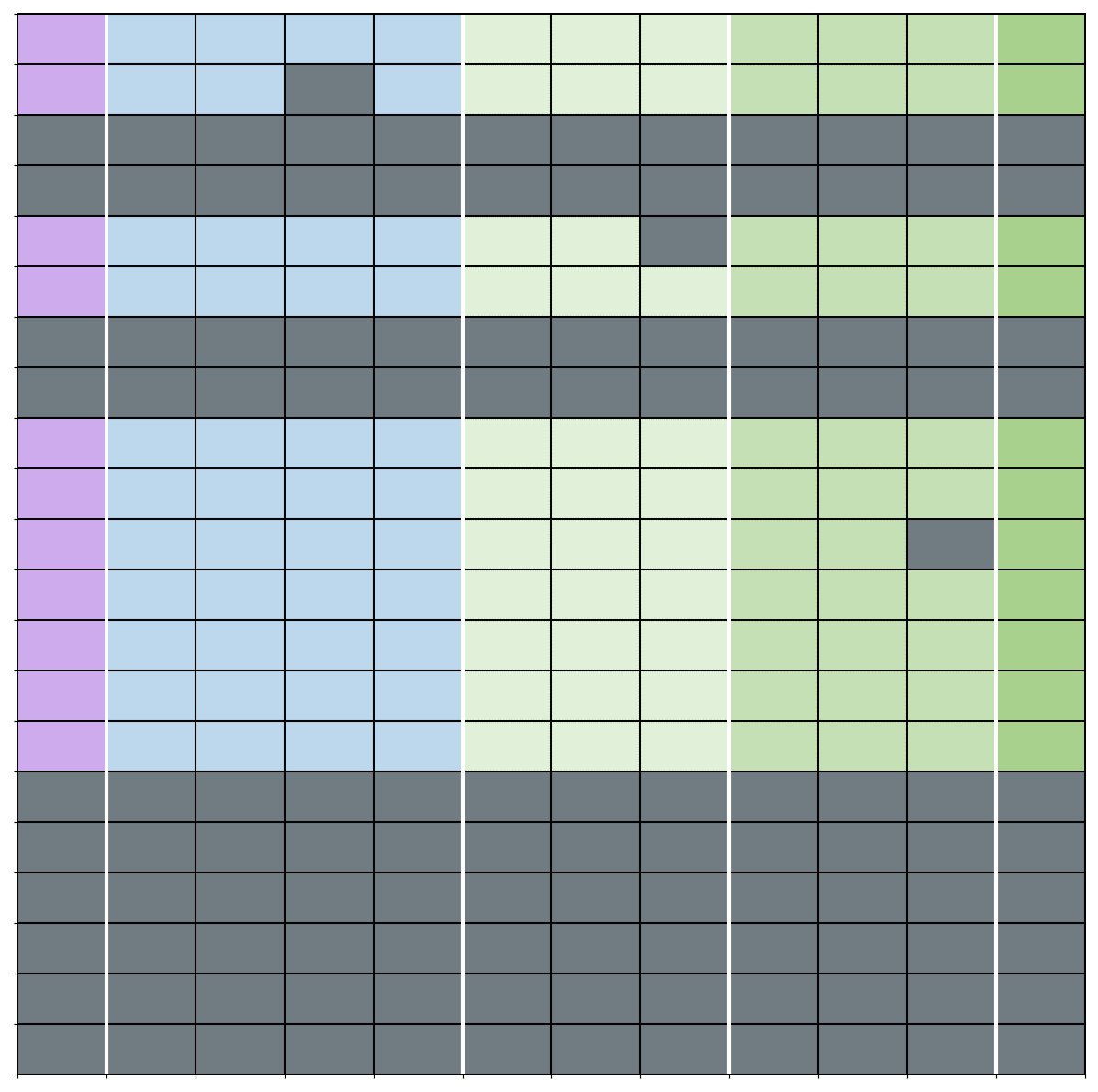}\hfill
    \includegraphics[width=.16\linewidth]{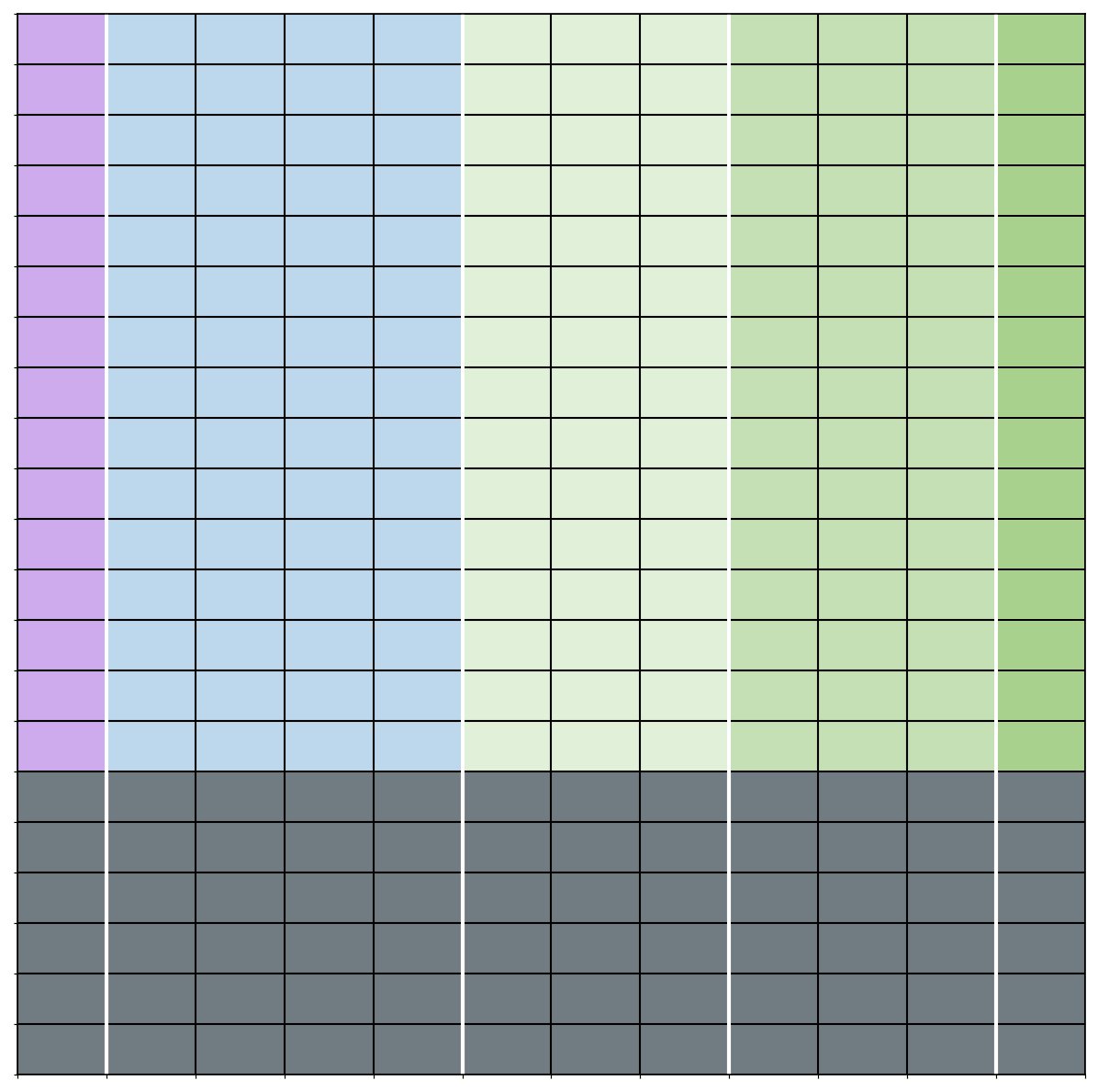}
  \end{subfigure}
  \caption{
  Visualization of a 50-step generation process, which progresses from left to right. Intermediate results are shown at every 10th step, with each synthesized scene displayed alongside a mask that highlights the fixed elements in color.
  }
\label{fig:sampling}
\end{figure}

\newpage
\newpage
\begin{figure}
  \centering
  
  \begin{subfigure}{\textwidth}
    \centering
    \includegraphics[width=.15\linewidth]{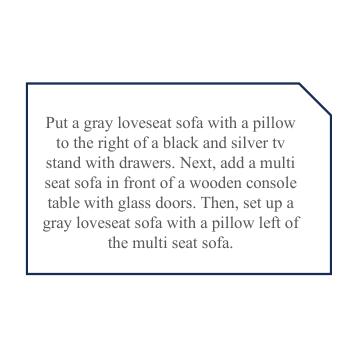}\hfill
    \includegraphics[width=.15\linewidth]{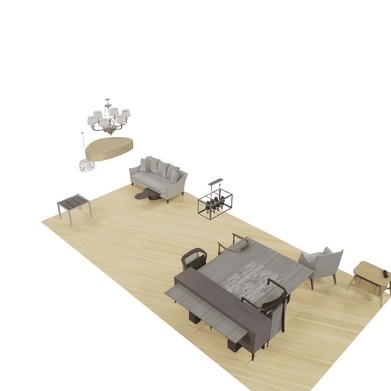}\hfill
    \includegraphics[width=.15\linewidth]{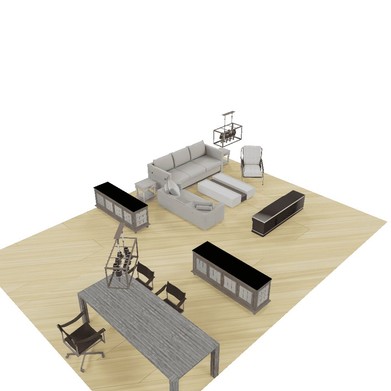}\hfill
    \includegraphics[width=.15\linewidth]{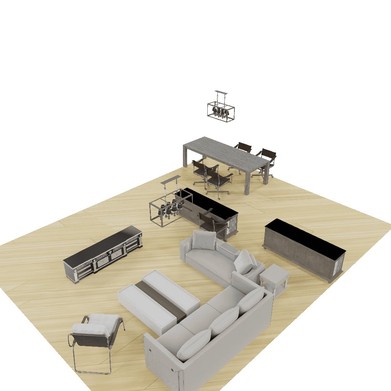}\hfill
    \includegraphics[width=.15\linewidth]{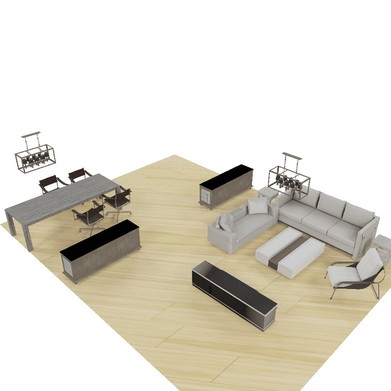}
  \end{subfigure}
  
 \begin{subfigure}{\textwidth} 
    \centering
    \includegraphics[width=.15\linewidth]{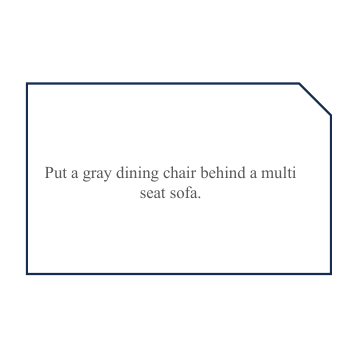}\hfill
    \includegraphics[width=.15\linewidth]{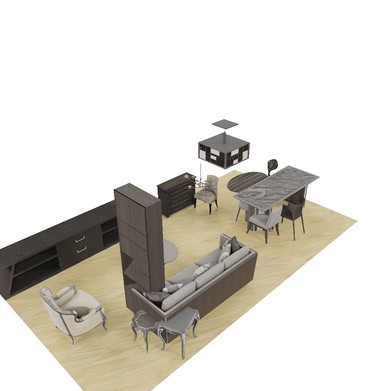}\hfill
    \includegraphics[width=.15\linewidth]{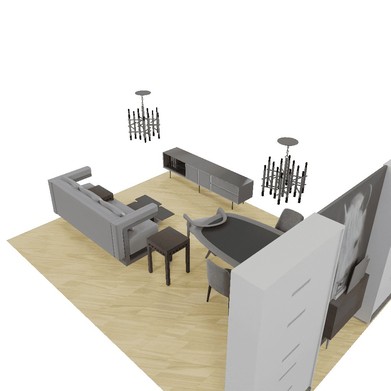}\hfill
    \includegraphics[width=.15\linewidth]{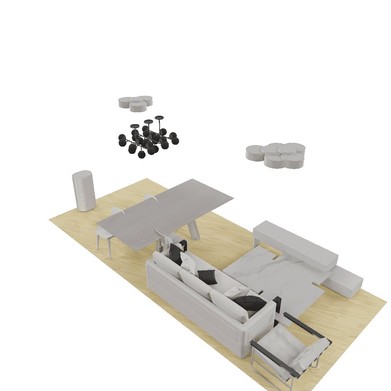}\hfill
    \includegraphics[width=.15\linewidth]{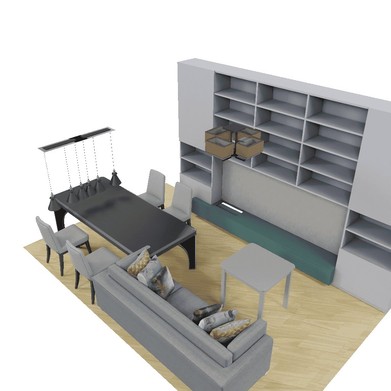}
  \end{subfigure}
  
  \begin{subfigure}{\textwidth} 
    \centering
    \includegraphics[width=.15\linewidth]{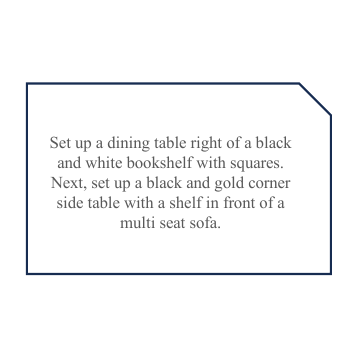}\hfill
    \includegraphics[width=.15\linewidth]{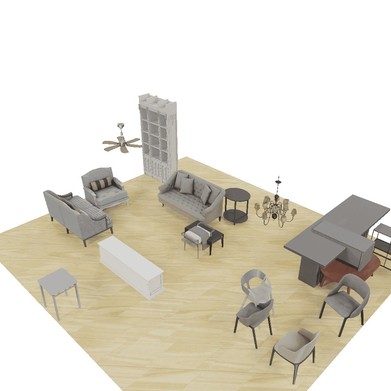}\hfill
    \includegraphics[width=.15\linewidth]{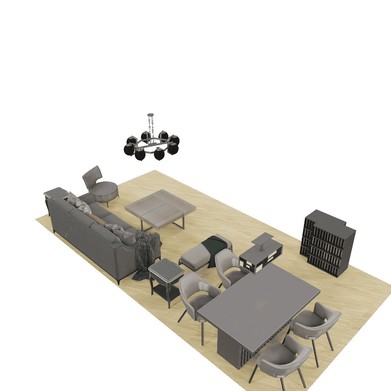}\hfill
    \includegraphics[width=.15\linewidth]{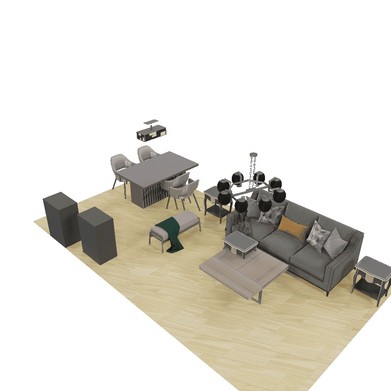}
    \includegraphics[width=.15\linewidth]{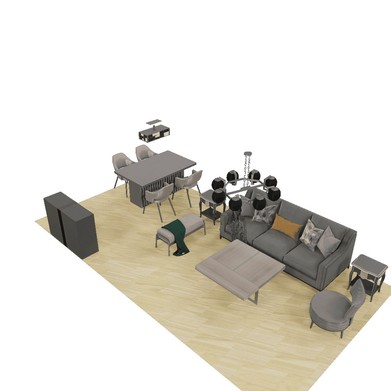}
  \end{subfigure}

  \begin{subfigure}{\textwidth} 
    \centering
    \includegraphics[width=.15\linewidth]{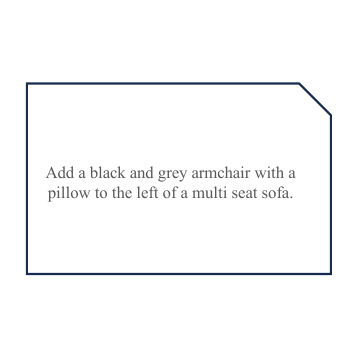}\hfill
    \includegraphics[width=.15\linewidth]{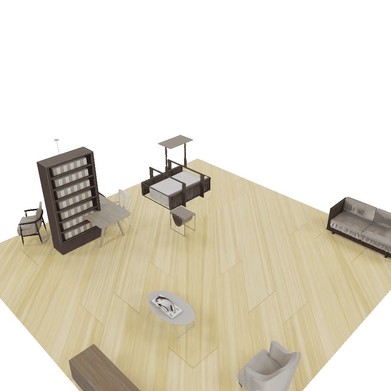}\hfill
    \includegraphics[width=.15\linewidth]{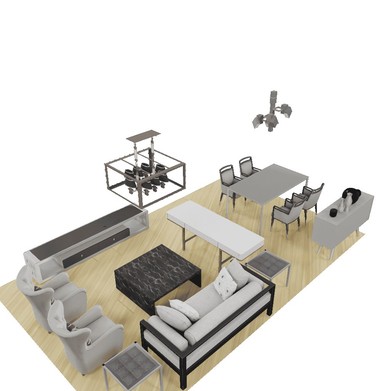}
    \includegraphics[width=.15\linewidth]{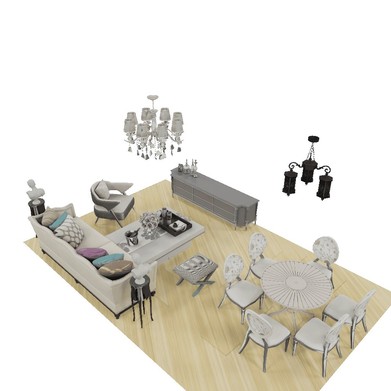}\hfill
    \includegraphics[width=.15\linewidth]{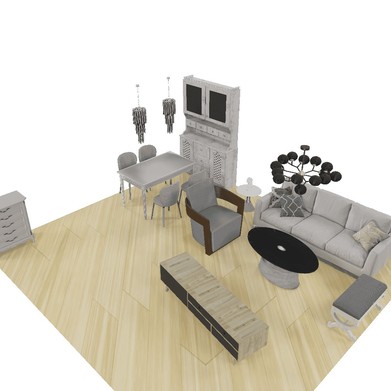}
  \end{subfigure}

  \begin{subfigure}{\textwidth} 
    \centering
    \includegraphics[width=.15\linewidth]{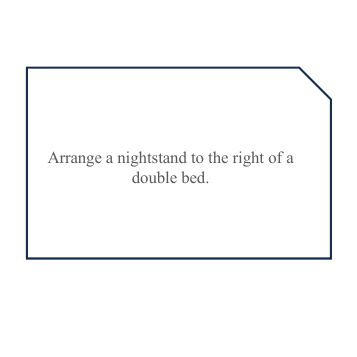}\hfill
    \includegraphics[width=.15\linewidth]{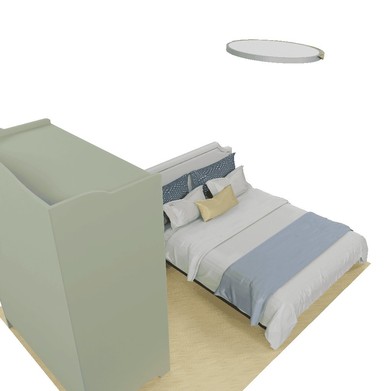}\hfill
    \includegraphics[width=.15\linewidth]{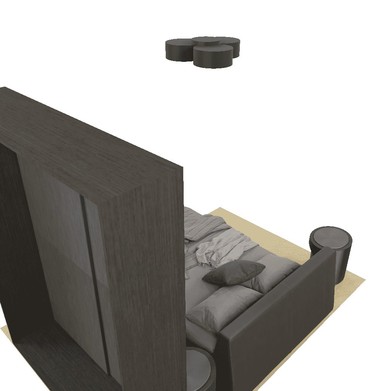}\hfill
    \includegraphics[width=.15\linewidth]{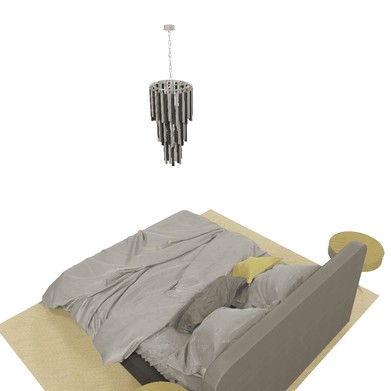}\hfill
    \includegraphics[width=.15\linewidth]{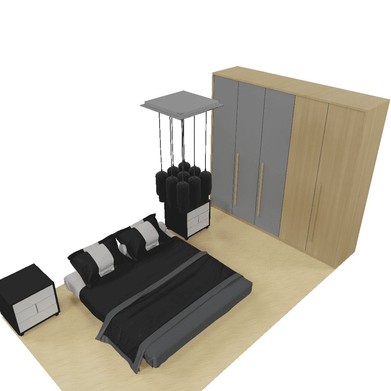}
  \end{subfigure}

  \begin{subfigure}{\textwidth} 
    \centering
    \includegraphics[width=.15\linewidth]{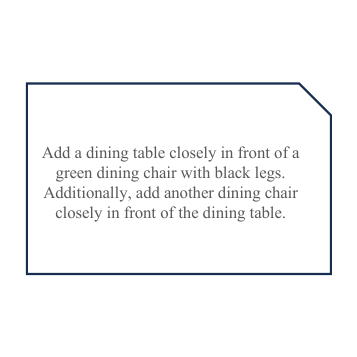}\hfill
    \includegraphics[width=.15\linewidth]{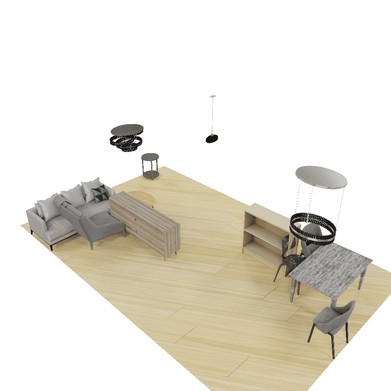}\hfill
    \includegraphics[width=.15\linewidth]{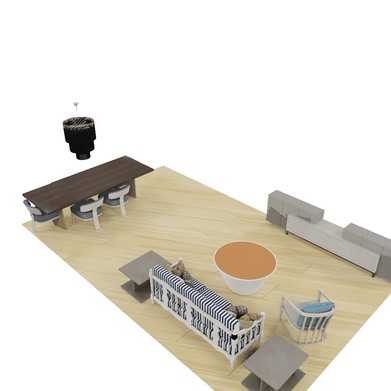}\hfill
    \includegraphics[width=.15\linewidth]{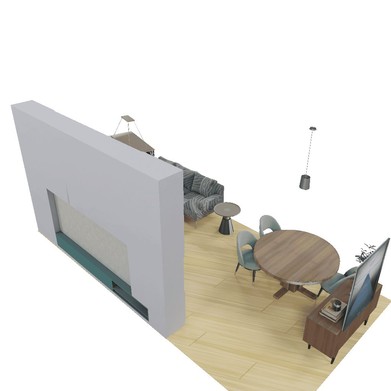}\hfill
    \includegraphics[width=.15\linewidth]{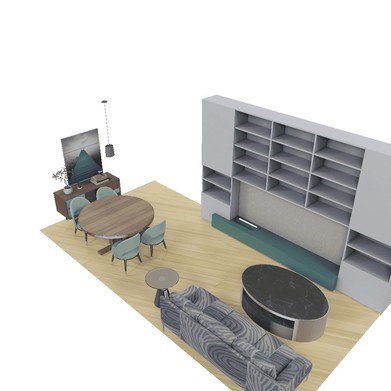}
  \end{subfigure}

    \begin{subfigure}{\textwidth} 
    \centering
    \includegraphics[width=.15\linewidth]{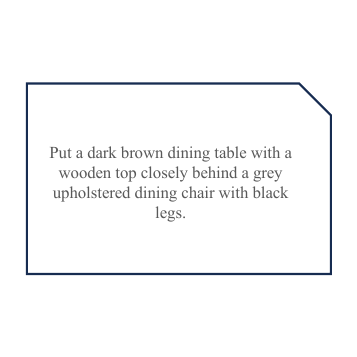}\hfill
    \includegraphics[width=.15\linewidth]{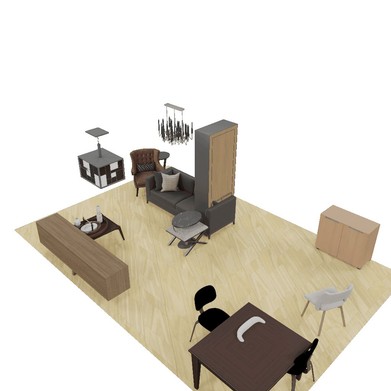}\hfill
    \includegraphics[width=.15\linewidth]{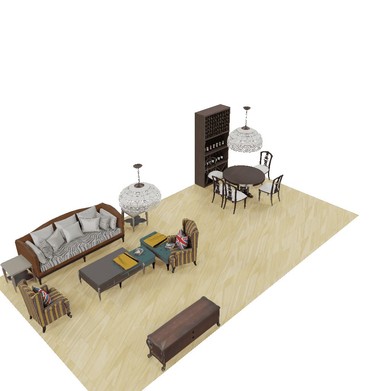}\hfill
    \includegraphics[width=.15\linewidth]{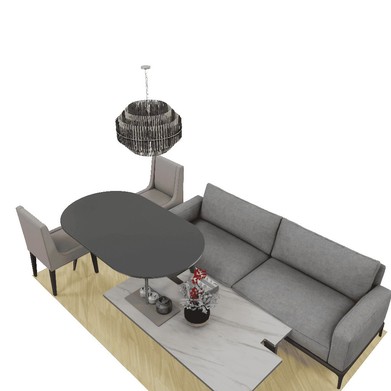}\hfill
    \includegraphics[width=.15\linewidth]{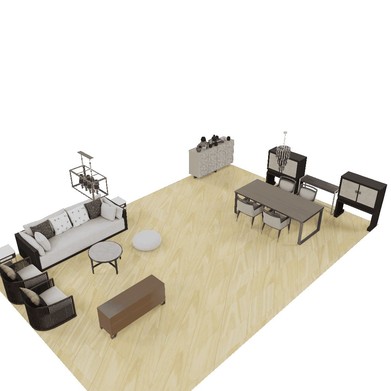}
  \end{subfigure}

  \begin{subfigure}{\textwidth} 
    \centering
    \includegraphics[width=.15\linewidth]{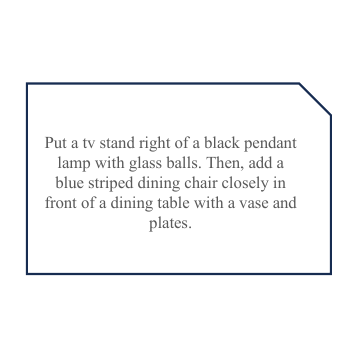}\hfill
    \includegraphics[width=.15\linewidth]{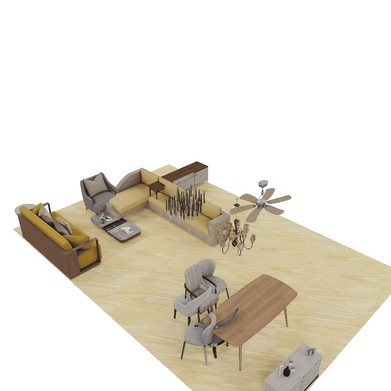}\hfill
    \includegraphics[width=.15\linewidth]{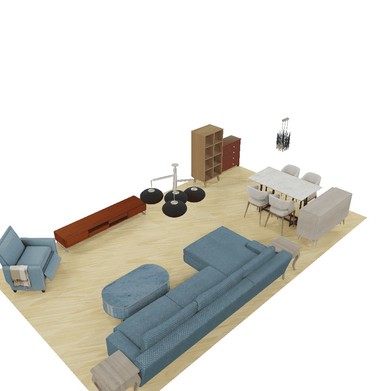}\hfill
    \includegraphics[width=.15\linewidth]{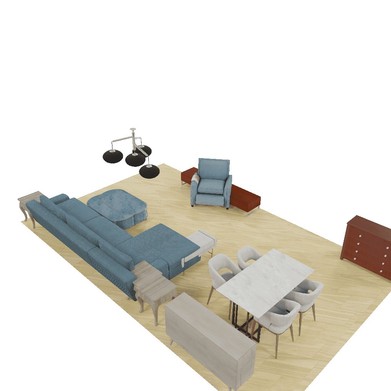}\hfill
    \includegraphics[width=.15\linewidth]{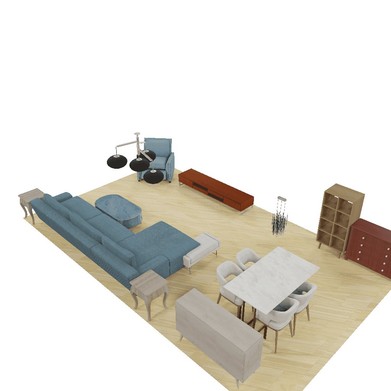}
  \end{subfigure}

\makebox[.15\linewidth][c]{(a) Instruction}\hfill
\makebox[.15\linewidth][c]{(b) ATISS}\hfill
\makebox[.15\linewidth][c]{(c) DiffuScene}\hfill
\makebox[.15\linewidth][c]{(d) InstructScene}\hfill
\makebox[.15\linewidth][c]{(e) Ours}

  \caption{ Comparison of qualitative results for text-to-scene generation.}
  \label{text-to-scene}
\end{figure}
\newpage
\begin{figure}
  \centering
  
  \begin{subfigure}{\textwidth}
    \centering
    \includegraphics[width=.19\linewidth]{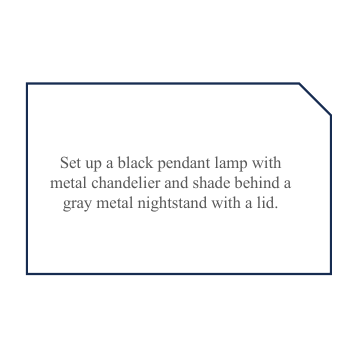}\hfill
    \includegraphics[width=.19\linewidth]{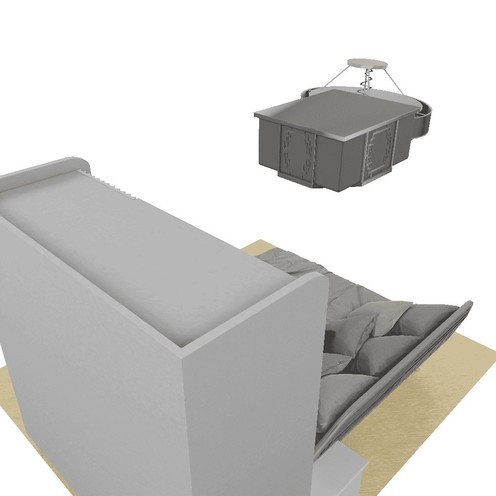}\hfill
    \includegraphics[width=.19\linewidth]{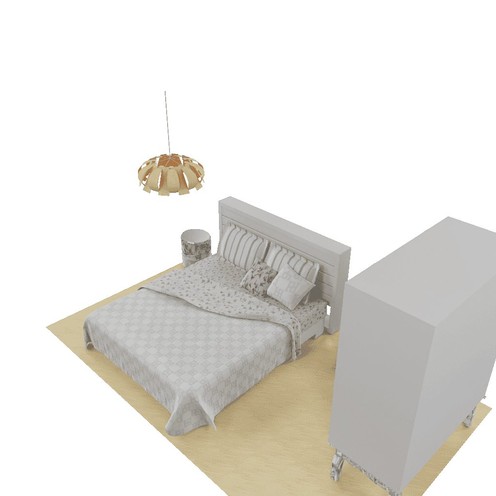}\hfill
    \includegraphics[width=.19\linewidth]{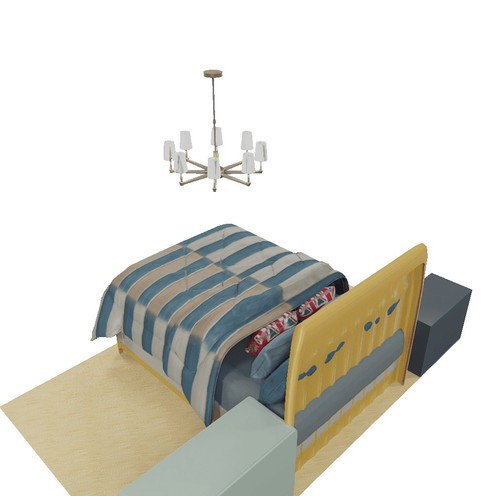}\hfill
    \includegraphics[width=.19\linewidth]{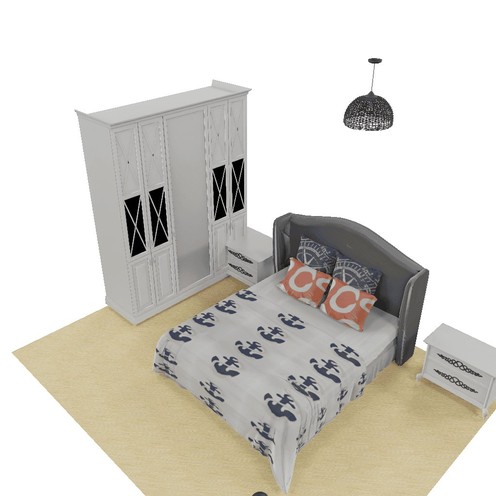}
  \end{subfigure}
  
  \vspace{2pt}

 \begin{subfigure}{\textwidth}
    \centering
    \includegraphics[width=.19\linewidth]{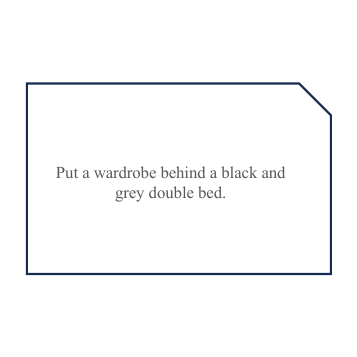}\hfill
    \includegraphics[width=.19\linewidth]{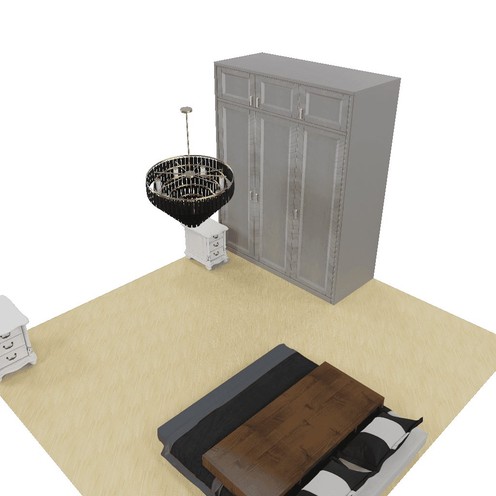}\hfill
    \includegraphics[width=.19\linewidth]{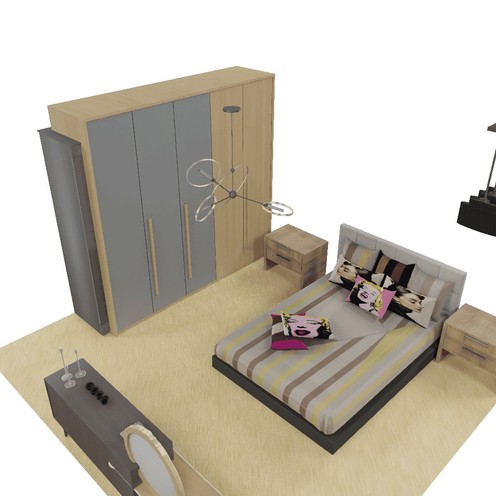}\hfill
    \includegraphics[width=.19\linewidth]{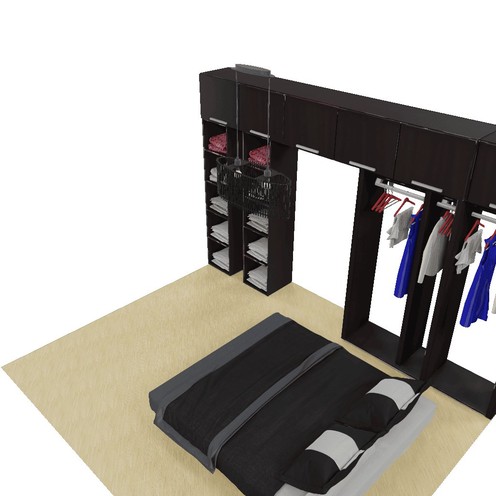}\hfill
    \includegraphics[width=.19\linewidth]{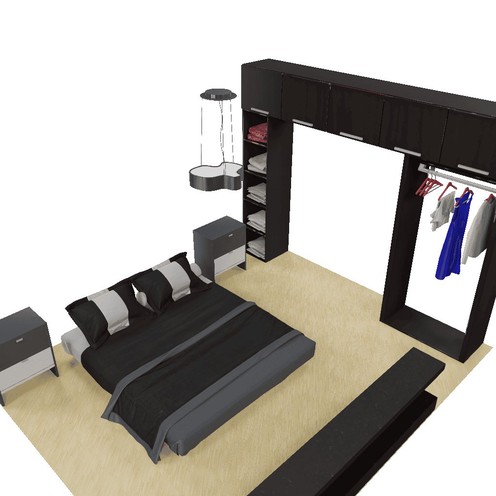}
  \end{subfigure}
  
  \vspace{2pt}
  \begin{subfigure}{\textwidth}
    \centering
    \includegraphics[width=.19\linewidth]{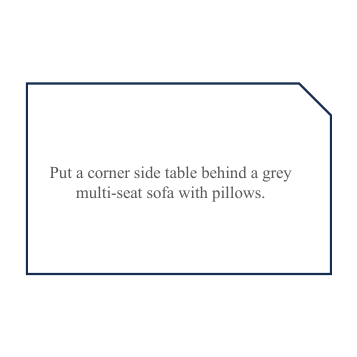}\hfill
    \includegraphics[width=.19\linewidth]{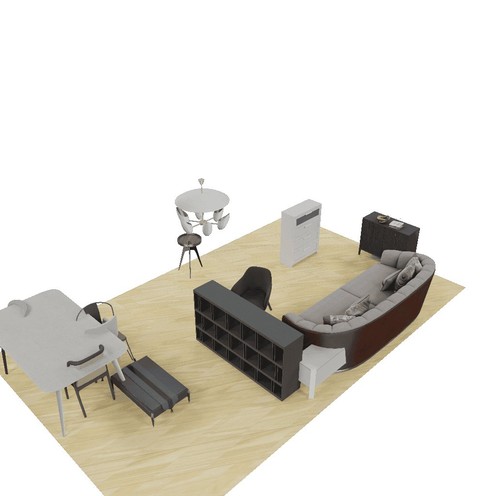}\hfill
    \includegraphics[width=.19\linewidth]{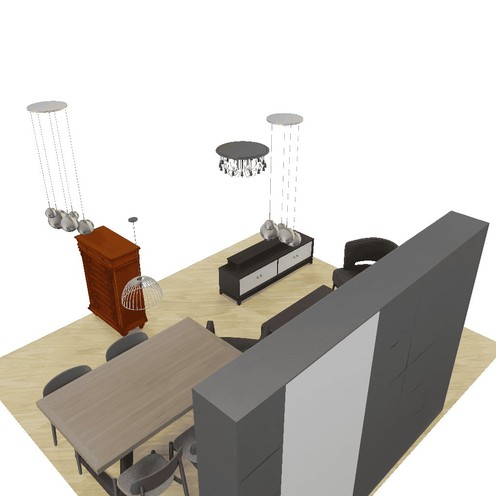}\hfill
    \includegraphics[width=.19\linewidth]{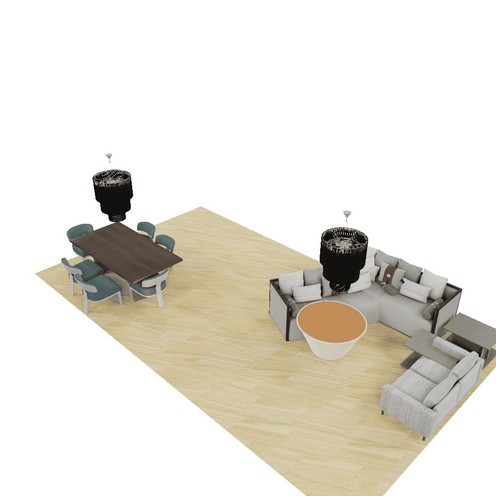}
    \includegraphics[width=.19\linewidth]{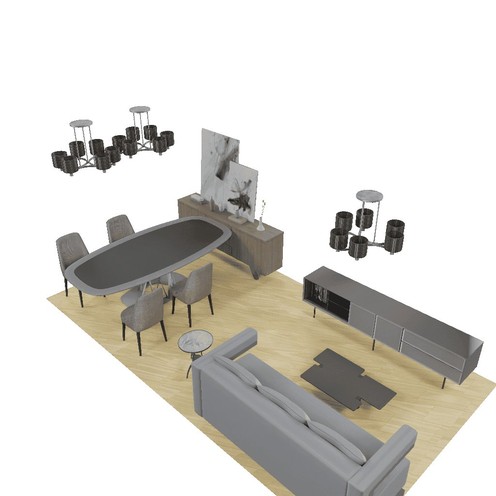}
  \end{subfigure}

  \vspace{2pt}
  \begin{subfigure}{\textwidth}
    \centering
    \includegraphics[width=.19\linewidth]{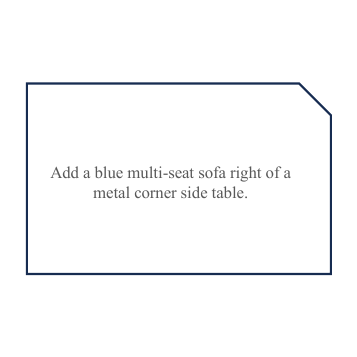}\hfill
    \includegraphics[width=.19\linewidth]{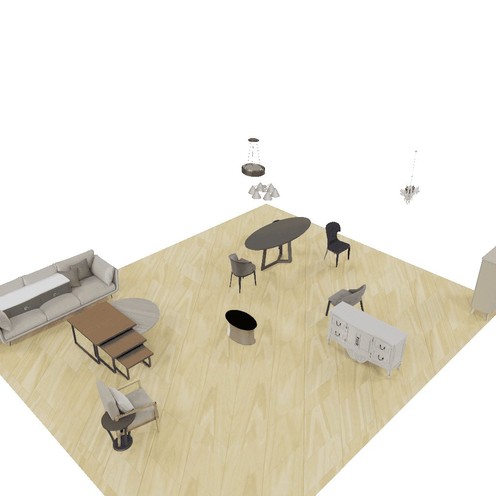}\hfill
    \includegraphics[width=.19\linewidth]{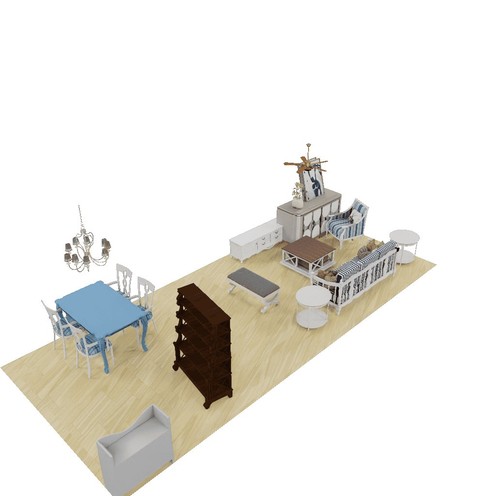}
    \includegraphics[width=.19\linewidth]{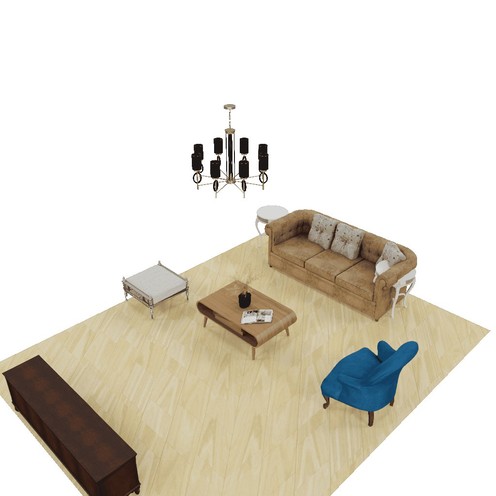}\hfill
    \includegraphics[width=.19\linewidth]{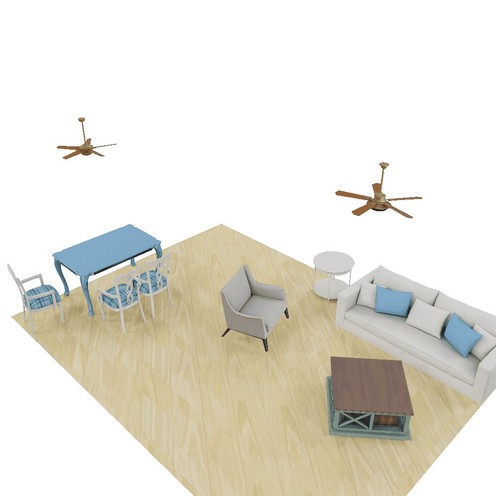}
  \end{subfigure}

  \vspace{2pt}
  \begin{subfigure}{\textwidth}
    \centering
    \includegraphics[width=.19\linewidth]{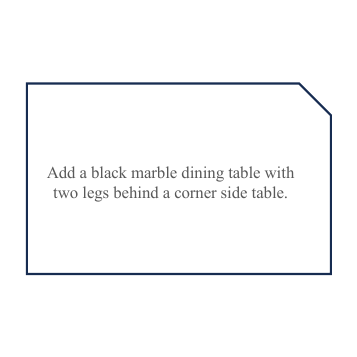}\hfill
    \includegraphics[width=.19\linewidth]{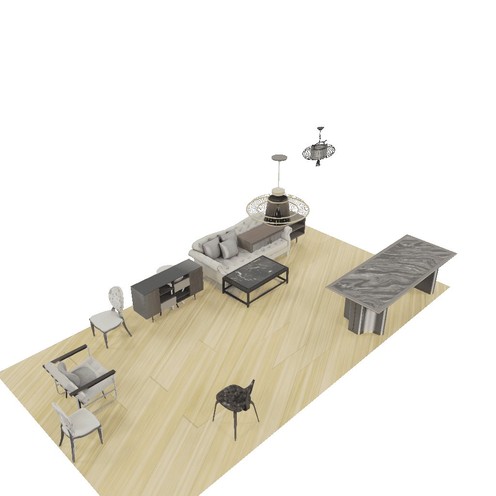}\hfill
    \includegraphics[width=.19\linewidth]{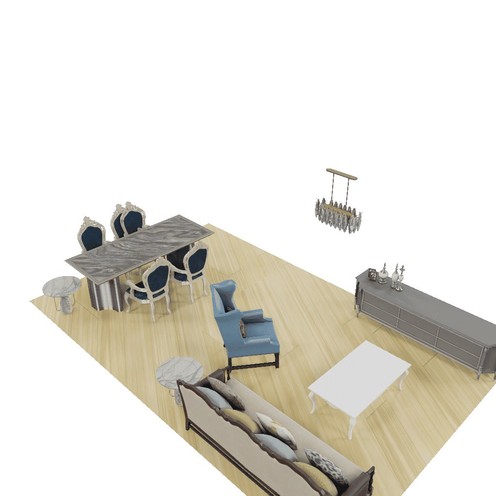}\hfill
    \includegraphics[width=.19\linewidth]{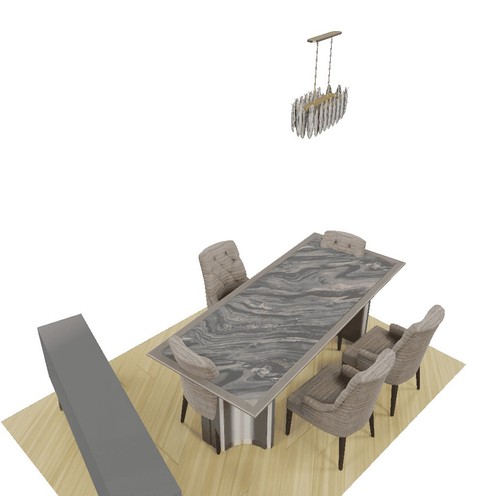}\hfill
    \includegraphics[width=.19\linewidth]{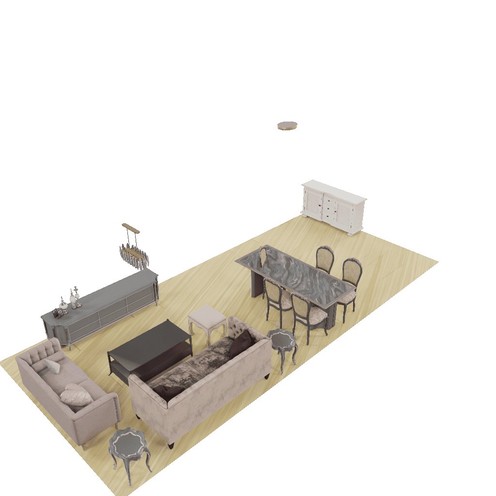}
  \end{subfigure}

  \vspace{2pt}
  \begin{subfigure}{\textwidth}
    \centering
    \includegraphics[width=.19\linewidth]{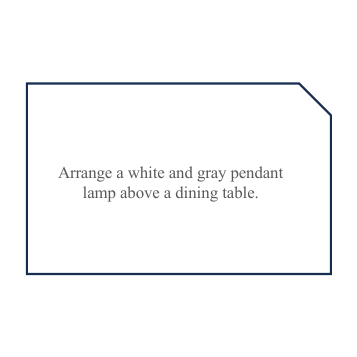}\hfill
    \includegraphics[width=.19\linewidth]{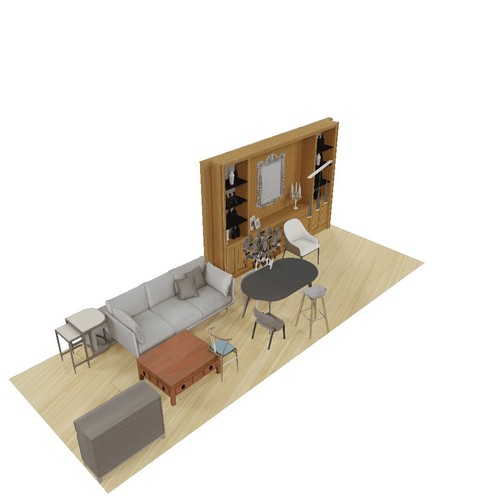}\hfill
    \includegraphics[width=.19\linewidth]{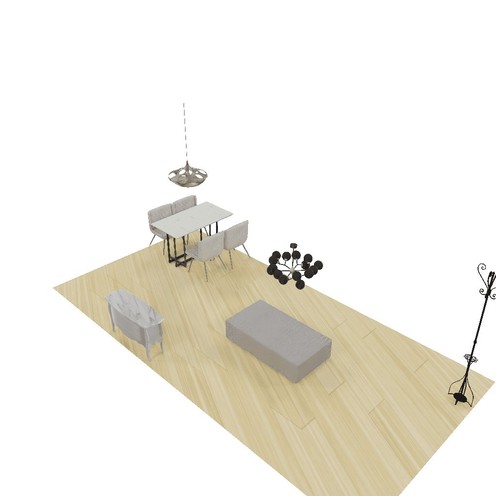}\hfill
    \includegraphics[width=.19\linewidth]{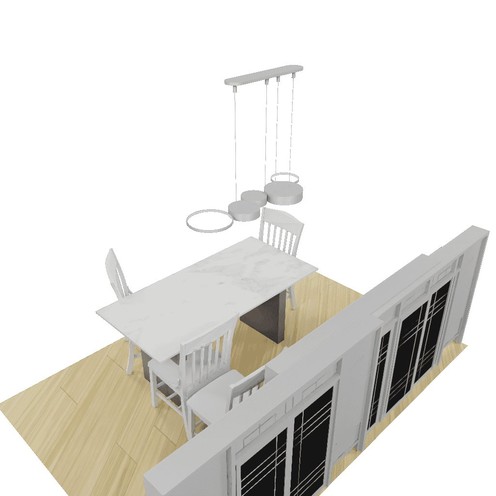}\hfill
    \includegraphics[width=.19\linewidth]{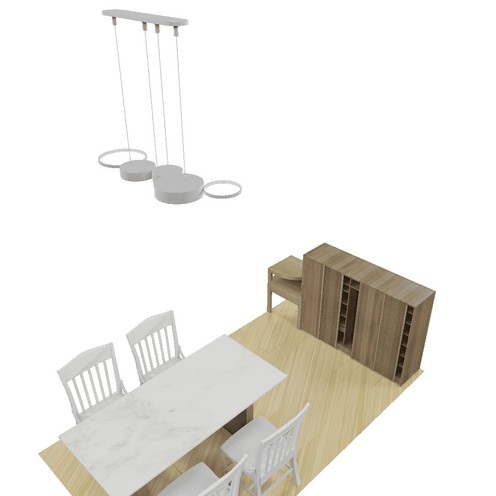}
  \end{subfigure}

  \makebox[.19\linewidth][c]{(a) Instruction}\hfill
\makebox[.19\linewidth][c]{(b) ATISS}\hfill
\makebox[.19\linewidth][c]{(c) DiffuScene}\hfill
\makebox[.19\linewidth][c]{(d) InstructScene}\hfill
\makebox[.19\linewidth][c]{(e) Ours}

  \caption{
  Comparison of qualitative results for 1 relation constraint in instruction.
  }
  \label{tri1}
\end{figure}

\newpage
\begin{figure}
  \centering
  
  \begin{subfigure}{\textwidth}
    \centering
    \includegraphics[width=.19\linewidth]{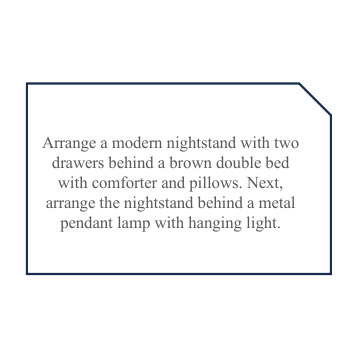}\hfill
    \includegraphics[width=.19\linewidth]{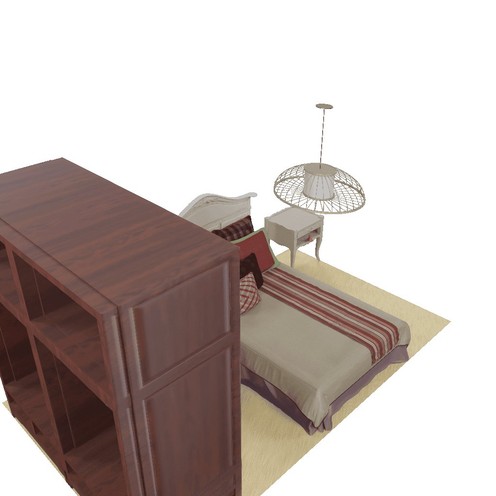}\hfill
    \includegraphics[width=.19\linewidth]{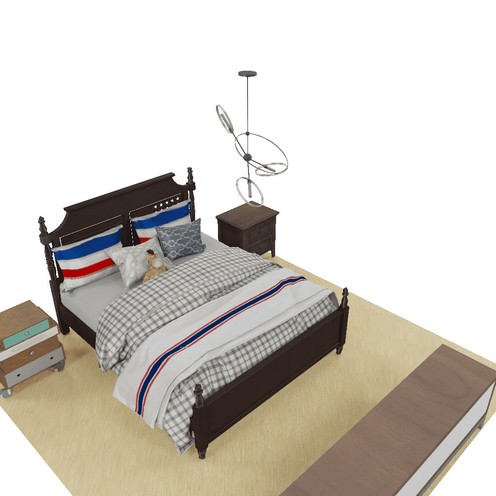}\hfill
    \includegraphics[width=.19\linewidth]{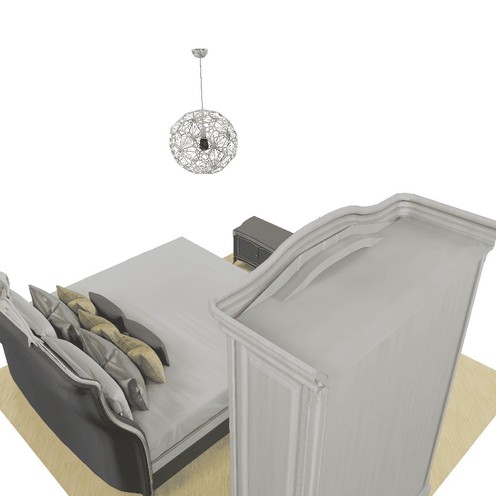}\hfill
    \includegraphics[width=.19\linewidth]{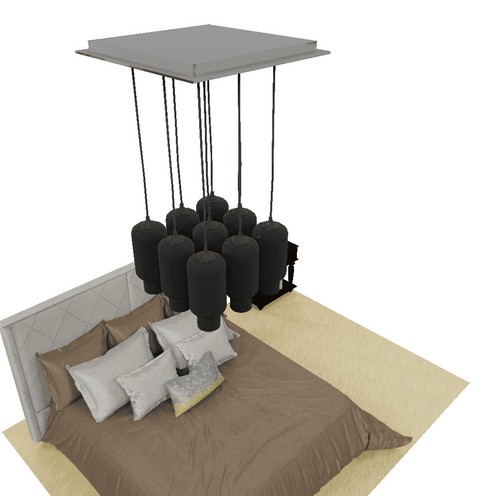}
  \end{subfigure}
  
  \vspace{2pt}

 \begin{subfigure}{\textwidth}
    \centering
    \includegraphics[width=.19\linewidth]{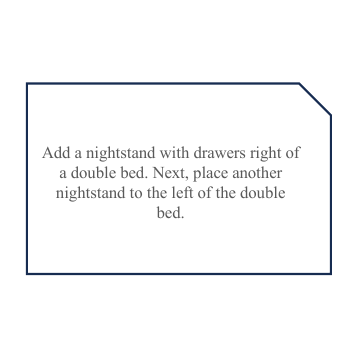}\hfill
    \includegraphics[width=.19\linewidth]{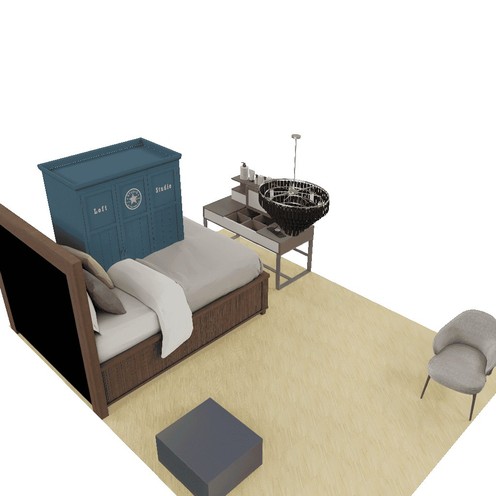}\hfill
    \includegraphics[width=.19\linewidth]{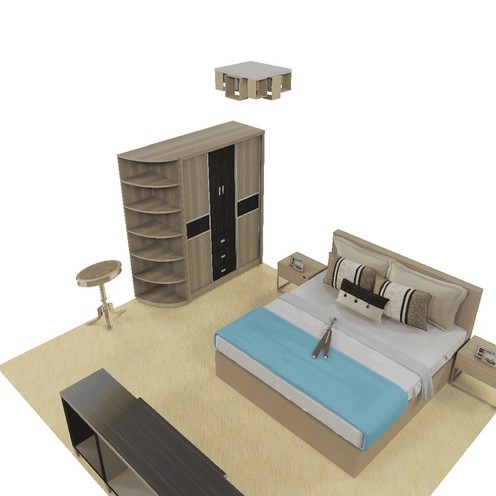}\hfill
    \includegraphics[width=.19\linewidth]{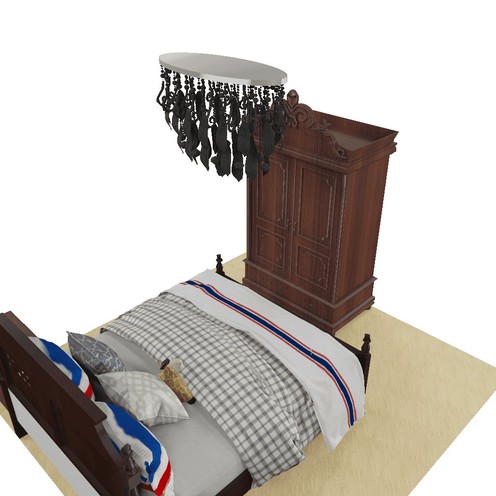}\hfill
    \includegraphics[width=.19\linewidth]{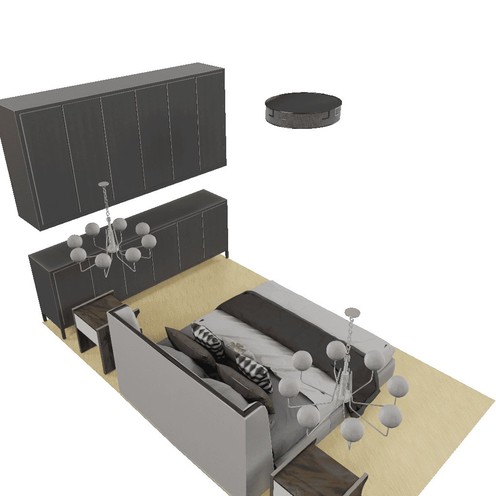}
  \end{subfigure}
  
  \vspace{2pt}
  \begin{subfigure}{\textwidth}
    \centering
    \includegraphics[width=.19\linewidth]{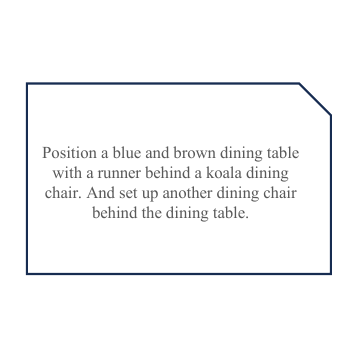}\hfill
    \includegraphics[width=.19\linewidth]{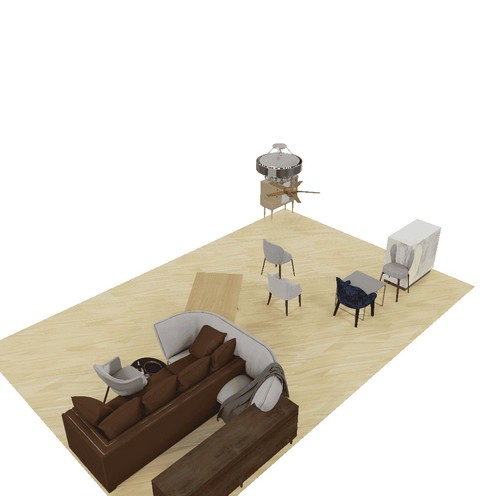}\hfill
    \includegraphics[width=.19\linewidth]{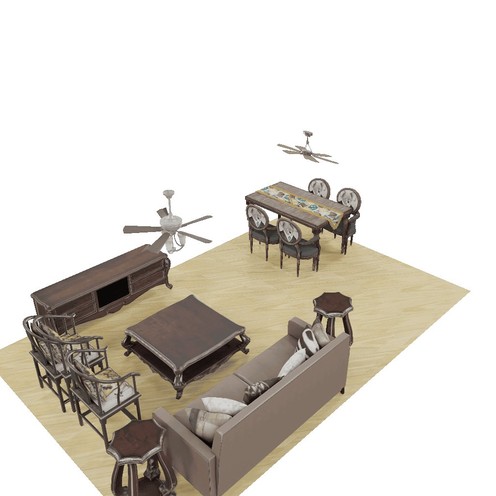}\hfill
    \includegraphics[width=.19\linewidth]{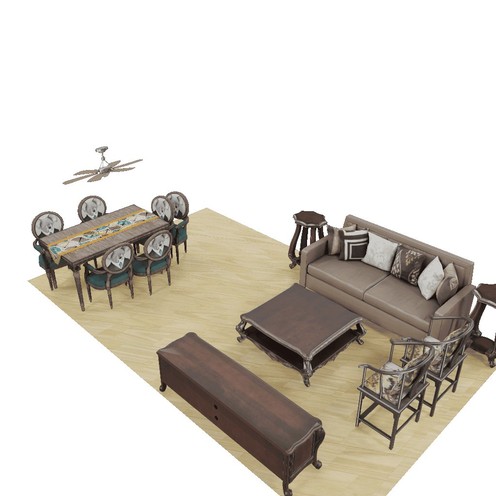}
    \includegraphics[width=.19\linewidth]{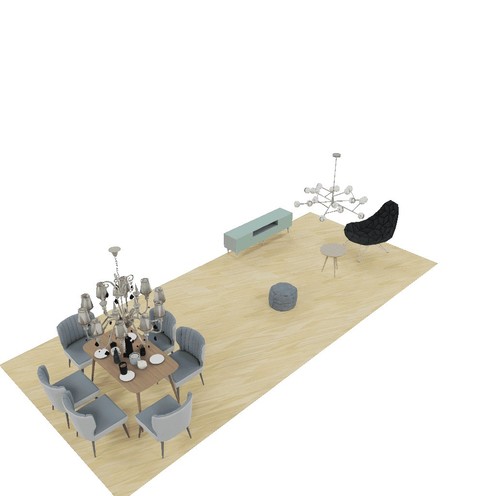}
  \end{subfigure}

  \vspace{2pt}
  \begin{subfigure}{\textwidth}
    \centering
    \includegraphics[width=.19\linewidth]{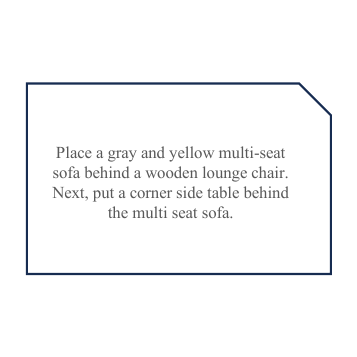}\hfill
    \includegraphics[width=.19\linewidth]{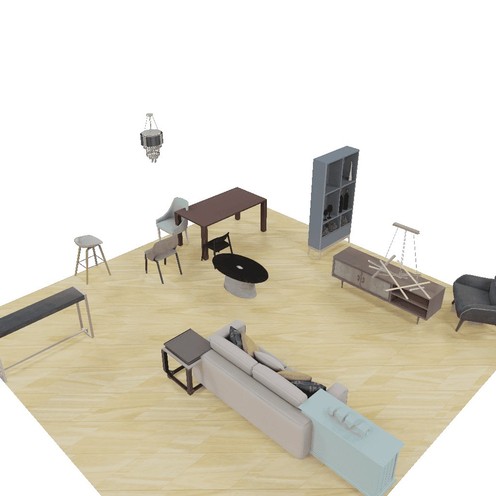}\hfill
    \includegraphics[width=.19\linewidth]{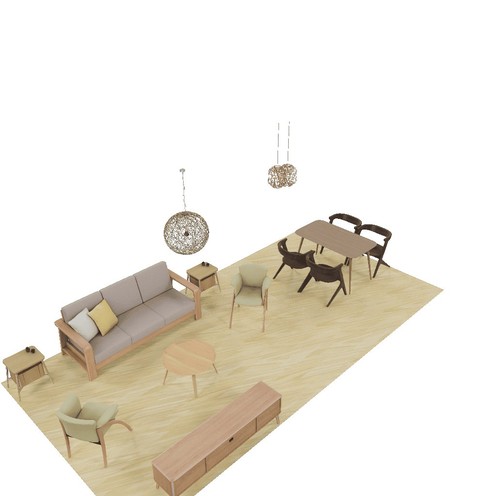}
    \includegraphics[width=.19\linewidth]{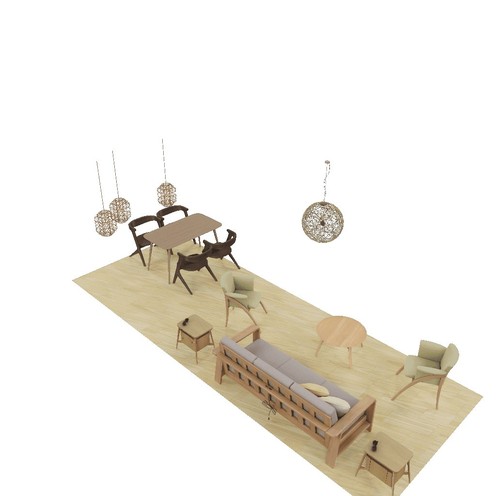}\hfill
    \includegraphics[width=.19\linewidth]{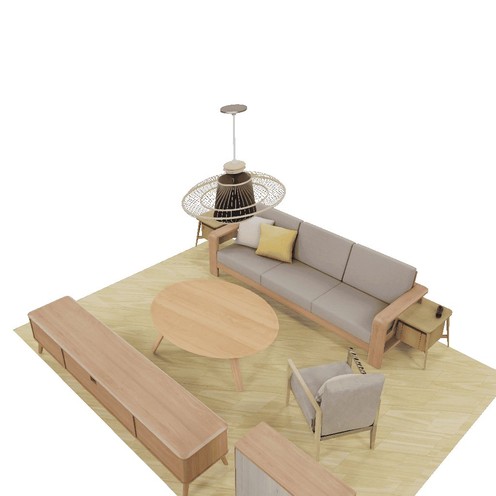}
  \end{subfigure}

  \vspace{2pt}
  \begin{subfigure}{\textwidth}
    \centering
    \includegraphics[width=.19\linewidth]{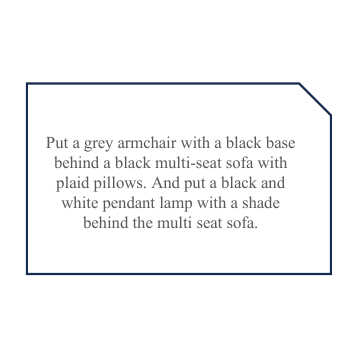}\hfill
    \includegraphics[width=.19\linewidth]{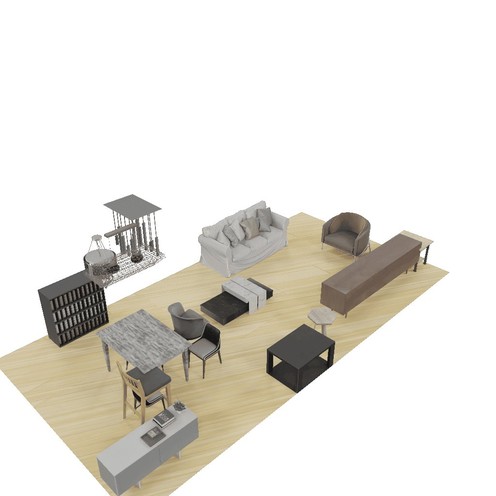}\hfill
    \includegraphics[width=.19\linewidth]{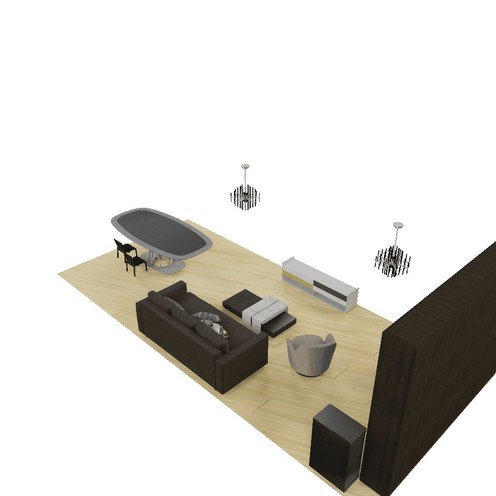}\hfill
    \includegraphics[width=.19\linewidth]{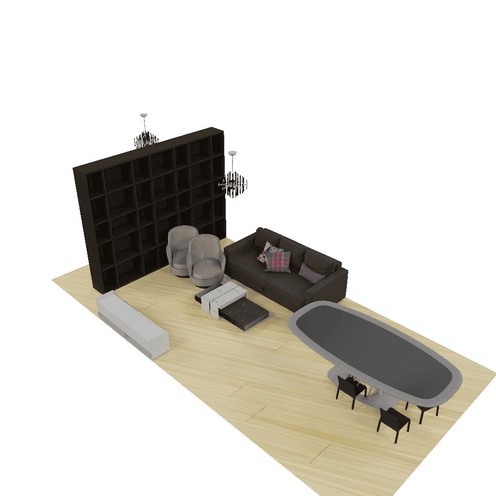}\hfill
    \includegraphics[width=.19\linewidth]{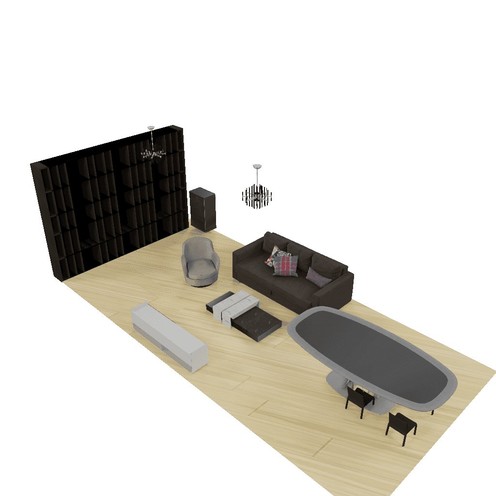}
  \end{subfigure}

  \vspace{2pt}
  \begin{subfigure}{\textwidth}
    \centering
    \includegraphics[width=.19\linewidth]{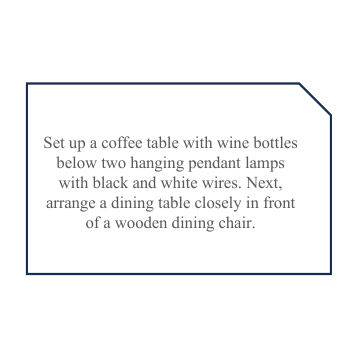}\hfill
    \includegraphics[width=.19\linewidth]{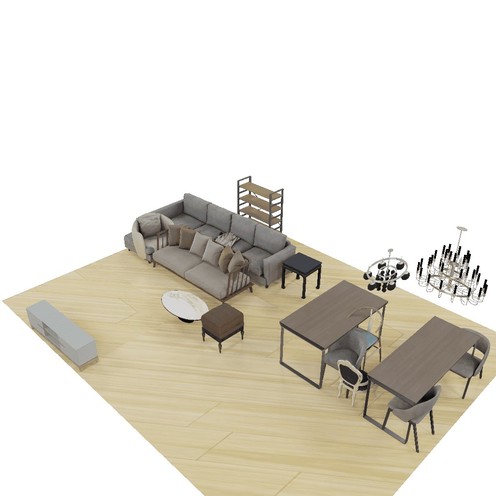}\hfill
    \includegraphics[width=.19\linewidth]{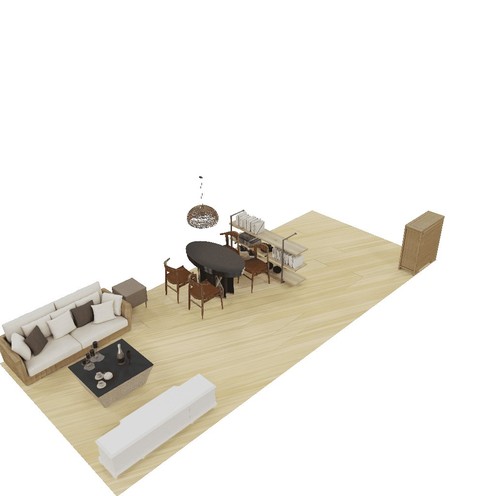}\hfill
    \includegraphics[width=.19\linewidth]{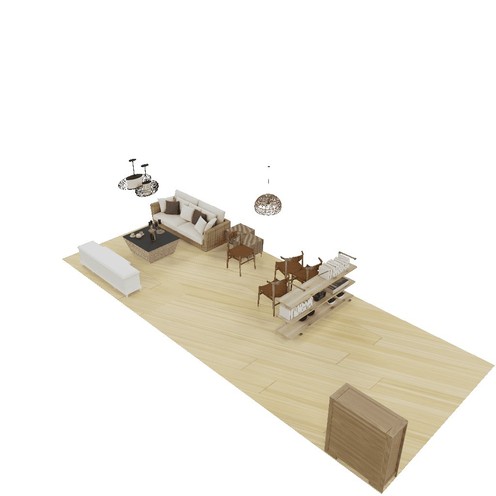}\hfill
    \includegraphics[width=.19\linewidth]{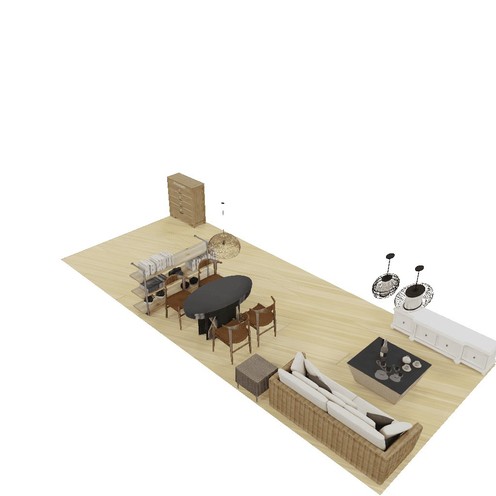}
  \end{subfigure}
  \makebox[.19\linewidth][c]{(a) Instruction}\hfill
\makebox[.19\linewidth][c]{(b) ATISS}\hfill
\makebox[.19\linewidth][c]{(c) DiffuScene}\hfill
\makebox[.19\linewidth][c]{(d) InstructScene}\hfill
\makebox[.19\linewidth][c]{(e) Ours}
  \caption{
  Comparison of qualitative results for 2 relation constraints in instruction.
  }
  \label{tri2}
\end{figure}

\newpage
\begin{figure}
  \centering
  
  \begin{subfigure}{\textwidth}
    \centering
    \includegraphics[width=.19\linewidth]{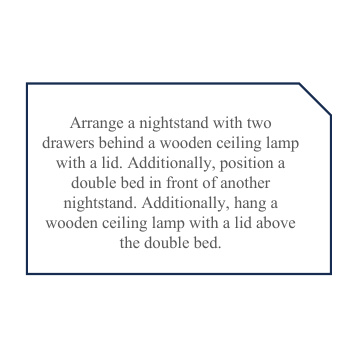}\hfill
    \includegraphics[width=.19\linewidth]{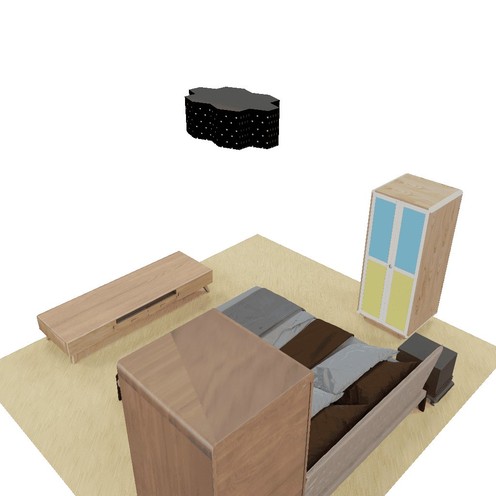}\hfill
    \includegraphics[width=.19\linewidth]{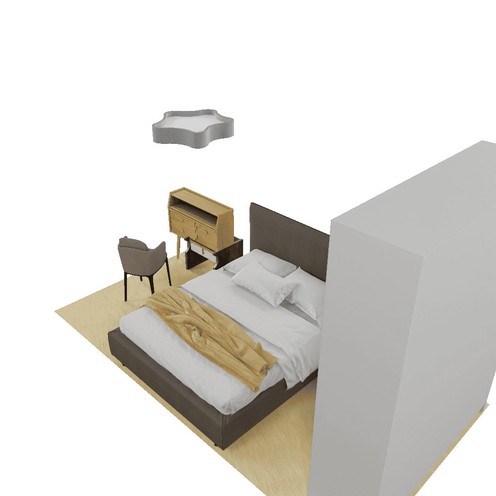}\hfill
    \includegraphics[width=.19\linewidth]{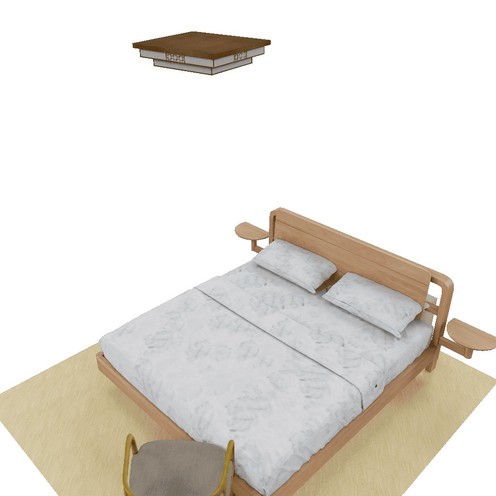}\hfill
    \includegraphics[width=.19\linewidth]{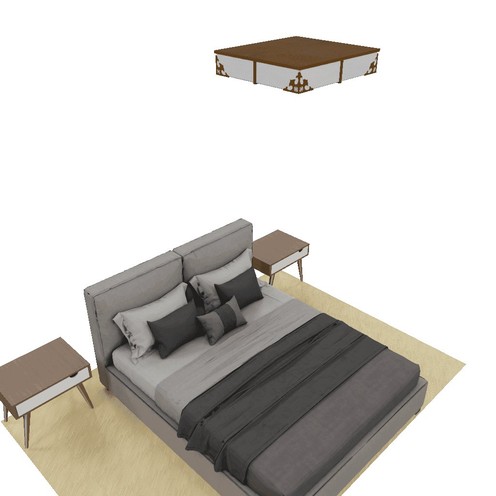}
  \end{subfigure}
  
  \vspace{2pt}

 \begin{subfigure}{\textwidth}
    \centering
    \includegraphics[width=.19\linewidth]{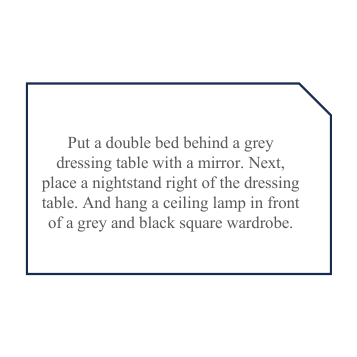}\hfill
    \includegraphics[width=.19\linewidth]{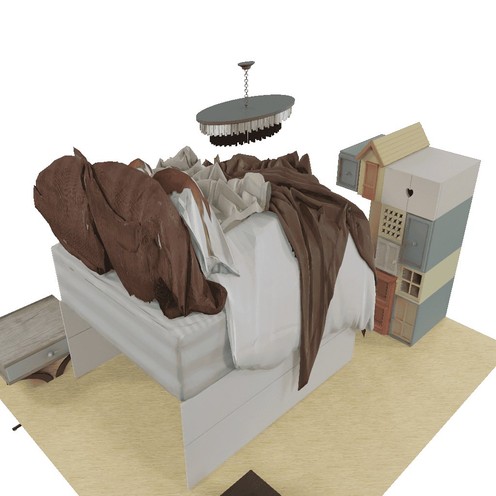}\hfill
    \includegraphics[width=.19\linewidth]{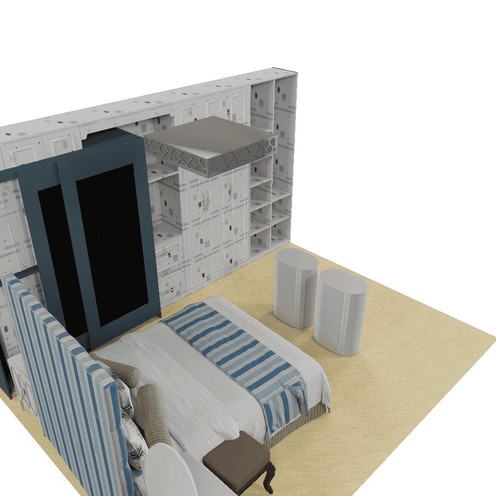}\hfill
    \includegraphics[width=.19\linewidth]{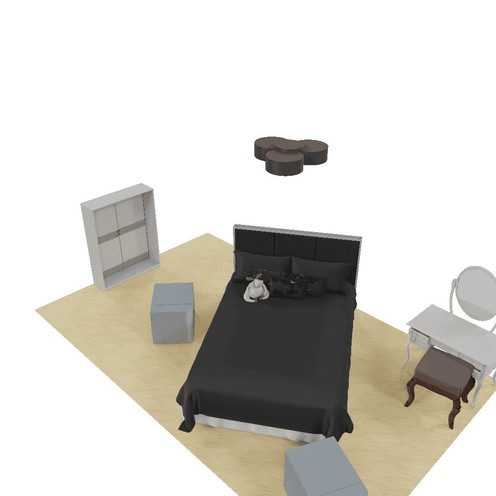}\hfill
    \includegraphics[width=.19\linewidth]{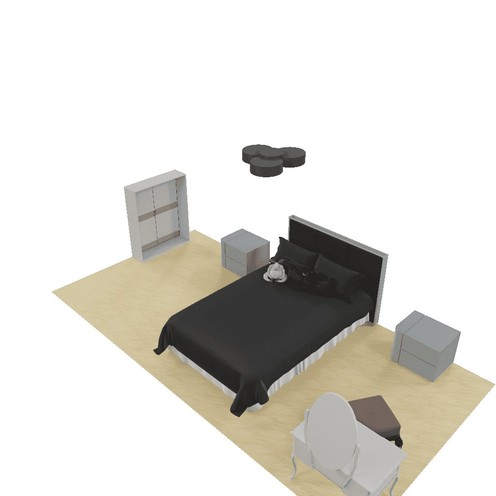}
  \end{subfigure}

  \vspace{2pt}
  \begin{subfigure}{\textwidth}
    \centering
    \includegraphics[width=.19\linewidth]{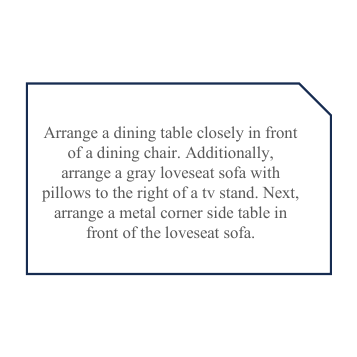}\hfill
    \includegraphics[width=.19\linewidth]{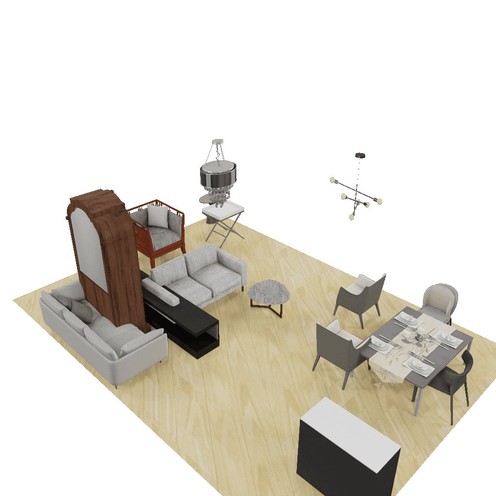}\hfill
    \includegraphics[width=.19\linewidth]{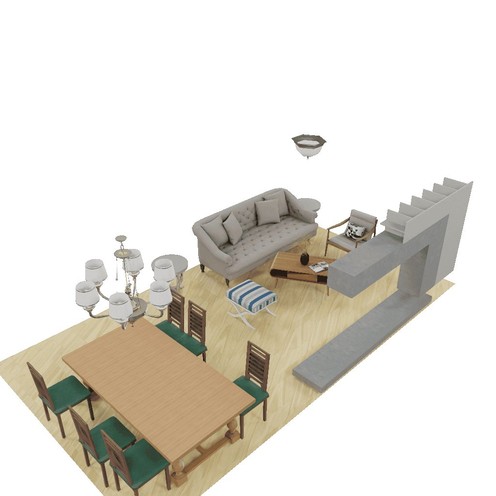}\hfill
    \includegraphics[width=.19\linewidth]{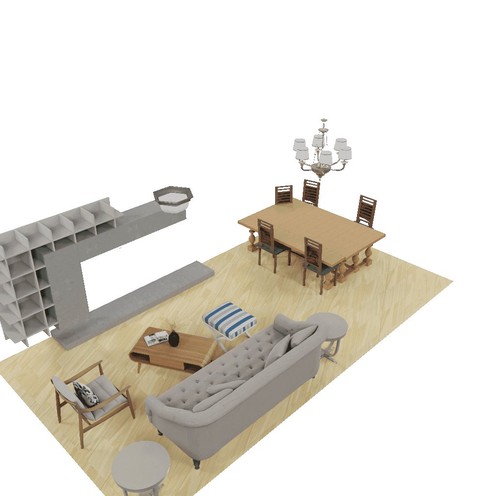}
    \includegraphics[width=.19\linewidth]{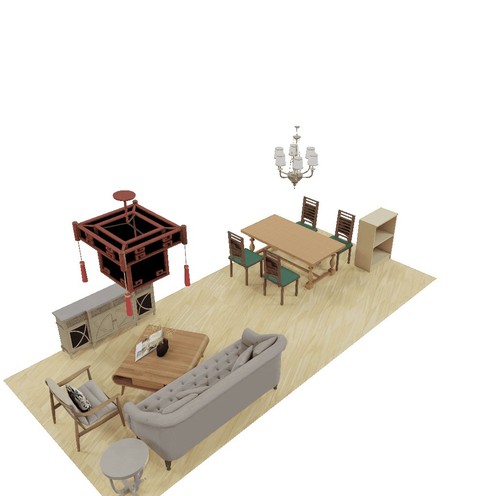}
  \end{subfigure}

  \vspace{2pt}
  \begin{subfigure}{\textwidth}
    \centering
    \includegraphics[width=.19\linewidth]{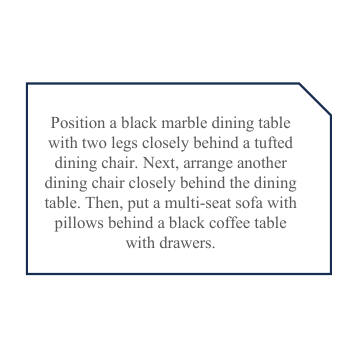}\hfill
    \includegraphics[width=.19\linewidth]{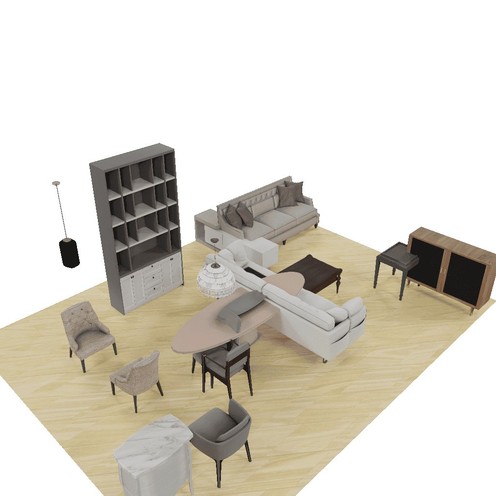}\hfill
    \includegraphics[width=.19\linewidth]{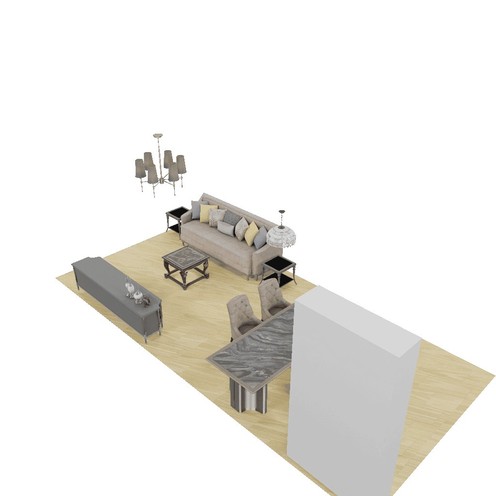}
    \includegraphics[width=.19\linewidth]{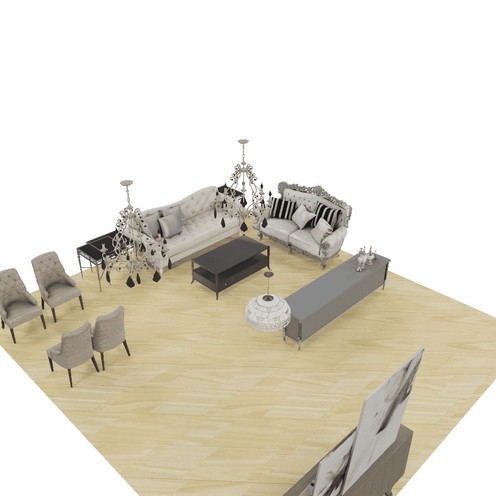}\hfill
    \includegraphics[width=.19\linewidth]{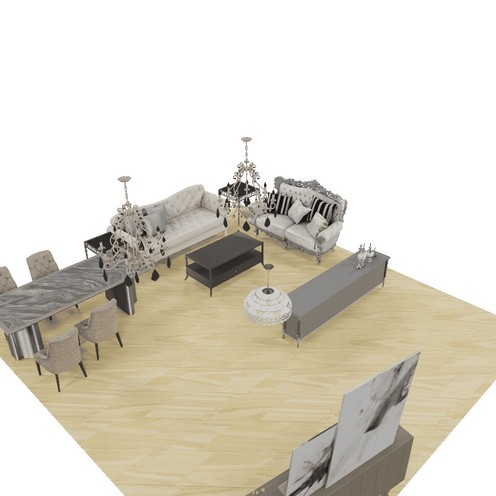}
  \end{subfigure}

  \vspace{2pt}
  \begin{subfigure}{\textwidth}
    \centering
    \includegraphics[width=.19\linewidth]{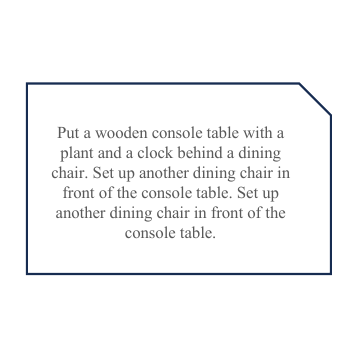}\hfill
    \includegraphics[width=.19\linewidth]{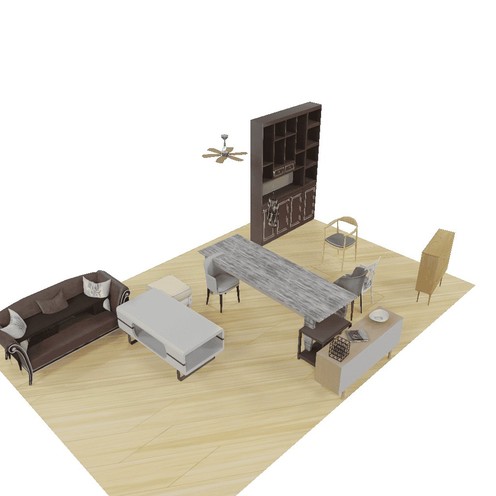}\hfill
    \includegraphics[width=.19\linewidth]{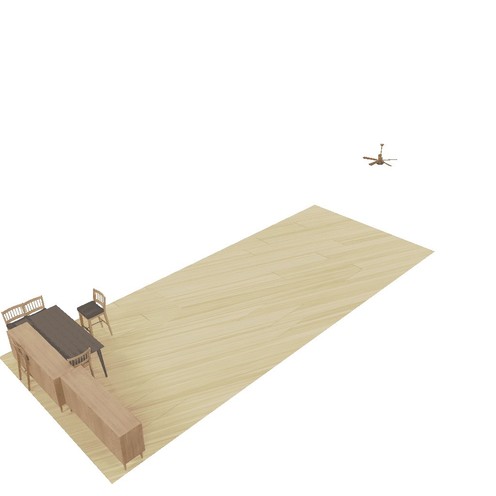}\hfill
    \includegraphics[width=.19\linewidth]{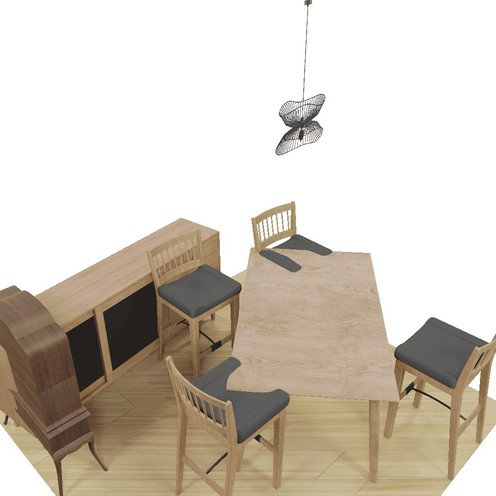}\hfill
    \includegraphics[width=.19\linewidth]{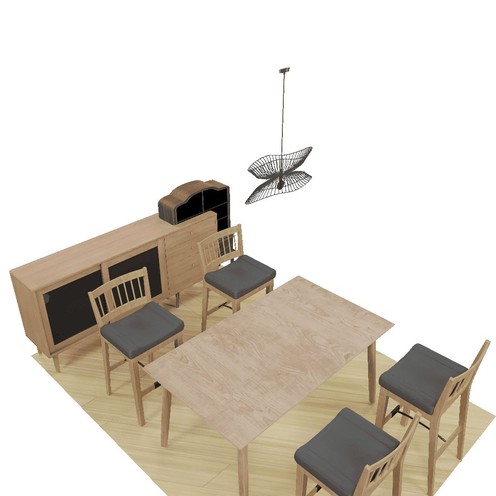}
  \end{subfigure}

  \vspace{2pt}
  \begin{subfigure}{\textwidth}
    \centering
    \includegraphics[width=.19\linewidth]{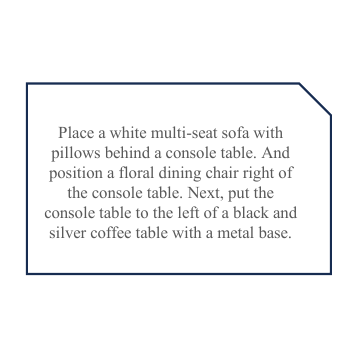}\hfill
    \includegraphics[width=.19\linewidth]{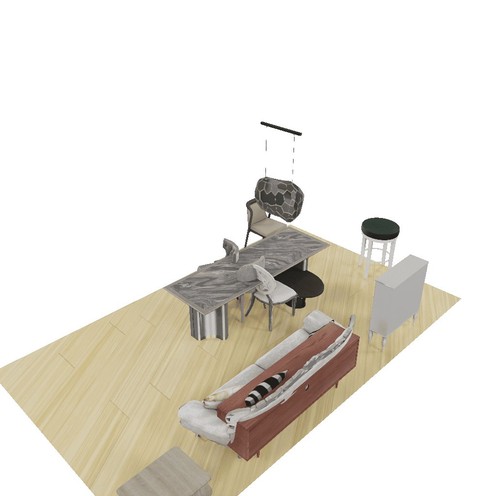}\hfill
    \includegraphics[width=.19\linewidth]{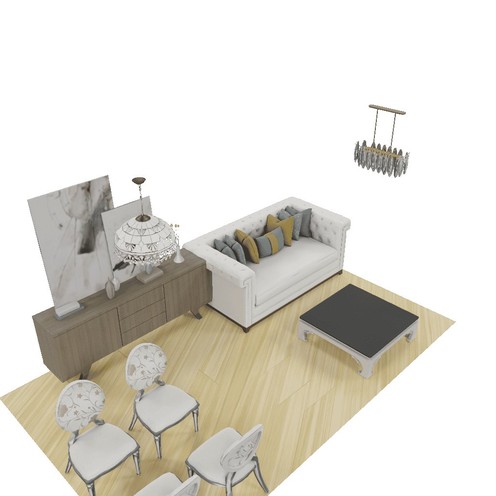}\hfill
    \includegraphics[width=.19\linewidth]{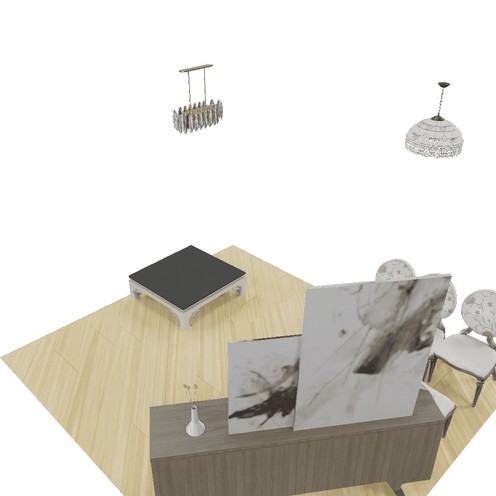}\hfill
    \includegraphics[width=.19\linewidth]{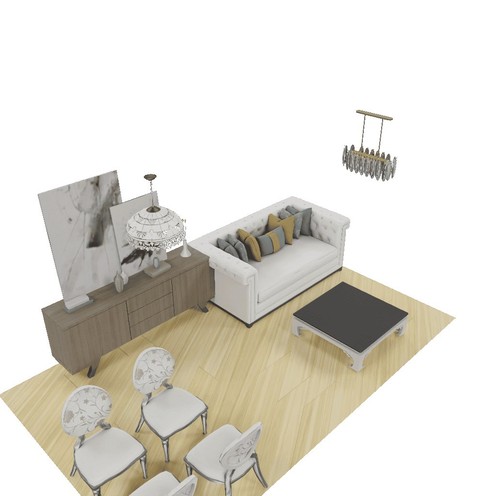}
  \end{subfigure}
  
  \makebox[.19\linewidth][c]{(a) Instruction}\hfill
\makebox[.19\linewidth][c]{(b) ATISS}\hfill
\makebox[.19\linewidth][c]{(c) DiffuScene}\hfill
\makebox[.19\linewidth][c]{(d) InstructScene}\hfill
\makebox[.19\linewidth][c]{(e) Ours}
  \caption{
  Comparison of qualitative results for 3 relation constraints in instruction.
  }
  \label{tri3}
\end{figure}

\newpage
\begin{figure}
  \centering
  
  \begin{subfigure}{\textwidth}
    \centering
    \includegraphics[width=.19\linewidth]{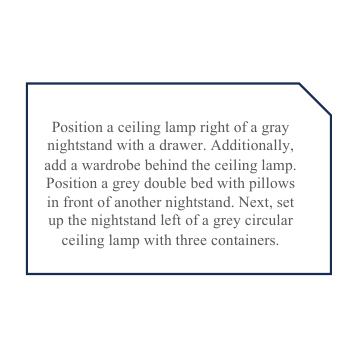}\hfill
    \includegraphics[width=.19\linewidth]{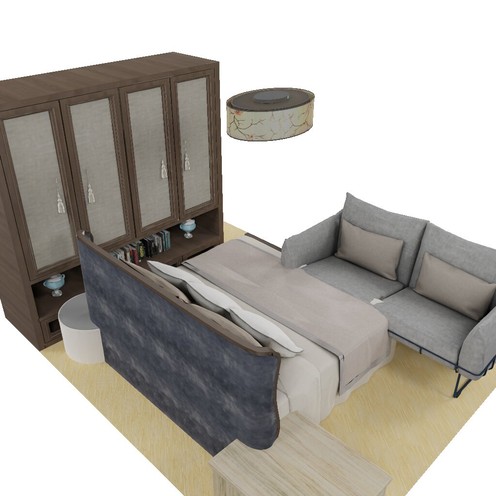}\hfill
    \includegraphics[width=.19\linewidth]{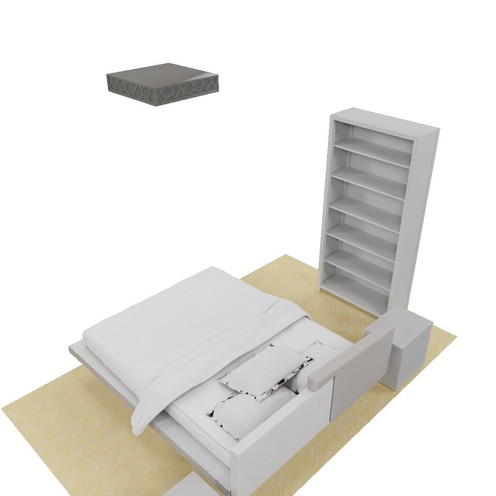}\hfill
    \includegraphics[width=.19\linewidth]{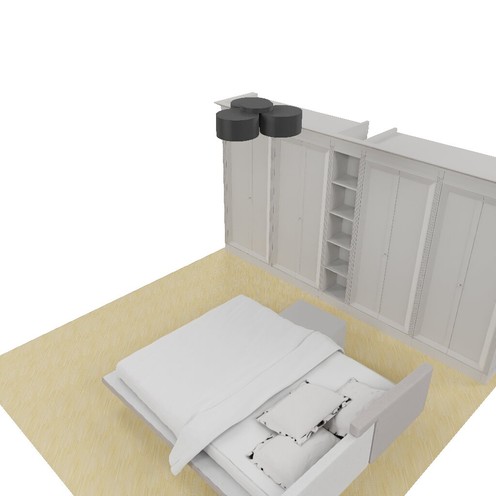}\hfill
    \includegraphics[width=.19\linewidth]{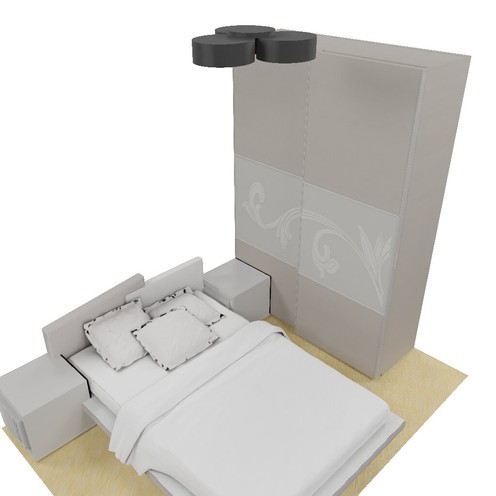}
  \end{subfigure}
  
  \vspace{2pt}

 \begin{subfigure}{\textwidth}
    \centering
    \includegraphics[width=.19\linewidth]{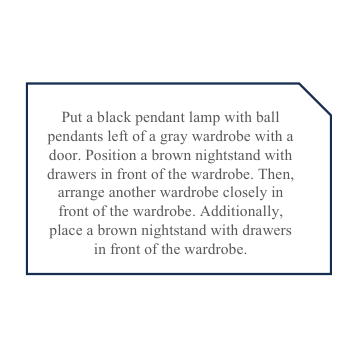}\hfill
    \includegraphics[width=.19\linewidth]{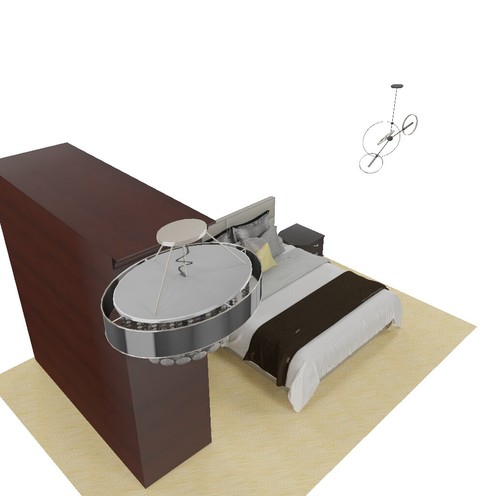}\hfill
    \includegraphics[width=.19\linewidth]{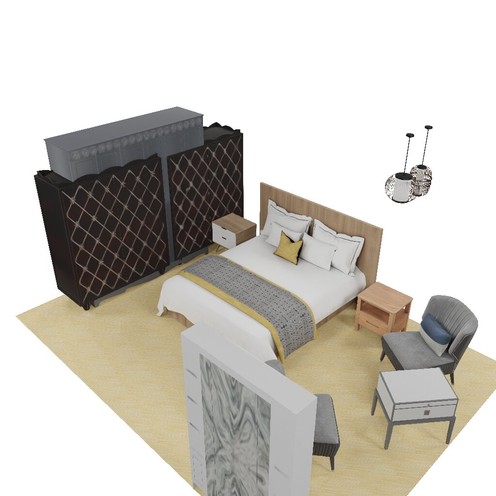}\hfill
    \includegraphics[width=.19\linewidth]{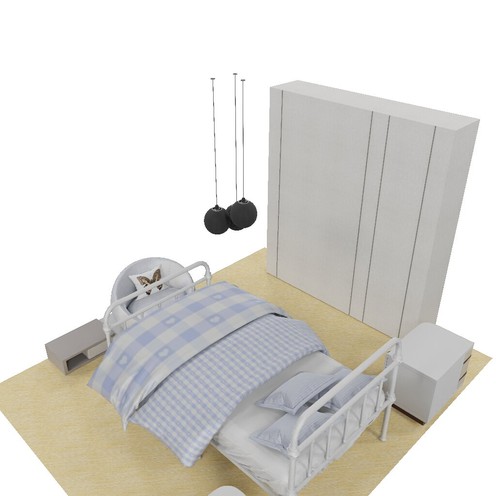}\hfill
    \includegraphics[width=.19\linewidth]{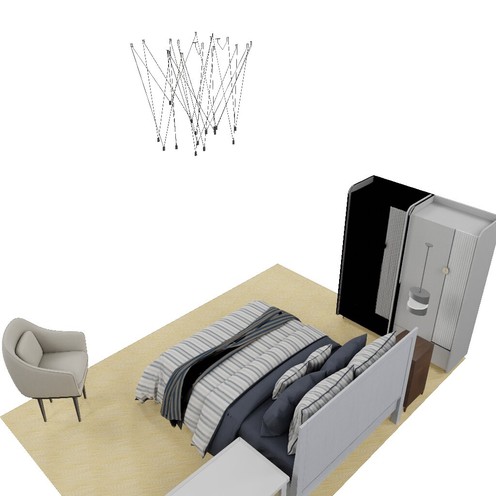}
  \end{subfigure}

  \vspace{2pt}
  \begin{subfigure}{\textwidth}
    \centering
    \includegraphics[width=.19\linewidth]{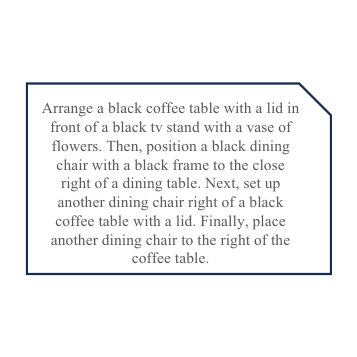}\hfill
    \includegraphics[width=.19\linewidth]{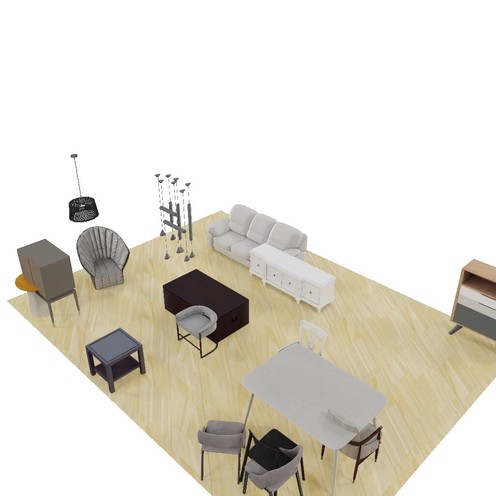}\hfill
    \includegraphics[width=.19\linewidth]{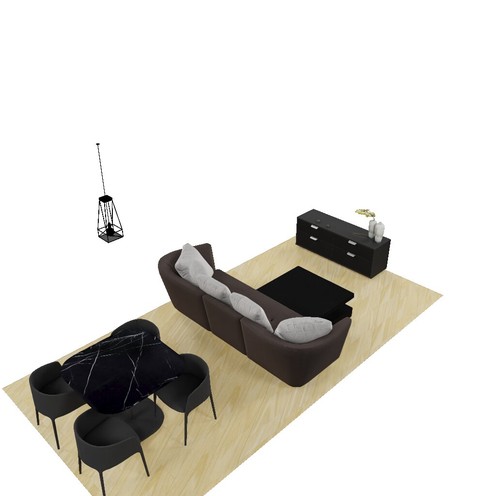}\hfill
    \includegraphics[width=.19\linewidth]{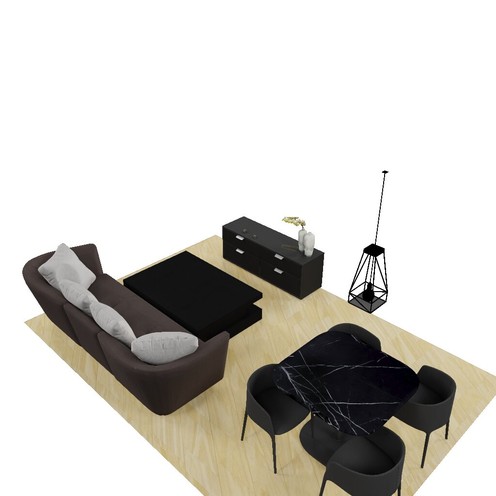}
    \includegraphics[width=.19\linewidth]{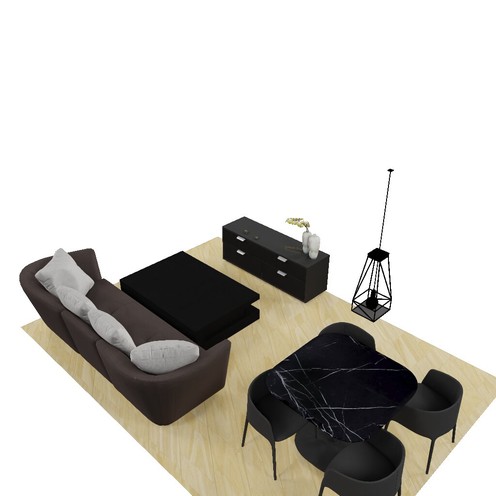}
  \end{subfigure}

  \vspace{2pt}
  \begin{subfigure}{\textwidth}
    \centering
    \includegraphics[width=.19\linewidth]{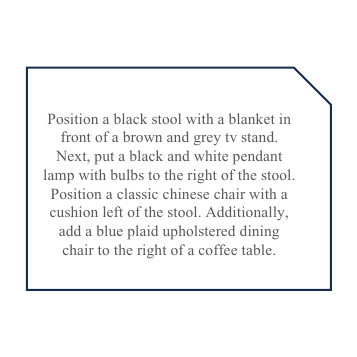}\hfill
    \includegraphics[width=.19\linewidth]{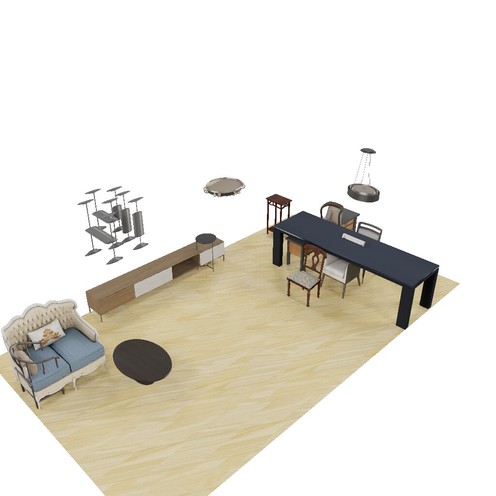}\hfill
    \includegraphics[width=.19\linewidth]{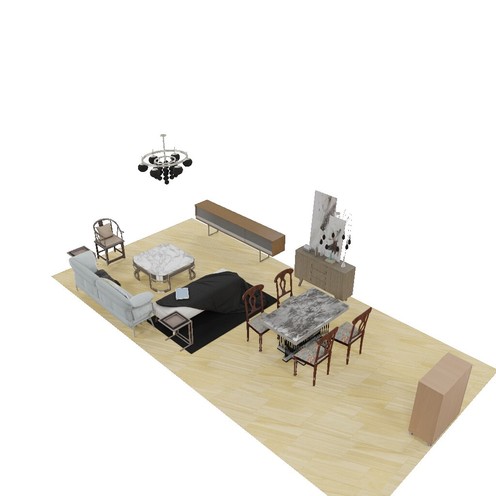}
    \includegraphics[width=.19\linewidth]{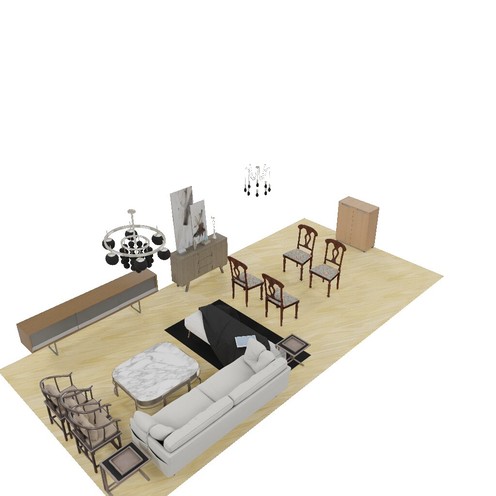}\hfill
    \includegraphics[width=.19\linewidth]{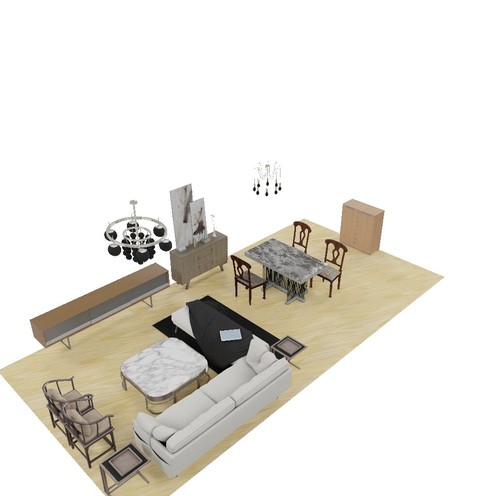}
  \end{subfigure}

  \vspace{2pt}
  \begin{subfigure}{\textwidth}
    \centering
    \includegraphics[width=.19\linewidth]{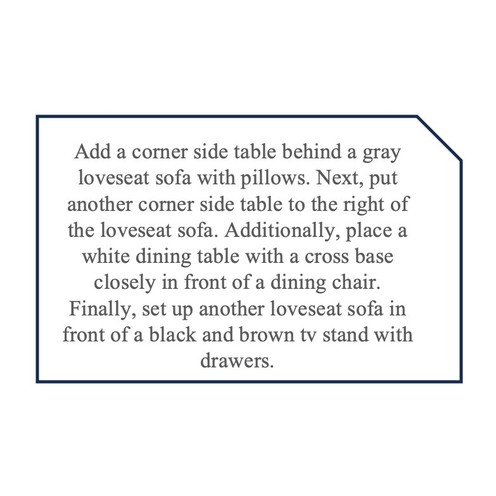}\hfill
    \includegraphics[width=.19\linewidth]{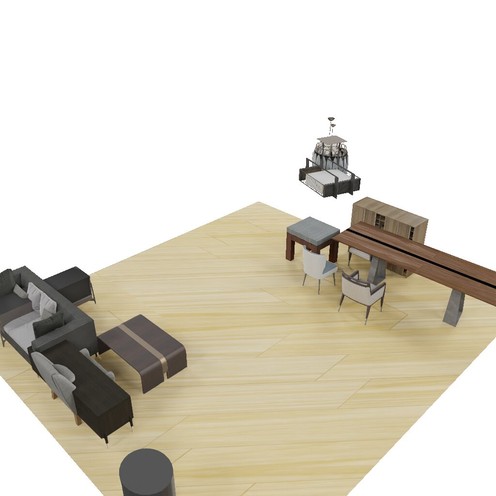}\hfill
    \includegraphics[width=.19\linewidth]{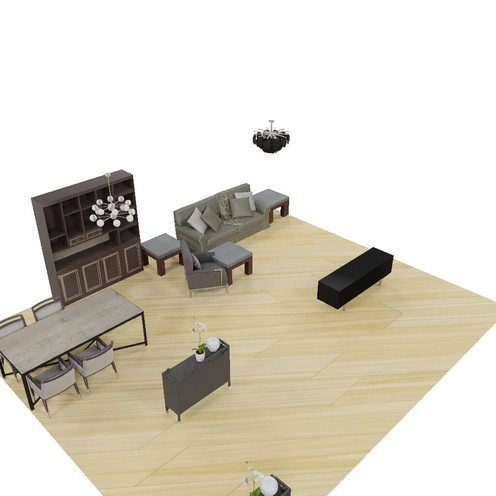}\hfill
    \includegraphics[width=.19\linewidth]{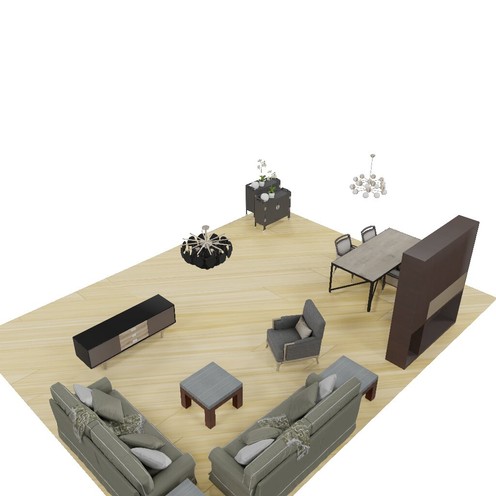}\hfill
    \includegraphics[width=.19\linewidth]{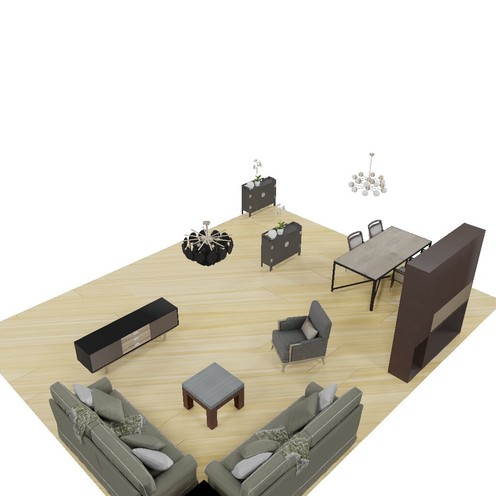}
  \end{subfigure}

  \vspace{2pt}
  \begin{subfigure}{\textwidth}
    \centering
    \includegraphics[width=.19\linewidth]{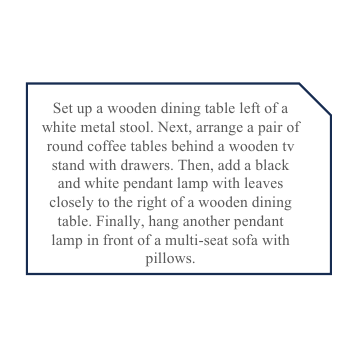}\hfill
    \includegraphics[width=.19\linewidth]{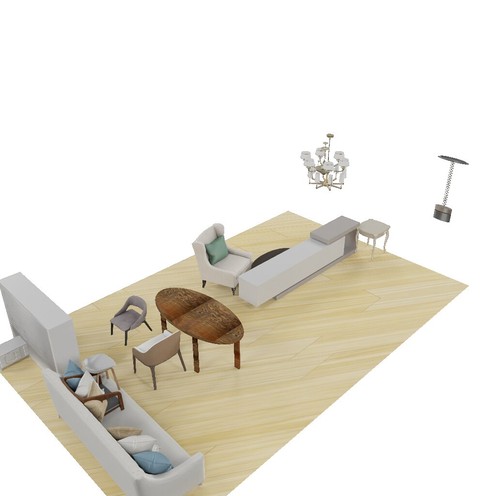}\hfill
    \includegraphics[width=.19\linewidth]{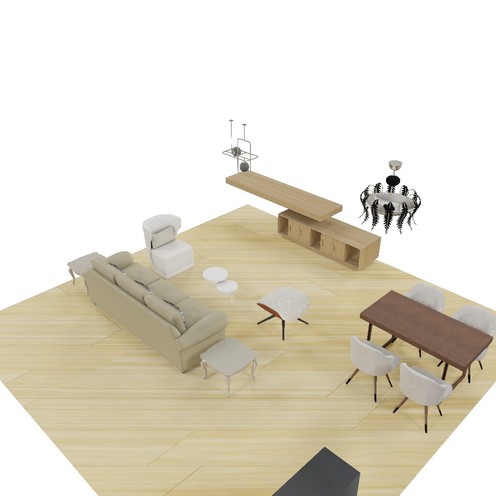}\hfill
    \includegraphics[width=.19\linewidth]{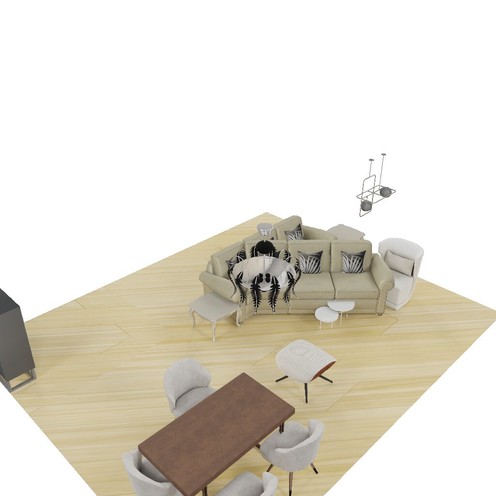}\hfill
    \includegraphics[width=.19\linewidth]{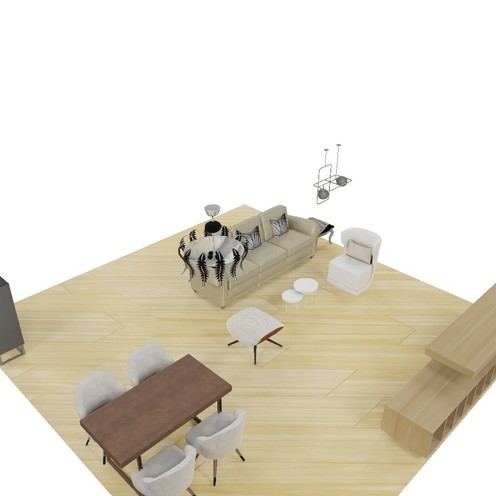}
  \end{subfigure}
  \makebox[.19\linewidth][c]{(a) Instruction}\hfill
\makebox[.19\linewidth][c]{(b) ATISS}\hfill
\makebox[.19\linewidth][c]{(c) DiffuScene}\hfill
\makebox[.19\linewidth][c]{(d) InstructScene}\hfill
\makebox[.19\linewidth][c]{(e) Ours}
  \caption{
  Comparison of qualitative results for 4 relation constraints in instruction.
  }
  \label{tri4}
\end{figure}

\newpage
\begin{figure}
  \centering

\begin{subfigure}[b]{.18\textwidth}\end{subfigure}\hfill
\begin{subfigure}[b]{.18\textwidth}
  \centering
  \includegraphics[width=\linewidth]{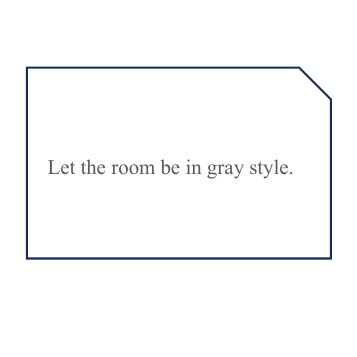}
  \caption*{(a) Instruction}
\end{subfigure}\hfill
\begin{subfigure}[b]{.18\textwidth}
  \centering
  \includegraphics[width=\linewidth]{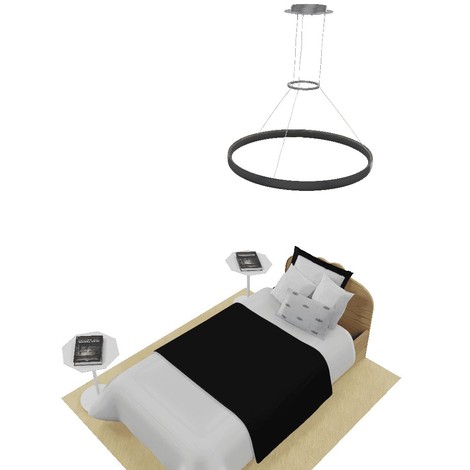}
  \caption*{(b) Original Scene}
\end{subfigure}\hfill
\begin{subfigure}[b]{.18\textwidth}\end{subfigure}

\begin{subfigure}[b]{.18\textwidth}
  \centering
  \includegraphics[width=\linewidth]{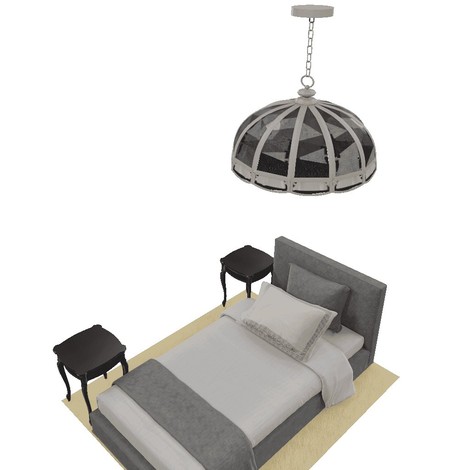}
  \caption*{(c) ATISS}
\end{subfigure}\hfill
\begin{subfigure}[b]{.18\textwidth}
  \centering
  \includegraphics[width=\linewidth]{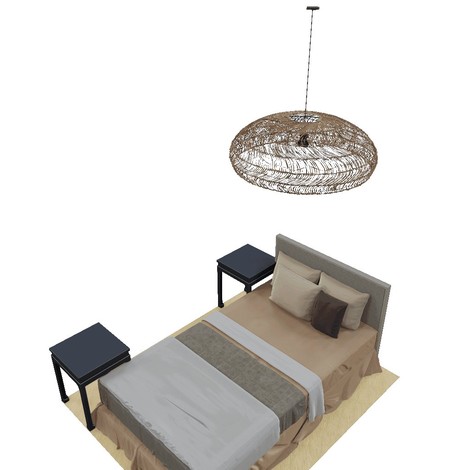}
  \caption*{(d) DiffuScene}
\end{subfigure}\hfill
\begin{subfigure}[b]{.18\textwidth}
  \centering
  \includegraphics[width=\linewidth]{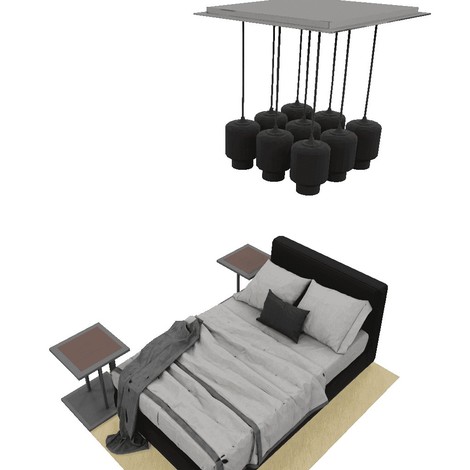}
  \caption*{(e) InstructScene}
\end{subfigure}\hfill
\begin{subfigure}[b]{.18\textwidth}
  \centering
  \includegraphics[width=\linewidth]{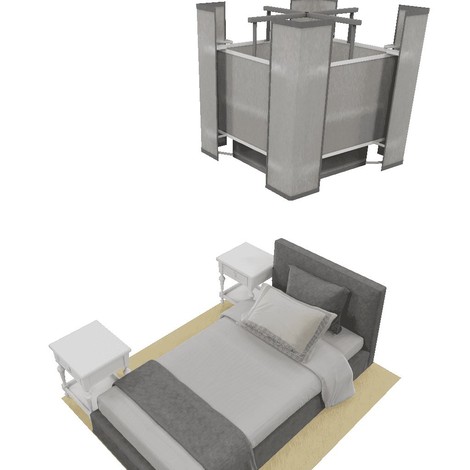}
  \caption*{(f) Ours}
\end{subfigure}

\begin{subfigure}[b]{.18\textwidth}\end{subfigure}\hfill
\begin{subfigure}[b]{.18\textwidth}
  \centering
  \includegraphics[width=\linewidth]{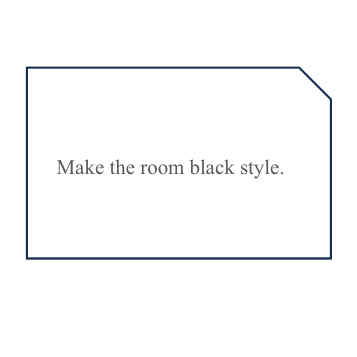}
  \caption*{(a) Instruction}
\end{subfigure}\hfill
\begin{subfigure}[b]{.18\textwidth}
  \centering
  \includegraphics[width=\linewidth]{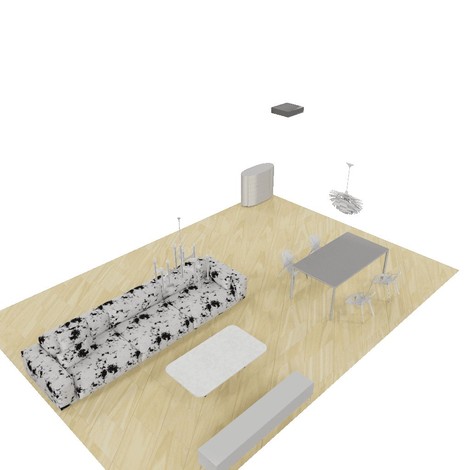}
  \caption*{(b) Original Scene}
\end{subfigure}\hfill
\begin{subfigure}[b]{.18\textwidth}\end{subfigure}

\begin{subfigure}[b]{.18\textwidth}
  \centering
  \includegraphics[width=\linewidth]{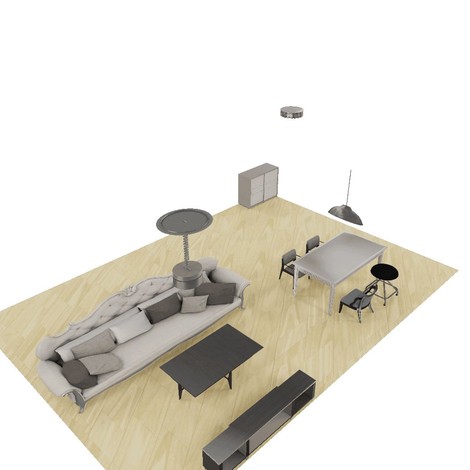}
  \caption*{(c) ATISS}
\end{subfigure}\hfill
\begin{subfigure}[b]{.18\textwidth}
  \centering
  \includegraphics[width=\linewidth]{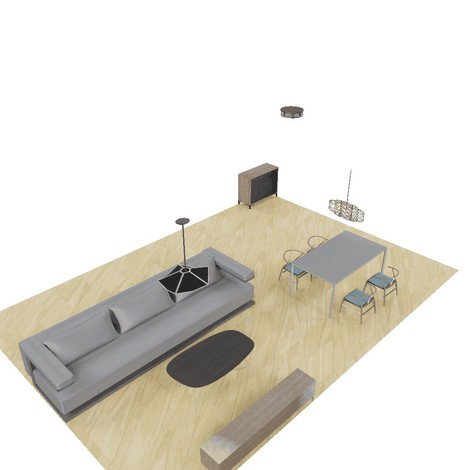}
  \caption*{(d) DiffuScene}
\end{subfigure}\hfill
\begin{subfigure}[b]{.18\textwidth}
  \centering
  \includegraphics[width=\linewidth]{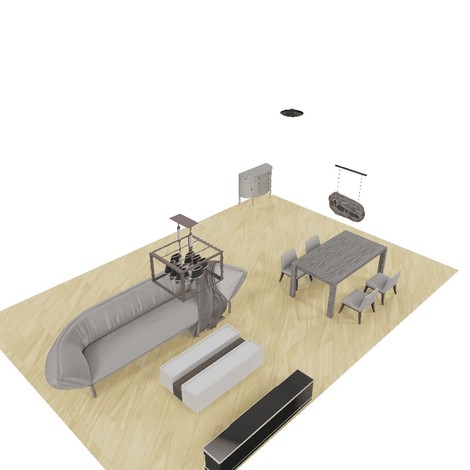}
  \caption*{(e) InstructScene}
\end{subfigure}\hfill
\begin{subfigure}[b]{.18\textwidth}
  \centering
  \includegraphics[width=\linewidth]{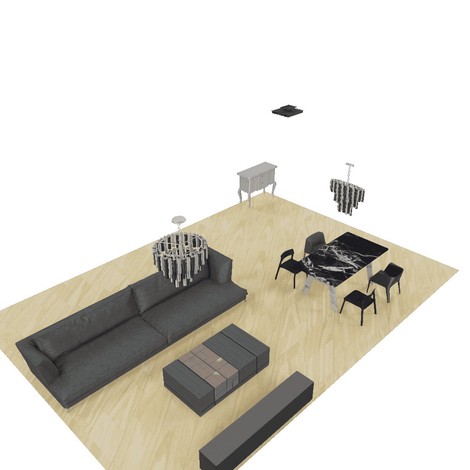}
  \caption*{(f) Ours}
\end{subfigure}

\begin{subfigure}[b]{.18\textwidth}\end{subfigure}\hfill
\begin{subfigure}[b]{.18\textwidth}
  \centering
  \includegraphics[width=\linewidth]{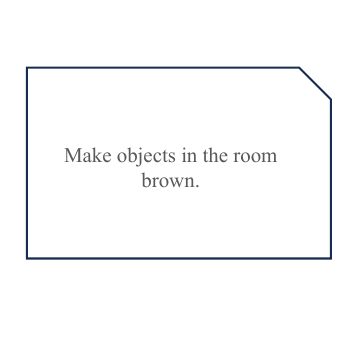}
  \caption*{(a) Instruction}
\end{subfigure}\hfill
\begin{subfigure}[b]{.18\textwidth}
  \centering
  \includegraphics[width=\linewidth]{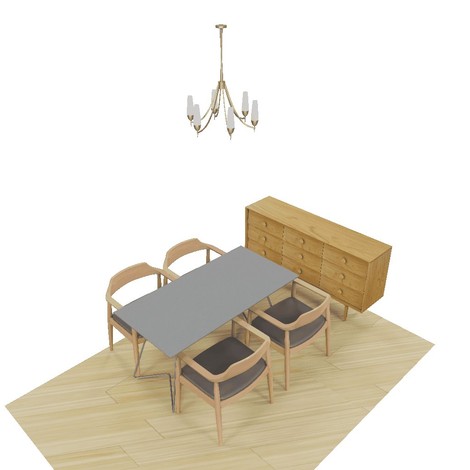}
  \caption*{(b) Original Scene}
\end{subfigure}\hfill
\begin{subfigure}[b]{.18\textwidth}\end{subfigure}

\begin{subfigure}[b]{.18\textwidth}
  \centering
  \includegraphics[width=\linewidth]{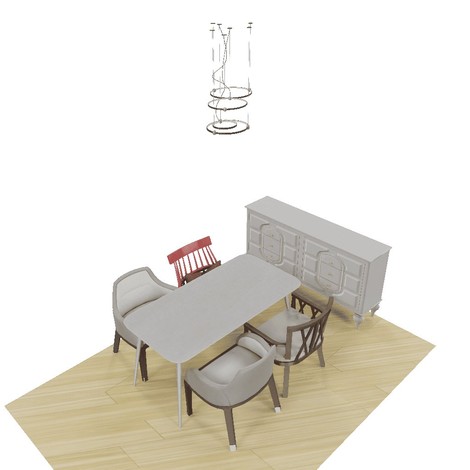}
  \caption*{(c) ATISS}
\end{subfigure}\hfill
\begin{subfigure}[b]{.18\textwidth}
  \centering
  \includegraphics[width=\linewidth]{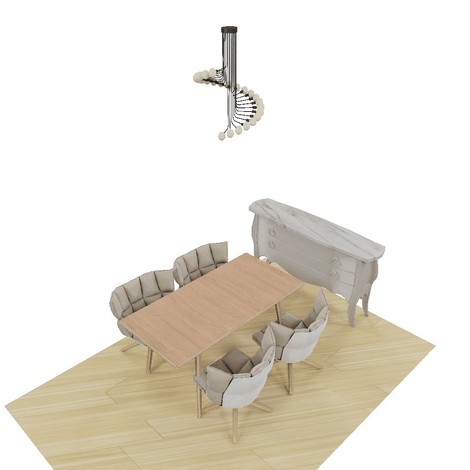}
  \caption*{(d) DiffuScene}
\end{subfigure}\hfill
\begin{subfigure}[b]{.18\textwidth}
  \centering
  \includegraphics[width=\linewidth]{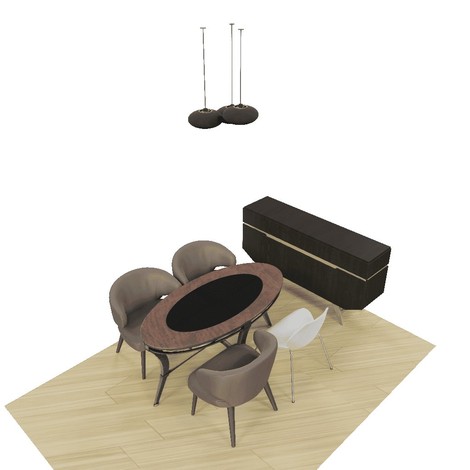}
  \caption*{(e) InstructScene}
\end{subfigure}\hfill
\begin{subfigure}[b]{.18\textwidth}
  \centering
  \includegraphics[width=\linewidth]{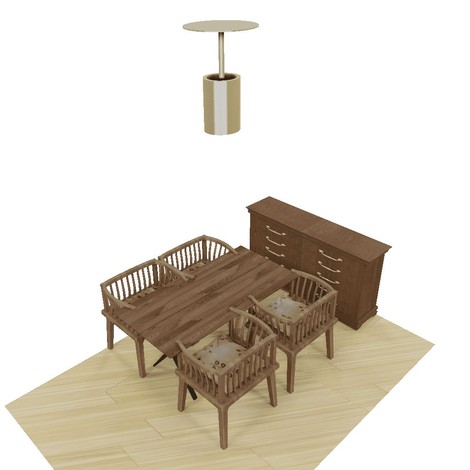}
  \caption*{(f) Ours}
\end{subfigure}

  \caption{ Comparison of qualitative results for stylization task.}
  \label{Stylization}
\end{figure}
\begin{figure}[t]
  \centering
  \begin{subfigure}[b]{.32\textwidth}
    \centering
    \includegraphics[width=\linewidth]{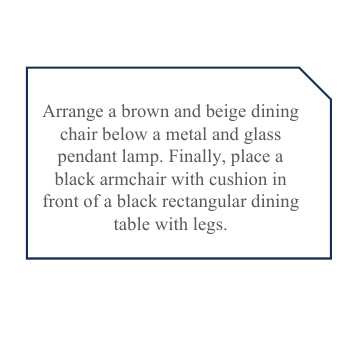}
    \caption*{(a) Instruction}
  \end{subfigure}\hfill
  \begin{subfigure}[b]{.36\textwidth}
    \centering
    \includegraphics[width=\linewidth]{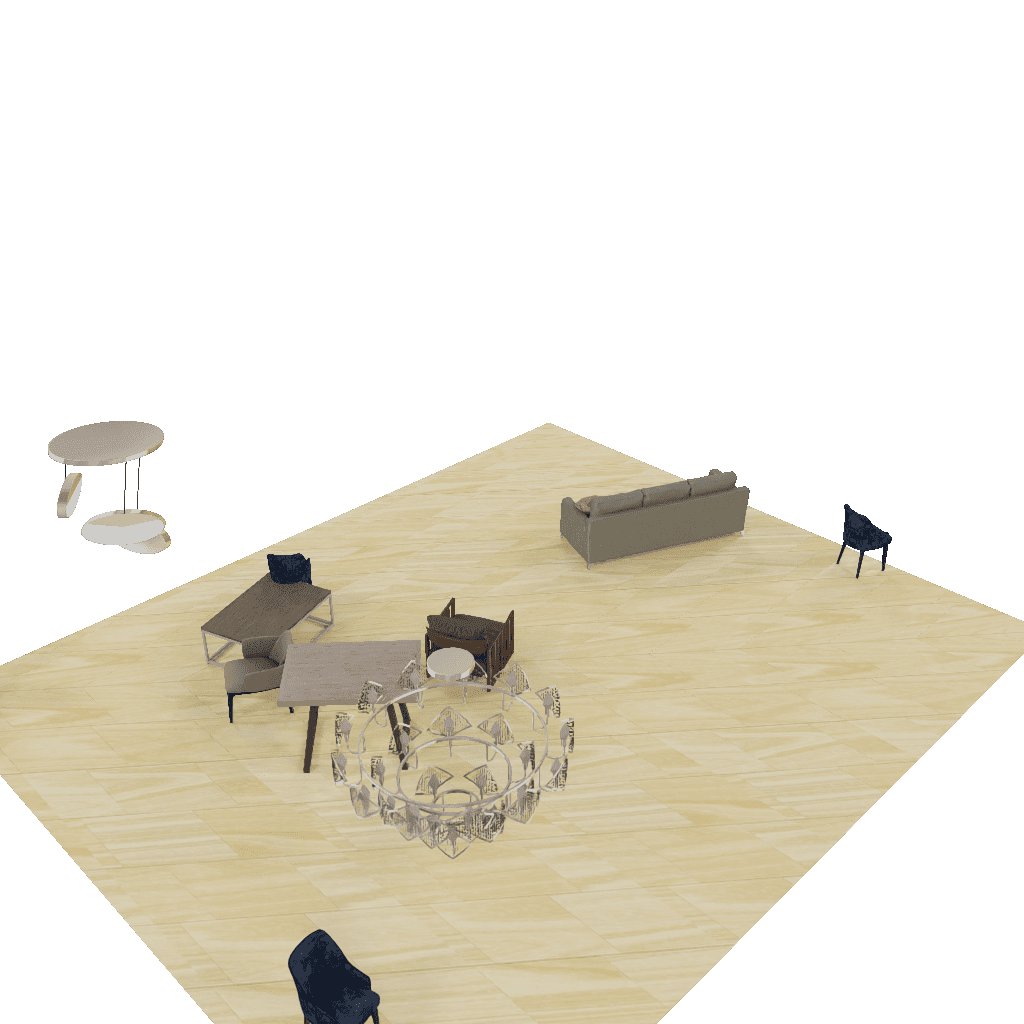}
    \caption*{(b) Messy Scene}
  \end{subfigure}

  \vspace{0pt}
  \begin{subfigure}[b]{.36\textwidth}
    \centering
    \includegraphics[width=\linewidth]{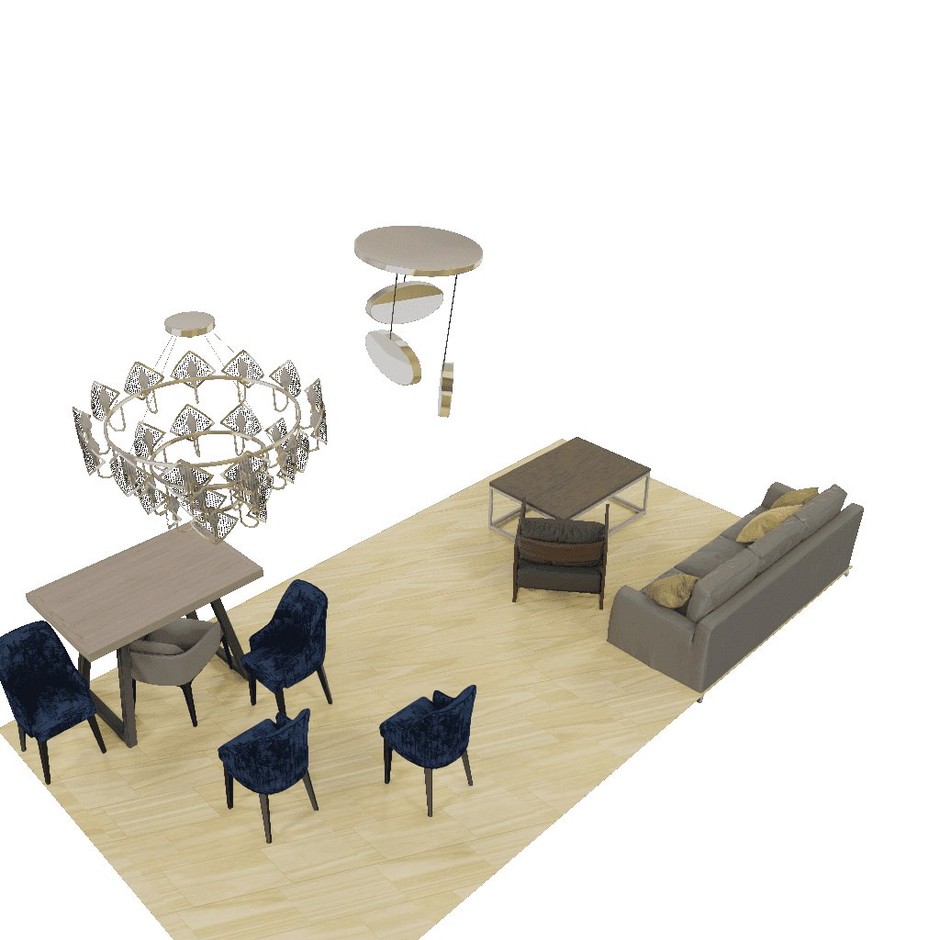}
    \caption*{(c) ATISS}
  \end{subfigure}\hfill
  \begin{subfigure}[b]{.36\textwidth}
    \centering
    \includegraphics[width=\linewidth]{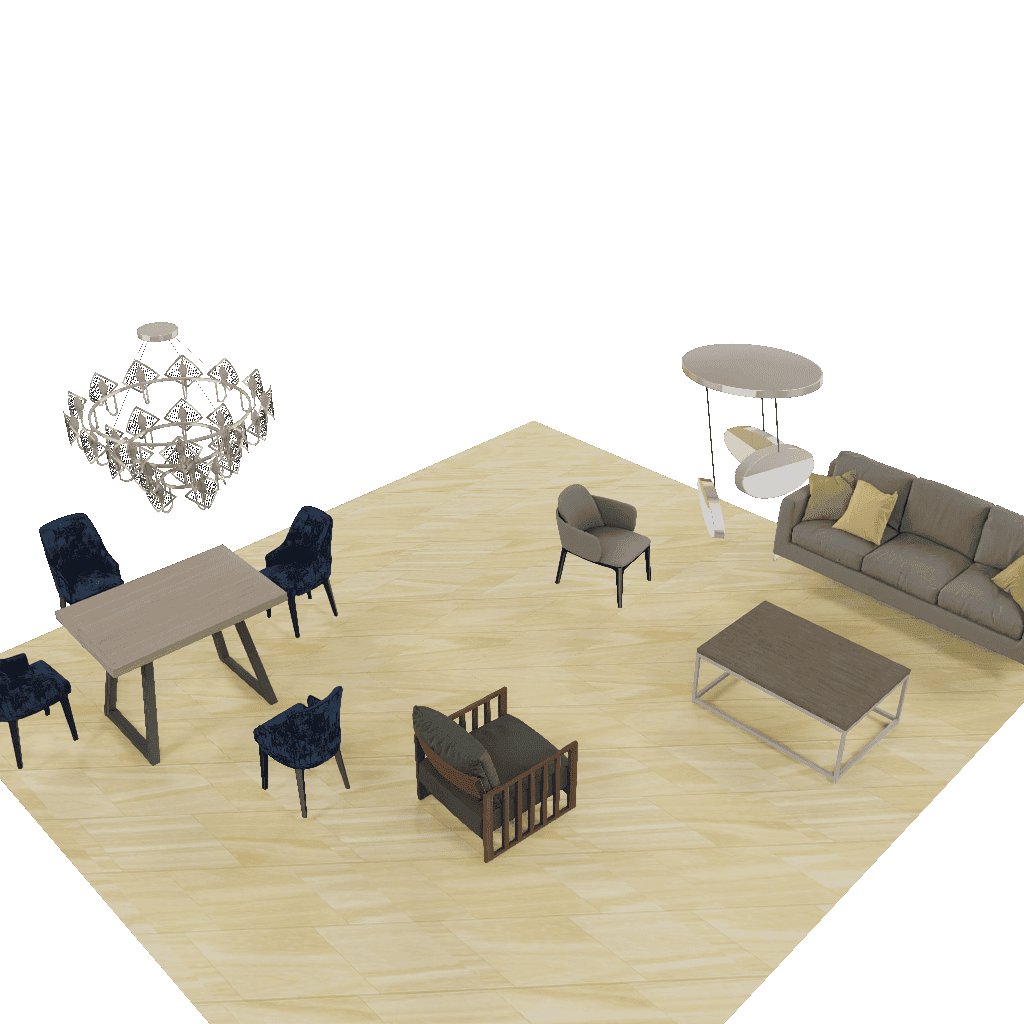}
    \caption*{(d) DiffuScene}
  \end{subfigure}

  \vspace{0pt}
  \begin{subfigure}[b]{.36\textwidth}
    \centering
    \includegraphics[width=\linewidth]{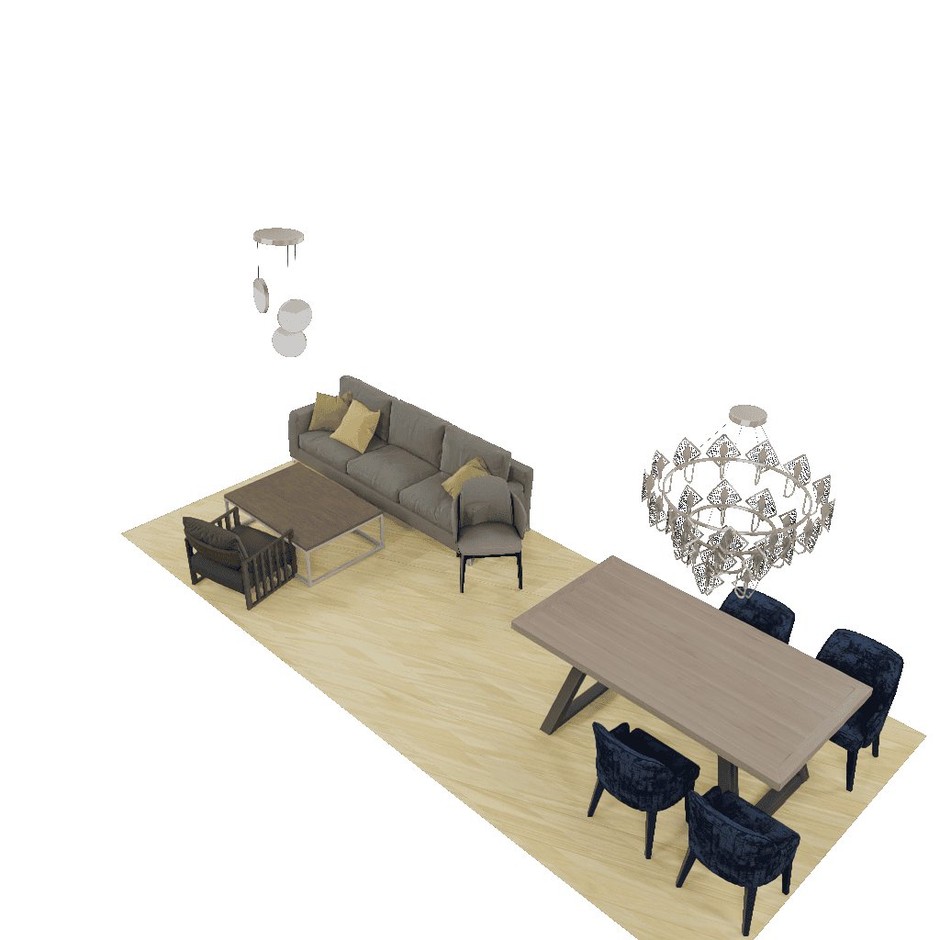}
    \caption*{(e) InstructScene}
  \end{subfigure}\hfill
  \begin{subfigure}[b]{.36\textwidth}
    \centering
    \includegraphics[width=\linewidth]{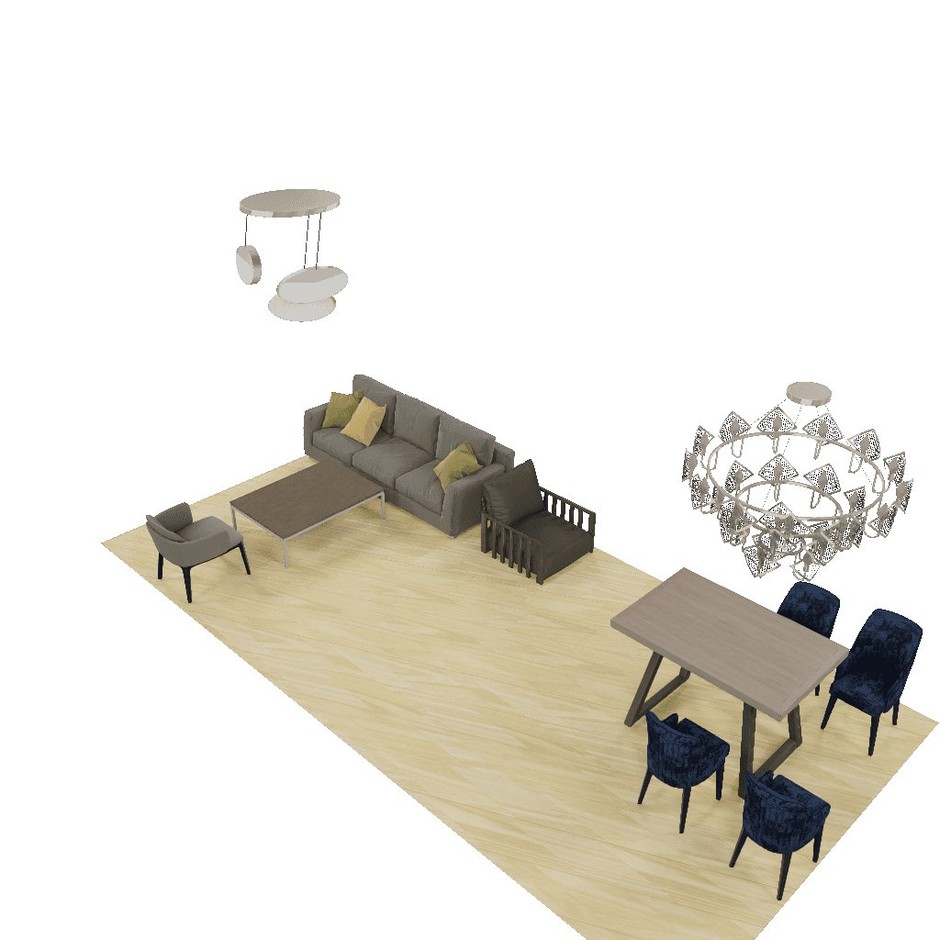}
    \caption*{(f) Ours}
  \end{subfigure}

  \caption{Comparison of qualitative results for rearrangement task.}
  \label{Rearrangement}
\end{figure}

\begin{figure}[t]
  \centering
  \begin{subfigure}[b]{.32\textwidth}
    \centering
    \includegraphics[width=\linewidth]{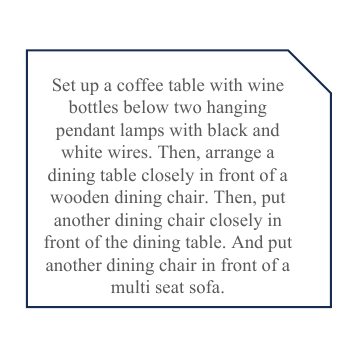}
    \caption*{(a) Instruction}
  \end{subfigure}\hfill
  \begin{subfigure}[b]{.36\textwidth}
    \centering
    \includegraphics[width=\linewidth]{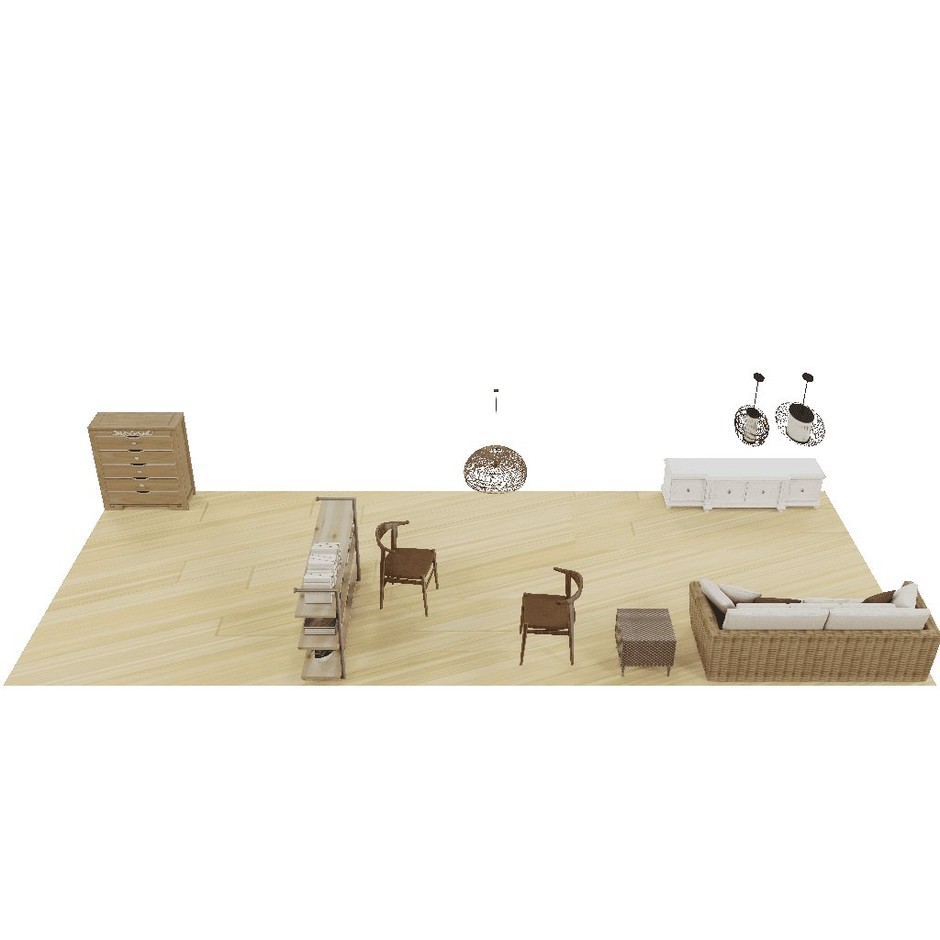}
    \caption*{(b) Partial Scene}
  \end{subfigure}

  \vspace{0pt}
  \begin{subfigure}[b]{.36\textwidth}
    \centering
    \includegraphics[width=\linewidth]{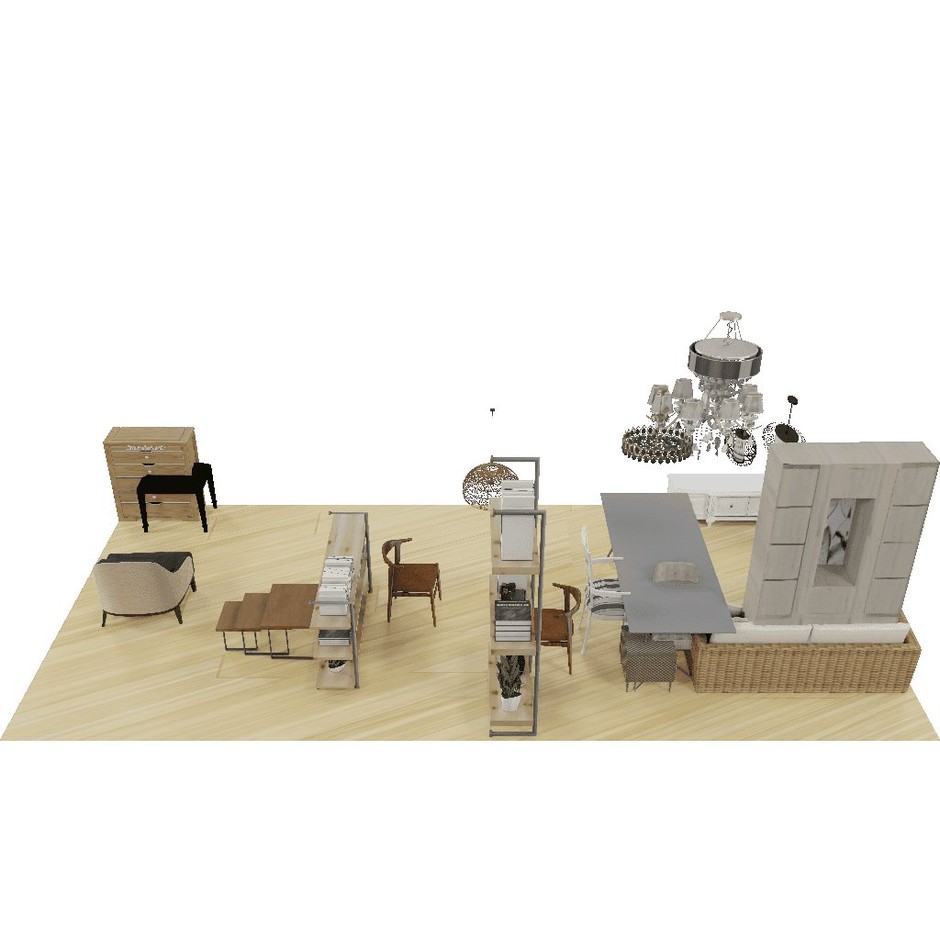}
    \caption*{(c) ATISS}
  \end{subfigure}\hfill
  \begin{subfigure}[b]{.36\textwidth}
    \centering
    \includegraphics[width=\linewidth]{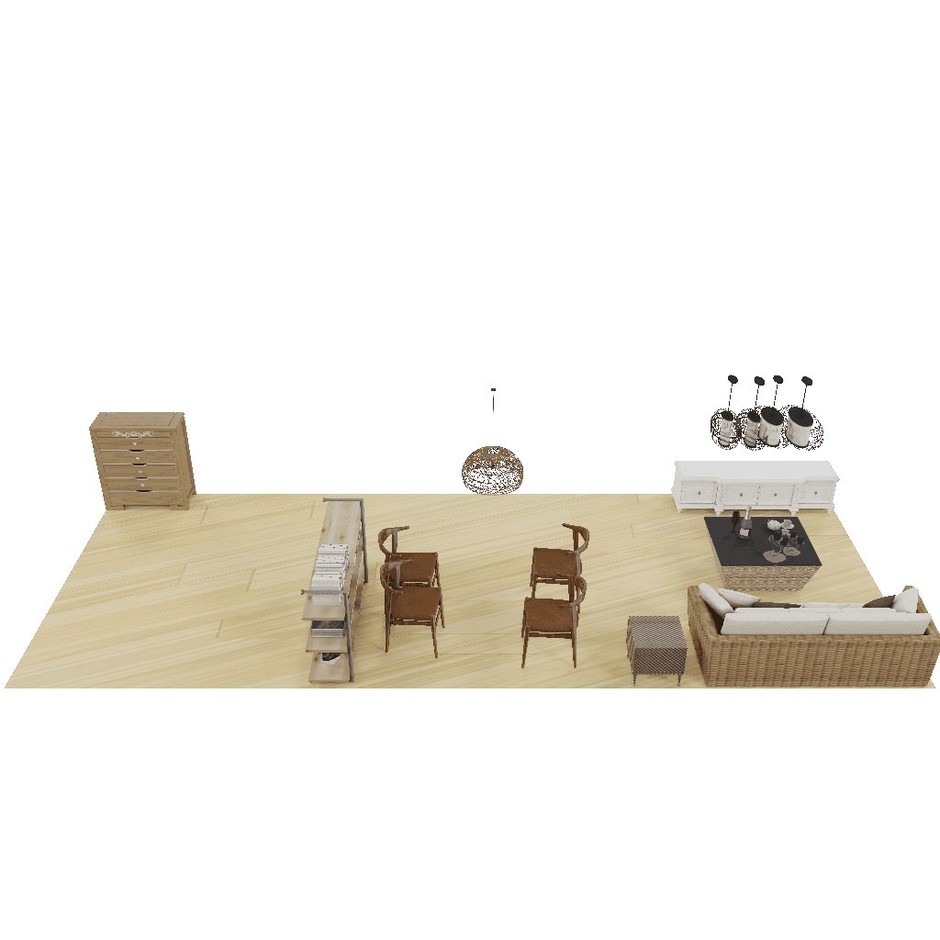}
    \caption*{(d) DiffuScene}
  \end{subfigure}

  \vspace{0pt}
  \begin{subfigure}[b]{.36\textwidth}
    \centering
    \includegraphics[width=\linewidth]{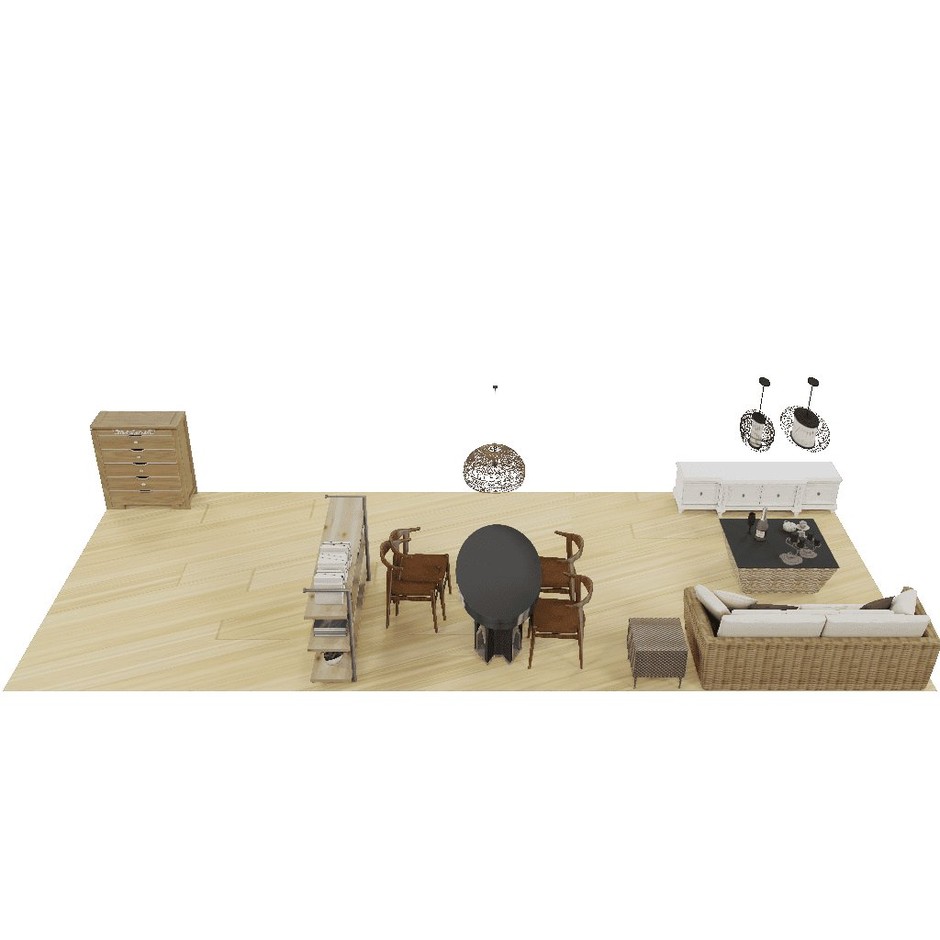}
    \caption*{(e) InstructScene}
  \end{subfigure}\hfill
  \begin{subfigure}[b]{.36\textwidth}
    \centering
    \includegraphics[width=\linewidth]{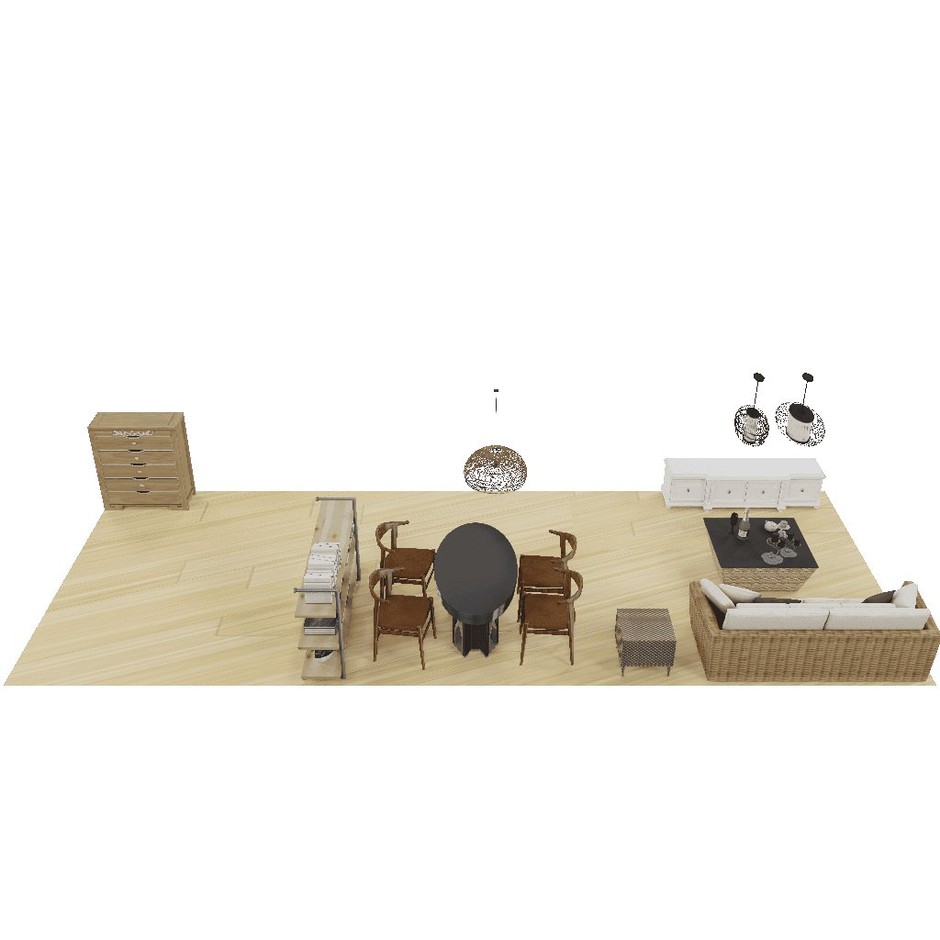}
    \caption*{(f) Ours}
  \end{subfigure}

  \caption{Comparison of qualitative results for completion task.}
  \label{Completion}
\end{figure}

\newpage
\begin{figure}
  \centering

\begin{subfigure}[b]{.30\textwidth}
  \centering
  \includegraphics[width=\linewidth]{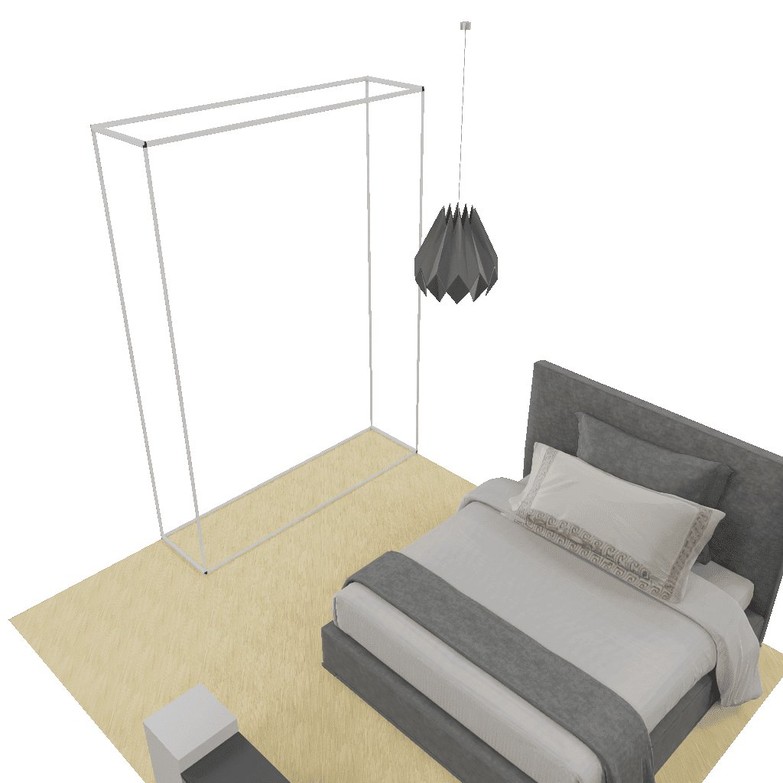}
\end{subfigure}\hfill
\begin{subfigure}[b]{.30\textwidth}
  \centering
  \includegraphics[width=\linewidth]{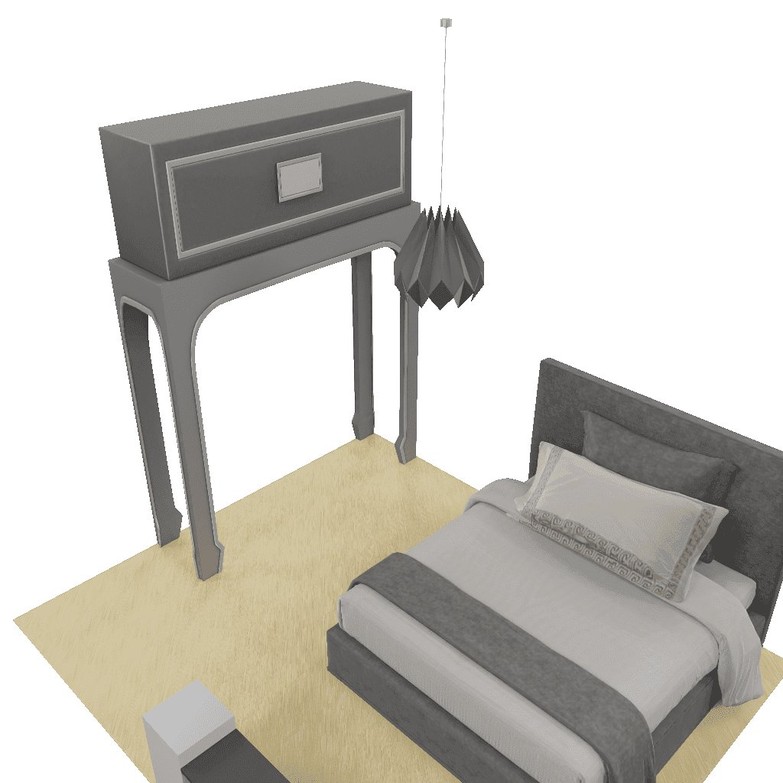}
\end{subfigure}\hfill
\begin{subfigure}[b]{.30\textwidth}
  \centering
  \includegraphics[width=\linewidth]{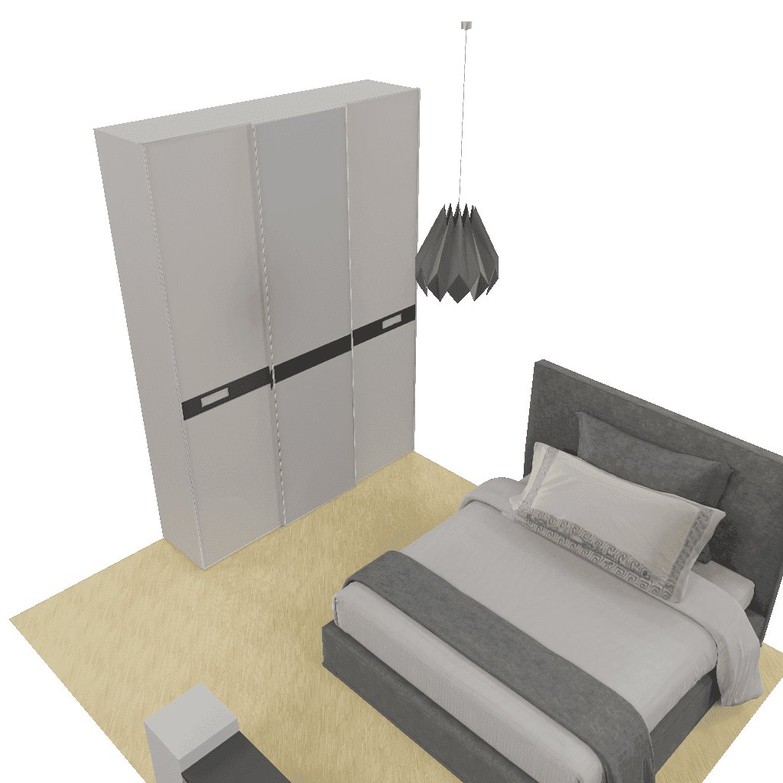}
\end{subfigure}\hfill

\begin{subfigure}[b]{.30\textwidth}
  \centering
  \includegraphics[width=\linewidth]{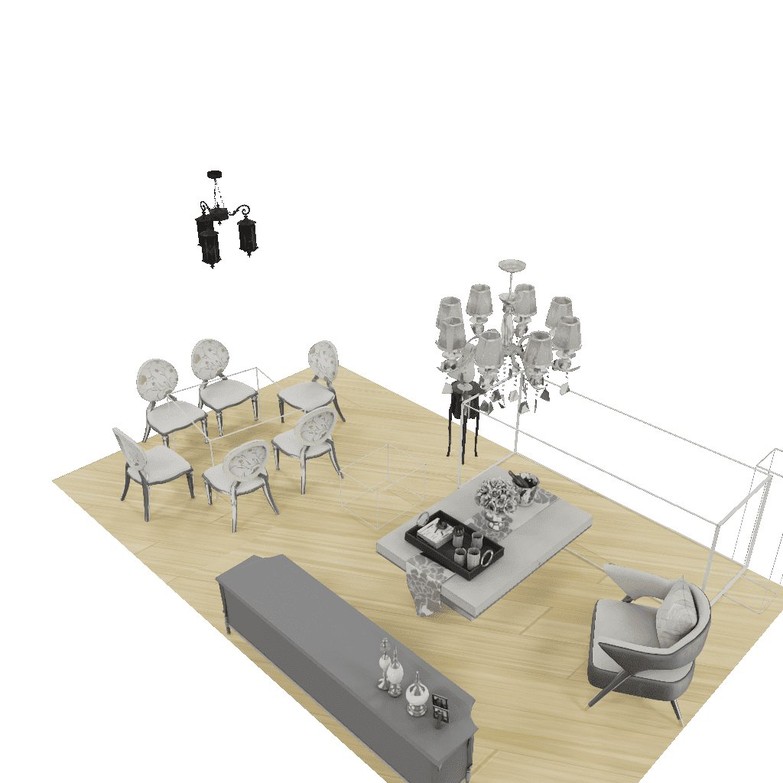}
\end{subfigure}\hfill
\begin{subfigure}[b]{.30\textwidth}
  \centering
  \includegraphics[width=\linewidth]{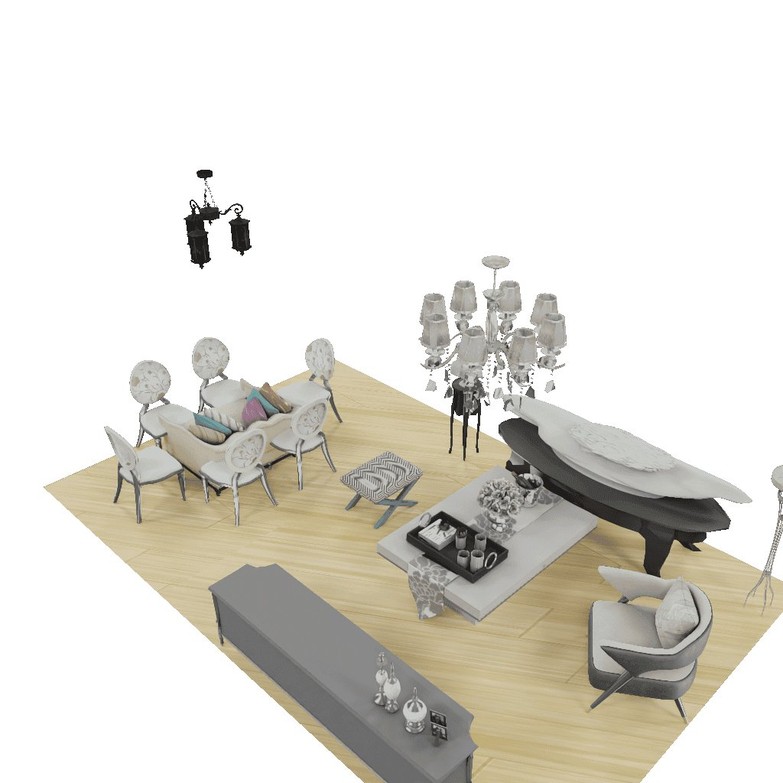}
\end{subfigure}\hfill
\begin{subfigure}[b]{.30\textwidth}
  \centering
  \includegraphics[width=\linewidth]{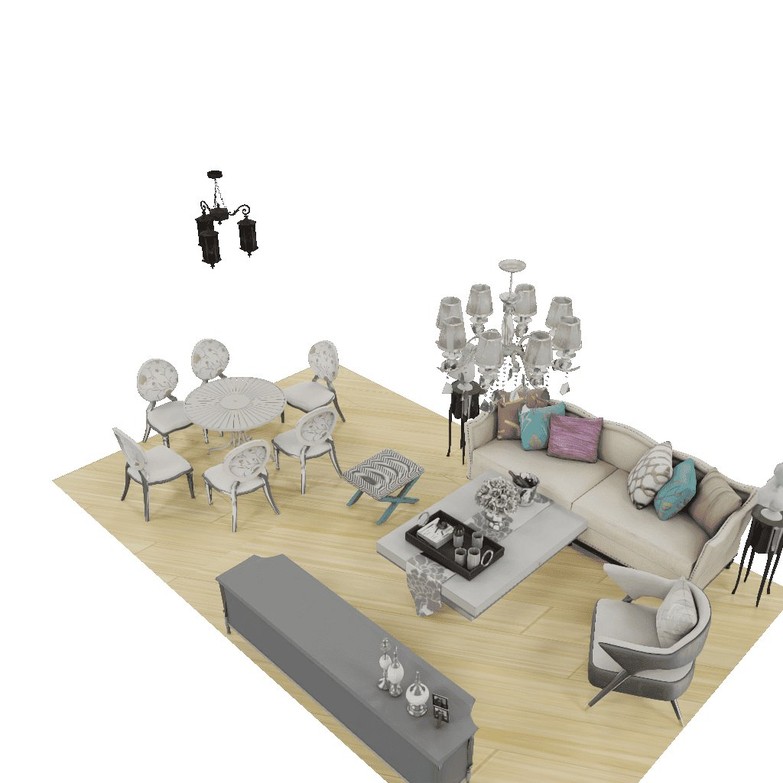}
\end{subfigure}\hfill

\begin{subfigure}[b]{.30\textwidth}
  \centering
  \includegraphics[width=\linewidth]{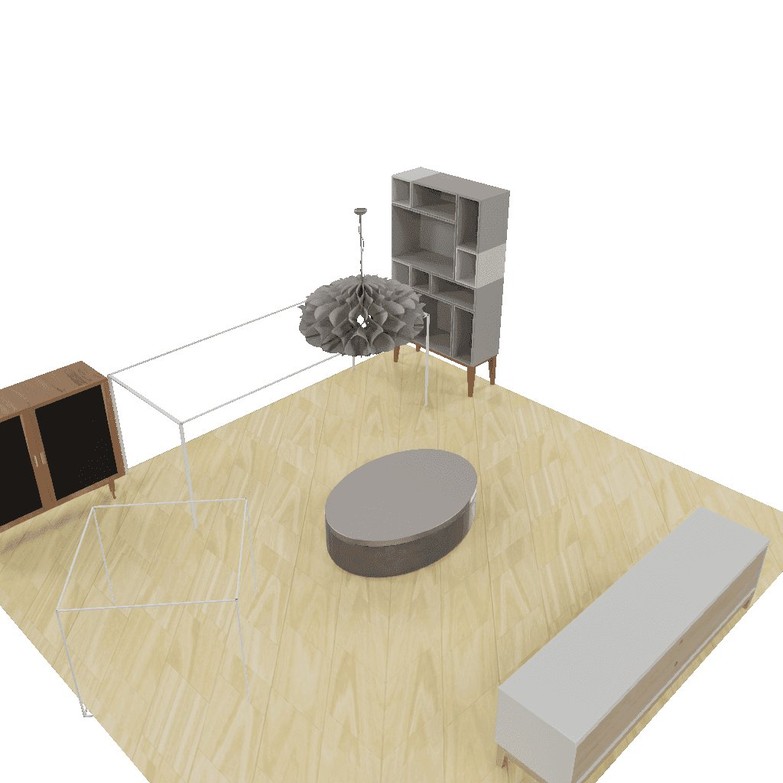}
\end{subfigure}\hfill
\begin{subfigure}[b]{.30\textwidth}
  \centering
  \includegraphics[width=\linewidth]{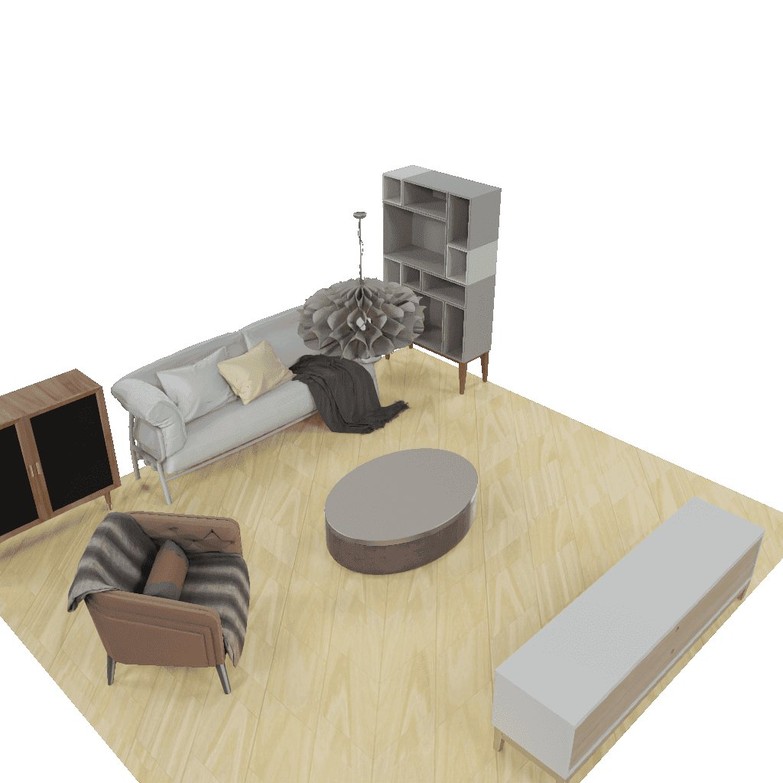}
\end{subfigure}\hfill
\begin{subfigure}[b]{.30\textwidth}
  \centering
  \includegraphics[width=\linewidth]{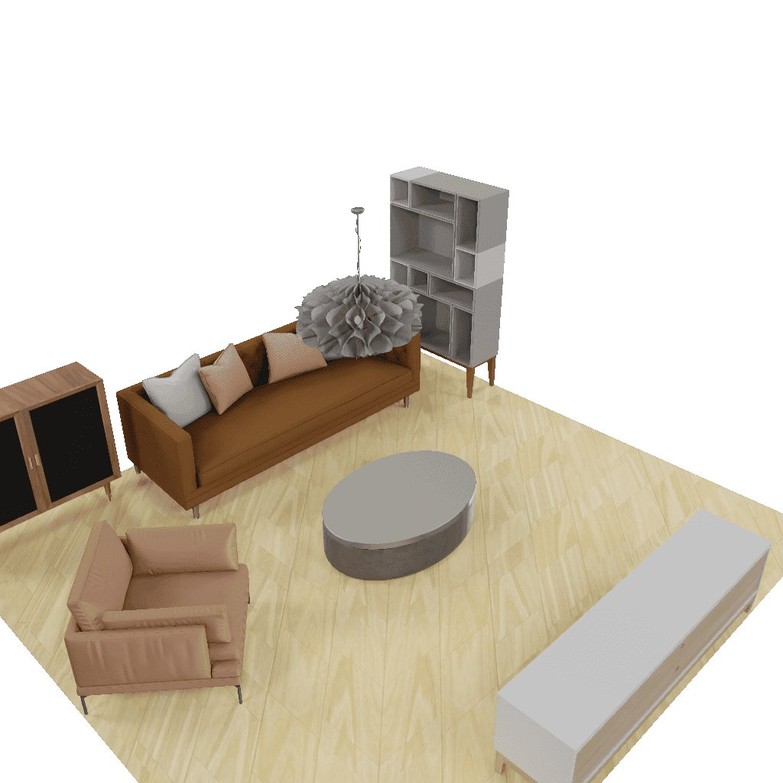}
\end{subfigure}\hfill

\begin{subfigure}[b]{.30\textwidth}
  \centering
  \includegraphics[width=\linewidth]{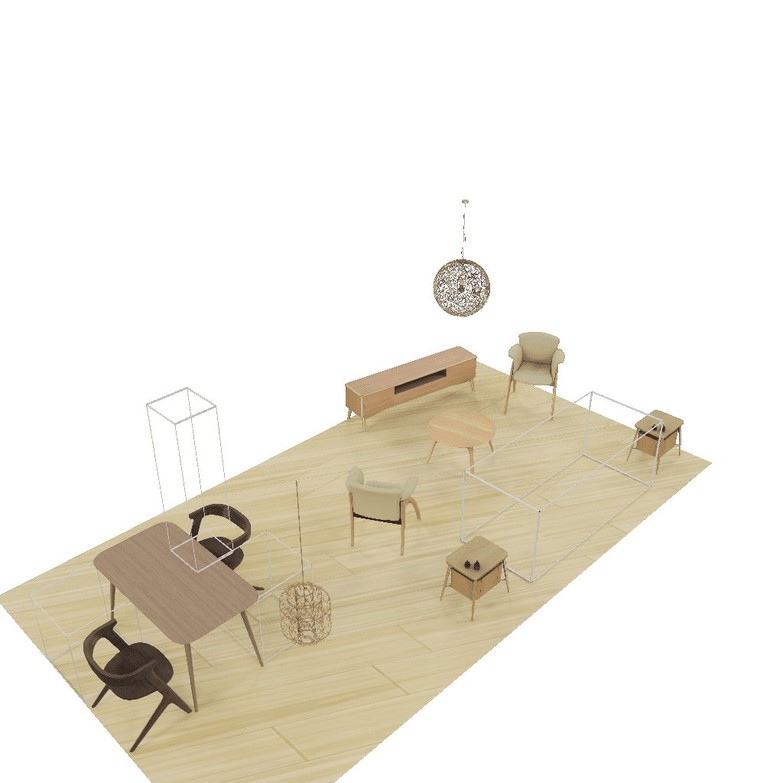}
  \caption*{(a) Empty layout}
\end{subfigure}\hfill
\begin{subfigure}[b]{.30\textwidth}
  \centering
  \includegraphics[width=\linewidth]{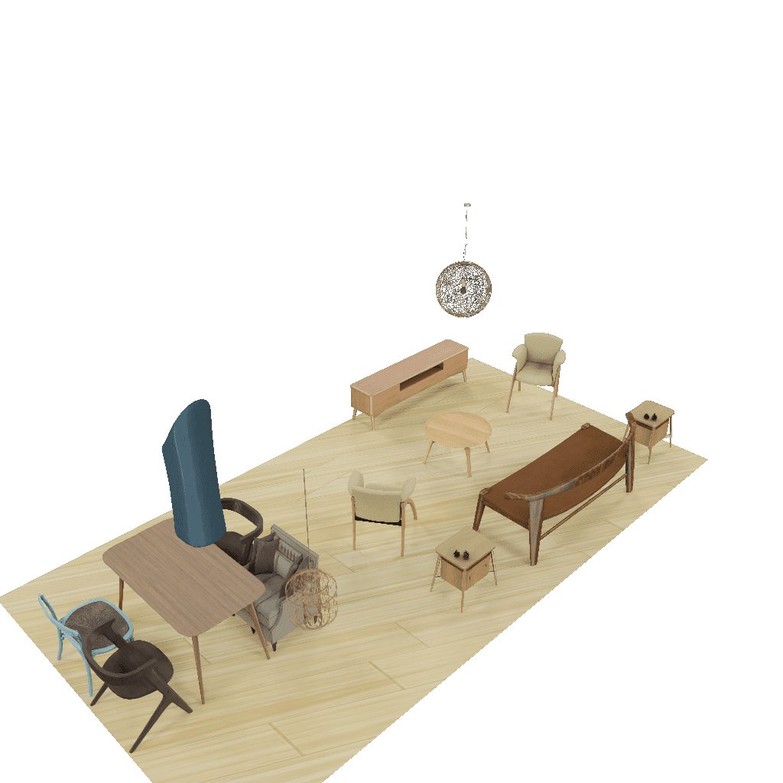}
  \caption*{(b) DiffuScene}
\end{subfigure}\hfill
\begin{subfigure}[b]{.30\textwidth}
  \centering
  \includegraphics[width=\linewidth]{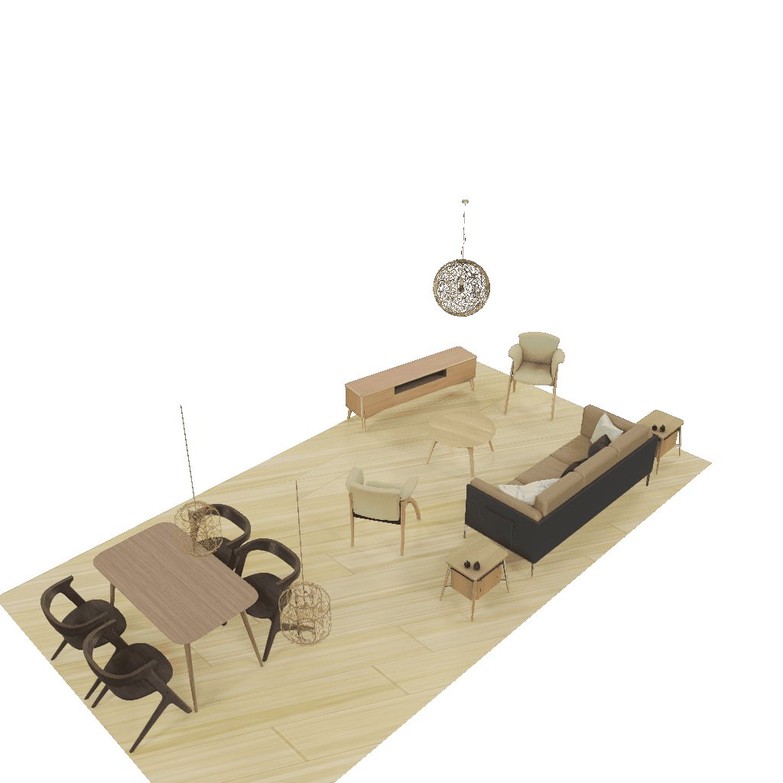}
  \caption*{(b) Ours}
\end{subfigure}\hfill

  \caption{ Comparison of qualitative results for layout-to-object task.}
  \label{layouto}
\end{figure}
\begin{figure}[t]
  \centering

  \includegraphics[width=.18\linewidth]{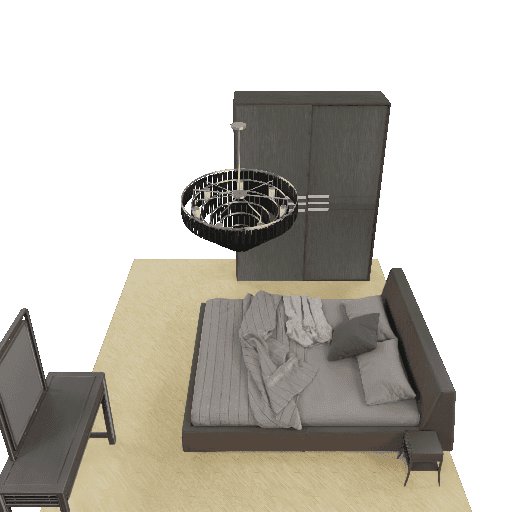}\hfill
  \includegraphics[width=.18\linewidth]{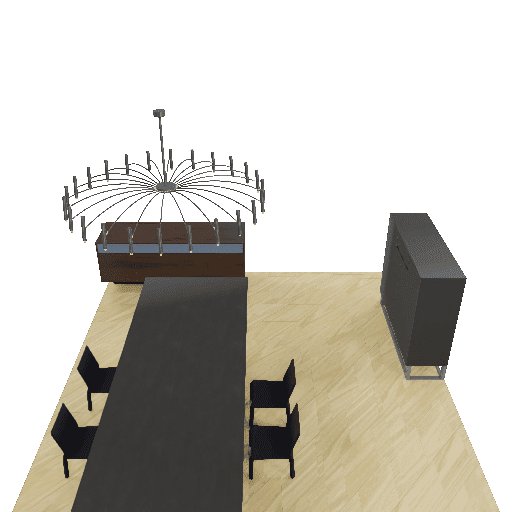}\hfill
  \includegraphics[width=.18\linewidth]{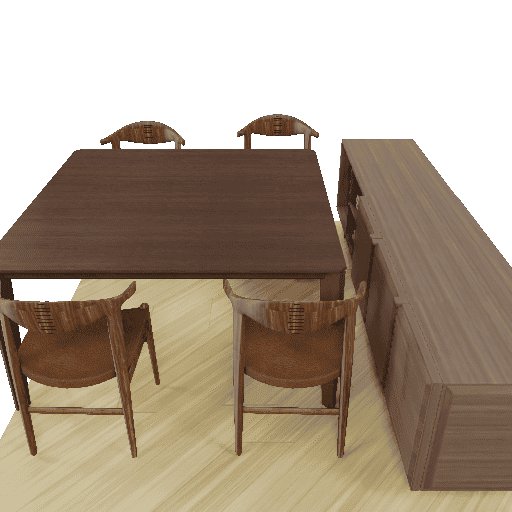}\hfill
  \includegraphics[width=.18\linewidth]{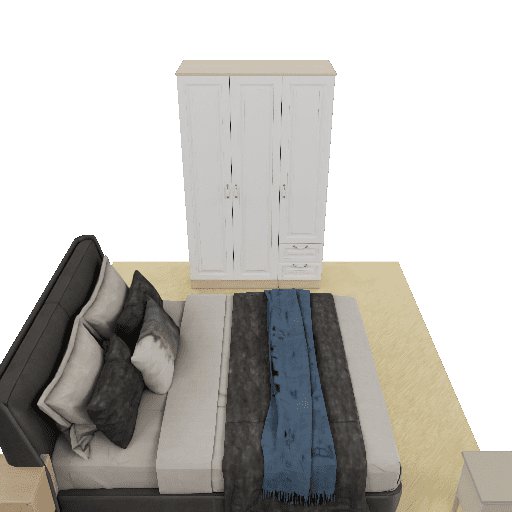}\hfill
  \includegraphics[width=.18\linewidth]{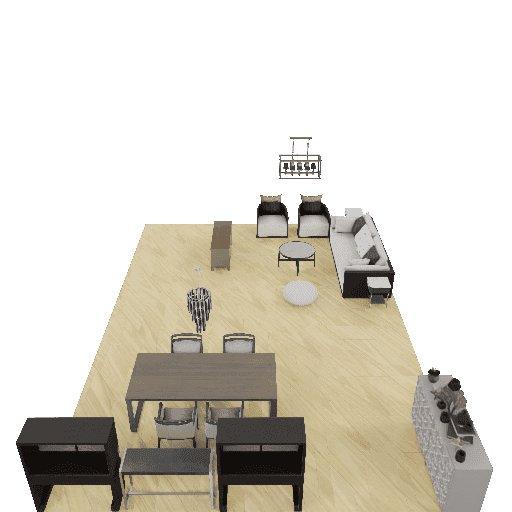}

  \vspace{6pt}  

  \includegraphics[width=.18\linewidth]{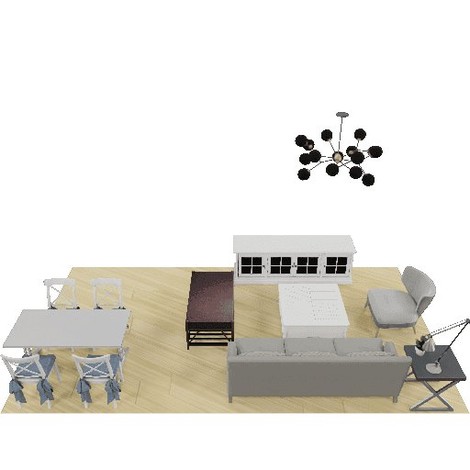}\hfill
  \includegraphics[width=.18\linewidth]{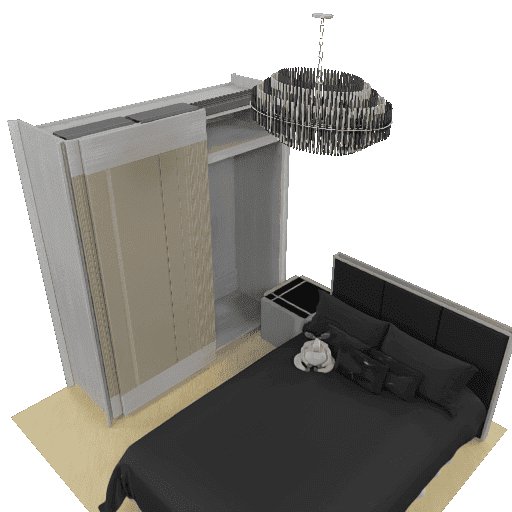}\hfill
  \includegraphics[width=.18\linewidth]{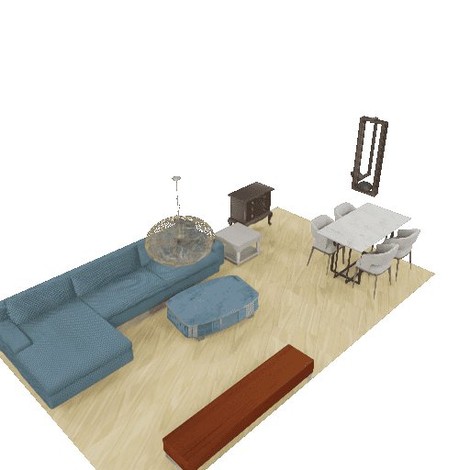}\hfill
  \includegraphics[width=.18\linewidth]{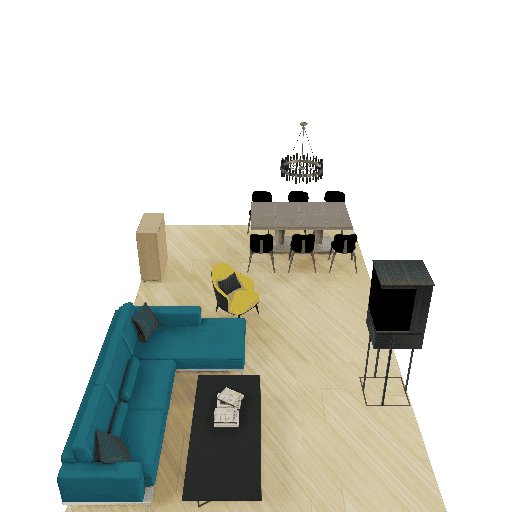}\hfill
  \includegraphics[width=.18\linewidth]{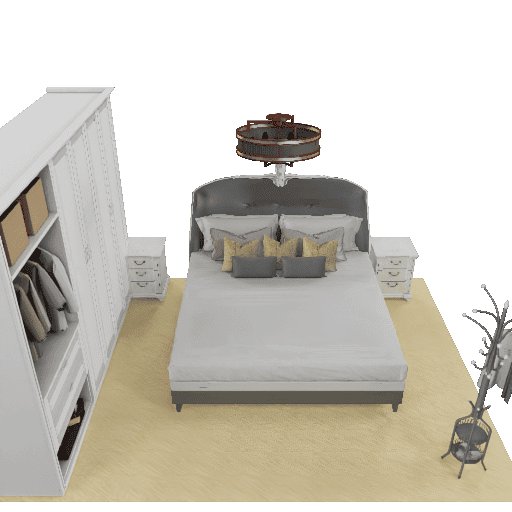}

  \vspace{6pt}  
  \includegraphics[width=.18\linewidth]{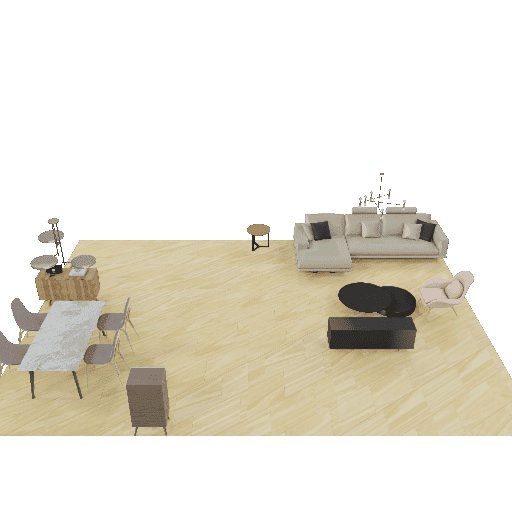}\hfill
  \includegraphics[width=.18\linewidth]{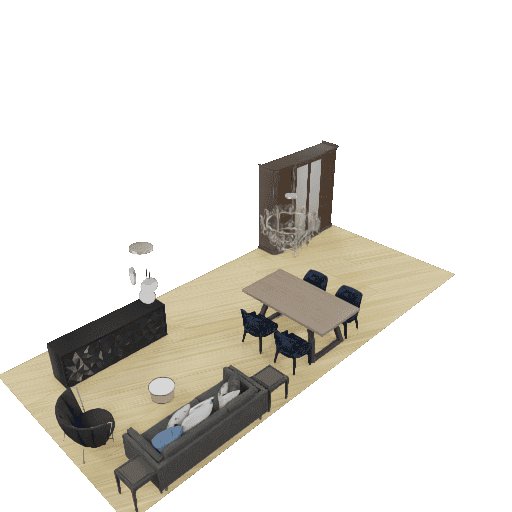}\hfill
  \begin{subfigure}[b]{.18\linewidth}
    \centering
    \includegraphics[width=\linewidth]{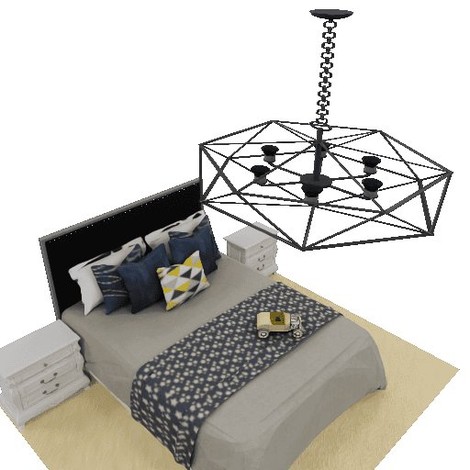}
    \caption*{(a) Bedroom}
  \end{subfigure}\hfill
  \begin{subfigure}[b]{.18\linewidth}
    \centering
    \includegraphics[width=\linewidth]{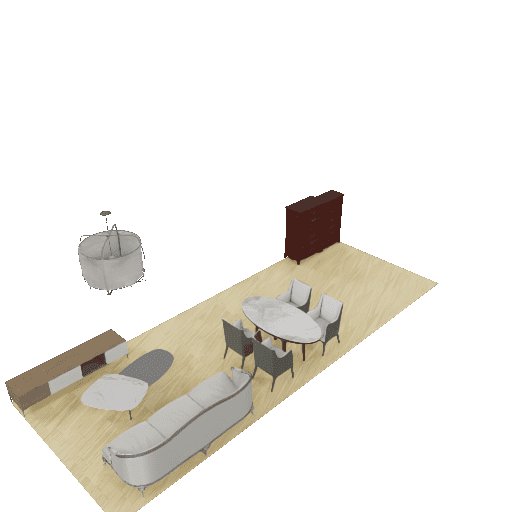}
    \caption*{(b) Living Room}
  \end{subfigure}\hfill
  \begin{subfigure}[b]{.18\linewidth}
    \centering
    \includegraphics[width=\linewidth]{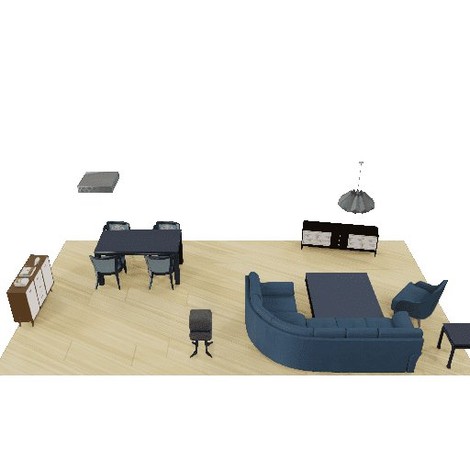}
    \caption*{(c) Dining Room}
  \end{subfigure}

  \caption{Unconditional generation results across three room types.}
  \label{fig:uncond_rooms}
\end{figure}

\begin{figure}[t]
  \centering
\vspace{-10pt}
\hfill
\begin{subfigure}[b]{.20\textwidth}
  \centering
  \includegraphics[width=\linewidth]{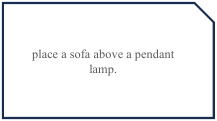}
\end{subfigure}\hfill\hfill
\begin{subfigure}[b]{.20\textwidth}
  \centering
  \includegraphics[width=\linewidth]{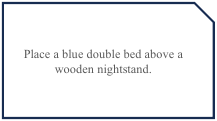}
\end{subfigure}\hfill\hfill
\begin{subfigure}[b]{.20\textwidth}
  \centering
  \includegraphics[width=\linewidth]{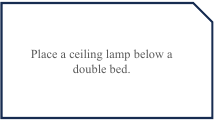}
\end{subfigure}\hfill

\hfill
\begin{subfigure}[b]{.29\textwidth}
  \centering
  \includegraphics[width=\linewidth]{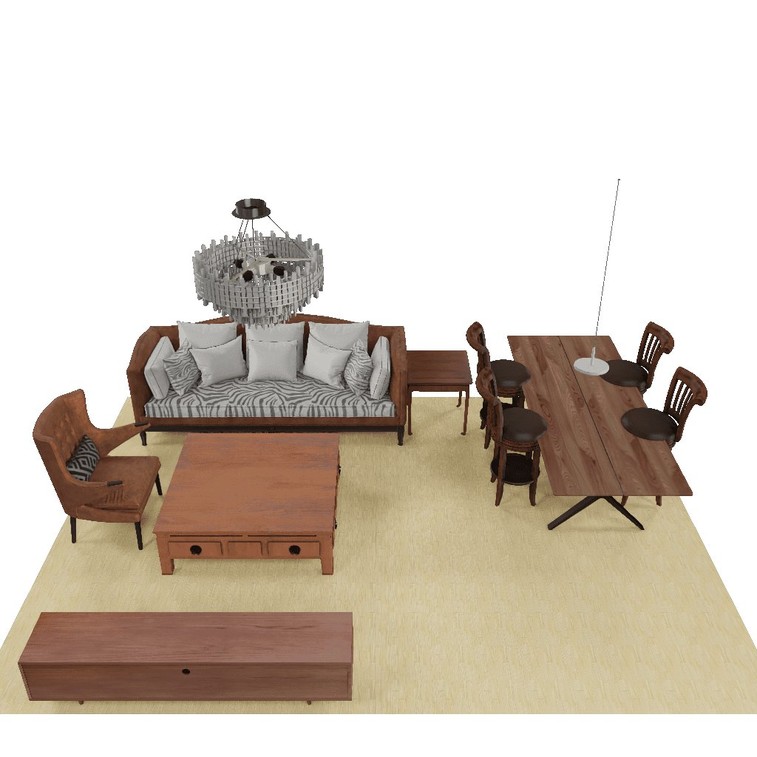}
\end{subfigure}\hfill\hfill
\begin{subfigure}[b]{.29\textwidth}
  \centering
  \includegraphics[width=\linewidth]{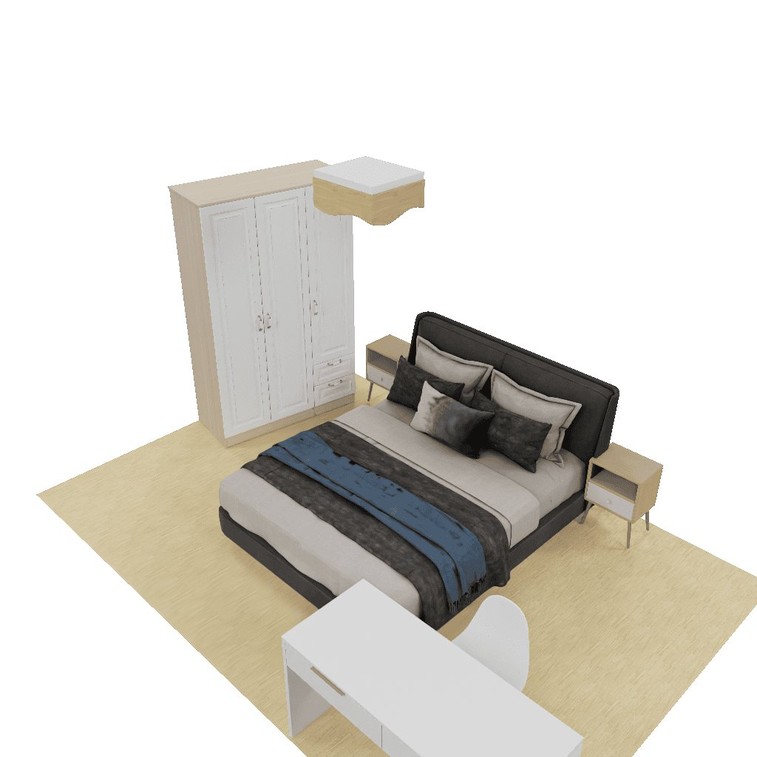}
\end{subfigure}\hfill\hfill
\begin{subfigure}[b]{.29\textwidth}
  \centering
  \includegraphics[width=\linewidth]{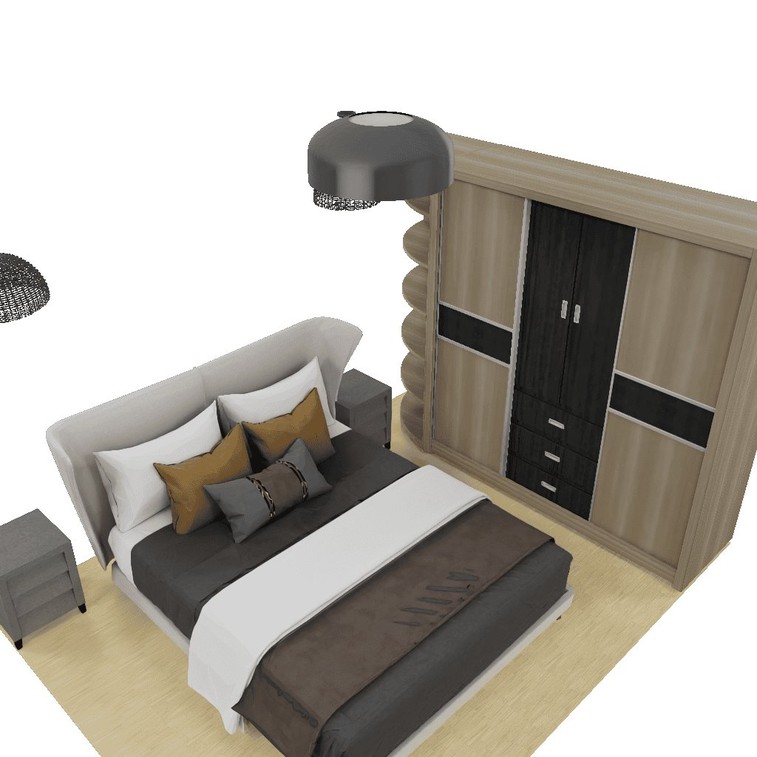}
\end{subfigure}\hfill
\vspace{-8pt}
  \caption{Qualitative results under physically implausible instructions. SceneNAT prioritizes layout plausibility rather than enforcing physically invalid relations.}
  \label{arbitration}
\end{figure}

\begin{figure}[t]
  \centering
\vspace{-6pt}
\hfill
\begin{subfigure}[b]{.20\textwidth}
  \centering
  \includegraphics[width=\linewidth]{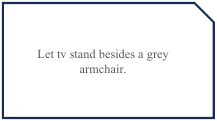}
\end{subfigure}\hfill\hfill
\begin{subfigure}[b]{.20\textwidth}
  \centering
  \includegraphics[width=\linewidth]{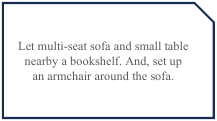}
\end{subfigure}\hfill
\begin{subfigure}[b]{.20\textwidth}
  \centering
  \includegraphics[width=\linewidth]{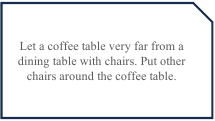}
\end{subfigure}\hfill

\hfill
\begin{subfigure}[b]{.29\textwidth}
  \centering
  \includegraphics[width=\linewidth]{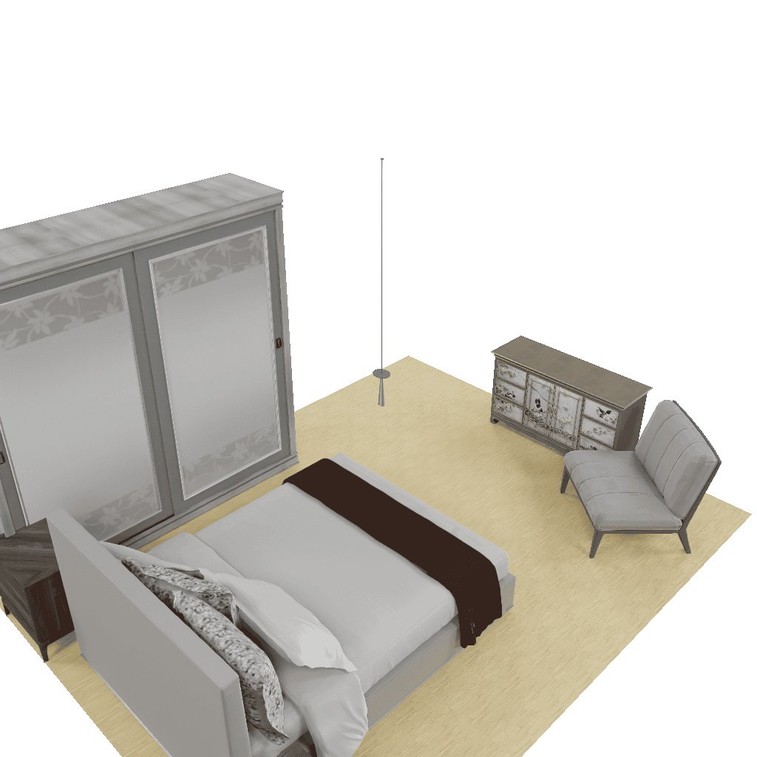}
\end{subfigure}\hfill\hfill
\begin{subfigure}[b]{.29\textwidth}
  \centering
  \includegraphics[width=\linewidth]{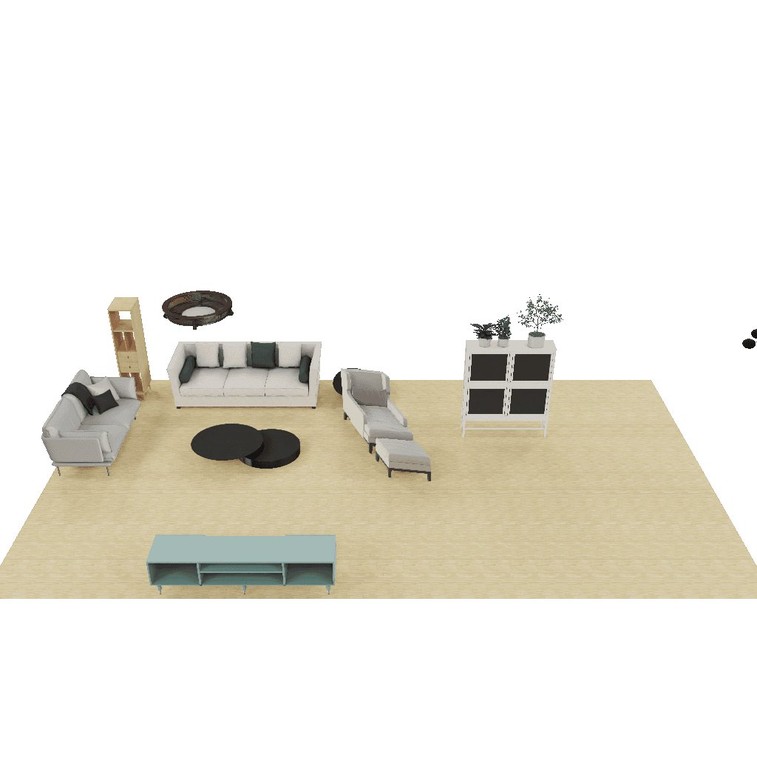}
\end{subfigure}\hfill\hfill
\begin{subfigure}[b]{.29\textwidth}
  \centering
  \includegraphics[width=\linewidth]{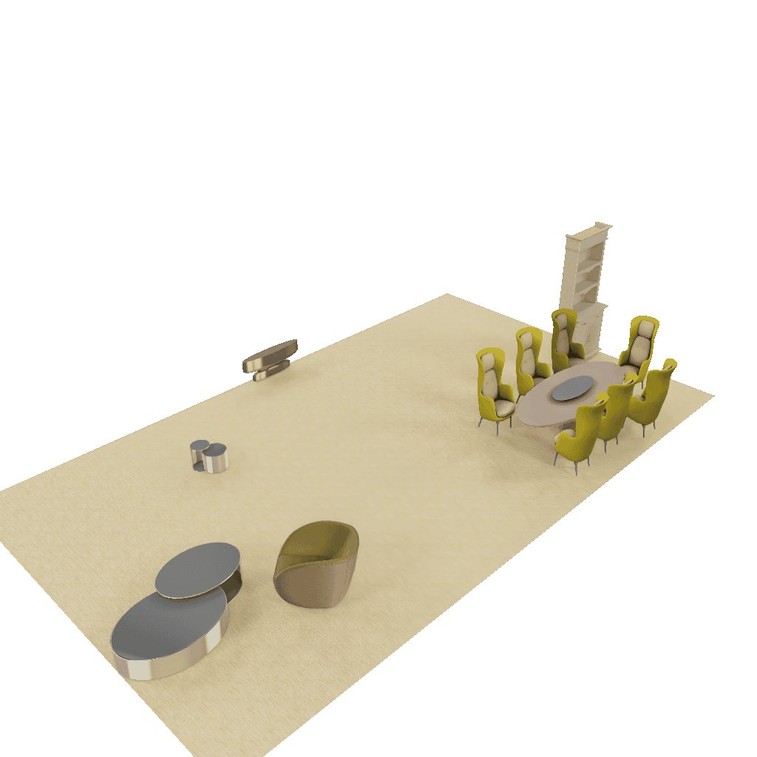}
\end{subfigure}\hfill
\vspace{2pt}

\hfill
\begin{subfigure}[b]{.20\textwidth}
  \centering
  \includegraphics[width=\linewidth]{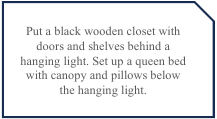}
\end{subfigure}\hfill\hfill
\begin{subfigure}[b]{.20\textwidth}
  \centering
  \includegraphics[width=\linewidth]{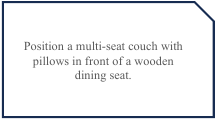}
\end{subfigure}\hfill\hfill
\begin{subfigure}[b]{.20\textwidth}
  \centering
  \includegraphics[width=\linewidth]{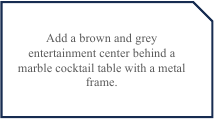}
\end{subfigure}\hfill

\hfill
\begin{subfigure}[b]{.29\textwidth}
  \centering
  \includegraphics[width=\linewidth]{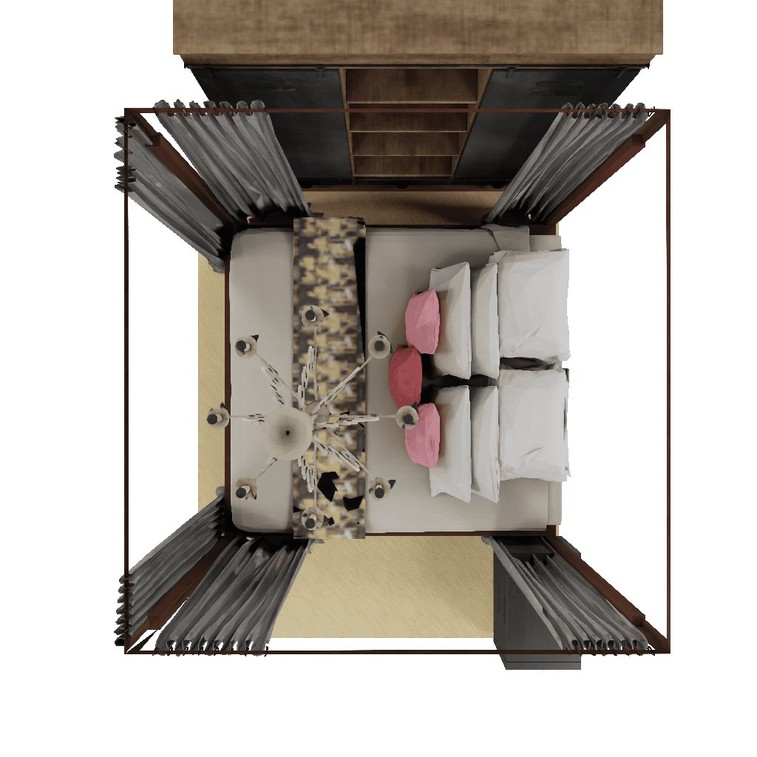}
\end{subfigure}\hfill\hfill
\begin{subfigure}[b]{.29\textwidth}
  \centering
  \includegraphics[width=\linewidth]{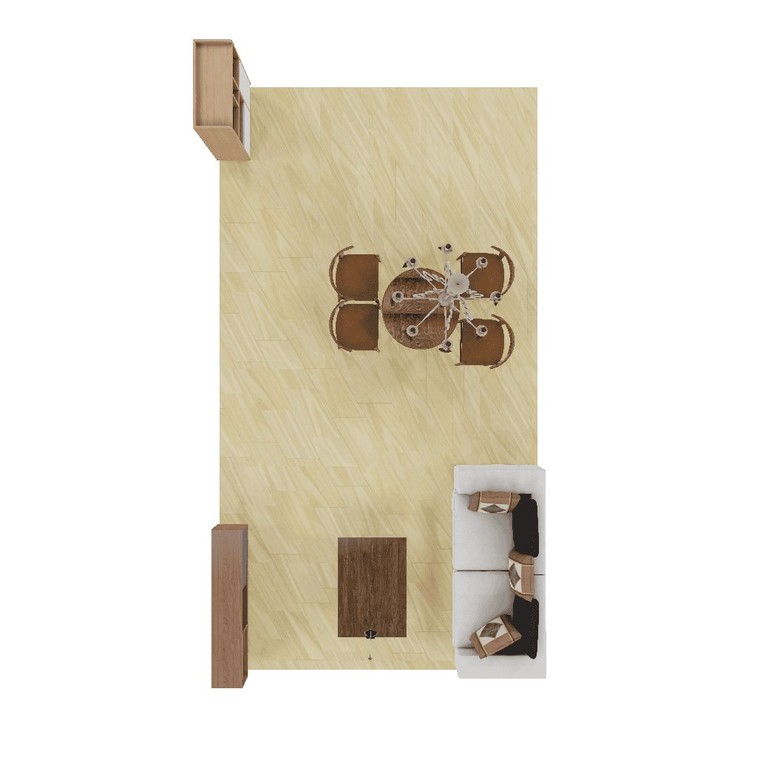}
\end{subfigure}\hfill\hfill
\begin{subfigure}[b]{.29\textwidth}
  \centering
  \includegraphics[width=\linewidth]{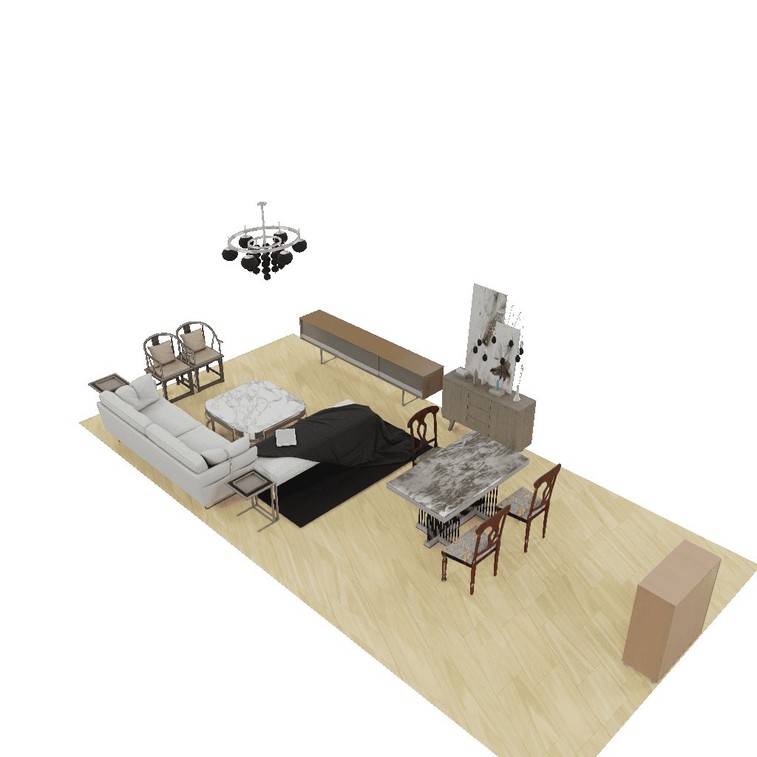}
\end{subfigure}\hfill
  \caption{Generalization to open-vocabulary instructions unseen during training. \textbf{Top:} unseen spatial predicates. \textbf{Bottom:} unseen object attributes.}
  \label{openvoca}
\end{figure}

\newpage
\begin{figure}[t]
\centering

\subcaptionbox{Role and Functionality\label{wild:a}}[1.0\linewidth]{
    \centering
    \begin{subfigure}[c]{.21\textwidth}
      \centering
      \includegraphics[width=\linewidth]{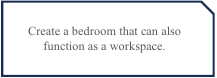}
    \end{subfigure}\hfill
    \begin{subfigure}[c]{.27\textwidth}
      \centering
      \includegraphics[width=\linewidth]{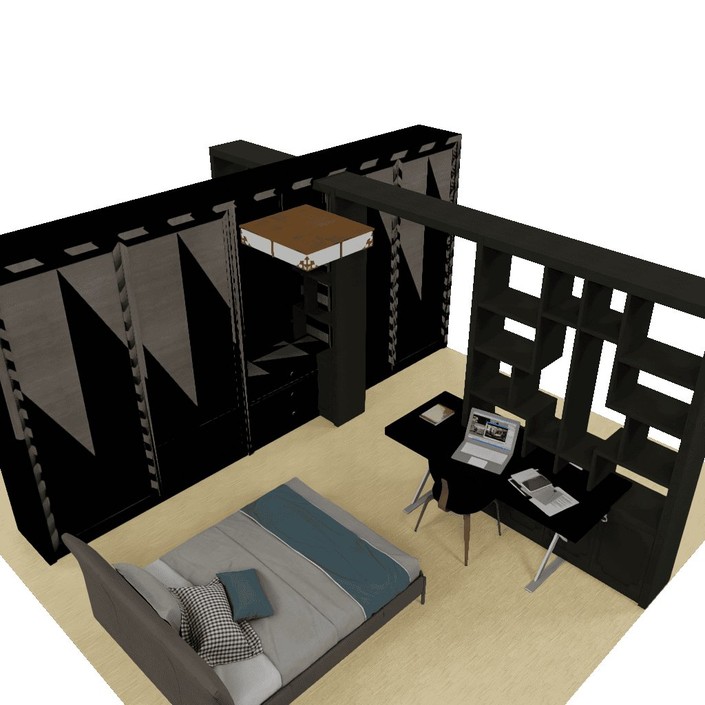}
    \end{subfigure}\hfill
    \begin{subfigure}[c]{.21\textwidth}
      \centering
      \includegraphics[width=\linewidth]{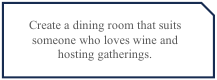}
    \end{subfigure}\hfill
    \begin{subfigure}[c]{.27\textwidth}
      \centering
      \includegraphics[width=\linewidth]{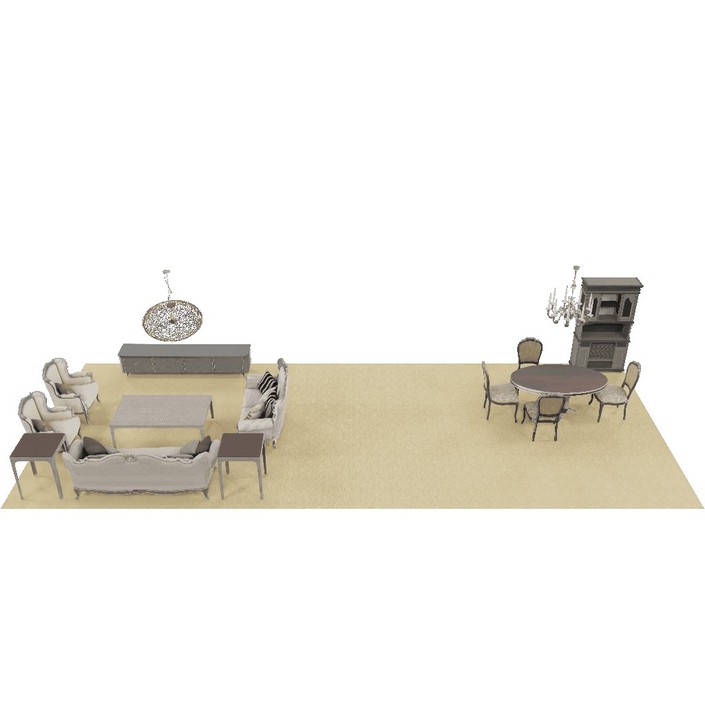}
    \end{subfigure}
}
\medskip

\subcaptionbox{Aesthetic and Abstract Descriptions\label{wild:b}}[1.0\linewidth]{
    \centering
    \begin{subfigure}[c]{.21\textwidth}
      \centering
      \includegraphics[width=\linewidth]{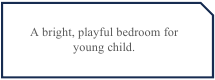}
    \end{subfigure}\hfill
    \begin{subfigure}[c]{.27\textwidth}
      \centering
      \includegraphics[width=\linewidth]{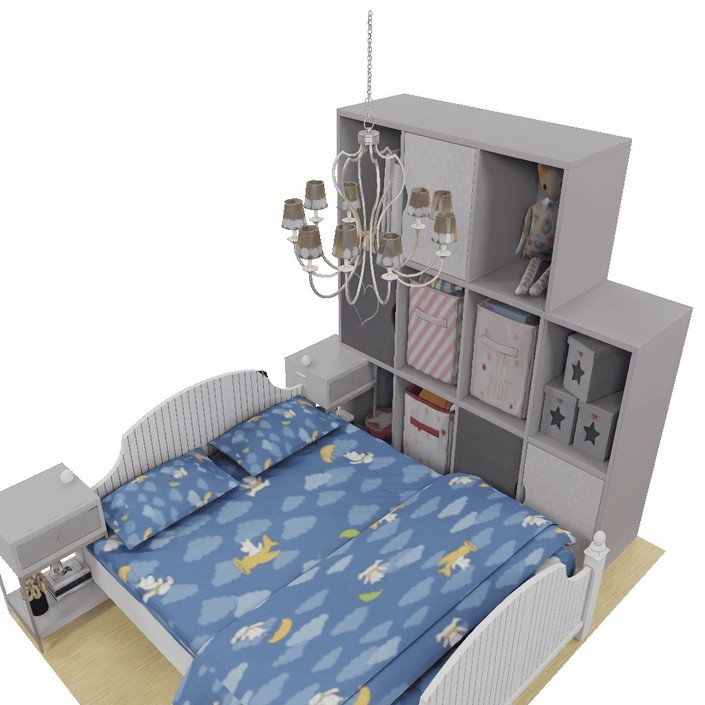}
    \end{subfigure}\hfill
    \begin{subfigure}[c]{.21\textwidth}
      \centering
      \includegraphics[width=\linewidth]{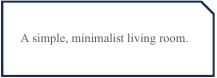}
    \end{subfigure}\hfill
    \begin{subfigure}[c]{.27\textwidth}
      \centering
      \includegraphics[width=\linewidth]{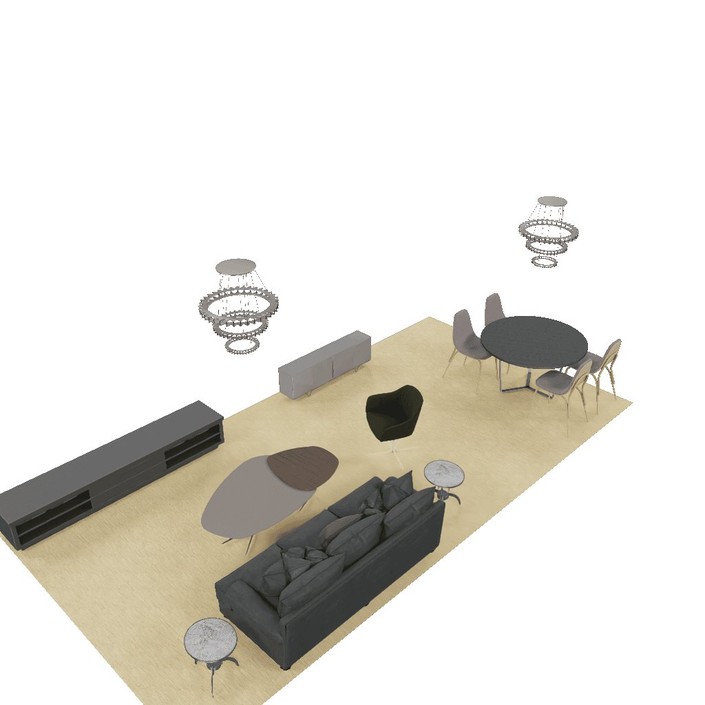}
    \end{subfigure}
}
\medskip

\subcaptionbox{Negative Constraints\label{wild:c}}[1.0\linewidth]{
    \centering
    \begin{subfigure}[c]{.21\textwidth}
      \centering
      \includegraphics[width=\linewidth]{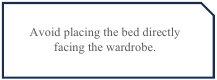}
    \end{subfigure}\hfill
    \begin{subfigure}[c]{.27\textwidth}
      \centering
      \includegraphics[width=\linewidth]{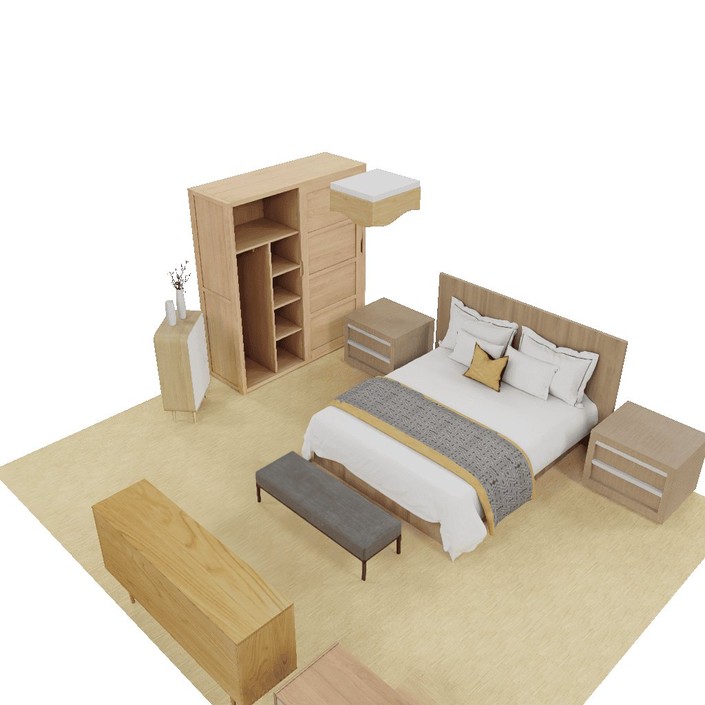}
    \end{subfigure}\hfill
    \begin{subfigure}[c]{.21\textwidth}
      \centering
      \includegraphics[width=\linewidth]{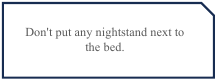}
    \end{subfigure}\hfill
    \begin{subfigure}[c]{.27\textwidth}
      \centering
      \includegraphics[width=\linewidth]{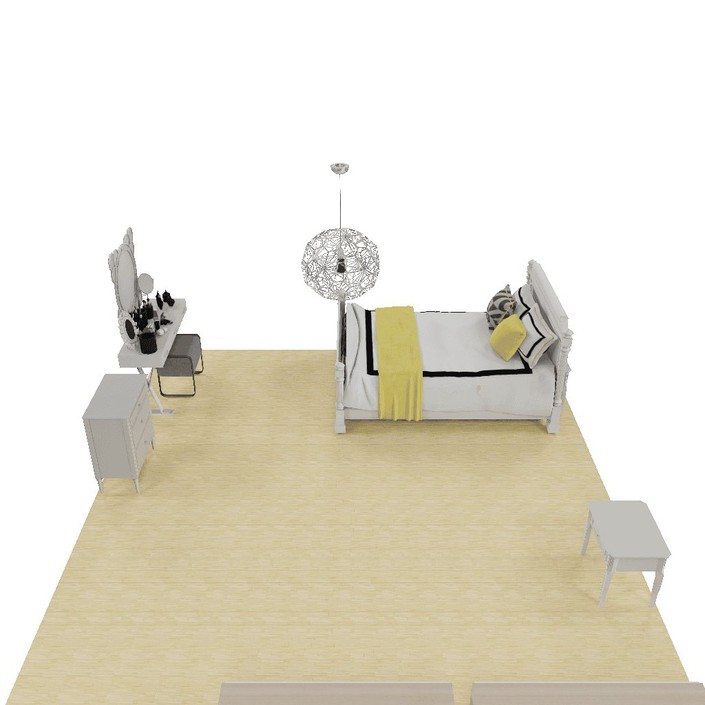}
    \end{subfigure}
}
\medskip

\subcaptionbox{High-Level Spatial Reasoning\label{wild:d}}[1.0\linewidth]{
    \centering
    \begin{subfigure}[c]{.21\textwidth}
      \centering
      \includegraphics[width=\linewidth]{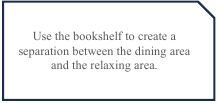}
    \end{subfigure}\hfill
    \begin{subfigure}[c]{.27\textwidth}
      \centering
      \includegraphics[width=\linewidth]{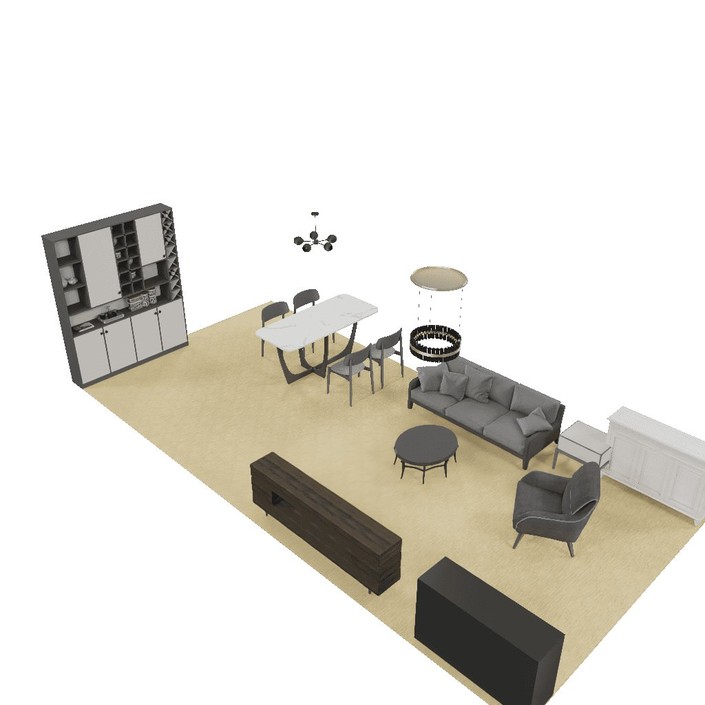}
    \end{subfigure}\hfill
    \begin{subfigure}[c]{.21\textwidth}
      \centering
      \includegraphics[width=\linewidth]{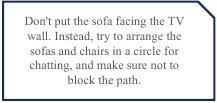}
    \end{subfigure}\hfill
    \begin{subfigure}[c]{.27\textwidth}
      \centering
      \includegraphics[width=\linewidth]{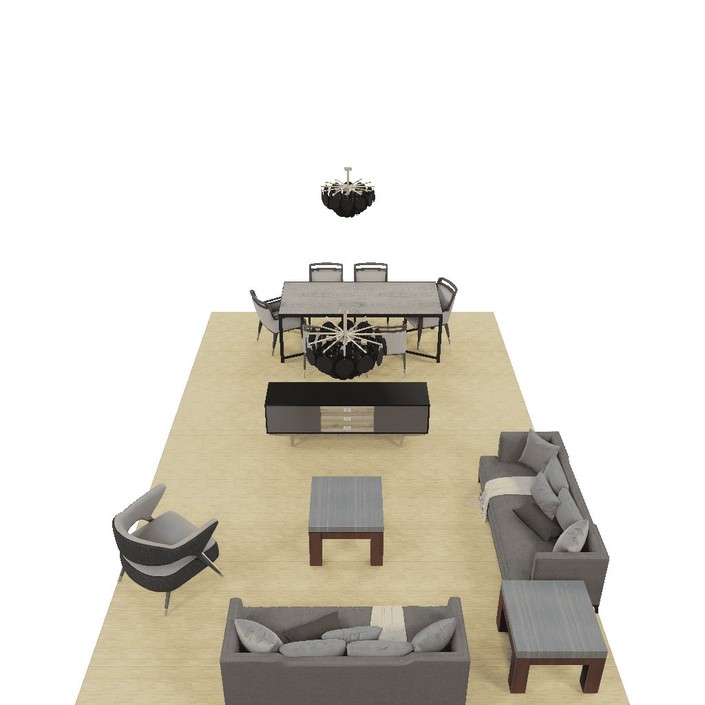}
    \end{subfigure}
}

\caption{Qualitative results of various types of ``in-the-wild" human instructions.}
\label{in-the-wild}
\end{figure}
\clearpage

\begin{figure}[t]
  \centering

\vspace{-20pt}  
\begin{subfigure}[b]{.16\textwidth}
  \centering
  \includegraphics[width=\linewidth]{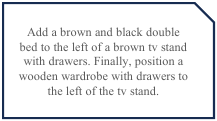}
\end{subfigure}\hfill
\begin{subfigure}[b]{.16\textwidth}
  \centering
  \includegraphics[width=\linewidth]{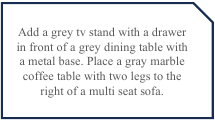}
\end{subfigure}\hfill
\begin{subfigure}[b]{.16\textwidth}
  \centering
  \includegraphics[width=\linewidth]{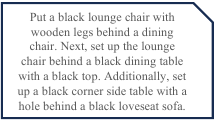}
\end{subfigure}\hfill
\begin{subfigure}[b]{.20\textwidth}
  \centering
  \includegraphics[width=\linewidth]{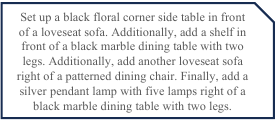}
\end{subfigure}

\begin{subfigure}[b]{.21\textwidth}
  \centering
  \includegraphics[width=\linewidth]{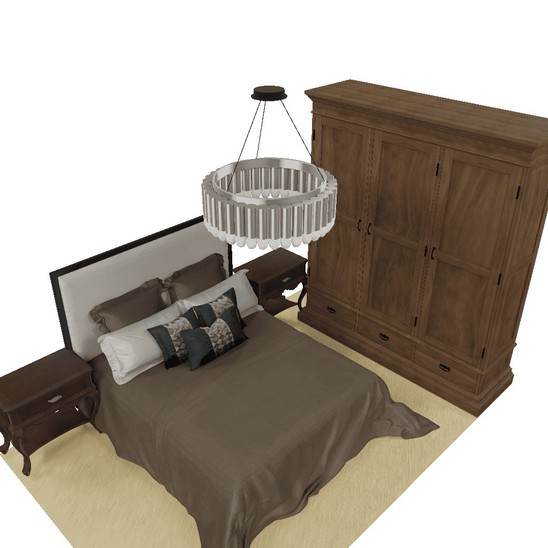}
\end{subfigure}\hfill
\begin{subfigure}[b]{.21\textwidth}
  \centering
  \includegraphics[width=\linewidth]{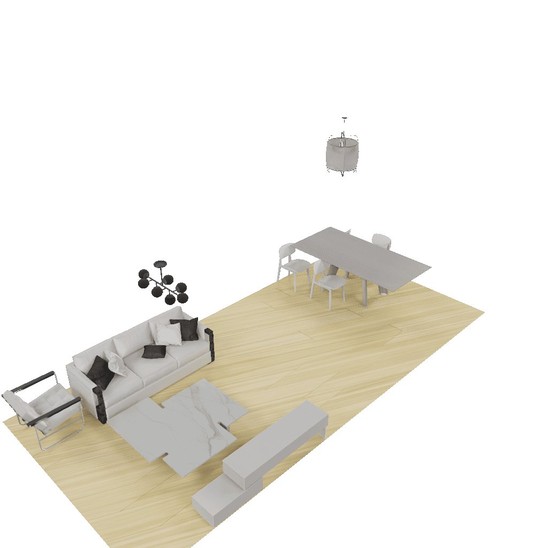}
\end{subfigure}\hfill
\begin{subfigure}[b]{.21\textwidth}
  \centering
  \includegraphics[width=\linewidth]{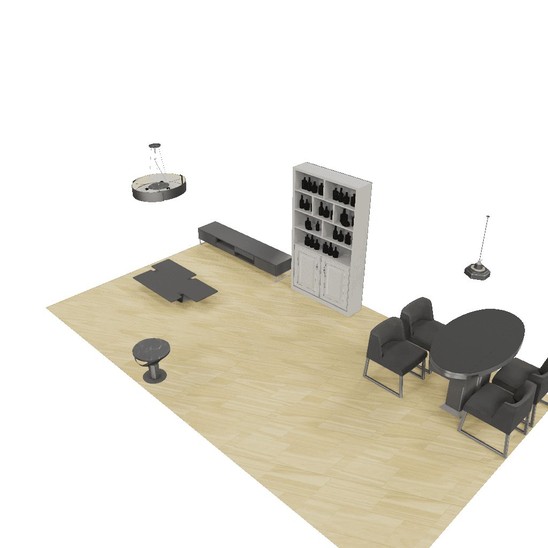}
\end{subfigure}\hfill
\begin{subfigure}[b]{.21\textwidth}
  \centering
  \includegraphics[width=\linewidth]{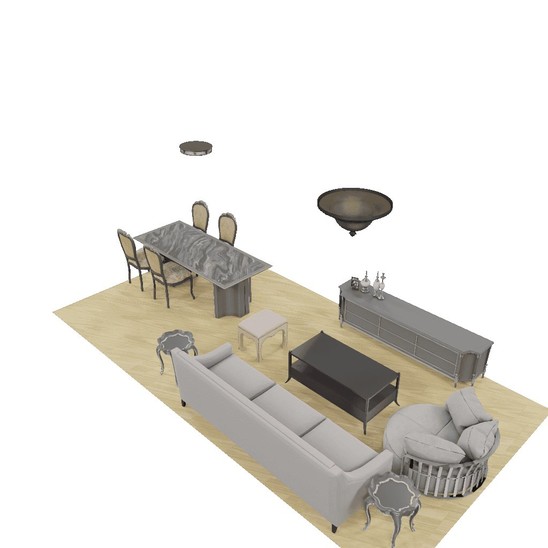}
\end{subfigure}

\caption{Failure case where the generated scene fails to satisfy the spatial relations specified in the instruction.}
\vspace{-20pt}  
  \label{fail_rel}
\end{figure}

\begin{figure}[t]
  \centering
\hfill
\begin{subfigure}[c]{.18\textwidth}
  \centering
  \includegraphics[width=\linewidth]{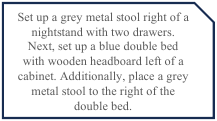}
\end{subfigure}\hfill\hfill
\begin{subfigure}[c]{.22\textwidth}
  \centering
  \includegraphics[width=\linewidth]{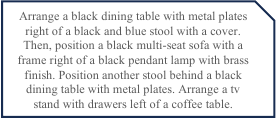}
\end{subfigure}\hfill

\hfill
\begin{subfigure}[b]{.32\textwidth}
  \centering
  \includegraphics[width=\linewidth]{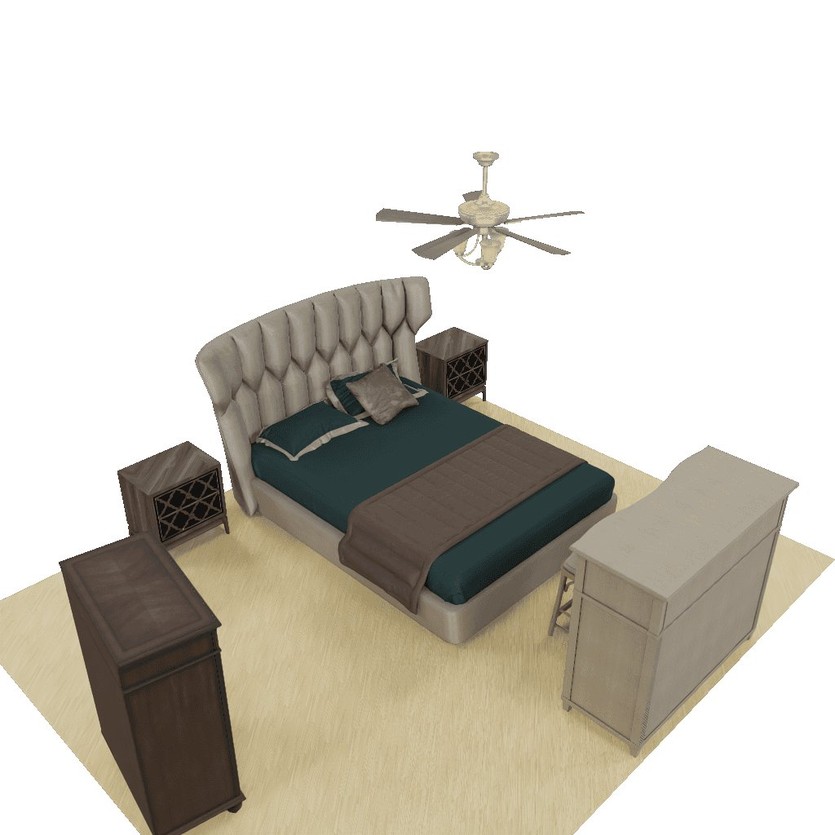}
\end{subfigure}\hfill\hfill
\begin{subfigure}[b]{.32\textwidth}
  \centering
  \includegraphics[width=\linewidth]{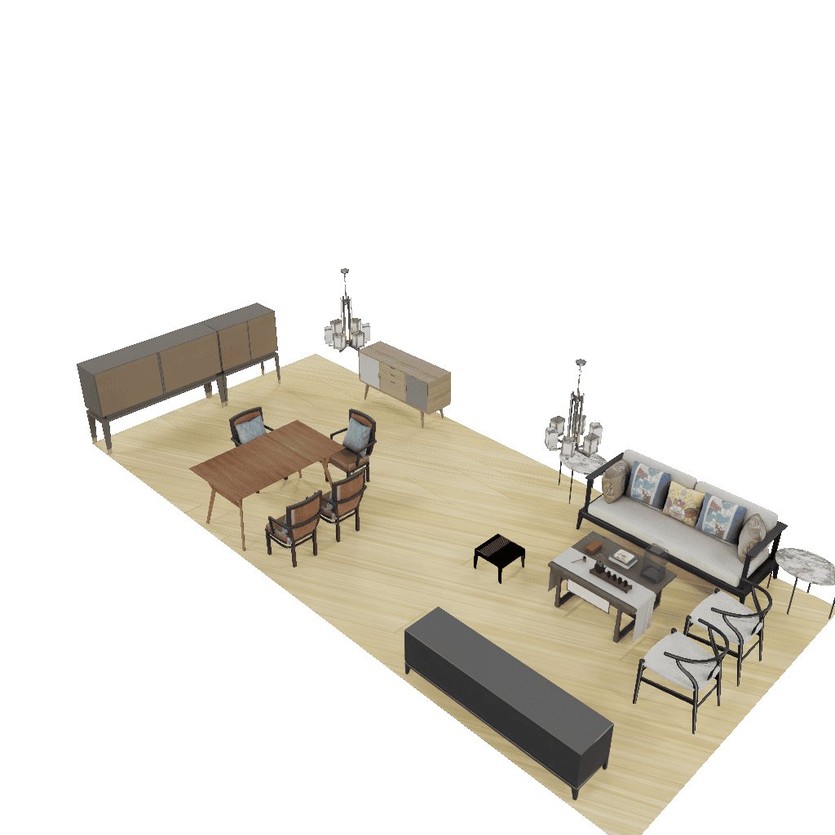}
\end{subfigure}\hfill

\hfill\begin{subfigure}[c]{.18\textwidth}
  \centering
  \includegraphics[width=\linewidth]{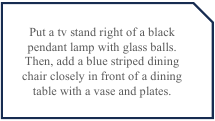}
\end{subfigure}\hfill\hfill
\begin{subfigure}[c]{.18\textwidth}
  \centering
  \includegraphics[width=\linewidth]{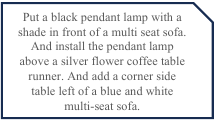}
\end{subfigure}
\vspace{2pt}

\hfill
\begin{subfigure}[b]{.32\textwidth}
  \centering
  \includegraphics[width=\linewidth]{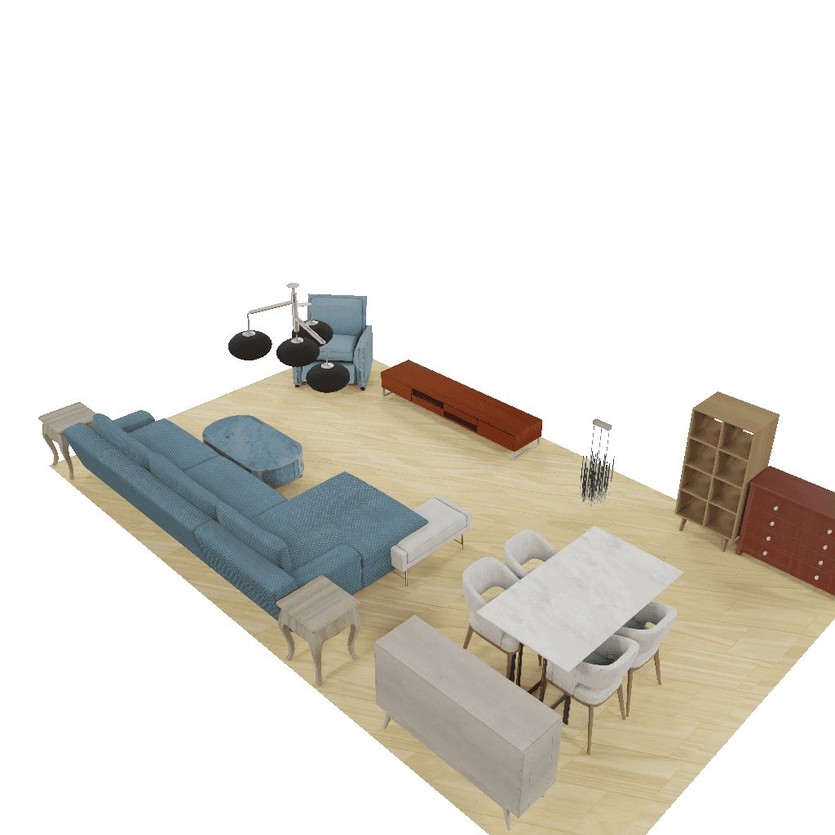}
\end{subfigure}\hfill\hfill
\begin{subfigure}[b]{.32\textwidth}
  \centering
  \includegraphics[width=\linewidth]{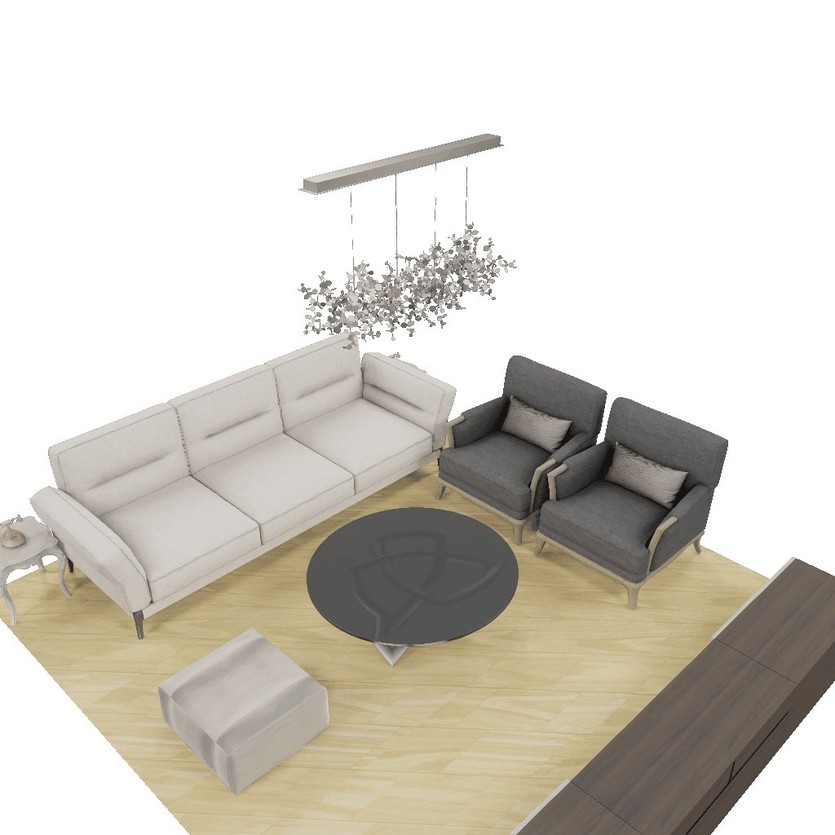}
\end{subfigure}
\caption{Failure case where the generated scene fails to satisfy the object attributes specified in the instruction.}
  \label{fail_desc}
\end{figure}

\newpage
\begin{figure}
  \centering

\begin{subfigure}[b]{.40\textwidth}
  \centering
  \includegraphics[width=\linewidth]{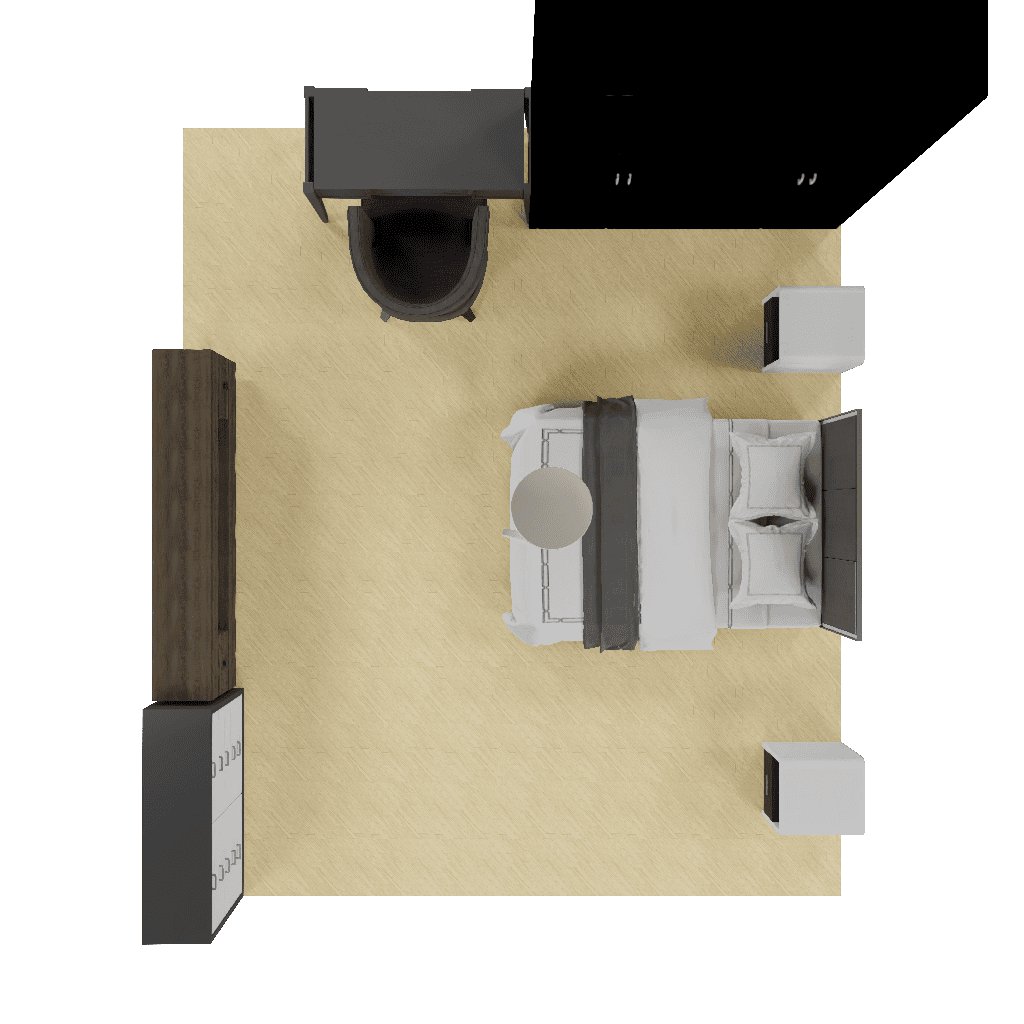}
\end{subfigure}\hfill
\begin{subfigure}[b]{.40\textwidth}
  \centering
  \includegraphics[width=\linewidth]{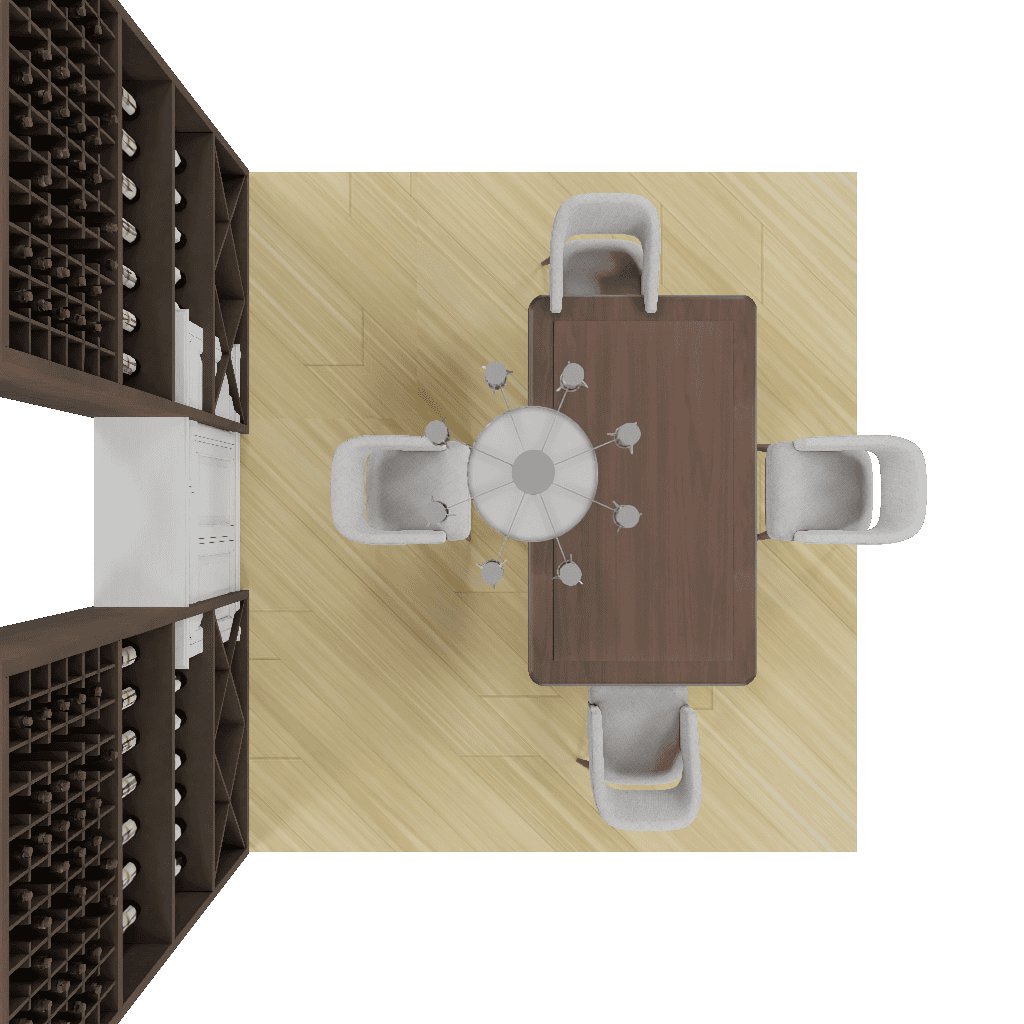}
\end{subfigure}\hfill

\begin{subfigure}[b]{.40\textwidth}
  \centering
  \includegraphics[width=\linewidth]{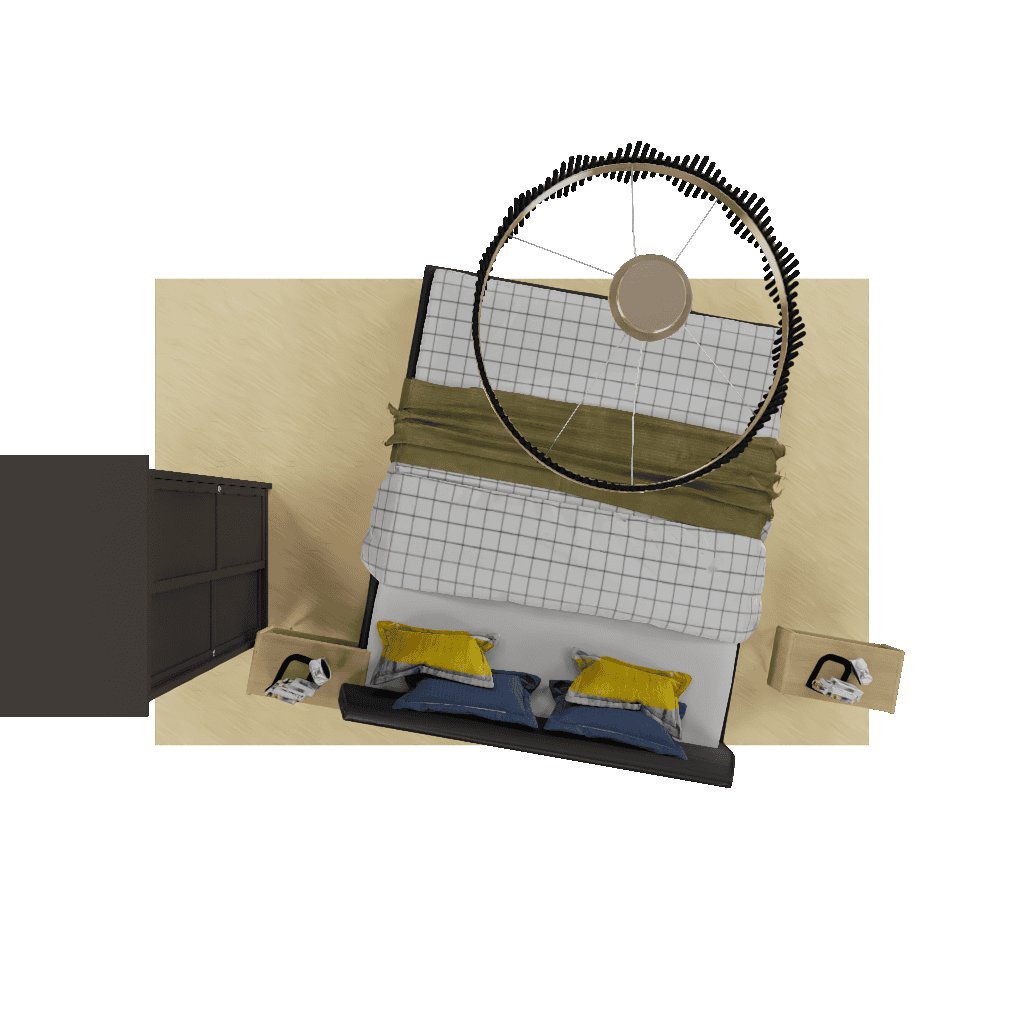}
\end{subfigure}\hfill
\begin{subfigure}[b]{.40\textwidth}
  \centering
  \includegraphics[width=\linewidth]{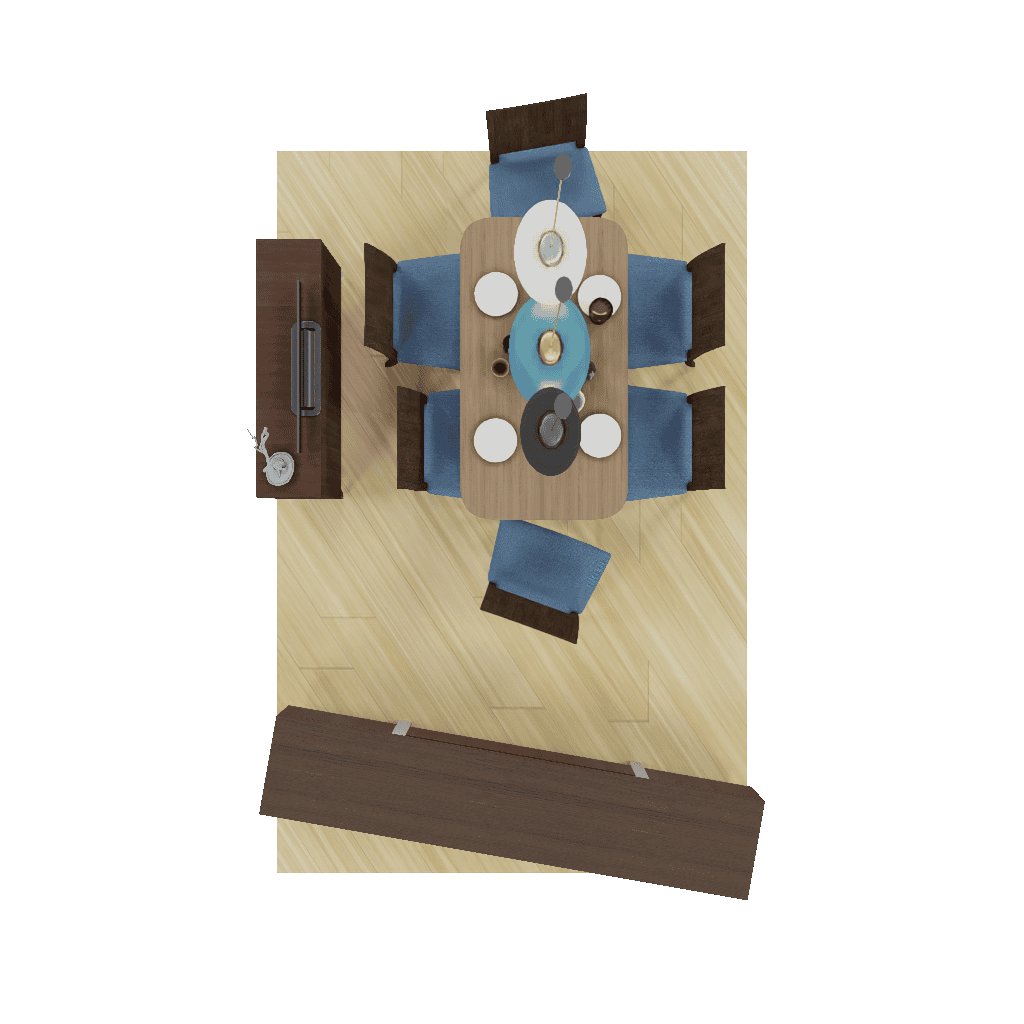}
\end{subfigure}\hfill

\begin{subfigure}[b]{.40\textwidth}
  \centering
  \includegraphics[width=\linewidth]{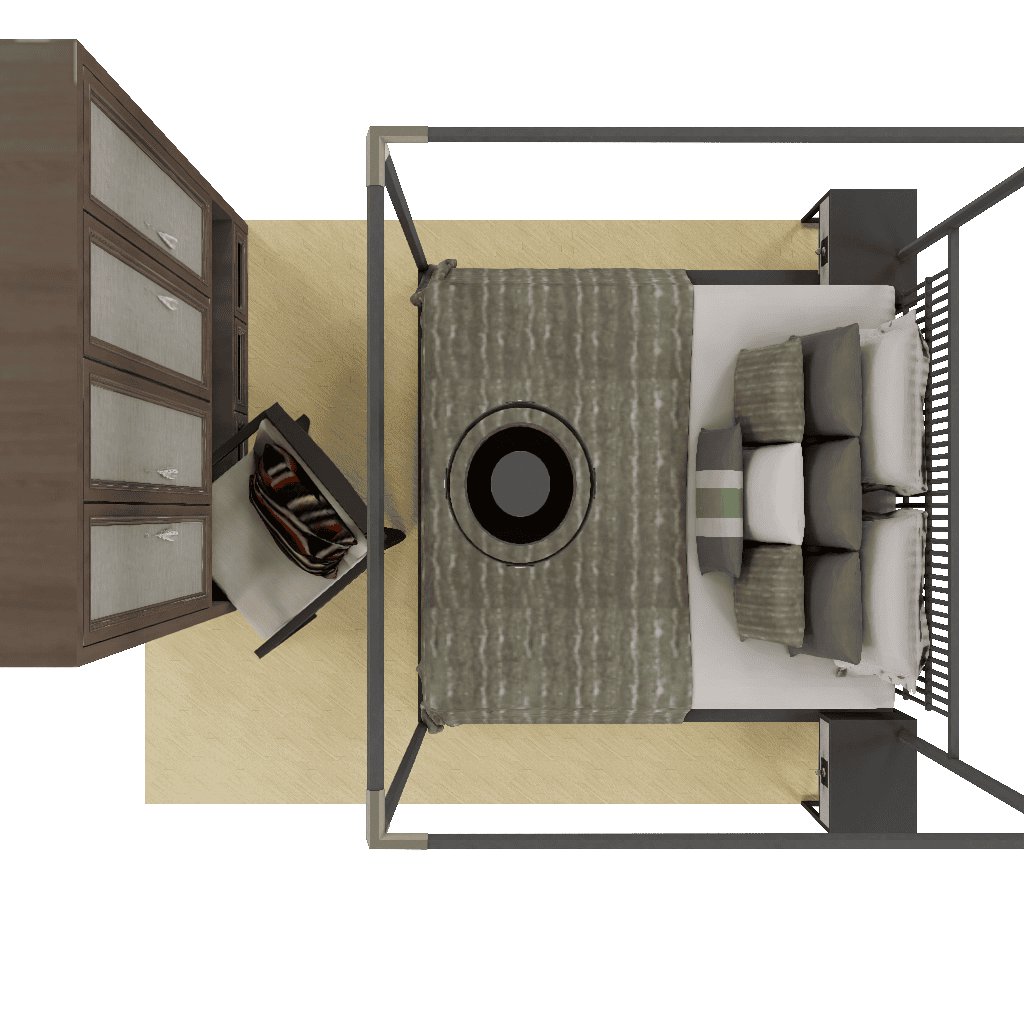}
\end{subfigure}\hfill
\begin{subfigure}[b]{.40\textwidth}
  \centering
  \includegraphics[width=\linewidth]{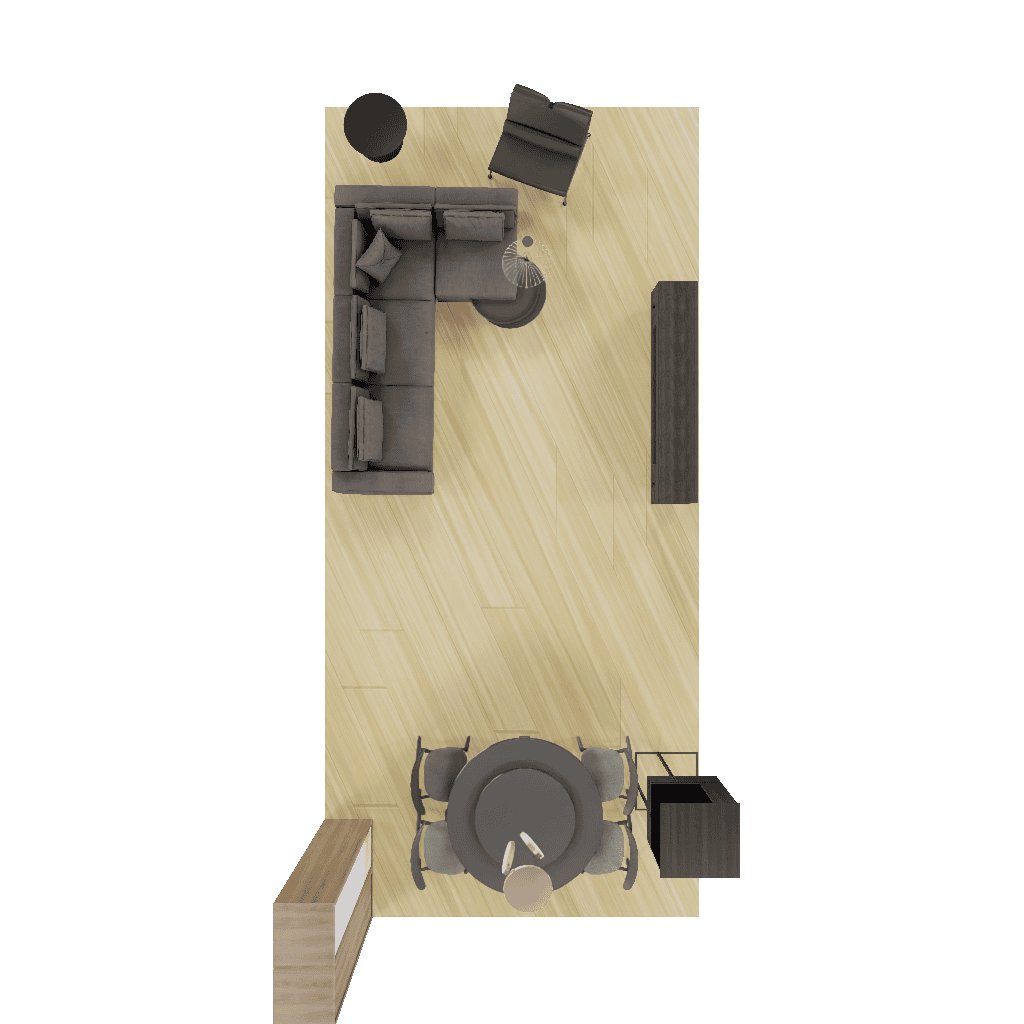}
\end{subfigure}\hfill

  \caption{Failure case where the generated scene fails to satisfy physical plausibility, exhibiting object collisions or incorrect scale and rotation.}
  \label{fail_layout}
\end{figure}

\end{document}